\documentclass[10pt,twocolumn,letterpaper]{article}

\usepackage{cvpr}              %

\usepackage[dvipsnames]{xcolor}

\usepackage[accsupp]{axessibility}

\usepackage{times}
\usepackage{epsfig}
\usepackage{graphicx}
\usepackage{amsmath}
\usepackage{amssymb}
\usepackage{lipsum}
\usepackage{setspace}

\usepackage{makecell}
\usepackage{multirow}
\usepackage{bbold}
\usepackage{adjustbox}
\usepackage{caption}
\usepackage{subcaption}
\usepackage{tikz}
\usepackage{comment}
\usepackage{booktabs}
\usepackage{array}
\usepackage{colortbl}
\usepackage[percent]{overpic}

\usepackage{pifont}
\newcommand{\cmark}{\ding{51}}
\newcommand{\xmark}{\ding{55}}

\definecolor{cvprblue}{rgb}{0.21,0.49,0.74}
\usepackage[pagebackref,breaklinks,colorlinks,citecolor=cvprblue]{hyperref}

\title{Task-aligned Part-aware Panoptic Segmentation \\ through Joint Object-Part Representations}

\author{Daan de Geus \quad\quad Gijs Dubbelman\\
Eindhoven University of Technology \\
{\tt\small \{d.c.d.geus, g.dubbelman\}@tue.nl}
}

\begin{document}
\maketitle
\begin{abstract}
Part-aware panoptic segmentation (PPS) requires (a) that each foreground object and background region in an image is segmented and classified, and (b) that all parts within foreground objects are segmented, classified and linked to their parent object. Existing methods approach PPS by separately conducting object-level and part-level segmentation. However, their part-level predictions are not linked to individual parent objects. Therefore, their learning objective is not aligned with the PPS task objective, which harms the PPS performance. To solve this, and make more accurate PPS predictions, we propose Task-Aligned Part-aware Panoptic Segmentation (TAPPS). This method uses a set of shared queries to jointly predict (a) object-level segments, and (b) the part-level segments within those same objects. As a result, TAPPS learns to predict part-level segments that are linked to individual parent objects, aligning the learning objective with the task objective, and allowing TAPPS to leverage joint object-part representations. With experiments, we show that TAPPS considerably outperforms methods that predict objects and parts separately, and achieves new state-of-the-art PPS results.

\end{abstract}    
\section{Introduction}
\label{sec:intro}

To fully understand what is depicted in an image, it is important to consider concepts at different levels of abstraction. For rich scene understanding, we should not only recognize foreground objects (\eg, \textit{car}, \textit{human}) and background regions (\eg, \textit{sky}, \textit{ocean}), but also simultaneously identify the parts that constitute the objects (\eg, \textit{car-wheel}, \textit{human-arm}). In a step towards such comprehensive scene understanding, De Geus \etal introduced part-aware panoptic segmentation (PPS)~\cite{degeus2021pps}. 
The objective of this computer vision task is 
(1) to output a segmentation mask and class label for all thing objects and stuff regions in an image like for panoptic segmentation~\cite{kirillov2019ps} -- we call these \textit{object-level segments} -- and 
(2) to simultaneously segment and classify all parts \textit{within} each identified object. These are called \textit{part-level segments} and should be explicitly linked to an object-level segment, establishing a part-whole relation.

\begin{figure}[t]
	\centering
    	\includegraphics[width=1.\linewidth]{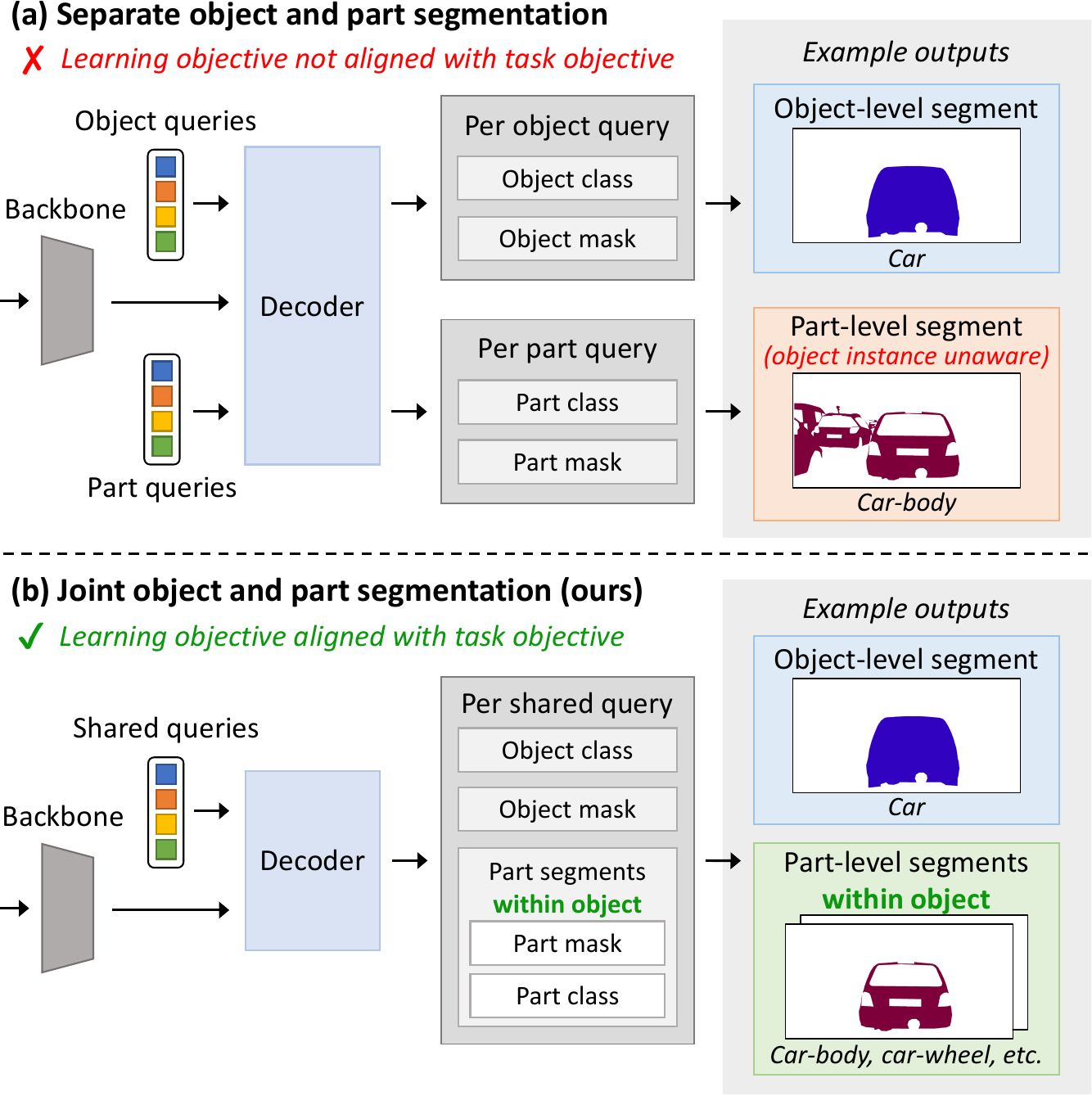}
    \vspace{-15pt}
	\caption{\textbf{Task-aligned part-aware panoptic segmentation.} \textbf{(a)} Existing works separately predict object-level segments and \textit{object-instance-unaware} part-level segments. \textbf{(b)} In this work, we predict objects and parts jointly, using a set of shared queries. This allows our method to predict parts within individual object segments, aligning its learning objective with the PPS task objective.}
\label{fig:eyecatcher}
\vspace{-5pt}
\end{figure}

Current state-of-the-art methods~\cite{li2022ppf,li2023ppf++} address PPS by using two different sets of learnable queries to separately predict object-level and part-level segments, see \cref{fig:eyecatcher}\textcolor{red}{a}. While these methods outperform earlier baselines~\cite{degeus2021pps}, they have one main limitation: their learning objective is not aligned with the task objective. Where the PPS task objective is to predict parts within each individual object-level segment, these existing works conduct part-level semantic segmentation, predicting part-level masks that cover multiple objects (see \cref{fig:eyecatcher}\textcolor{red}{a}). In other words, these networks are not optimized for the PPS task, but instead solve the surrogate subtasks of object-level panoptic segmentation and part-level semantic segmentation. As a result, to assign parts to individual objects, these methods require post-processing. Additionally, we hypothesize that there are several other negative consequences: (1) These networks learn a conflicting feature representation. They learn that object-level thing instances should be separated, but also that the parts of these object-level instances should be grouped together. We expect that this harms the ability of the networks to separate instances. (2) Predicting objects and parts separately may cause incompatible predictions (\eg, a \textit{car-window} part with a \textit{bicycle} object), which makes it unclear which prediction to trust, and requires further post-processing. (3) The networks encode information about objects and their parts separately, whereas this information is potentially complementary. See \cref{sec:method:prelims:mask_cls_pps} for more details on these limitations.

In this work, we aim to design a \textbf{simple} network for PPS that overcomes these limitations and thereby makes more accurate PPS predictions. To achieve this, we propose the Task-Aligned Part-aware Panoptic Segmentation (TAPPS) method. Instead of using separate queries for objects and parts, TAPPS uses one set of shared queries to jointly represent objects and the parts they contain. Specifically, each of these queries learns to represent at most one object-level segment, for which it predicts (1) a segmentation mask and object-level class, and (2) the 
segmentation masks and classes for all part-level segments within this object. This is visualized in \cref{fig:eyecatcher}\textcolor{red}{b} and explained in \cref{sec:method}.

With this approach, TAPPS explicitly predicts part-level segments per individual object. Therefore, the network's learning objective is now aligned with the PPS task objective, and TAPPS is directly optimized for the PPS task. As a result, both object- and part-level segmentation are learned in an `object-instance-aware' manner. This removes the conflict in the learned feature representations, allowing for better instance separability. Additionally, we hypothesize that predicting objects and parts from the same query reduces incompatibilities between object and part predictions, as they are made using the same information. Moreover, we use TAPPS to go even further, and explicitly only predict part segments compatible with predicted object segments. This enforces full object-part compatibility and simplifies the part segmentation task. Finally, the network can now encode complementary object-level and part-level information in a shared query, giving it a richer representation for making object- and part-level segmentation predictions.

With experiments, detailed in \cref{sec:experiments}, we show that TAPPS outperforms the baseline with separate object and part queries in multiple aspects: (1) The part segmentation quality within identified objects is significantly better, which shows the positive impact of using joint object-part representations and simplifying the part segmentation task. (2) The object instance segmentation performance is considerably improved, demonstrating the benefits of learning parts in an object-instance-aware manner. Together, this yields a large overall improvement with respect to the baseline, and causes TAPPS to considerably outperform existing works across different benchmarks, achieving new state-of-the-art results. See \cref{sec:results} for more extensive results.

To summarize, we make the following contributions:
\begin{itemize}
    \item We propose TAPPS, a simple approach for PPS that aligns the learning objective with the task objective, facilitating object instance separability and enabling joint object-part representations for more accurate PPS predictions.
    \item We use the shared object-part queries to constrain TAPPS to only predict part segments that are compatible with the object class, enforcing object-part compatibility and simplifying the part segmentation task.
    \item We experimentally show the effectiveness of TAPPS across a range of datasets and network configurations.
\end{itemize}
The code for TAPPS is made publicly available through \url{https://tue-mps.github.io/tapps/}.
\section{Related work}
\label{sec:rel_work}

\paragraph{Part-aware panoptic segmentation.}

Part-aware panoptic segmentation~\cite{degeus2021pps} is an image segmentation task introduced for scene understanding at multiple abstraction levels. It extends panoptic segmentation~\cite{kirillov2019ps} by requiring (a) object-level panoptic segmentation, and (b) part-level segmentation \textit{within} object-level segments. Most existing works do not predict parts per object-level segment, but instead make separate predictions for panoptic segmentation and the surrogate task of part-level semantic segmentation -- or \textit{part segmentation}~\cite{li2022ppf,li2023ppf++,jagadeesh2022jppf}. JPPF~\cite{jagadeesh2022jppf} proposes a single-network approach with a shared encoder followed by separate heads for semantic, instance, and part segmentation. To merge the predictions by these heads to the PPS format, JPPF proposes a new rule-based fusion strategy that outperforms the originally introduced merging strategy for PPS~\cite{degeus2021pps}. Panoptic-PartFormer and its extension Panoptic-PartFormer++~\cite{li2022ppf,li2023ppf++} tackle PPS with a Transformer-based approach that independently makes panoptic segmentation and part segmentation predictions using separate sets of learnable queries for thing, stuff and part segments. 
Alternatively, ViRReq~\cite{tang2022virreq} is a paradigm in which PPS is predicted in a cascading fashion, by first segmenting objects and then segmenting the parts within these objects `by request', but it requires multiple networks to achieve this.

In contrast to these existing approaches, we propose a method that jointly predicts object-level segments and the part-level segments within these objects. This aligns the learning objective with the task objective, and yields improved PPS performance. For more details, see \cref{sec:method}.

\paragraph{Part-level image segmentation.}
Part-level segmentation is also widely studied beyond PPS. Most works address part segmentation, \ie, semantic segmentation for parts~\cite{zhao2019bsanet,singh2022float,tan2021partparsing,michieli2020gmnet,sun2023vlpart,wei2023ovparts,liu2022udapart,li2022deephierarchical,gong2019graphonomy,gong2018cihp,li2018multi,li2020self,li2017holistic,liang2018lip, ruan2019devil,wang2020hierarchical,zhao2018mhp}. For more complete scene understanding, other works also take into account object instances, like PPS does, with instance-aware part segmentation~\cite{li2020self,ruan2019devil,yang2021hierrcnn,yang2020rprcnn,yang2019parsingrcnn,zhang2022aiparsing,pan2023ops}. However, these works do not consider background stuff classes, which PPS does consider. Alternatively, UPerNet~\cite{xiao2018upernet} separately predicts part-level and object-level semantic segmentation, including background classes, but without any instance awareness. Another work has extended PPS by also predicting the relations between objects~\cite{qi2023aims}, training a model on annotations from different datasets. In this work, we address the PPS task as introduced by De Geus \etal~\cite{degeus2021pps}, and we compare our approach to state-of-the-art alternatives.

\section{Method}
\label{sec:method}

\subsection{PPS task definition}
\label{sec:method:task_def}

Part-aware panoptic segmentation (PPS)~\cite{degeus2021pps} requires consistent image segmentation across two abstraction levels: the object level and part level. For object-level segmentation, following panoptic segmentation~\cite{kirillov2019ps}, an image should be divided into $N$ object-level segments, where $N$ varies per image. Each segment $s_{i}$ consists of a binary mask $\mathbf{M}^{\textrm{obj}}_{i}$ and an object class label $c^{\textrm{obj}}_{i}$. Thing classes require a segment per object instance, and stuff classes require a single segment per class. Next, per segment $s_i$, PPS requires a set of $K_{i}$ part-level segments. Each part-level segment $s^{\textrm{pt}}_{i,j}$ consists of a binary part-level mask $\mathbf{M}^{\textrm{pt}}_{i,j}$ and a part-level class $c^{\textrm{pt}}_{i,j}$ that are compatible with the object-level mask and class. Specifically, the part-level mask 
must be a subset of the object-level mask,
and the part-level class 
should be one of the part-level classes defined for the object-level class.
For instance, if $c^{\textrm{obj}}_{i}$ is \textit{car}, then $c^{\textrm{pt}}_{i,j}$ cannot be \textit{human-head} but it can be \textit{car-window}. For all parts, a single part-level segment is required for each part class within an object.

To summarize, each image should be divided into a set of segments $\mathcal{S} = \{{s_{i}}\}^{N}_{i=1} = \{(\mathbf{M}^{\textrm{obj}}_{i}, c^{\textrm{obj}}_{i}, \mathcal{P}_{i})\}^{N}_{i=1}$, where $\mathcal{P}_{i} = \{{s^{\textrm{pt}}_{i,j}}\}^{{K}_{i}}_{j=1} = \{(\mathbf{M}^{\textrm{pt}}_{i,j}, c^{\textrm{pt}}_{i,j})\}^{{K}_{i}}_{j=1}$ is a set of part-level segments compatible with the object-level segment in $s_i$. In practice, part-level classes are only defined for some object-level classes. When segment $s_i$ has an object class that does not have parts, then $K_{i} = 0$ and $s_{i} = (\mathbf{M}^{\textrm{obj}}_{i}, c^{\textrm{obj}}_{i})$.

\subsection{Preliminaries}
\label{sec:method:prelims}
Like existing state-of-the-art PPS approaches~\cite{li2022ppf,li2023ppf++}, our work uses the powerful \textit{mask classification} meta-architecture that is the foundation of many state-of-the-art segmentation models~\cite{cheng2021maskformer,cheng2022mask2former,zhang2021knet,jain2023oneformer,wang2021maxdeeplab}. Here, we first explain this mask classification paradigm in more detail. Then, we explain how existing works use it for PPS, and what the limitations are of these approaches.

\subsubsection{Mask classification framework}
The concept of mask classification is to predict a set of $N^{\textrm{q}}$ object-level segments, \ie, a set of $N^{\textrm{q}}$ pixel-level masks $\mathbf{\hat{M}}^{\textrm{obj}}$ and corresponding class labels $\hat{c}^{\textrm{obj}}$. To make these predictions, mask classification networks output two components: high-resolution features $\mathbf{F} \in \mathbb{R}^{{E}\times{H}\times{W}}$ and queries $\mathbf{Q} \in \mathbb{R}^{N^{\textrm{q}}\times{E}}$, where $H$ and $W$ are the height and width of the features, and $E$ is the feature and query dimensionality. Each of the $N^{\textrm{q}}$ queries is used to predict the class label and segmentation mask of at most one object-level segment. The class predictions are a function of only the queries, and the mask predictions are a function of both the queries $\mathbf{Q}$ and the high-resolution features $\mathbf{F}$. To output these high-resolution features, the input image is first fed into a \textit{backbone} to extract multi-scale features, \eg, ResNet~\cite{he2016resnet} or Swin~\cite{liu2021swin}. Subsequently, these features are further refined and upsampled to high-resolution features by a \textit{pixel decoder}, \eg, Semantic FPN~\cite{lin2017fpn,kirillov2019panopticfpn}. In the other part of the network, the queries $\mathbf{Q}$ are generated by processing learnable queries $\mathbf{Q^0}$ using a \textit{transformer decoder}, which applies self-attention across queries and cross-attention with image features. Through bipartite matching between $N^{\textrm{q}}$ predicted and $N$ ground-truth segments, each query is assigned to at most one ground-truth segment. As $N^{\textrm{q}}$ is not always equal to $N$, some queries may not be assigned to a ground-truth segment. If so, they do not receive any supervision for segmentation, and will learn a `no object' class. In this work, we build upon the Mask2Former~\cite{cheng2022mask2former} instantiation of this meta-architecture.

\begin{figure*}[t]
	\centering
    	\includegraphics[width=1.\linewidth]{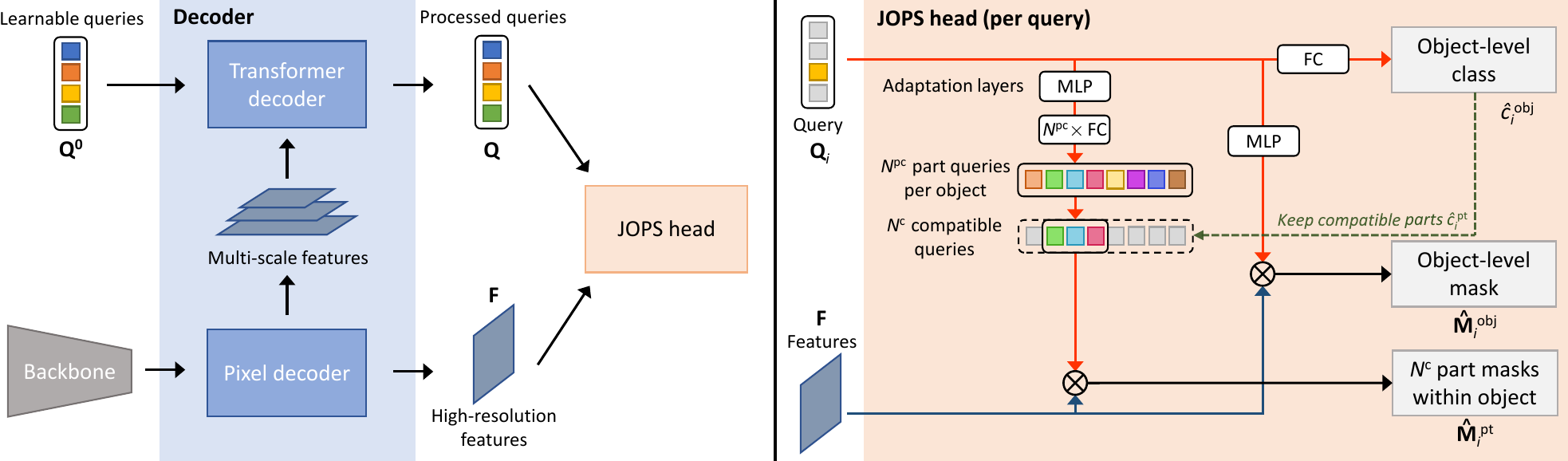}
    \vspace{-14pt}
	\caption{\textbf{Network architecture}. Left: The overall TAPPS architecture. A set of learnable queries and features from a backbone are fed into a pixel decoder and transformer decoder to generate high-resolution features and processed queries. Right: These queries and features are fed into the JOPS head, which predicts for each shared query (a) an object-level class, (b) an object-level segmentation mask, and (c) a set of part-level masks for the part-level classes compatible with the object-level class. Operator $\otimes$ denotes a matrix multiplication.}
\label{fig:overall_architecture}
\end{figure*}

\subsubsection{Mask classification for PPS}
\label{sec:method:prelims:mask_cls_pps}
With the aforementioned meta-architecture, we can conduct object-level segmentation, and thereby solve one aspect of the PPS task definition. To solve the full task, the object-level segments should also be further segmented into parts. 

Current state-of-the-art approaches~\cite{li2022ppf,li2023ppf++} achieve this by introducing an additional set of queries, the part-level queries depicted in \cref{fig:eyecatcher}\textcolor{red}{a}. Each of these part-level queries learns to represent a part-level segment. However, in contrast to the PPS task definition, these part-level segments are not explicitly linked to an object-level segment. Instead, these part-level segments represent an entire part-level class, \ie, their masks contain all pixels belonging to one part-level class across multiple object segments.

This means that the learning objective of these networks is not aligned with the PPS task objective. As a result, they are not directly optimized for PPS, but instead for two surrogate tasks: object-level panoptic segmentation and part-level semantic segmentation. In addition to necessitating post-processing, we hypothesize that this has several negative consequences:
(1) By learning parts in an `object-instance-unaware' manner, the network learns conflicting representations in the features, $\mathbf{F}$. On the one hand, different thing instances belong to different object-level segments, requiring distinct features. On the other hand, the parts of these instances belong to the same part-level segment, requiring similar features. We hypothesize that this conflict harms the ability of the network to learn to separate object-level thing instances.
(2) Separately predicting objects and parts in this way may also cause incompatible predictions. These incompatibilities make it unclear for downstream processes which prediction to trust. To obtain predictions compliant with the PPS definition, rule-based post-processing is required, which discards incompatible predictions~\cite{degeus2021pps}. This is undesired because the discarded information is potentially correct. (3) By representing object- and part-level information with separate queries, the network cannot encode complementary information about an individual object and its parts, which it could use to make more informed predictions at these abstraction levels.

\subsection{Task-aligned PPS}
\label{sec:method:jops}

To overcome the limitations of state-of-the-art methods and achieve a better PPS performance, the objective of this work is to design a \textbf{simple} PPS approach that uses the powerful mask classification framework, but that aligns its learning objective with the PPS task objective, and explicitly predicts part-level segments within individual objects. To achieve this, we should (a) generate a unique query for each object-part combination, \ie, for each part-level segment within an object-level segment, and (b) explicitly link these part-level segments to their parent objects, as PPS requires. To accomplish this, we propose Task-Aligned Part-aware Panoptic Segmentation (TAPPS), a method that jointly represents object-level segments and their parts with a set of shared queries, and generates `per-object' part queries from each shared query. Each of these queries can then be used to predict one part-level segment within an object-level segment, and is automatically linked to its parent object segment.

\subsubsection{Overall architecture}
\label{sec:method:jops:overall}

In \cref{fig:overall_architecture}, we visualize the network architecture of TAPPS. First, we initialize a set of $N^{\textrm{q}}$ learnable queries, $\mathbf{Q^0} \in \mathbb{R}^{N^{\textrm{q}} \times E}$. Each of these queries learns to represent one object-level segment, but also the part-level segments belonging to this object (\eg, \textit{car3} and also \textit{car3-wheels}, \textit{car3-windows}, etc.). Following the mask classification framework described in \cref{sec:method:prelims}, 
these learnable queries $\mathbf{Q^0}$ and the features from a backbone are fed into a decoder, which generates a set of processed queries $\mathbf{Q}$ and high-resolution features $\mathbf{F}$. Subsequently, these processed queries and features enter the Joint Object and Part Segmentation (JOPS) head. For each query $\mathbf{Q}_{i} \in \mathbb{R}^{E}$, this head predicts (a) an object-level class, (b) an object-level segmentation mask, and (c) a set of part-level segmentation masks and classes for the parts within this object-level segment.

With this approach, we solve the limitations of existing works. Following the PPS task definition, we predict part-level segments per individual object-level segment. (1) As a result, we learn object-instance-aware representations for both parts and objects, improving the ability of the network to separate instances. (2) By predicting objects and parts jointly, the compatibility between both sets of predictions greatly improves. Moreover, in the JOPS head, we ensure that we only predict compatible parts, ensuring full object-part compatibility (see \cref{sec:method:jops:head}). (3) The joint object-part representations allow the network to fully leverage the complementary information from both abstraction levels.

\subsubsection{JOPS head}
\label{sec:method:jops:head}
Within the JOPS head, we predict an object-level segment and compatible part-level segments for each shared query. For the object-level predictions, we follow Mask2Former~\cite{cheng2022mask2former}: (1) For the object-level class, we feed each processed query $\mathbf{Q}_{i}$ through a single fully connected layer to predict a score for each possible class, and obtain the predicted class $\hat{c}^{\textrm{obj}}_{i}$ by picking the highest-scoring class. (2) For the object-level mask, we feed the query $\mathbf{Q}_{i}$ through a 3-layer MLP and generate a mask by taking the product of the resulting mask queries $\mathbf{Q}^{\textrm{m}}_{i}$ and the features $\mathbf{F}$, and applying a sigmoid activation, yielding $\mathbf{\hat{M}}^{\textrm{obj}}_{i} \in {[0,1]}^{{H}\times{W}}$.

To predict part-level segments using the same features $\mathbf{F}$, we need to generate queries for each part class within an object-level segment. To do so, we first apply an MLP to each query $\mathbf{Q}_{i}$ to \textit{adapt} it for part-level segmentation, and then apply $N^{\textrm{pc}}$ different fully-connected layers, where $N^{\textrm{pc}}$ is the total number of part-level classes in the dataset. This results in $\mathbf{Q}^{\textrm{pt}}_{i} \in \mathbb{R}^{{N^{\textrm{pc}}}\times{E}}$, a set of `per-object' part-level queries, where each query always corresponds to a fixed, predetermined part class. We could then take the product of $\mathbf{Q}^{\textrm{pt}}_{i}$ and the features $\mathbf{F}$ to generate a segmentation mask for all part-level classes. However, we already have an object-level class prediction for each query, and we know that only a subset of all $N^{\textrm{pc}}$ part classes is compatible with a certain object class. Therefore, we propose to only predict and supervise part-level masks for those compatible part-level classes. This simplifies the part segmentation task that the network needs to learn, as it no longer has to learn to predict `empty' segmentation masks for incompatible part classes. Concretely, as visualized in \cref{fig:overall_architecture}, we identify the object-level class $\hat{c}^{\textrm{obj}}_{i}$ for each query $\mathbf{Q}_{i}$, and keep only the part-level queries $\mathbf{Q}^{\textrm{pt,c}}_{i} \in \mathbb{R}^{{N^{\textrm{c}}}\times{E}}$ that correspond to the $N^{\textrm{c}}$ part classes that are compatible with the object-level class. Then, we compute the product of these remaining queries $\mathbf{Q}^{\textrm{pt,c}}_{i}$ and the features $\mathbf{F}$, and apply a sigmoid activation to generate compatible part segmentation masks $\mathbf{\hat{M}}_{i}^{\textrm{pt}} \in {[0,1]}^{{N^{\textrm{c}}}\times{H}\times{W}}$. As each per-object part query corresponds to a fixed part class, we also know the part classes $\hat{c}_{i}^{\textrm{pt}}$. Note that the number $N^{\textrm{c}}$ depends on the object-level class and will therefore vary per query $\mathbf{Q}_{i}$.

\subsubsection{Training}
\label{sec:method:jops:training}

To assign each query to at most one ground-truth object-level segment with corresponding part-level segments during training, we apply bipartite matching based on the predicted and ground-truth object-level classes and masks. 

To supervise the object-level segments, we use the cross-entropy loss for the classes, and both the Dice~\cite{milletari2016dice} and cross-entropy loss for the segmentation masks. Together, these losses form the object-level loss $L_{\textrm{obj}}$. For the part-level segmentation masks, we also use the Dice and cross-entropy loss, together forming part-level loss $L_{\textrm{pt}}$. We balance the losses with weights $\lambda_{\textrm{obj}}$ and $\lambda_{\textrm{pt}}$ to calculate the total loss
\begin{equation}
    L = {\lambda_{\textrm{obj}}}{L_{\textrm{obj}}} + {\lambda_{\textrm{pt}}}{L_{\textrm{pt}}}.
\end{equation}
We find that using $\lambda_{\textrm{obj}} = \lambda_{\textrm{pt}} = 1$ provides the best balance; see the supplementary material for more details.
\section{Experimental setup}
\label{sec:experiments}

\paragraph{Datasets.} We use the two PPS benchmarks for evaluation.

\textit{Cityscapes Panoptic Parts} (Cityscapes-PP)~\cite{degeus2021pps} extends the original Cityscapes dataset~\cite{cordts2016cityscapes} with part-level labels. It consists of street scene images from several cities. It has labels for 19 object classes (11 stuff; 8 thing). Part classes are defined and labeled for 5 object-level thing classes; there are 23 part classes in total. We train on the \textit{train} split (2975 images), and evaluate on the \textit{val} split (500 images).

\textit{Pascal Panoptic Parts} (Pascal-PP)~\cite{degeus2021pps}, which combines existing labels for Pascal VOC~\cite{chen2014pascalpart, everingham2010pascal, mottaghi14pascalcontext}, consists of a wide range of scenes and classes. There are 59 object-level classes (39 stuff; 20 thing), and part-level classes are defined for 15 thing classes. Unless otherwise indicated, we use the default part class definition with 57 part classes in total~\cite{degeus2021pps}. To evaluate a more challenging setting, we additionally evaluate on Pascal-PP-107, which uses the 107 non-background classes from the Pascal-Part-108 definition introduced by Michieli~\etal \cite{michieli2020gmnet} for part segmentation. For both class definitions, we train on the \textit{training} split (4998 images), and evaluate on the \textit{validation} split (5105 images).

\paragraph{Baseline.}
Our baseline is a version of TAPPS that uses the same network architecture, but uses a separate set of 100 additional queries for part-level segmentation (see also \cref{fig:eyecatcher}\textcolor{red}{a}). Comparing the results of the baseline in \cref{tab:results:main} to existing methods in \cref{tab:results:sota}, we find that our baseline outperforms state-of-the-art approaches that also use separate part-level queries~\cite{li2023ppf++,li2022ppf}, indicating that it is a strong baseline. See the supplementary material for more details.

\begin{table*}[t]
\centering
\begin{subtable}[h]{0.49\linewidth}
    \newcolumntype{C}{>{\centering}p{0.023\textwidth}}
    \adjustbox{width=1.\linewidth}{
    \begin{tabular}{lccccccc}
    \toprule
    \multirow{2}[2]{*}{Method} & \multicolumn{3}{c}{PartPQ} & PartSQ & \multicolumn{3}{c}{PQ}  \\
    \cmidrule(lr){2-4} \cmidrule(lr){5-5} \cmidrule(lr){6-8}
    & Pt & No pt & All & Pt & Th & St & All \\
    \midrule
    \multicolumn{8}{c}{\textit{ImageNet pre-training}} \\
    \midrule
    Baseline
    & 55.2 & 38.8 & 42.9 & 71.5 & 62.0 & 37.3 & 45.7 \\
    \textbf{TAPPS (ours)}
    & \textbf{59.6} & \textbf{39.4} & \textbf{44.6} & \textbf{74.3} & \textbf{65.6} & \textbf{37.6} & \textbf{47.1} \\
    \midrule
    \multicolumn{8}{c}{\textit{COCO pre-training}} \\
    \midrule
    Baseline
    & 64.8 & 49.9 & 53.7 & 73.1 & 74.8 & 48.1 & 57.1  \\
    \textbf{TAPPS (ours)}
    & \textbf{67.2} & \textbf{50.4} & \textbf{54.7} & \textbf{75.1} & \textbf{75.7} & \textbf{48.5} & \textbf{57.7}  \\
    \bottomrule
    \end{tabular}
    }
    \caption{\textbf{Pascal-PP \textit{validation}}~\cite{chen2014pascalpart, degeus2021pps, everingham2010pascal, mottaghi14pascalcontext}.}
    \label{tab:results:main:pascal} 
\end{subtable}
\hfill
\begin{subtable}[h]{0.49\linewidth}
    \newcolumntype{C}{>{\centering}p{0.023\textwidth}}
    \adjustbox{width=1.0\linewidth}{
    \begin{tabular}{lccccccc}
    \toprule
    \multirow{2}[2]{*}{Method} & \multicolumn{3}{c}{PartPQ} & PartSQ & \multicolumn{3}{c}{PQ}  \\
    \cmidrule(lr){2-4} \cmidrule(lr){5-5} \cmidrule(lr){6-8}
    & Pt & No pt & All & Pt & Th & St & All \\
    \midrule
    \multicolumn{8}{c}{\textit{ImageNet pre-training}} \\
    \midrule
    Baseline
    & 46.6 & 62.6 & 58.4 & 66.1 & 53.8 & 67.1 & 61.5 \\
    \textbf{TAPPS (ours)}
    & \textbf{48.7} & \textbf{63.1} & \textbf{59.3} & \textbf{66.8} & \textbf{55.6} & \textbf{67.3} & \textbf{62.4} \\
    \midrule
    \multicolumn{8}{c}{\textit{COCO pre-training}} \\
    \midrule
    Baseline
    & 48.2 & 64.8 & 60.4 & 66.7 & 56.1 & 69.0 & 63.6  \\
    \textbf{TAPPS (ours)}
    & \textbf{48.9} & \textbf{65.7} & \textbf{61.3} & \textbf{66.9} & \textbf{57.2} & \textbf{69.7} & \textbf{64.4}  \\
    \bottomrule
    \end{tabular}
    }
    \caption{\textbf{Cityscapes-PP \textit{val}}~\cite{cordts2016cityscapes,degeus2021pps}.}
    \label{tab:results:main:cs}
\end{subtable}
\vspace{-5pt}

\caption{\textbf{Main results.} We compare TAPPS to a strong baseline that uses separate sets of queries to predict object- and part-level segments, instead of predicting object- and part-level segments jointly like TAPPS (see \cref{sec:experiments}).}
\label{tab:results:main}
\end{table*}

\paragraph{Evaluation metrics.}
The part-aware panoptic segmentation performance is evaluated using the default Part-aware Panoptic Quality (PartPQ) metric~\cite{degeus2021pps}. It captures both the ability to recognize and segment object-level segments (\ie, stuff regions and thing instances), and the ability to further segment the identified object-level segments into part-level masks. The PartPQ per object-level class $c$ is given by
\begin{equation}
\textrm{PartPQ}_{c} = \frac{\sum_{(p,g) \in \textit{TP}_c}\textrm{IoU\textsubscript{p}}(p,g)}{|\textit{TP}_c| + \frac{1}{2}|\textit{FP}_c|+ \frac{1}{2}|\textit{FN}_c|},
\label{equ:pq-parts}
\end{equation}
where $\textit{TP}_c$, $\textit{FP}_c$, and $\textit{FN}_c$ are the sets of true positive, false positive and false negative segments, respectively. A pair of predicted and ground-truth segments $(p,g)$ of the same class $c$ is part of $\textit{TP}_c$ if the Intersection-over-Union (IoU) of their object-level masks is larger than 0.5. If a ground-truth segment is not identified, it is part of $\textit{FN}_c$; if a prediction is incorrect, it is part of $\textit{FP}_c$. The IoU\textsubscript{p} term captures the segmentation performance within identified object-level segments. For object-level classes for which part classes are defined, it calculates the part-level mIoU. Otherwise, it uses the object-level IoU. We report the mean PartPQ over all classes, but also separately for object-level classes with parts (PartPQ\textsuperscript{Pt}) and without parts (PartPQ\textsuperscript{NoPt}).

To individually assess the ability of methods to conduct part-level segmentation within identified objects, we report the Part Segmentation Quality for object-level classes that have parts, PartSQ\textsuperscript{Pt}. Following De Geus \etal~\cite{degeus2021pps}, per object-level class $c$, the PartSQ\textsuperscript{Pt} calculates the average part segmentation mIoU within object-level segments:
\begin{equation}
    \textrm{PartSQ}^{\textrm{Pt}}_{{c}} = \frac{{\sum_{(p,g) \in \textit{TP}_c}\textrm{IoU\textsubscript{p}}(p,g)}}{|\textit{TP}_c|}.
\end{equation}

To evaluate the ability of networks to conduct object-level panoptic segmentation, we report the Panoptic Quality (PQ) metric~\cite{kirillov2019ps}. Similarly to the PartPQ, we report the average PQ over all classes, but also over all thing classes (PQ\textsuperscript{Th}) and stuff classes (PQ\textsuperscript{St}) separately.

\paragraph{Implementation details.}
TAPPS is built on top of the publicly available code of state-of-the-art panoptic segmentation network Mask2Former~\cite{cheng2022mask2former}. For all datasets, we use a batch size of 16 images, and train on 4 Nvidia A100 GPUs. To optimize TAPPS, we use AdamW~\cite{loshchilov2019adamw}, a weight decay of 0.05, and a polynomial learning rate decay schedule with an initial learning rate of $10^{-4}$ and a power of 0.9. For Cityscapes-PP, we train for 90k iterations and apply conventional data augmentation steps~\cite{cheng2022mask2former,li2022ppf}: random flip, random resize with a factor between 0.5 and 2.0, and finally a random crop of 512$\times$1024 pixels. For Pascal-PP, we train for 60k iterations in case of ImageNet pre-training, but for only 10k iterations in case of COCO pre-training, to prevent overfitting. Following state-of-the-art panoptic segmentation implementations on COCO~\cite{lin2014coco,cheng2022mask2former}, we apply large-scale jittering with a scale between 0.1 and 2.0 followed by a random crop of 1024$\times$1024 pixels. During inference, we resize the image such that the shortest side is 800 pixels. Note that we use exactly the same training and testing settings for both TAPPS and the baseline. For more implementation details, see the supplementary material.

\section{Results}
\label{sec:results}

\begin{table*}[t]
\centering
\adjustbox{width=.835\linewidth}{
\begin{tabular}{llcccccc}
\toprule
\multirow{2}[2]{*}{\text{Method}} & \multirow{2}[2]{*}{\text{Backbone}} & \multirow{2}[2]{*}{\shortstack{\text{Pre-}\\\text{training}}} & 
\multicolumn{3}{c}{\text{PartPQ}}  & \text{PartSQ} & \multicolumn{1}{c}{\text{PQ}} \\
\cmidrule(lr){4-6} \cmidrule(lr){7-7} \cmidrule{8-8}
 &  &  &
Pt & No Pt & All & Pt & All\\
\midrule

\multicolumn{8}{c}{Pascal-PP \textit{validation}} \\
\midrule
JPPF~\cite{jagadeesh2022jppf} &  EfficientNet-B5 &
I & 48.3 & 26.9 & 32.2 & -  & - \\
\textbf{TAPPS (ours)} & RN-50 & 
I & \textbf{59.6} & \textbf{39.4} & \textbf{44.6} & \textbf{74.3} & \textbf{47.1} \\

\midrule
Panoptic-PartFormer$^\dagger$~\cite{li2022ppf} &  RN-50 &
I,C & 56.1 & 38.8 & 43.2 & 66.8  & 47.6 \\
Panoptic-PartFormer++$^\dagger$~\cite{li2023ppf++} &  RN-50 &
I,C & 52.6 & 42.6 & 45.1 & 60.4 & 51.6 \\
\textbf{TAPPS (ours)} & RN-50 & 
I,C & \textbf{67.2} & \textbf{50.4} & \textbf{54.7} & \textbf{75.1} & \textbf{57.7} \\

\midrule
Panoptic-PartFormer$^\dagger$~\cite{li2022ppf} &  Swin-B &
I,C & 64.3 & 50.6 & 54.1 & 70.8 & 58.1 \\
Panoptic-PartFormer++$^\dagger$~\cite{li2023ppf++} &  Swin-B &
I,C & 48.9 & 52.1 & 51.3 & 53.0  & 59.8 \\
\textbf{TAPPS (ours)} & Swin-B & 
I,C & \textbf{72.2} & \textbf{56.3} & \textbf{60.4} & \textbf{78.1} & \textbf{63.0} \\

\midrule

\multicolumn{8}{c}{Cityscapes-PP \textit{val}} \\
\midrule

Panoptic-PartFormer~\cite{li2022ppf} &  RN-50 &
I & - & - & 54.5 & - & 57.8 \\
Panoptic-PartFormer++~\cite{li2023ppf++} &  RN-50 &
I & - & - & 57.5 & - & 61.6 \\
JPPF~\cite{jagadeesh2022jppf} &  EfficientNet-B5 &
I & 47.7 & \textbf{63.8} & \textbf{59.6} & -  & - \\
\textbf{TAPPS (ours)} & RN-50 & 
I & \textbf{48.7} & 63.1 & 59.3 & \textbf{66.8} & \textbf{62.4} \\

\midrule
Panoptic-PartFormer~\cite{li2022ppf} &  RN-50 &
I,C & 43.9 & 62.4 & 57.5 & 60.1 & 61.6\\
Panoptic-PartFormer++~\cite{li2023ppf++} &  RN-50 &
I,C & 42.5 & 65.1 & 59.2 & - & 63.6 \\
\textbf{TAPPS (ours)} & RN-50 & 
I,C & \textbf{48.9} & \textbf{65.7} & \textbf{61.3} & \textbf{66.9} & \textbf{64.4} \\
\midrule

Panoptic-PartFormer~\cite{li2022ppf} &  Swin-B &
I,C & 45.6 & 67.8 & 62.0 & 59.0 & 66.6 \\
Panoptic-PartFormer++~\cite{li2023ppf++} &  Swin-B &
I,C & 46.0 & 68.2 & 62.3 & - & 68.0 \\
Panoptic-PartFormer++~\cite{li2023ppf++} &  ConvNeXt-B &
I,C & 46.4 & \textbf{69.1} & 63.1 & -  & \textbf{68.2}  \\
SegFormer-B5 + CondInst + BPR ~\cite{tang2021bpr,tang2022virreq,tian2020condinst,xie2021segformer} &  MiT-B5 + RN-50 &
I,C* & 48.6 & 67.5 & 62.5 & -& - \\
\textbf{TAPPS (ours)} & Swin-B & 
I,C & \textbf{53.0} & 69.0 & \textbf{64.8} & \textbf{68.0} & 68.0 \\
\bottomrule

\end{tabular}
}
\vspace{-4pt}
\caption{\textbf{Comparison with state of the art}. Evaluation on the Cityscapes-PP and Pascal-PP benchmarks~\cite{cordts2016cityscapes, degeus2021pps, everingham2010pascal}. RN-50 is ResNet-50~\cite{he2016resnet}. Other backbones are EfficientNet-B5~\cite{tan2019efficientnet}, MiT-B5~\cite{xie2021segformer}, Swin-B~\cite{liu2021swin} and ConvNeXt-B~\cite{liu2022convnext}. I = ImageNet~\cite{russakovsky2015imagenet}, C = COCO panoptic~\cite{lin2014coco}, C* = COCO pre-training for instance segmentation. $^\dagger$PartPQ scores for these existing methods have been re-evaluated using official code~\cite{degeus2021pps} and are higher than originally reported on Pascal-PP~\cite{li2022ppf,li2023ppf++}, see supplementary material for more details.}
\label{tab:results:sota}
\vspace{-6pt}
\end{table*}

\subsection{Main results}
\label{sec:results:main}
First, we compare TAPPS to the strong baseline that uses separate sets of queries for object-level panoptic segmentation and part-level semantic segmentation (see \cref{sec:experiments}). The results in \cref{tab:results:main:pascal} demonstrate that TAPPS significantly outperforms the baseline on Pascal-PP in several aspects. Most importantly, the PartPQ for object-level classes with parts (PartPQ\textsuperscript{Pt}) is +4.4 or +2.4 higher, depending on the pre-training strategy. Looking in more detail, we find that this increase is caused by two individual improvements: (1) The part-level segmentation quality (PartSQ\textsuperscript{Pt}) is considerably higher than for the baseline. This shows the positive impact of having a joint representation for objects and parts, and simplifying the part segmentation task by only allowing compatible predictions (see also \cref{sec:results:ablations}). (2) The panoptic quality for thing classes (PQ\textsuperscript{Th}) also sees a substantial increase. Here, it should be noted that (a) all the object-level classes with parts are thing classes, and (b) thing classes require instance separation. Thus, this improvement does not only show the benefit of learning objects and parts jointly, but also indicates that learning parts in an object-instance-aware manner indeed improves the ability of the network to separate object instances (see also \cref{sec:results:additional}). 

For Cityscapes-PP, we see similar results in \cref{tab:results:main:cs}. Again, we observe improvements on the PartPQ\textsuperscript{Pt}, PartSQ\textsuperscript{Pt} and PQ\textsuperscript{Th} metrics. In this case, the absolute improvements are slightly smaller. This is expected because this dataset contains significantly fewer classes and very similar images, making it easier to learn the PPS task and limiting the potential gains that can still be obtained by TAPPS.

\subsection{Comparison with state of the art}
\label{sec:results:sota}
In \cref{tab:results:sota}, we compare TAPPS to existing state-of-the-art methods across different datasets, backbones, and pre-training settings. On both Pascal-PP and Cityscapes-PP, TAPPS significantly outperforms existing work. Most importantly, it consistently scores higher on the PartPQ, PartPQ\textsuperscript{Pt}, and PartSQ\textsuperscript{Pt} metrics. TAPPS is only slightly outperformed by JPPF~\cite{jagadeesh2022jppf} on Cityscapes-PP with ImageNet pre-training, but we note that JPPF uses the EfficientNet-B5~\cite{tan2019efficientnet} backbone, which is much more powerful than the ResNet-50~\cite{he2016resnet} used by TAPPS. Moreover, we do outperform JPPF by a large margin on the Pascal-PP dataset, showing the strength of TAPPS on more complex datasets. Overall, we achieve new state-of-the-art results on both datasets, obtaining PartPQ scores of 60.4 and 64.8 on Pascal-PP and Cityscapes-PP, improvements of +6.3 and +1.7, respectively. In the supplementary material, we compare qualitative examples of TAPPS, our baseline, and existing approaches, and we also show typical failure cases.

\subsection{Ablations}
\label{sec:results:ablations}

\begin{table}[t]
\centering
\adjustbox{width=1.0\linewidth}{
\begin{tabular}{cccccccc}
\toprule
\multicolumn{2}{c}{Predict compat.~parts}
 & \multicolumn{3}{c}{PartPQ} & PartSQ & PQ & 
\multirow{2}[2]{*}{\shortstack[c]{Obj. w/o \\ conflicts}} \\
\cmidrule(lr){1-2} \cmidrule(lr){3-5} \cmidrule(lr){6-6} \cmidrule{7-7}
 Training & Testing & Pt & No Pt & All & Pt & All & \\
\midrule
\multicolumn{8}{c}{\textit{ImageNet pre-training}} \\
\midrule
\multicolumn{2}{l}{\textit{Baseline}} & \textit{55.2} & \textit{38.8} & \textit{42.9} & \textit{71.5} & \textit{45.7} & \textit{66.1\%} \\
\xmark & \xmark & 57.9 & 39.2 & 43.9 & 72.9 & 46.6 & 99.7\% \\
\xmark & \cmark &
58.0 & 39.2 & 43.9 & 72.9 & 46.6 & \textbf{100.0\%}\\
 \rowcolor{gray!30}
\cmark & \cmark &
\textbf{59.6} & \textbf{39.4} & \textbf{44.6} & \textbf{74.3} & \textbf{47.1} & \textbf{100.0\%} \\
\midrule

\multicolumn{8}{c}{\textit{COCO pre-training}} \\
\midrule
\multicolumn{2}{l}{\textit{Baseline}} & \textit{64.8} & \textit{49.9} & \textit{53.7} & \textit{73.1} & \textit{57.1} & \textit{64.6\%} \\
\xmark & \xmark & 66.3 & \textbf{50.5} & 54.6 & 74.0 & \textbf{57.9} & 99.5\% \\
\xmark & \cmark &
66.4 & \textbf{50.5} & 54.6 & 74.0 & \textbf{57.9} & \textbf{100.0\%} \\
 \rowcolor{gray!30}
\cmark & \cmark &
\textbf{67.2} & 50.4 & \textbf{54.7} & \textbf{75.1} & 57.7 & \textbf{100.0\%} \\

\bottomrule
\end{tabular}
}
\vspace{-5pt}
\caption{\textbf{Predicting only compatible part classes during training and testing.} Evaluated on Pascal-PP~\cite{chen2014pascalpart, degeus2021pps, everingham2010pascal, mottaghi14pascalcontext}.}
\label{tab:results:compatible_only}

\vspace{-6pt}

\end{table}

\quad \textbf{Predicting only compatible parts.} \quad
Using shared queries for object- and part-level segmentation enables TAPPS to only allow part segmentation predictions that are compatible with the query's object-level class (see \cref{sec:method:jops:head}). In \cref{tab:results:compatible_only}, we evaluate the effect of predicting only compatible parts during both training and testing. In addition to the main metrics, we also assess the percentage of predicted objects for which there are \textit{no} conflicting part-level predictions.
The results show that, even when allowing incompatible predictions, TAPPS has significantly fewer object-part conflicts than the baseline. Naturally, when predicting only compatible parts during testing, all object-part conflicts are removed, eliminating the need for post-processing. However, this does not yield a big improvement in terms of the segmentation quality. When we also apply this during training, we do observe a part segmentation quality (PartSQ\textsuperscript{Pt}) improvement. This shows that simplifying the part segmentation task during training leads to improved performance, as we hypothesized in \cref{sec:method:jops:head}.

\begin{table}[t]
\centering
\adjustbox{width=0.85\linewidth}{
\begin{tabular}{cccccc}
\toprule
\multirow{2}[2]{*}{\shortstack{Adaptation layers}} & \multicolumn{3}{c}{PartPQ} & PartSQ & PQ\\
\cmidrule(lr){2-4} \cmidrule(lr){5-5}\cmidrule(lr){6-6}
& Pt & No Pt & All & Pt & All\\
\midrule
- & 
67.3 & 49.6 & 54.1 & 75.2 & 57.1 \\
1$\times$ FC &  
67.1 & 50.3 & 54.6 & 75.1 & 57.6 \\
 \rowcolor{gray!30}
2-layer MLP &  
67.2 & 50.4 & 54.7 & 75.1 & 57.7 \\
3-layer MLP &  
67.1 & 50.4 & 54.6 & 75.1 & 57.7 \\
\bottomrule
\end{tabular}
}
\vspace{-5pt}
\caption{\textbf{JOPS head architecture (\cref{fig:overall_architecture}).} Evaluated on Pascal-PP~\cite{chen2014pascalpart, degeus2021pps, everingham2010pascal, mottaghi14pascalcontext}, with pre-training on COCO panoptic~\cite{lin2014coco}.}
\label{tab:results:part_heads}
\end{table}
\textbf{JOPS head architecture.} \quad
In \cref{tab:results:part_heads}, we evaluate the impact of using different numbers of adaptation layers in the JOPS head \textit{before} applying the $N^{\textrm{pc}}$ fully-connected (FC) layers to generate the part queries (see \cref{fig:overall_architecture}). We find that one or two adaptation layers are necessary to generate object- and part-level representations that are sufficiently distinct to perform their respective tasks accurately; adding any more layers does not yield further improvements.

\subsection{Additional analyses}
\label{sec:results:additional}

\begin{table}[t]
\centering
    \begin{subtable}[h]{0.49\linewidth}
        \centering
        \adjustbox{width=1.\linewidth}{
        \begin{tabular}{lcc}
        \toprule
        Method & 
        mIoU\textsuperscript{Th} &
        PQ\textsuperscript{Th}
        \\
        \midrule
        Baseline & 69.8 & 62.0 \\
        \rowcolor{gray!30}
        TAPPS (ours) & 69.9 & 65.6 \\
        \bottomrule 
        \end{tabular}
        }
        \caption{\textbf{Pascal-PP~\cite{everingham2010pascal,degeus2021pps,chen2014pascalpart,mottaghi14pascalcontext}.}}
    \end{subtable}
    \hfill
    \begin{subtable}[h]{0.49\linewidth}
        \centering
        \adjustbox{width=1.\linewidth}{
        \begin{tabular}{lcc}
        \toprule
        Method & 
        mIoU\textsuperscript{Th} &
        PQ\textsuperscript{Th}
        \\  
        \midrule
        Baseline & 77.6 & 53.8 \\
        \rowcolor{gray!30}
        TAPPS (ours) & 78.2 & 55.6 \\
        \bottomrule
        \end{tabular}
        }
        \caption{\textbf{Cityscapes-PP~\cite{cordts2016cityscapes,degeus2021pps}.}}
    \end{subtable}
\vspace{-8pt}
\caption{\textbf{Performance for \textit{things}.} ImageNet pre-training~\cite{russakovsky2015imagenet}.}
\label{tab:results:things}
\vspace{-5pt}
\end{table}

\quad \textbf{Instance separability.} \quad
\cref{tab:results:main} showed that TAPPS improves the PQ\textsuperscript{Th}, \ie, the ability to recognize, segment and classify object instances. To assess if this is due to better object recognition or improved instance separability, we group all thing instances of the same object-level class together and evaluate the instance-agnostic semantic segmentation performance with the mIoU\textsuperscript{Th}. If the PQ\textsuperscript{Th} improvement were due to improved recognition, we expect the mIoU\textsuperscript{Th} to improve too. However, the results in \cref{tab:results:things} show only a minor mIoU\textsuperscript{Th} improvement. This indicates that the PQ\textsuperscript{Th} gain is not due to better recognition, but mainly results from a better ability to separate instances, showing the benefit of learning parts in an object-instance-aware manner.

\textbf{Performance on Pascal-PP-107.} \quad
To assess the performance of TAPPS in a more complex setting, we evaluate it on Pascal-PP-107, which has 107 part-level classes instead of 57. The results on this dataset, reported in \cref{tab:results:pascal107}, show once more that TAPPS consistently improves the part segmentation and thing segmentation performance with respect to the baseline, in this case leading to a PartPQ\textsuperscript{Pt} improvement of +4.1 or +2.2, depending on the pre-training strategy. This demonstrates that TAPPS is also effective when the PPS task becomes more complex.  

\begin{table}[t]
\centering
\adjustbox{width=1.0\linewidth}{
\begin{tabular}{lccccccc}
\toprule
\multirow{2}[2]{*}{Method} & \multicolumn{3}{c}{PartPQ} & PartSQ & \multicolumn{3}{c}{PQ}  \\
\cmidrule(lr){2-4} \cmidrule(lr){5-5} \cmidrule(lr){6-8}
& Pt & No Pt & All & Pt & Th & St & All \\
\midrule
\multicolumn{8}{c}{\textit{ImageNet pre-training}} \\
\midrule
Baseline
& 45.7 & 38.9 & 40.6 & 59.3 & 62.1 & 37.6 & 45.9 \\
\textbf{TAPPS (ours)}
& \textbf{49.8} & \textbf{39.3} & \textbf{42.0} & \textbf{61.7} & \textbf{65.8} & \textbf{37.7} & \textbf{47.2} \\
\midrule
\multicolumn{8}{c}{\textit{COCO pre-training}} \\
\midrule
Baseline
& 53.2 & 49.8 & 50.7 & 59.9 & 75.2 & \textbf{48.1} & 57.2  \\
\textbf{TAPPS (ours)}
& \textbf{55.4} & \textbf{50.0} & \textbf{51.4} & \textbf{62.1} & \textbf{75.6} & \textbf{48.1} & \textbf{57.4}  \\
\bottomrule
\end{tabular}
}
\vspace{-5pt}
\caption{\textbf{107 part-level categories.} Results on the more challenging Pascal-PP-107 dataset~\cite{everingham2010pascal,chen2014pascalpart,mottaghi14pascalcontext,michieli2020gmnet,degeus2021pps}.}
\label{tab:results:pascal107}
\end{table}

\begin{table}[t]
\centering
\adjustbox{width=.75\linewidth}{
\begin{tabular}{lccccccc}
\toprule
\multirow{2}[2]{*}{\shortstack[l]{Method}} & \multicolumn{3}{c}{PartPQ} & PartSQ & PQ \\
\cmidrule(lr){2-4} \cmidrule(lr){5-5} \cmidrule{6-6}
& Pt & No Pt & All & Pt & All \\
\midrule
Dynamic
& 65.1 & \textbf{50.5} & 54.2 & 73.1 & \textbf{57.7} \\
 \rowcolor{gray!30}
Fixed
& \textbf{67.2} & 50.4 & \textbf{54.7} & \textbf{75.1} & \textbf{57.7} \\
\midrule
\end{tabular}
}
\vspace{-5pt}
\caption{\textbf{Fixed or dynamic part segmentation.} Evaluated on Pascal-PP~\cite{chen2014pascalpart, degeus2021pps, everingham2010pascal, mottaghi14pascalcontext}, with pre-training on COCO panoptic~\cite{lin2014coco}.}
\label{tab:results:fixed_vs_dynamic}
\vspace{-9pt}
\end{table}
\textbf{Fixed vs.~dynamic part segmentation.} \quad As explained in \cref{sec:method:jops:head}, TAPPS uses a fixed fully connected layer for each part class to generate the corresponding `per-object' part query. Alternatively, it is possible to predict part masks and classes dynamically within each object segment, using a set of dynamic queries like we do for object-level segmentation. In \cref{tab:results:fixed_vs_dynamic}, we compare our \textit{fixed} part segmentation setting with this \textit{dynamic} approach, which is explained in more detail in the supplementary material. We find that our fixed approach results in a better PartSQ\textsuperscript{Pt} and therefore PartPQ\textsuperscript{Pt} performance. We hypothesize that the dynamic approach performs worse because it makes the part segmentation task unnecessarily complex, as the network needs to learn to assign segments to queries dynamically, and additionally predict a class label.

\section{Conclusion}
\label{sec:discussion}

With experiments, we have shown that TAPPS considerably outperforms methods that predict objects and parts separately, by improving the object instance separability, part segmentation quality, and object-part compatibility. Importantly, these improvements can be attributed to the fact that TAPPS is directly optimized for the PPS task, using a set of shared queries to jointly predict objects and corresponding parts. With our promising findings, we hope to inspire future research towards even more complete scene understanding, \eg, image segmentation at even more abstraction levels, potentially with more flexible class hierarchies.

\vspace{-10pt}

{
\begin{spacing}{1.0} 
\small \paragraph{Acknowledgements.} 
This work is supported by Eindhoven Engine, NXP Semiconductors, and Brainport Eindhoven. This work made use of the Dutch national e-infrastructure with the support of the SURF Cooperative using grant no.~EINF-5302, which is financed by the Dutch Research Council (NWO).
\end{spacing}
}

\appendix

\section*{Appendix}
\label{supp:sec:overview}

In the appendix, we provide the following additional material:
\begin{itemize}
    \item In \cref{supp:sec:ablations}, we present the results of additional experiments, in which we evaluate the effect of different loss weights and data augmentation techniques, and assess the efficiency of TAPPS and other approaches.
    \item In \cref{supp:sec:exp_setup}, we provide more details of the experimental setup, including the implementation details of TAPPS and the strong baseline. 
    \item In \cref{supp:sec:qual_results}, we show examples of predictions by TAPPS on the Pascal-PP and Cityscapes-PP datasets, and qualitatively compare TAPPS to the strong baseline and existing work.
\end{itemize}
\section{Additional experiments}
\label{supp:sec:ablations}

\paragraph{Loss weights.}
\label{supp:sec:ablations:loss}
In \cref{supp:tab:results:loss_ablations}, we show the impact of using different loss weights to balance the losses for object-level segmentation and part-level segmentation, using the weights $\lambda_{\textrm{obj}}$ and $\lambda_{\textrm{pt}}$ (see Eq.~1 of the main manuscript). We find that balancing the losses with $\lambda_{\textrm{obj}}=\lambda_{\textrm{pt}}=1.0$ yields the best performance. As expected, the object-level segmentation performance, reflected in the PQ metric, drops when $\lambda_{\textrm{obj}}$ is decreased. Conversely, the part-level segmentation performance, reflected in the PartSQ\textsuperscript{Pt} metric, drops when $\lambda_{\textrm{pt}}$ is decreased.

\paragraph{Data augmentation techniques.}
\label{supp:sec:ablations:data_aug}
As explained in \cref{supp:sec:exp_setup:impl_details} of this document, we use large-scale jittering data augmentation for our experiments on Pascal-PP. However, the existing works Panoptic-PartFormer~\cite{li2022ppf} and Panoptic-PartFormer++~\cite{li2023ppf++} use less aggressive data augmentation techniques during training. To show that the difference in data augmentation techniques is not the main reason that TAPPS outperforms these methods, we train TAPPS with the same data augmentation techniques that are used by these methods, according to the official code repository of Panoptic-PartFormer. Specifically, we apply a random horizontal flip, and then directly resize the image such that the smallest side is 800 pixels. The results in \cref{supp:tab:results:ppf_data_aug} show that TAPPS still significantly outperforms both existing methods when using these data augmentation techniques. This shows that the improvement by TAPPS is mainly caused by the methodology and network architecture, and not by the data augmentation differences. Finally, we note that there are no differences in data augmentation techniques between our method and existing works on the Cityscapes-PP dataset, so these results in Tab.~2 of the main manuscript are directly comparable.

\begin{table}[t]
\centering
\adjustbox{width=0.85\linewidth}{
\begin{tabular}{ccccccc}
\toprule
\multirow{2}[2]{*}{$\lambda_{\textrm{pt}}$} & \multirow{2}[2]{*}{$\lambda_{\textrm{obj}}$} & \multicolumn{3}{c}{PartPQ} & PartSQ & \multicolumn{1}{c}{PQ}\\
\cmidrule(lr){3-5} \cmidrule(lr){6-6} \cmidrule(lr){7-7}
 & & Pt & No pt & All & Pt & All\\
\midrule
\rowcolor{gray!30}
1.0 & 1.0
& 67.2 & 50.4 & 54.7 & 75.1 & 57.7 \\
\midrule
0.5 & 1.0
& 67.0 & 50.1 & 54.4 & 75.0 & 57.5 \\
0.2 & 1.0
& 65.9 & 50.1 & 54.1 & 73.6 & 57.5 \\
0.1 & 1.0
& 64.9 & 50.2 & 53.9 & 72.0 & 57.6 \\
\midrule
1.0 & 0.5
& 67.3 & 50.2 & 54.5 & 75.5 & 57.4 \\
1.0 & 0.2
& 66.6 & 49.1 & 53.5 & 75.5 & 56.4 \\
1.0 & 0.1
& 66.3 & 48.0 & 52.7 & 75.6 &  55.5 \\
\bottomrule
\end{tabular}
}
\caption{\textbf{Loss weights.} Evaluated on Pascal-PP~\cite{chen2014pascalpart, degeus2021pps, everingham2010pascal, mottaghi14pascalcontext}, with pre-training on COCO panoptic segmentation~\cite{lin2014coco}.}
\label{supp:tab:results:loss_ablations}
\end{table}

\begin{table}[t]
\centering
\adjustbox{width=1.\linewidth}{
\begin{tabular}{lcccccc}
\toprule
\multirow{2}[2]{*}{Method} & \multirow{2}[2]{*}{Data augmentation} & \multicolumn{3}{c}{PartPQ} & PartSQ & \multicolumn{1}{c}{PQ}\\
\cmidrule(lr){3-5} \cmidrule(lr){6-6} \cmidrule(lr){7-7}
 & & Pt & No pt & All & Pt & All\\
\midrule
\textit{TAPPS (ours)} & \textit{Default}
& \textit{67.2} & \textit{50.4} & \textit{54.7} & \textit{75.1} & \textit{57.7} \\
\midrule
TAPPS (ours) & Flip \& resize~\cite{li2022ppf}
& \textbf{65.8} & \textbf{47.2} & \textbf{51.9} & \textbf{74.8} & \textbf{54.9} \\
Panoptic-PartFormer~\cite{li2022ppf} & Flip \& resize~\cite{li2022ppf}
& 56.1 & 38.8 & 43.2 & 66.8 & 47.6 \\
Panoptic-PartFormer++~\cite{li2023ppf++} & Flip \& resize~\cite{li2022ppf}
& 52.6 & 42.6 & 45.1 & 60.4 & 51.6 \\

\bottomrule
\end{tabular}
}
\caption{\textbf{Comparison with state-of-the-art methods using identical data augmentation techniques.} Evaluated on Pascal-PP~\cite{chen2014pascalpart, degeus2021pps, everingham2010pascal, mottaghi14pascalcontext}, with pre-training on COCO panoptic segmentation~\cite{lin2014coco}. For the default data augmentation techniques of TAPPS, see \cref{supp:sec:exp_setup:impl_details:general}.}
\label{supp:tab:results:ppf_data_aug}
\end{table}

\begin{table}[t]
\begin{subtable}[h]{1.\linewidth}
\centering
\small
\adjustbox{width=1.0\linewidth}{
\begin{tabular}{lccc}
\toprule
Method & Inference speed & Memory & \# Params \\
\midrule
Baseline & 14.9 fps & 2.1 GB & 44M \\
TAPPS (predict all parts) & 12.7 fps & 7.8 GB & 48M \\
\rowcolor{gray!30}
TAPPS (predict compat.~parts) & 16.1 fps & 1.7 GB & 48M \\
\bottomrule
\end{tabular}
}
\caption{\small \textbf{Different versions of TAPPS,} with ResNet-50 backbone.}
\label{supp:tab:efficiency_vs_tapps}
\end{subtable}
\vspace{5pt}

\begin{subtable}[h]{1.\linewidth}
\centering
\small
\adjustbox{width=1.0\linewidth}{
\begin{tabular}{lccc}
\toprule
Method & Inference speed & \# Params & PartPQ \\
\midrule
Panoptic-PartFormer~\cite{li2022ppf} & 11.5 fps & 40M & 43.2\\
\rowcolor{gray!30}
TAPPS (ours) & 16.1 fps & 48M & 54.7 \\
\bottomrule
\end{tabular}
}
\caption{\small \textbf{TAPPS vs.~existing work,}  with ResNet-50 backbone.}
\label{supp:tab:efficiency_vs_sota}
\end{subtable}
\label{supp:tab:efficiency}
\caption{\textbf{Efficiency.} 
We evaluate the average inference speed in frames per second (fps) and the maximum required GPU memory on the Pascal-PP \textit{val} set~\cite{chen2014pascalpart,degeus2021pps,everingham2010pascal,mottaghi14pascalcontext}, using an Nvidia A100 GPU.}
\end{table}

\paragraph{Efficiency.}
\label{supp:sec:ablations:efficiency}
In \cref{supp:tab:efficiency_vs_tapps}, we show the effect of predicting \textit{only compatible parts} on the model's efficiency. Most importantly, we observe that the default version of TAPPS, which only predicts the masks for the $N^{\textrm{c}}$ \textit{compatible} parts, is much more efficient than the version that predicts masks for \textit{all} $N^{\textrm{pc}}$ part classes, in terms of both inference speed and memory. Moreover, by only considering compatible parts, TAPPS is also more efficient than the baseline that uses \textit{separate} object-level and part-level queries. Another reason that TAPPS is more efficient than the baseline is that its part-level queries do not participate in the self-attention and cross-attention operations in the decoder, unlike those of the baseline. This shows the strength of the simplicity of TAPPS. 

In \cref{supp:tab:efficiency_vs_sota}, we compare TAPPS to existing method Panoptic-PartFormer~\cite{li2022ppf}. The results show that TAPPS obtains a considerably better PartPQ, while using only 8M more parameters, and even being much faster. 
\section{Experimental setup}
\label{supp:sec:exp_setup}

In this section, we provide further details about the experimental setup. First, in \cref{supp:sec:exp_setup:impl_details}, we describe the implementation details more extensively, for TAPPS, the strong baseline, and the TAPPS version that conducts dynamic part segmentation. Second, in \cref{supp:sec:exp_setup:eval_ppf}, we explain how we obtain the PartPQ scores for Panoptic-PartFormer~\cite{li2022ppf} and Panoptic-PartFormer++~\cite{li2023ppf++} on Pascal-PP.

\subsection{Implementation details}
\label{supp:sec:exp_setup:impl_details}

The most important implementation details have already been provided in Sec.~4 of the main manuscript. This subsection provides a more comprehensive overview of the implementation details.

\begin{figure*}[t]
	\centering
    	\includegraphics[width=1.\linewidth]{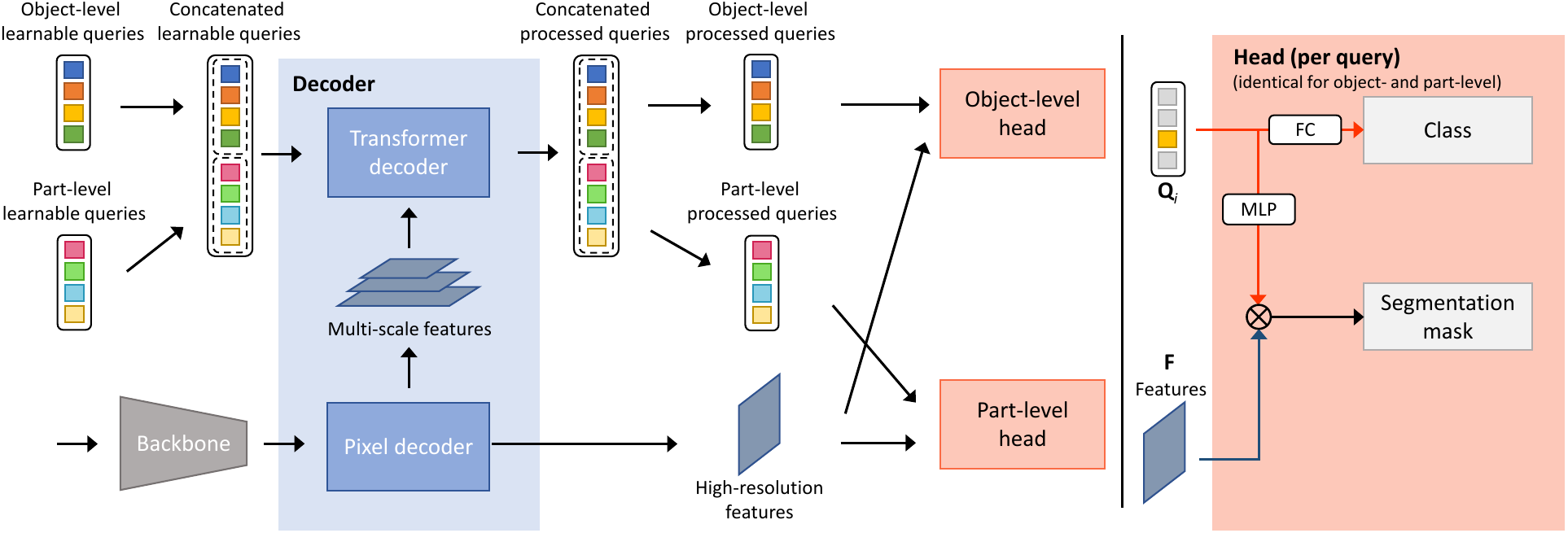}
    \vspace{-14pt}
	\caption{\textbf{Baseline network architecture.} Our strong baseline uses two separate sets of queries, one set for object-level segmentation and another set for part-level segmentation. Using these two sets of queries, this baseline network separately predicts object-level segments and object-unaware part-level segments. Operator $\otimes$ denotes a matrix multiplication.}
\label{supp:fig:baseline}
\end{figure*}

\subsubsection{General}
\label{supp:sec:exp_setup:impl_details:general}
This subsection describes the implementation details that apply to both TAPPS and the baseline. For completeness, we repeat some of the details already mentioned in the main manuscript.

Both TAPPS and the baseline are implemented on top of the publicly available code of Mask2Former~\cite{cheng2022mask2former}, which uses Detectron2~\cite{wu2019detectron2}. All experiments are conducted with a batch size of 16, using 4 Nvidia A100 GPUs in total. Following Mask2Former, we optimize all networks using AdamW~\cite{loshchilov2019adamw}, using a polynomial learning rate schedule with an initial learning rate of $10^{-4}$, a power of 0.9, and a weight decay of 0.05. When we apply ImageNet pre-training, we initialize the backbone with weights pre-trained on ImageNet-1K~\cite{russakovsky2015imagenet}. In case of COCO pre-training, we initialize both the backbone and the compatible decoder layers with weights pre-trained on COCO panoptic segmentation~\cite{lin2014coco,kirillov2019ps}; we use the weights provided in the official repository of Mask2Former. Like Mask2Former, TAPPS applies deep supervision~\cite{cheng2022mask2former}. This means that the segmentation masks and classes are predicted after each transformer layer in the decoder, and that a loss is calculated for these predictions at each of these layers. The overall loss is the sum of the total losses at all transformer layers. 

For experiments on \textit{Pascal-PP}~\cite{chen2014pascalpart,everingham2010pascal,mottaghi14pascalcontext,degeus2021pps}, we train for 60k iterations in case of ImageNet pre-training. In case of COCO pre-training, we train for 10k iterations, to avoid overfitting. Following state-of-the-art panoptic segmentation implementations on COCO~\cite{lin2014coco}, during training, we apply a random horizontal flip, followed by large-scale jittering with a scale between 0.1 and 2.0 and a random crop of 1024$\times$1024 pixels. During inference, we resize the image such that the shortest side is 800 pixels.

For experiments on \textit{Cityscapes-PP}~\cite{cordts2016cityscapes,degeus2021pps}, we train for 90k iterations for both ImageNet and COCO pre-training. We follow the conventional data augmentation steps for Cityscapes during training~\cite{li2022ppf,cheng2022mask2former}: random horizontal flip with a probability of 0.5, scaling the image with a random factor between 0.5 and 2.0, and finally a random crop of 512$\times$1024 pixels. During inference, we feed the full-resolution images of 1024$\times$2048 pixels.

\subsubsection{TAPPS}
\label{supp:sec:exp_setup:impl_details:tapps}
For TAPPS, we use $N^{\textrm{q}}=100$ shared queries. This is equal to the default number of object-level queries used by Mask2Former, because each shared query still represents only one object-level segment. Following Mask2Former, query embedding dimension $E=256$. By default, the adaptation layer in the JOPS head is an MLP with two fully connected layers with 256 input and output channels and a ReLU activation in between; see Tab.~4 of the main manuscript for ablations. 

During inference, TAPPS outputs PPS predictions without requiring rule-based post-processing, as it does not have to assign object-instance-unaware parts to individual objects, or resolve conflicts between object- and part-level predictions. Specifically, for each query, TAPPS simply outputs (a) an object-level class and mask, and (b) a part-level mask for each part-level class that is compatible with the predicted object-level class. For each pixel within the object-level mask, TAPPS keeps only the highest-scoring part mask prediction, applying \texttt{argmax}. This results in a set of compatible part-level segments that belong to an individual object-level segment. Applying this to all shared queries, we output a set of predictions that comply with the PPS task definition.

\subsubsection{Baseline}
\label{supp:sec:exp_setup:impl_details:baseline}
Like Mask2Former, our baseline uses 100 queries for object-level segmentation. Additionally, to also conduct part-level semantic segmentation, it uses 100 additional queries. In \cref{supp:fig:baseline}, we depict the network architecture for this strong baseline. As seen in this figure, both sets of queries are concatenated when entering the Transformer decoder, so there can be interaction between object-level and part-level queries through self-attention. Note that, although they are concatenated, these queries are not `mixed' or shared. The first 100 queries are still object-level queries, which learn to represent object-level segments, and the final 100 queries are part-level queries, which learn to represent object-instance-unaware part-level segments. At the end of the decoder, the queries are again split into two sets of queries, and fed into separate heads. The object-level head predicts the object-level segmentation masks and classes, resulting in object-level panoptic segmentation predictions. The part-level head predicts the part-level segmentation masks and classes, resulting in object-instance-unaware part segmentation predictions, \ie, semantic segmentation for part classes.

During training, we apply the cross-entropy loss to the object-level and part-level classes, and we use both the cross-entropy and Dice loss~\cite{milletari2016dice} for the object-level and part-level segmentation masks. The total loss is the sum of the object-level mask and classification losses, and the part-level mask and classification losses. 

During inference, because the part-level predictions are object-instance-unaware and not explicitly linked to individual objects, and because there can be incompatibilities between object- and part-level predictions, we apply the default rule-based merging strategy~\cite{degeus2021pps} used by existing work to generate the final PPS predictions.

\subsubsection{Dynamic part segmentation}
\label{supp:sec:exp_setup:impl_details:dynamic_jops}
\begin{figure}[t]
	\centering
    	\includegraphics[width=1.\linewidth]{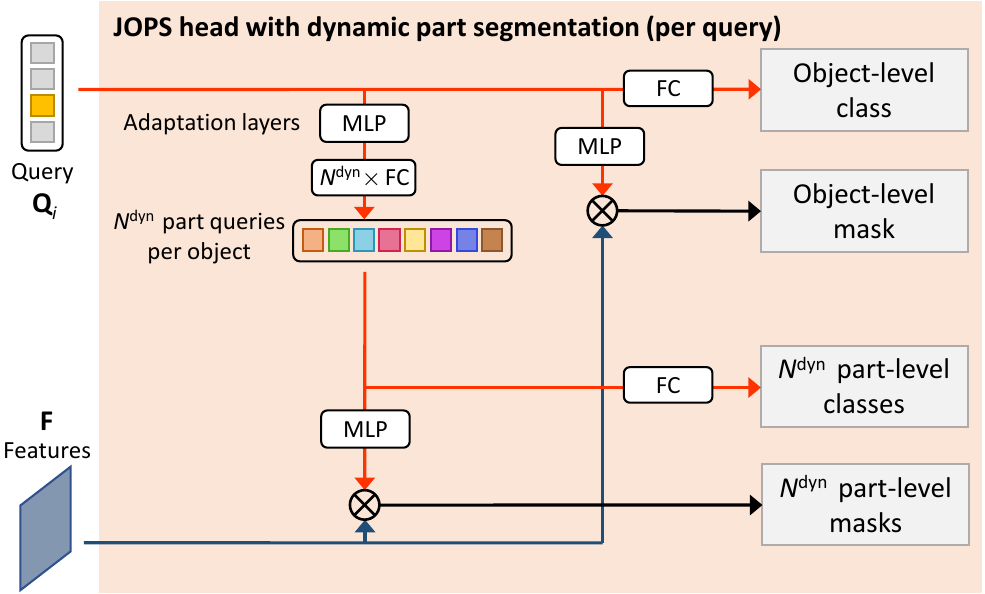}
	\caption{\textbf{Dynamic part segmentation.} When conducting dynamic part segmentation, the JOPS head uses $N^{\textrm{dyn}}$ fully-connected (FC) layers to generate $N^{\textrm{dyn}}$ per-object part queries. Each per-object part query dynamically learns to represent at most one part-level segment within an object. For each per-object part query, we predict (a) a part-level class and (b) a part-level mask.} 
\label{supp:fig:dynamic_partseg}
\vspace{-12pt}
\end{figure}

In Tab.~7 of the main manuscript, we compare our default version of TAPPS to a version that applies \textit{dynamic} part segmentation. By default, as explained in Sec.~3.3.2 of the main manuscript, we generate a set of \textit{fixed} per-object part queries in the JOPS head. That means that each per-object part query corresponds to a fixed, pre-determined part-level class, and that this query predicts a mask for this class. Alternatively, we can use a set of dynamic queries, which do not correspond to a fixed class, like we do for object-level segmentation. As depicted in \cref{supp:fig:dynamic_partseg}, for each query $\mathbf{Q}_{i}$, we apply $N^{\textrm{dyn}}$ fully-connected (FC) layers to generate a set of dynamic per-object part queries $\mathbf{Q}^{\textrm{dyn}}_{i} \in \mathbb{R}^{N^{\textrm{dyn}}\times{E}}$. As these queries are dynamic, they do not correspond to a fixed class, so for each of these queries we predict (a) a part-level class with a single fully-connected layer, and (b) a part-level segmentation mask by first applying a 3-layer MLP and then taking the product of the resulting mask queries with the features $\mathbf{F}$. We use $N^{\textrm{dyn}}=50$.

\begin{table*}[t]
    \centering
    \adjustbox{width=1.0\linewidth}{
    \begin{tabular}{llccccccc}
        \toprule
        \multirow{2}[2]{*}{Method} & \multirow{2}[2]{*}{Backbone} & \multirow{2}[2]{*}{Pre-training}  & \multicolumn{3}{c}{As originally reported~\cite{li2022ppf,li2023ppf++}} & \multicolumn{3}{c}{With official evaluation~\cite{degeus2021pps}} \\
        \cmidrule(lr){4-6} \cmidrule(lr){7-9}
        & & & PartPQ\textsuperscript{Pt} & PartPQ\textsuperscript{NoPt} & PartPQ & PartPQ\textsuperscript{Pt} & PartPQ\textsuperscript{NoPt} & PartPQ  \\
        \midrule
        Panoptic-PartFormer~\cite{li2022ppf} & ResNet-50~\cite{he2016resnet} & I,C & 
        -- & -- & 37.8 & 56.1 & 38.8 & 43.2 \\
        Panoptic-PartFormer++~\cite{li2023ppf++} & ResNet-50~\cite{he2016resnet} & I,C &
        -- & -- & 42.2 & 52.6 & 42.6 & 45.1 \\
        \midrule
        Panoptic-PartFormer~\cite{li2022ppf} & Swin-B~\cite{liu2021swin} & I,C & 
        -- & -- & 47.4 & 64.3 & 50.6 & 54.1 \\
        Panoptic-PartFormer++~\cite{li2023ppf++} & Swin-B~\cite{liu2021swin} & I,C &
        -- & -- & 49.3 & 48.9 & 52.1 & 51.3 \\
        \bottomrule
        
    \end{tabular}
    }
    \caption{\textbf{Re-evaluation of existing work on Pascal-PP~\cite{everingham2010pascal,chen2014pascalpart,mottaghi14pascalcontext,degeus2021pps}.} After discovering an evaluation bug in the official code of Panoptic-PartFormer~\cite{li2022ppf} which caused the PartPQ scores to be lower than they actually are, we re-evaluate the predictions by Panoptic-PartFormer~\cite{li2022ppf} and Panoptic-PartFormer++~\cite{li2023ppf++} using the official PPS evaluation repository~\cite{degeus2021pps}. We use these \textit{higher} correct numbers in our comparisons in the main manuscript. I = ImageNet~\cite{russakovsky2015imagenet}, C = COCO panoptic~\cite{lin2014coco} pre-training.}
    \label{supp:tab:results_ppf_eval}
\end{table*}

To supervise these part-level predictions during training, we assign each per-object part query to at most one part-level ground-truth segment using the same Hungarian matching algorithm we use for object-level segmentation~\cite{cheng2022mask2former}. This matching is applied separately within each object-level segment. If there is no matching ground-truth segment for a part query, we do not supervise the segmentation mask and supervise a `no-part' class label. The part class prediction by this dynamic version of TAPPS is supervised with a cross-entropy loss. The other losses remain the same.

During inference, the procedure is the same as for the default TAPPS. The only difference is the source of the part-level class prediction. This \textit{dynamic} version explicitly predicts it, whereas, for the default \textit{fixed} version, it is known because each per-object part query is associated with a predetermined part class.

\subsection{Evaluation of existing work}
\label{supp:sec:exp_setup:eval_ppf}
As mentioned in Tab.~2 of the main manuscript, the PartPQ scores of Panoptic-PartFormer~\cite{li2022ppf} and its extension Panoptic-PartFormer++~\cite{li2023ppf++} on Pascal-PP reported in our work are higher than the scores that these works originally reported~\cite{li2022ppf,li2023ppf++}. This is due to an evaluation bug that we discovered in the official code repository of Panoptic-PartFormer~\cite{li2022ppf}, which caused the resulting PartPQ scores to be lower than they actually are. Note that this bug only applies to Pascal-PP and not to Cityscapes-PP. We notified the authors of this bug, and they confirmed it. To assess whether this problem also occurred for Panoptic-PartFormer++, for which the code is not available, we requested the authors to send us the predictions by Panoptic-PartFormer++, so we could re-evaluate them on Pascal-PP. We are thankful that the authors have sent us these predictions. In \cref{supp:tab:results_ppf_eval}, we provide both the originally reported scores, and the scores after our re-evaluation given the official PPS evaluation repository~\cite{degeus2021pps}. For all methods, the overall PartPQ scores are higher when using the correct evaluation. Therefore, we compare against these higher values in Tab.~2 of the main manuscript.

\section{Qualitative results}
\label{supp:sec:qual_results}

\begin{figure*}[t]
\centering

\includegraphics[width=0.245\linewidth]{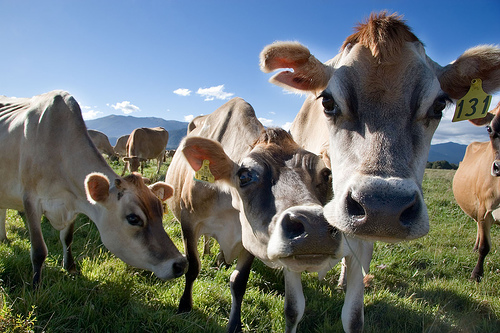}
\includegraphics[width=0.245\linewidth]{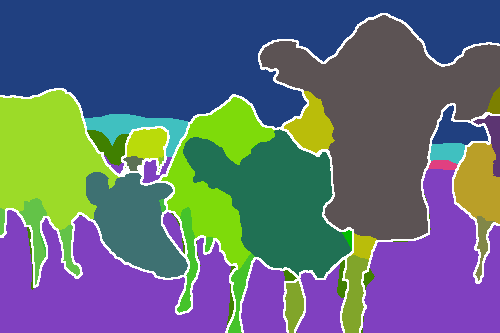}
\includegraphics[width=0.245\linewidth]{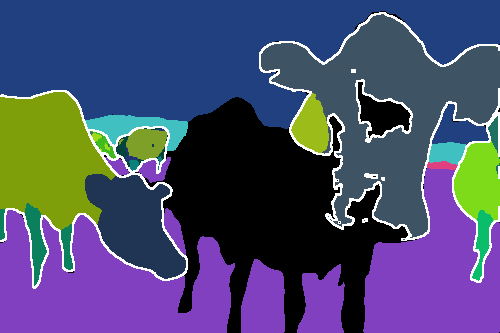}
\begin{tikzpicture}
    \node[anchor=south west,inner sep=0] (image) at (0,0) 
    {\includegraphics[width=0.245\linewidth]{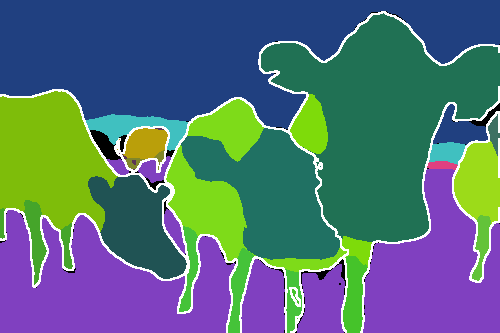}};
    \begin{scope}[x={(image.south east)},y={(image.north west)}]
    \draw[red,line width=0.5mm,rounded corners] (0.3,0.01) rectangle (0.84,0.84);
    \end{scope}
\end{tikzpicture}
\\

\includegraphics[width=0.245\linewidth]{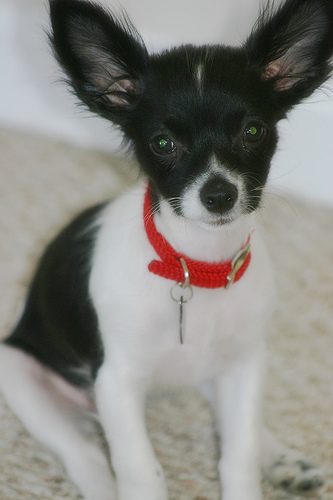}
\includegraphics[width=0.245\linewidth]{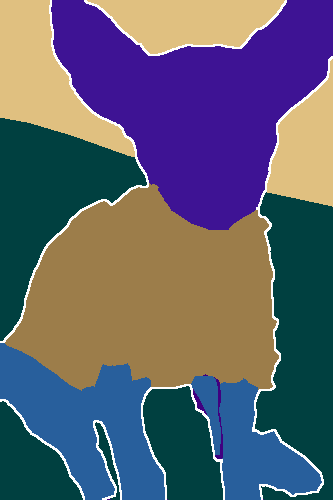}
\includegraphics[width=0.245\linewidth]{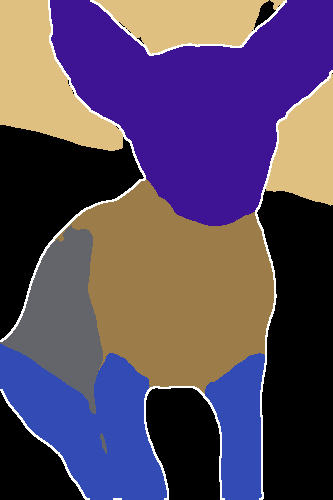}
\begin{tikzpicture}
    \node[anchor=south west,inner sep=0] (image) at (0,0) 
    {\includegraphics[width=0.245\linewidth]{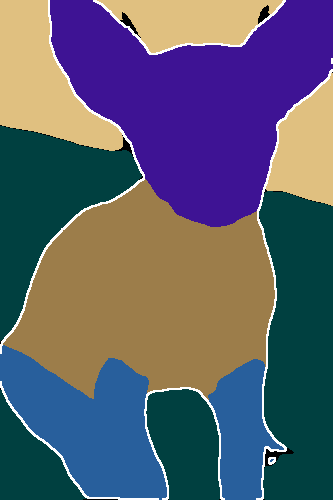}};
    \begin{scope}[x={(image.south east)},y={(image.north west)}]
    \draw[red,line width=0.5mm,rounded corners] (0.02,0.2) rectangle (0.3,0.6);
    \end{scope}
\end{tikzpicture}
\\

\includegraphics[width=0.245\linewidth]{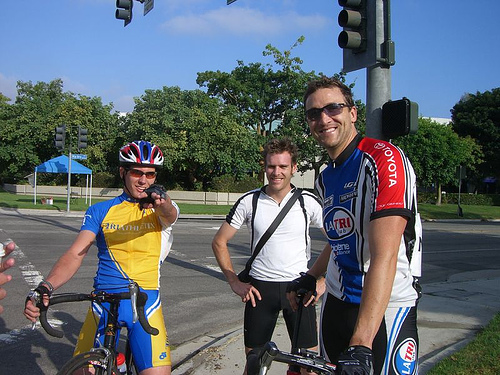}
\includegraphics[width=0.245\linewidth]{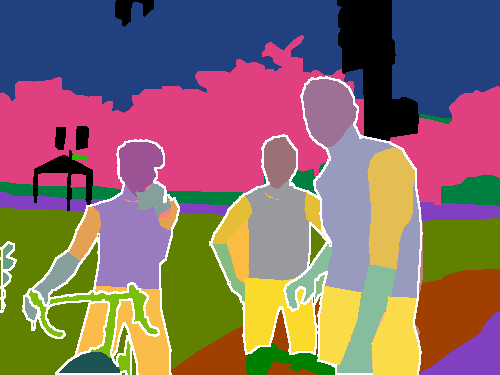}
\includegraphics[width=0.245\linewidth]{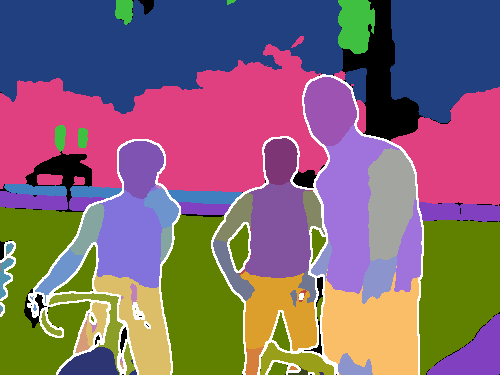}
\begin{tikzpicture}
    \node[anchor=south west,inner sep=0] (image) at (0,0) 
    {\includegraphics[width=0.245\linewidth]{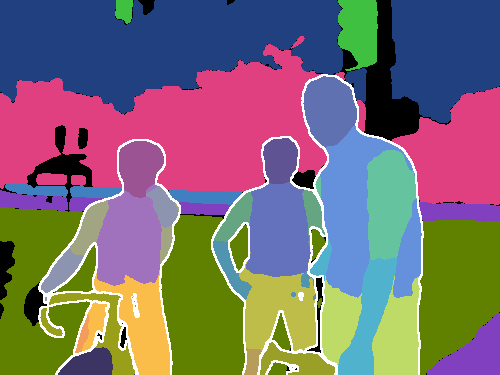}};
    \begin{scope}[x={(image.south east)},y={(image.north west)}]
    \draw[red,line width=0.5mm,rounded corners] (0.65,0.05) rectangle (0.85,0.35);
    \end{scope}
\end{tikzpicture}
\\

\includegraphics[width=0.245\linewidth, trim={0 0 4cm 0}, clip]{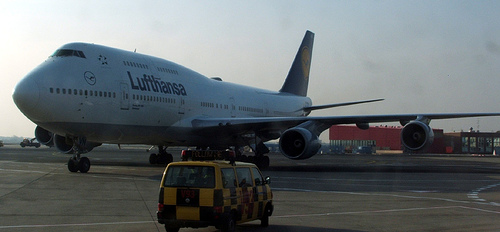}
\includegraphics[width=0.245\linewidth, trim={0 0 4cm 0}, clip]{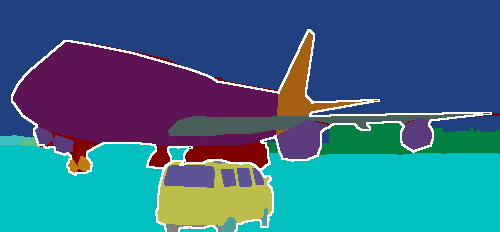}
\includegraphics[width=0.245\linewidth, trim={0 0 4cm 0}, clip]{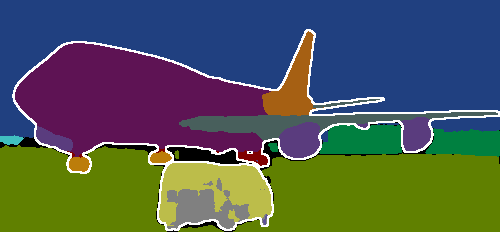}
\begin{tikzpicture}
    \node[anchor=south west,inner sep=0] (image) at (0,0) 
    {\includegraphics[width=0.245\linewidth, trim={0 0 4cm 0}, clip]{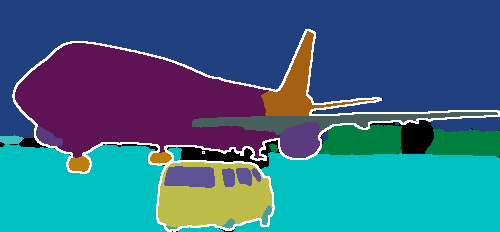}};
    \begin{scope}[x={(image.south east)},y={(image.north west)}]
    \draw[red,line width=0.5mm,rounded corners] (0.36,0.01) rectangle (0.75,0.35);
    \end{scope}
\end{tikzpicture}
\\

\includegraphics[width=0.245\linewidth]{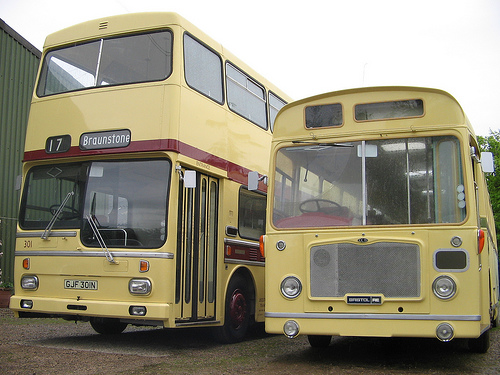}
\includegraphics[width=0.245\linewidth]{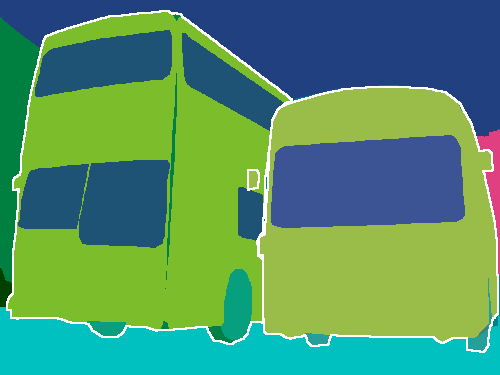}
\includegraphics[width=0.245\linewidth]{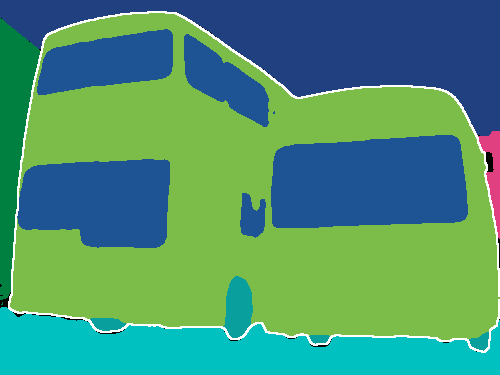}
\begin{tikzpicture}
    \node[anchor=south west,inner sep=0] (image) at (0,0) 
    {\includegraphics[width=0.245\linewidth]{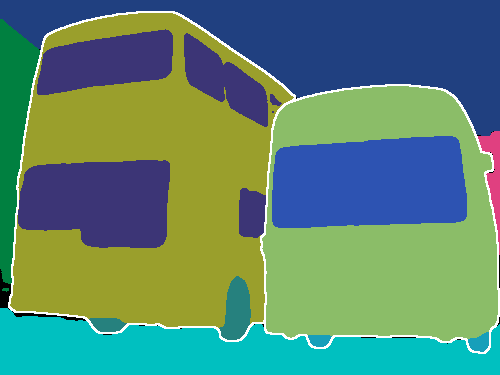}};
    \begin{scope}[x={(image.south east)},y={(image.north west)}]
    \draw[red,line width=0.5mm,rounded corners] (0.45,0.05) rectangle (0.6,0.8);
    \end{scope}
\end{tikzpicture}
\\

\begin{subfigure}[b]{0.245\textwidth}
 \centering
 \caption{Input image}
\end{subfigure}
\begin{subfigure}[b]{0.245\textwidth}
 \centering
 \caption{Ground truth}
\end{subfigure}
\begin{subfigure}[b]{0.245\textwidth}
 \centering
 \caption{Baseline}
\end{subfigure}
\begin{subfigure}[b]{0.245\textwidth}
 \centering
 \caption{TAPPS (ours)}
\end{subfigure}

\caption{\textbf{Qualitative examples of TAPPS and our strong baseline on Pascal-PP~\cite{chen2014pascalpart,everingham2010pascal,mottaghi14pascalcontext,degeus2021pps}.} Both networks use ResNet-50~\cite{he2016resnet} with COCO pre-training~\cite{lin2014coco}. White borders separate different object-level instances; color shades indicate different categories. Note that the colors of part-level categories are not identical across instances; there are different shades of the same color. In these examples, we can see how TAPPS improves both the instance separability and part segmentation quality with respect to the strong baseline. The red boxes indicate regions in which these differences are best visible. Best viewed digitally.}
\label{supp:fig:qual_results_baseline_ppp}
\end{figure*}
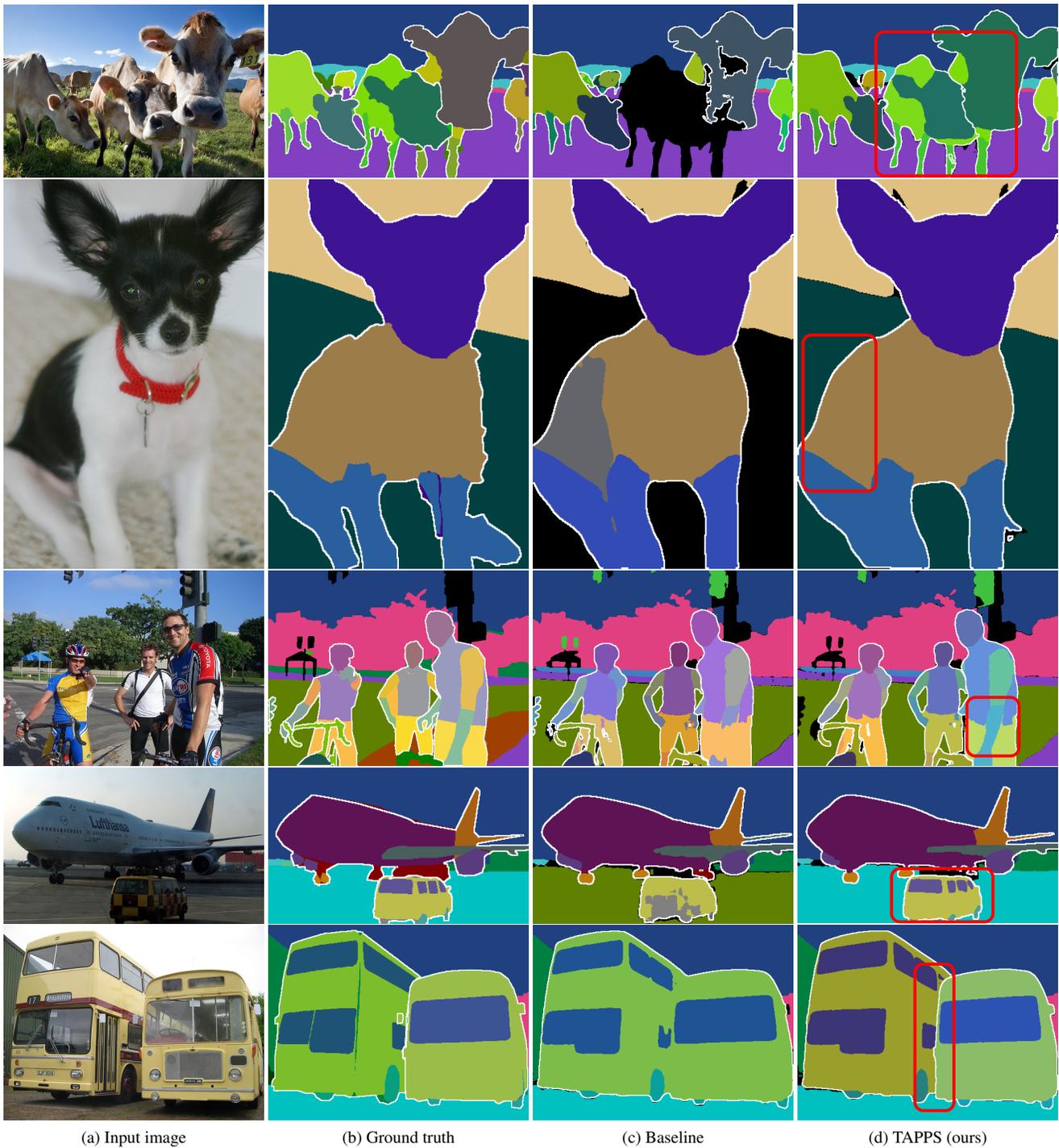

\begin{figure*}[t]
\centering

\includegraphics[width=0.245\linewidth, trim={6cm 4cm 6cm 0},clip]{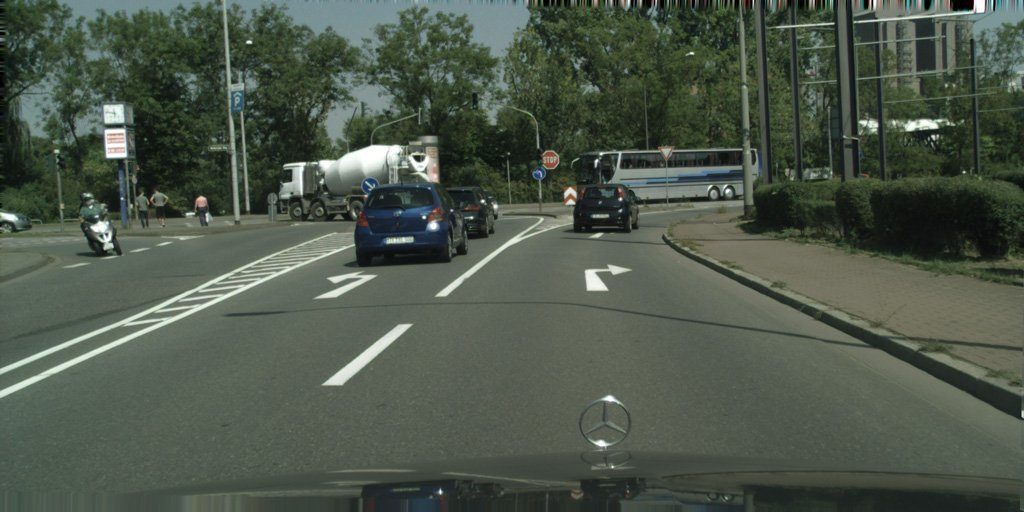}
\includegraphics[width=0.245\linewidth, trim={12cm 8cm 12cm 0},clip]{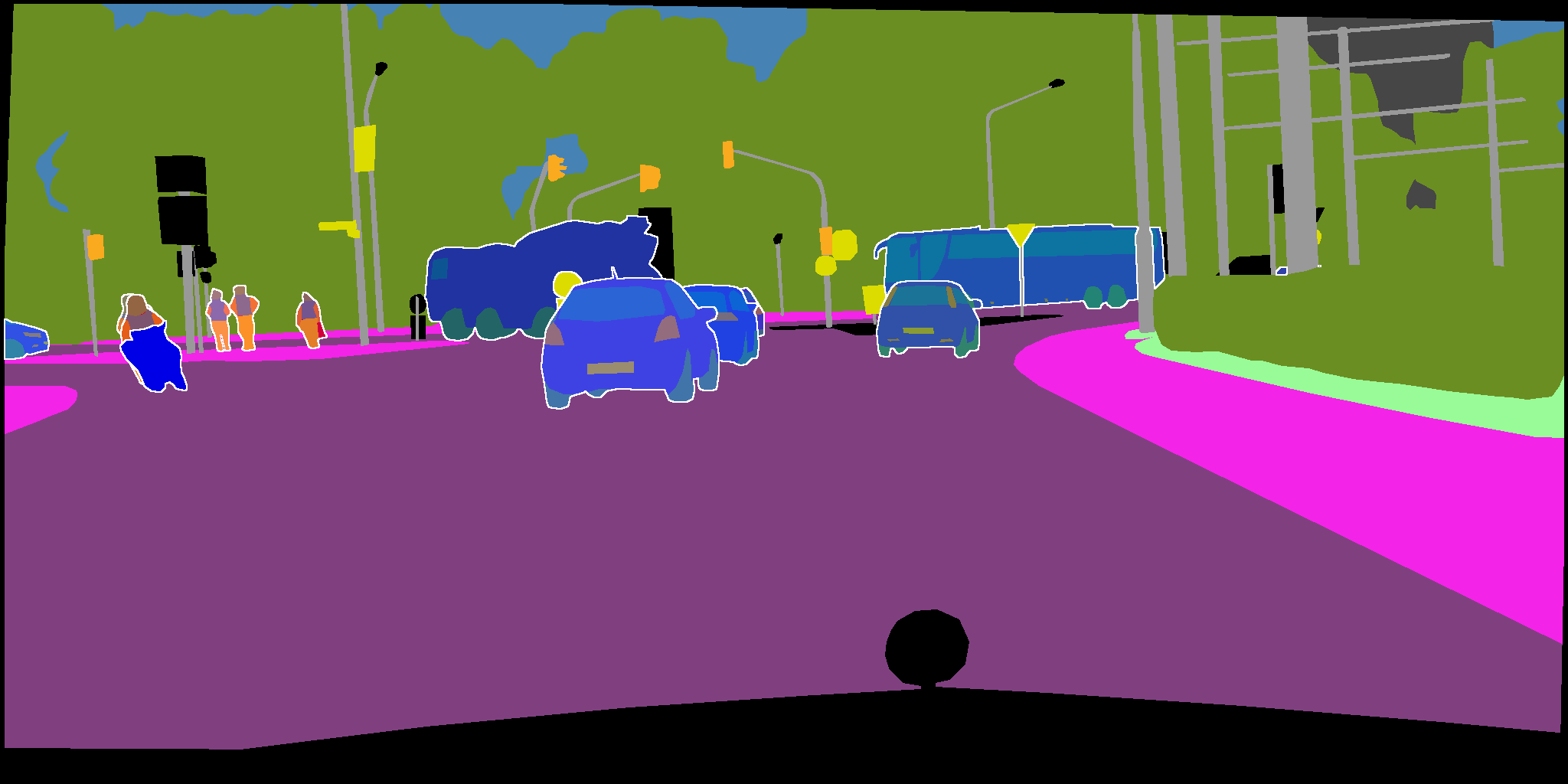}
\includegraphics[width=0.245\linewidth, trim={12cm 8cm 12cm 0},clip]{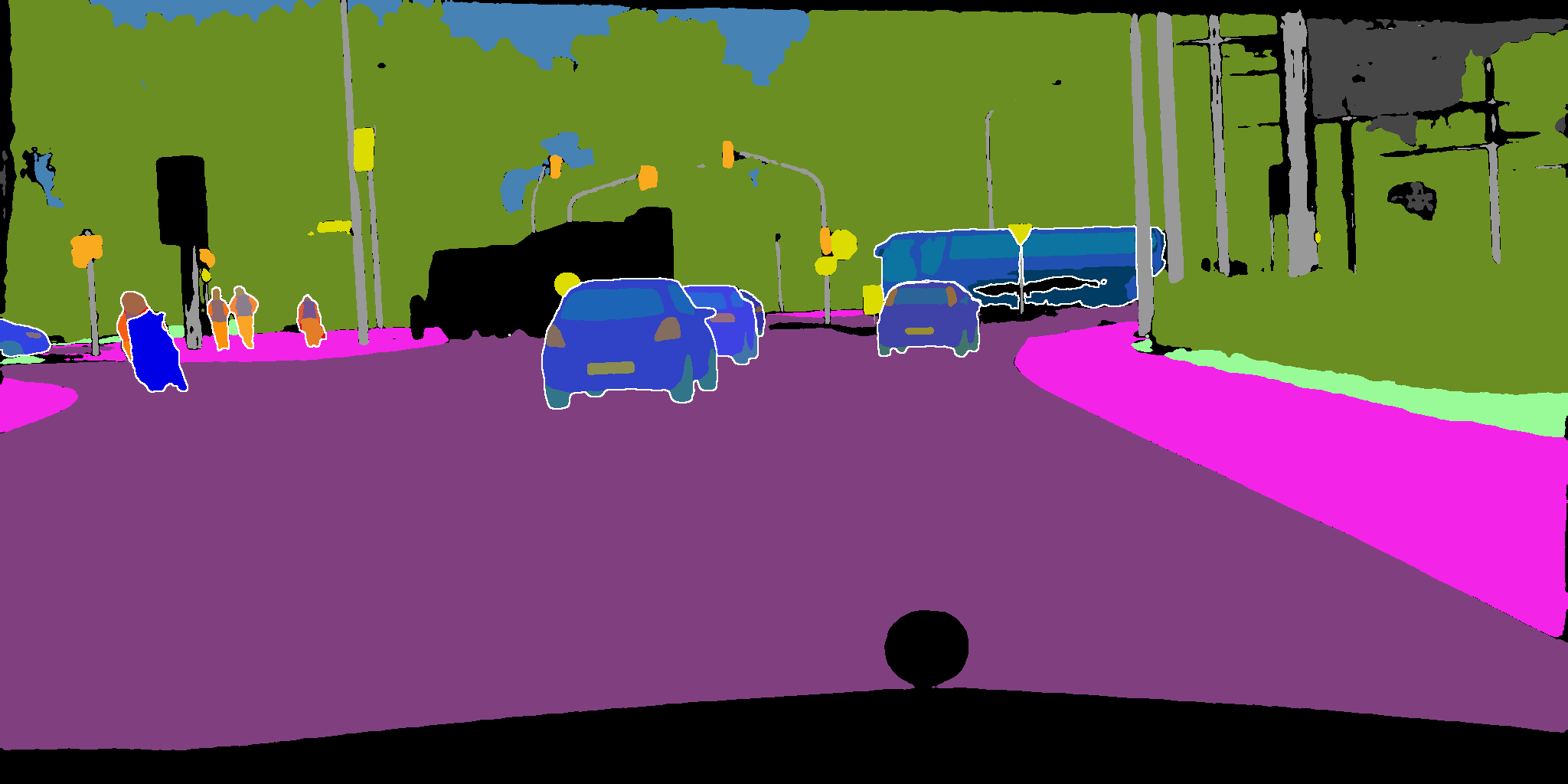}
\begin{tikzpicture}
    \node[anchor=south west,inner sep=0] (image) at (0,0) {\includegraphics[width=0.245\linewidth, trim={12cm 8cm 12cm 0},clip]{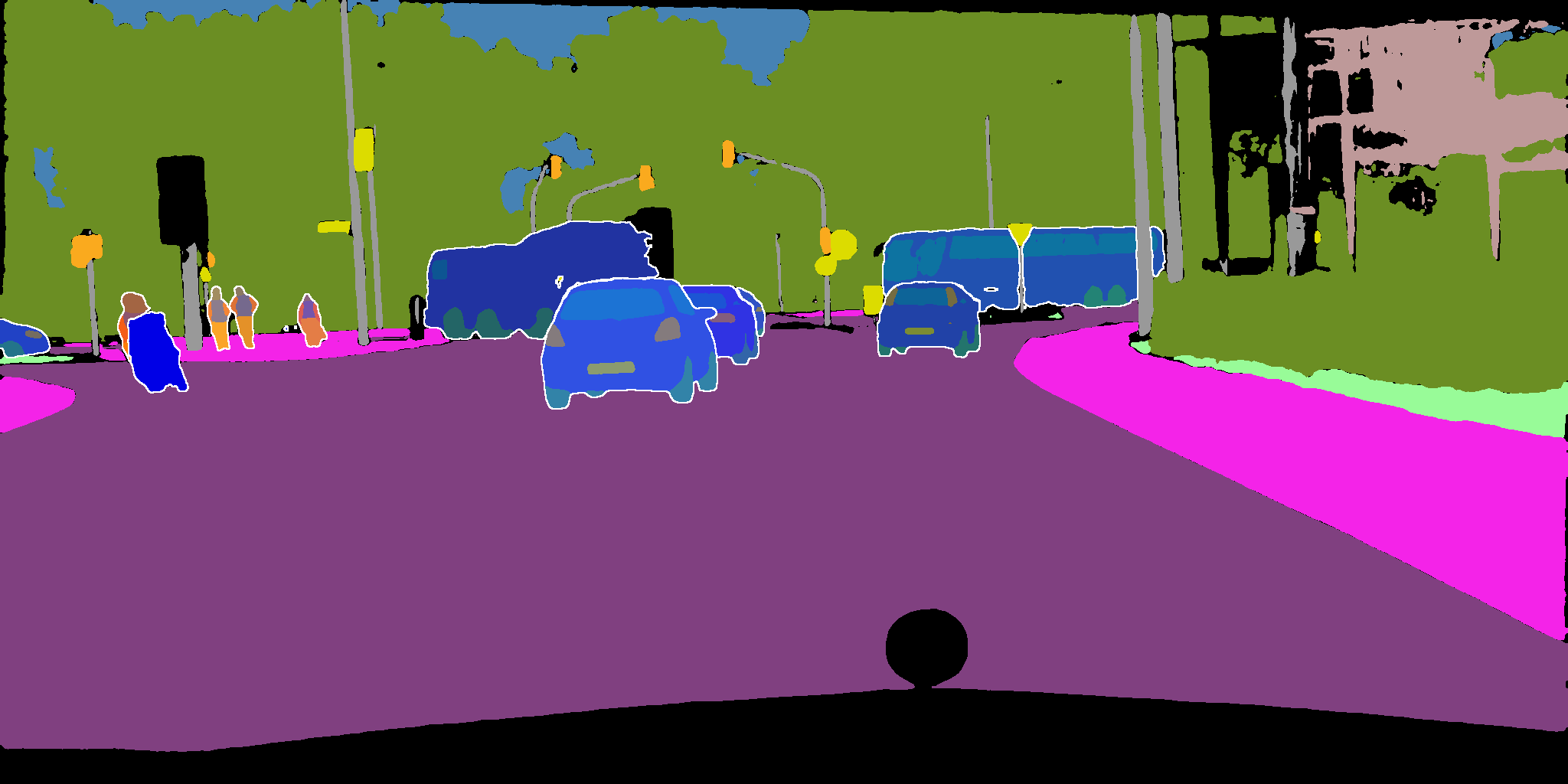}};
    \begin{scope}[x={(image.south east)},y={(image.north west)}]
        \draw[red,line width=0.5mm,rounded corners] (0.1,0.25) rectangle (0.9,0.7);
    \end{scope}
\end{tikzpicture}
\\

\includegraphics[width=0.245\linewidth, trim={0 2cm 8cm 0},clip]{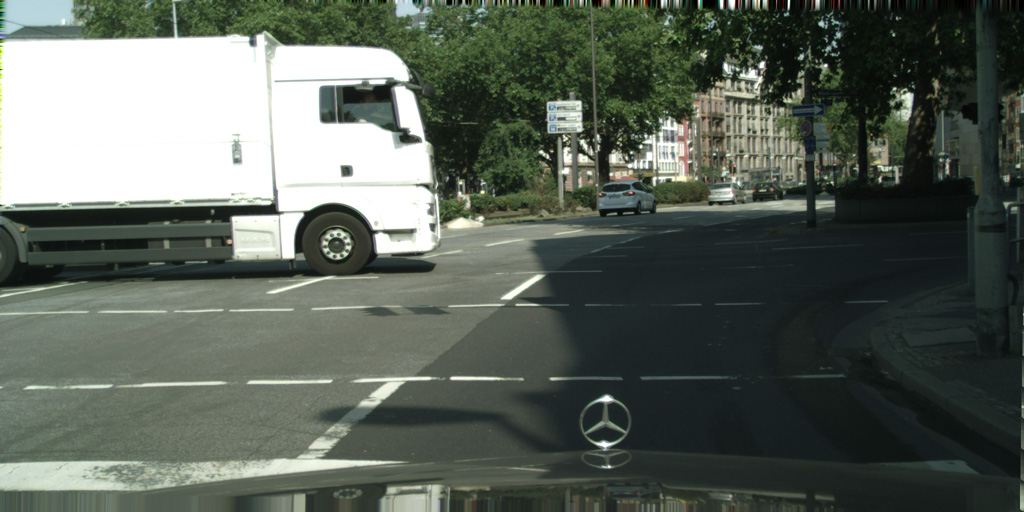}
\includegraphics[width=0.245\linewidth, trim={0 4cm 16cm 0},clip]{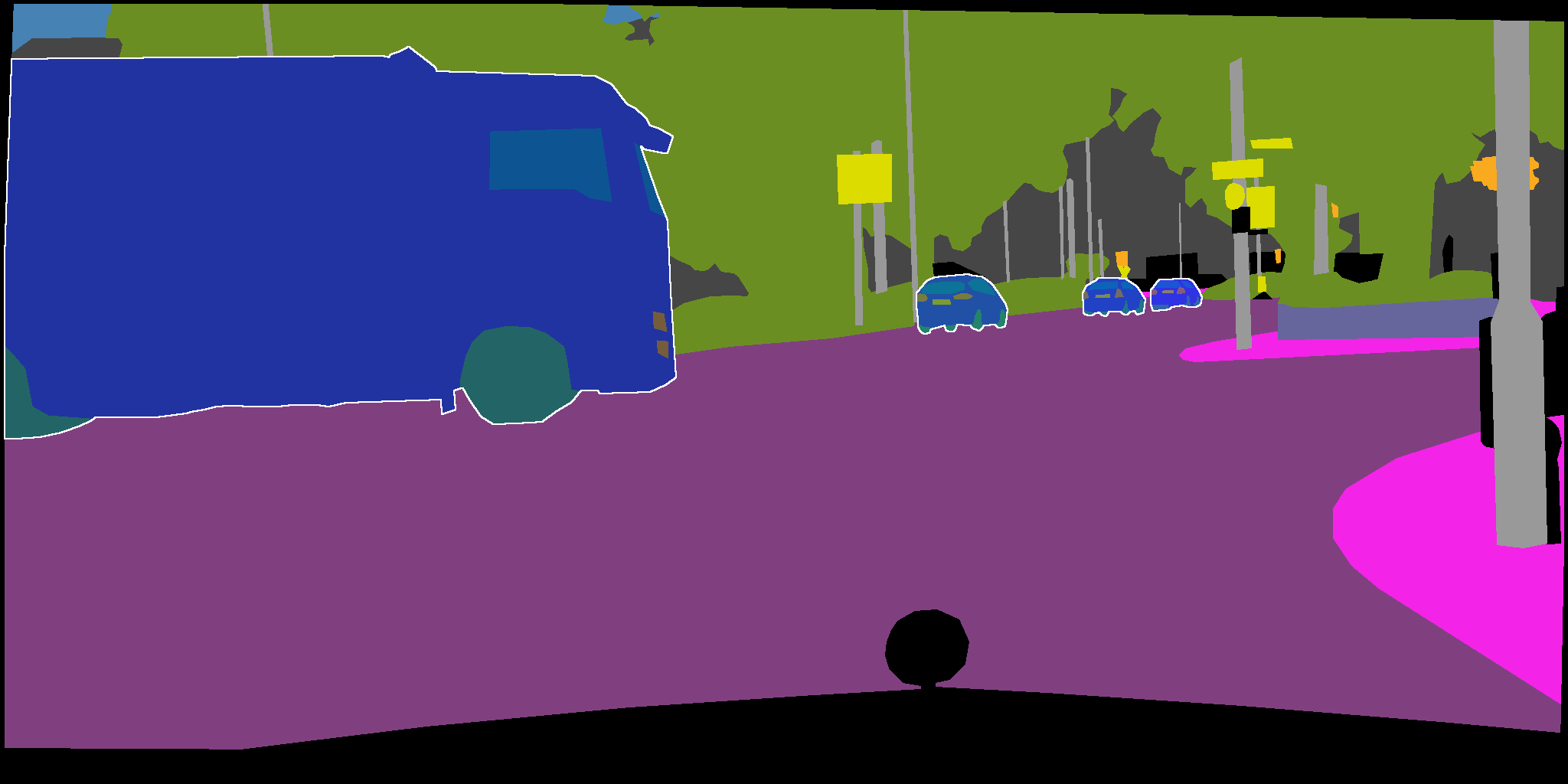}
\includegraphics[width=0.245\linewidth, trim={0 4cm 16cm 0},clip]{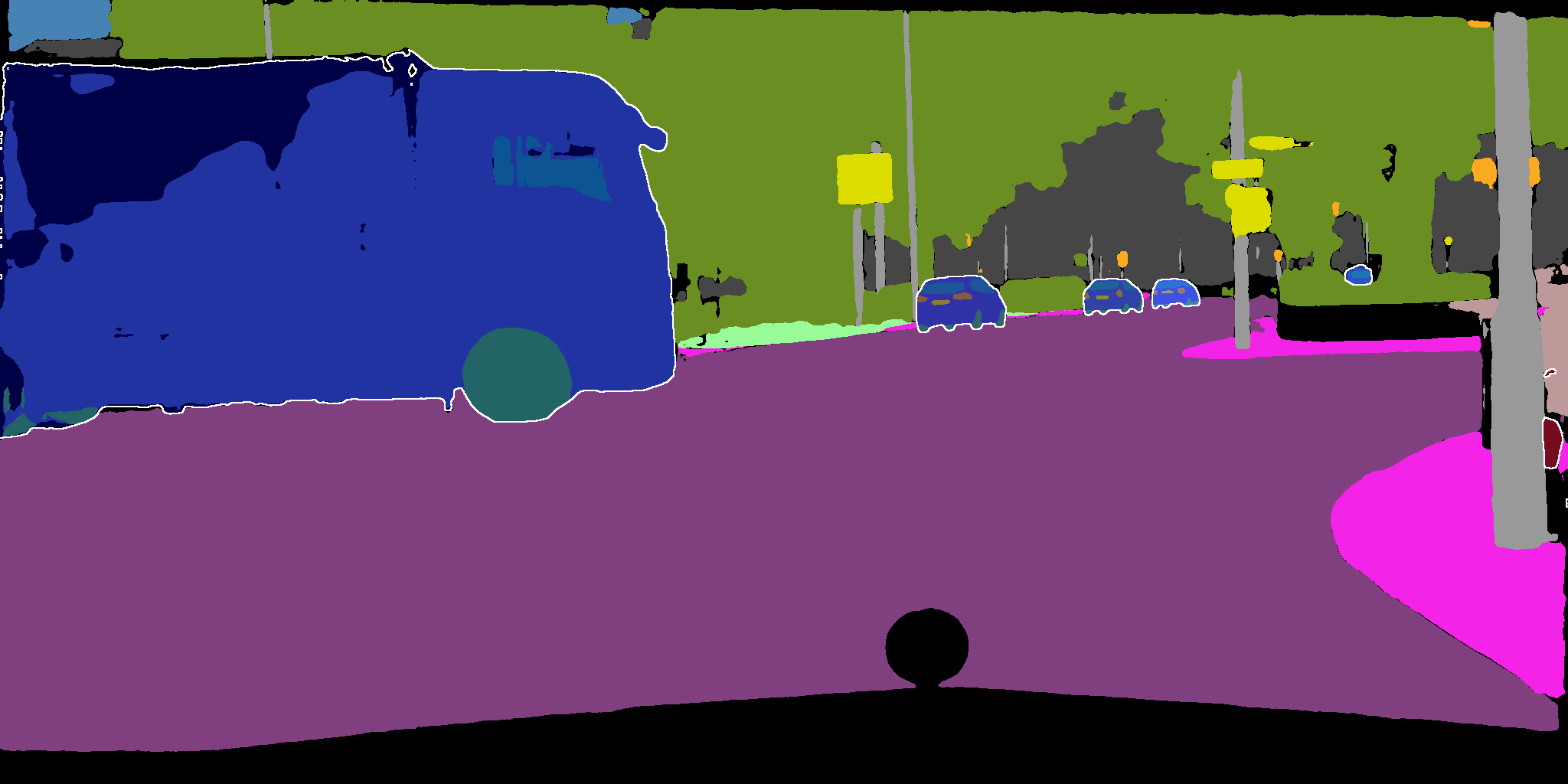}
\begin{tikzpicture}
    \node[anchor=south west,inner sep=0] (image) at (0,0) 
    {\includegraphics[width=0.245\linewidth, trim={0 4cm 16cm 0},clip]{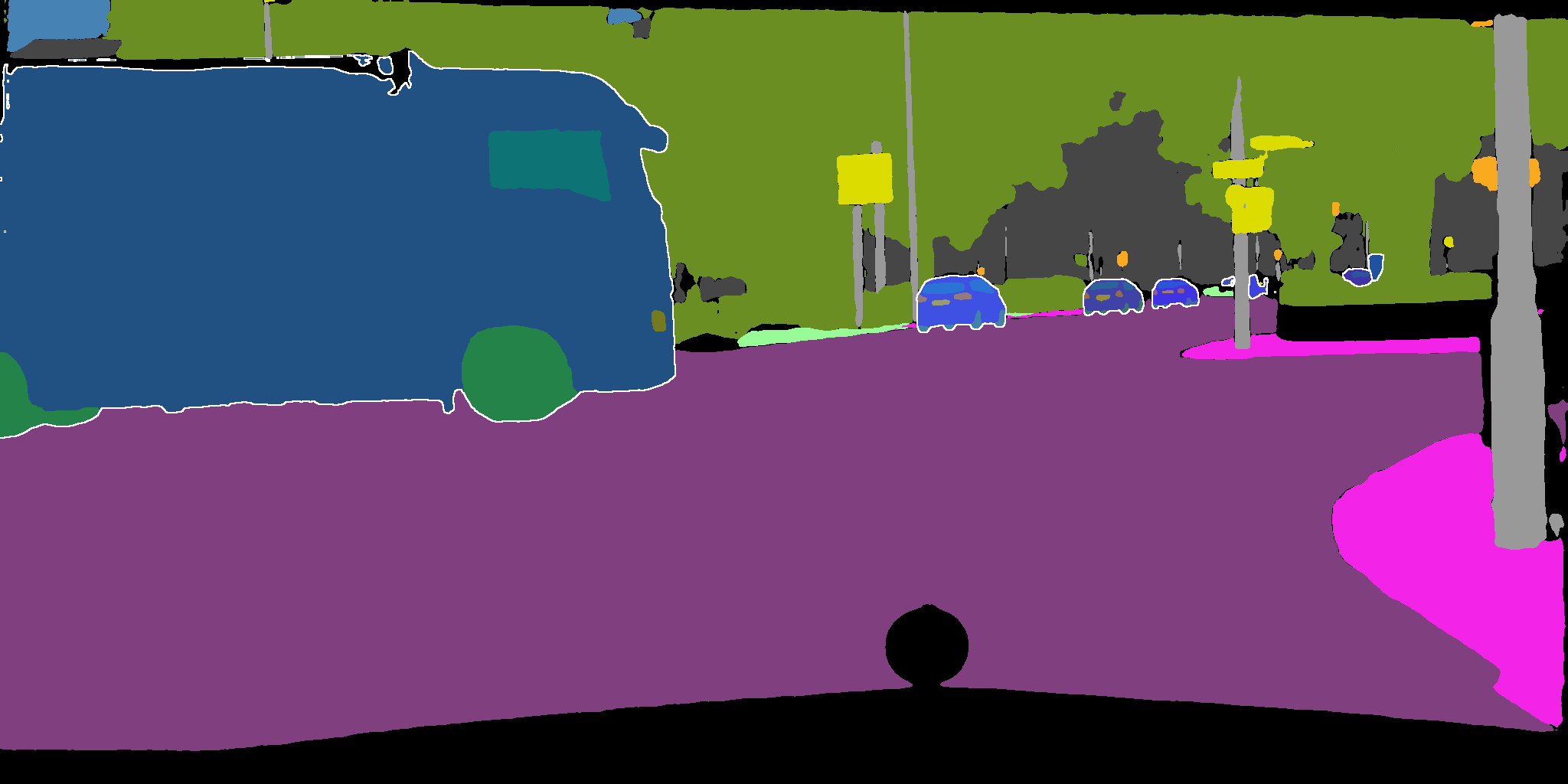}};
    \begin{scope}[x={(image.south east)},y={(image.north west)}]
        \draw[red,line width=0.5mm,rounded corners] (0.01,0.3) rectangle (0.65,0.98);
    \end{scope}
\end{tikzpicture}
\\

\includegraphics[width=0.245\linewidth, trim={2cm 4cm 12cm 0},clip]{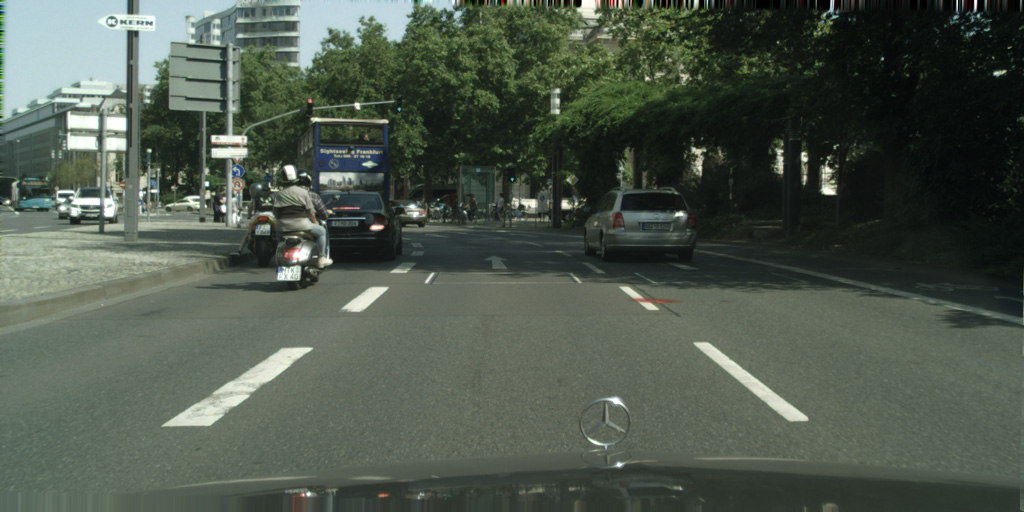}
\includegraphics[width=0.245\linewidth, trim={4cm 8cm 24cm 0},clip]{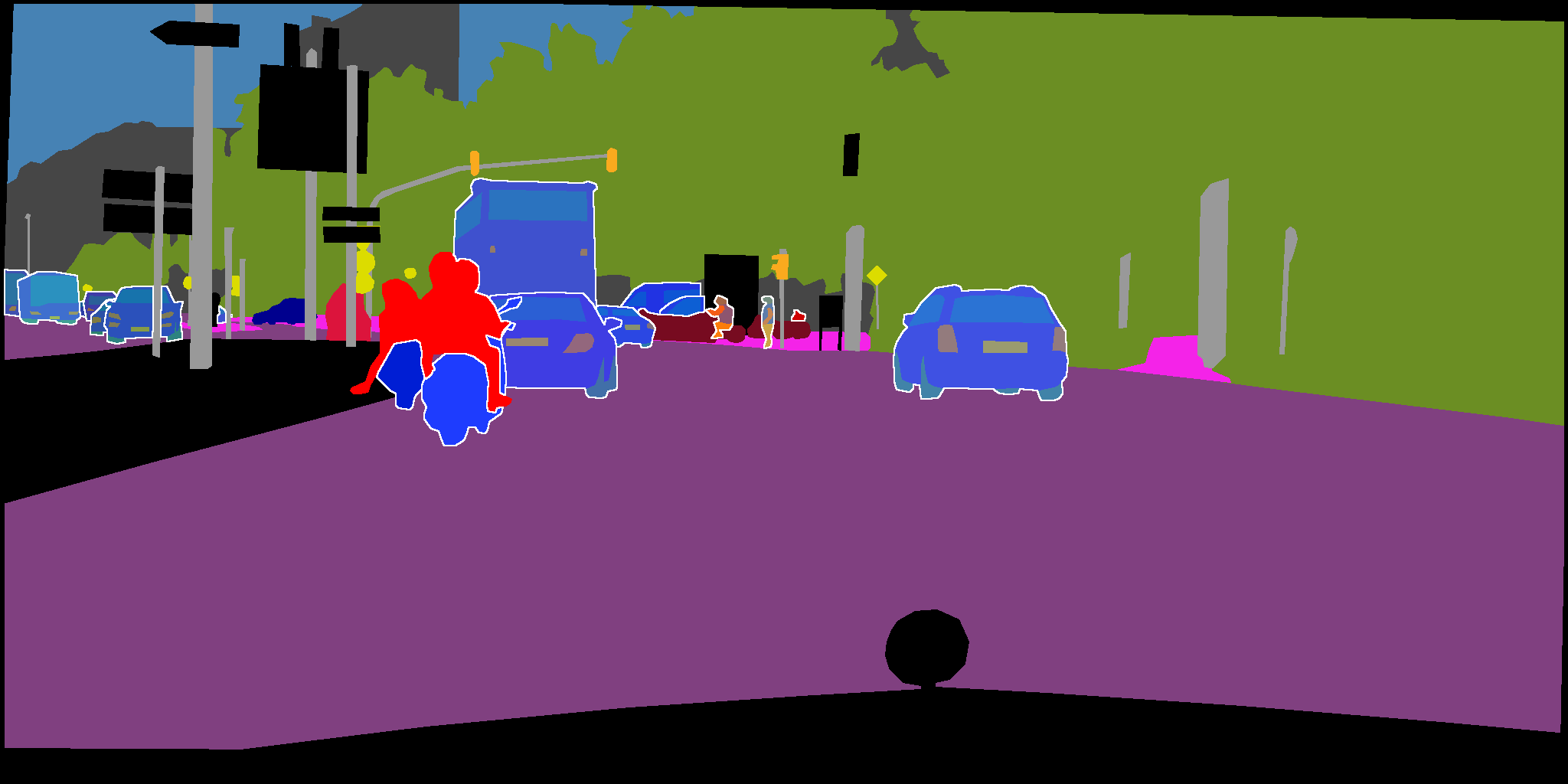}
\includegraphics[width=0.245\linewidth, trim={4cm 8cm 24cm 0},clip]{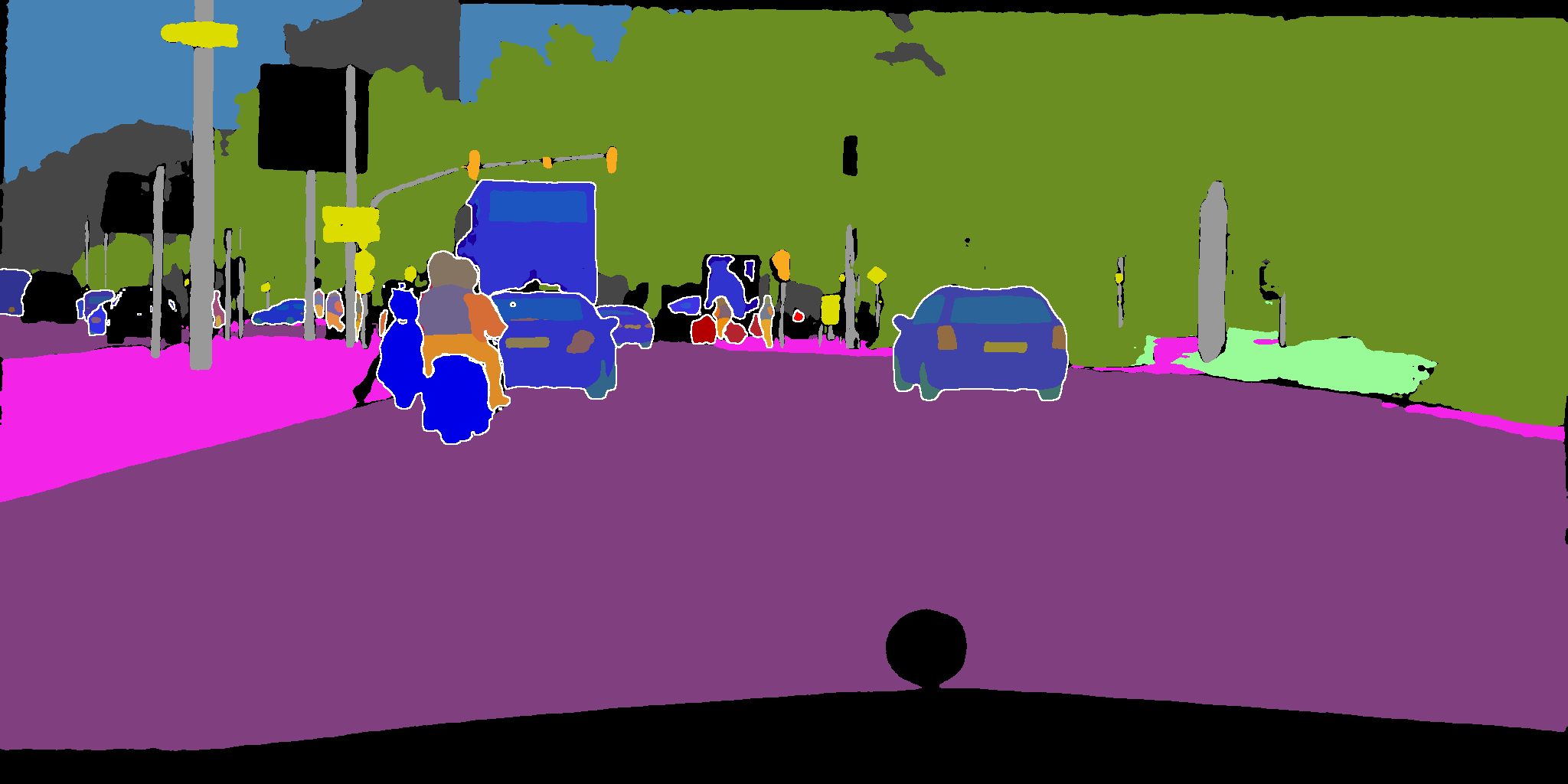}
\begin{tikzpicture}
    \node[anchor=south west,inner sep=0] (image) at (0,0) 
    {\includegraphics[width=0.245\linewidth, trim={4cm 8cm 24cm 0},clip]
    {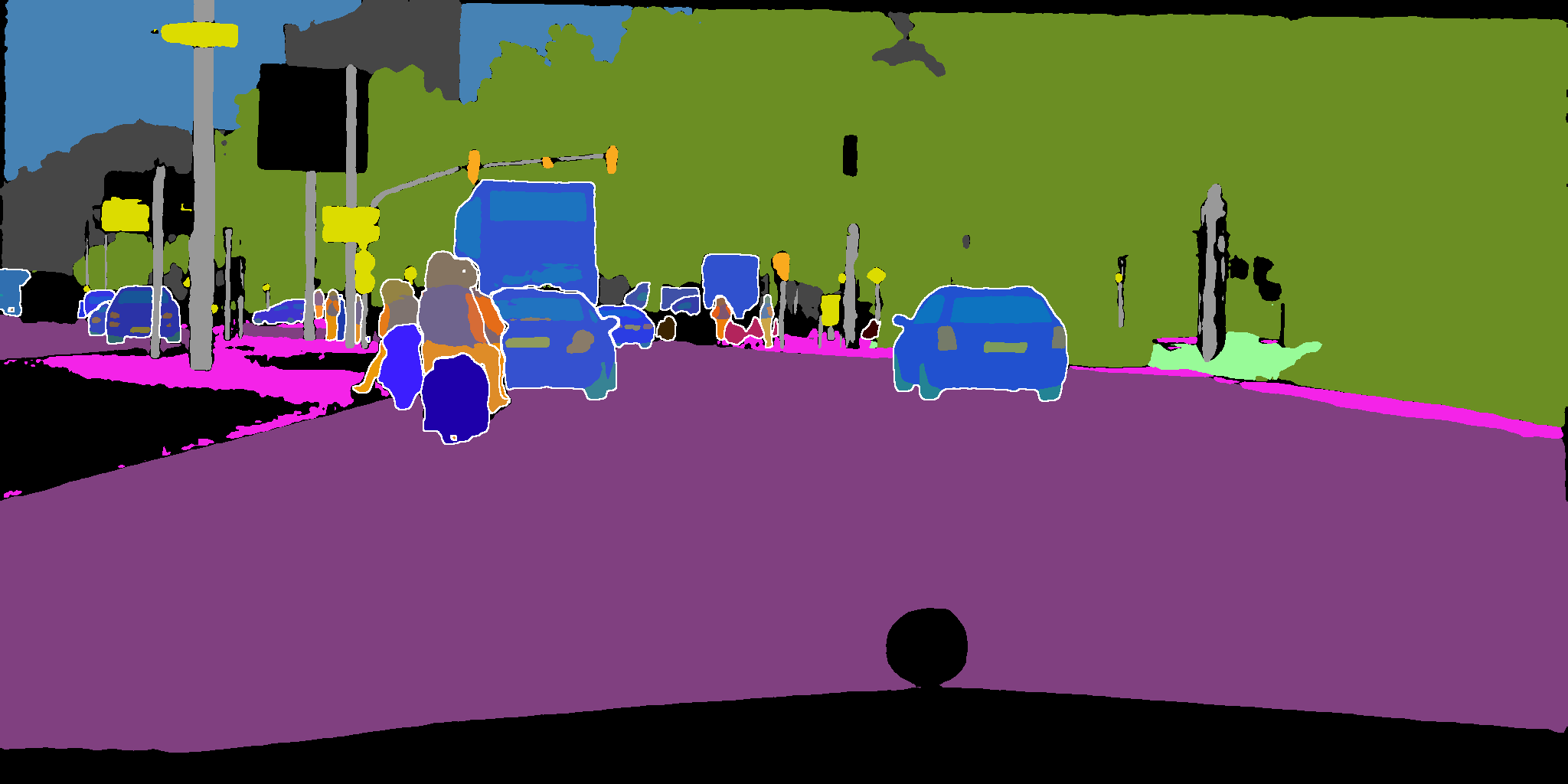}};
    \begin{scope}[x={(image.south east)},y={(image.north west)}]
        \draw[red,line width=0.5mm,rounded corners] (0.2,0.2) rectangle (0.6,0.75);
    \end{scope}
\end{tikzpicture}
\\

\includegraphics[width=0.245\linewidth, trim={8cm 2cm 0 0},clip]{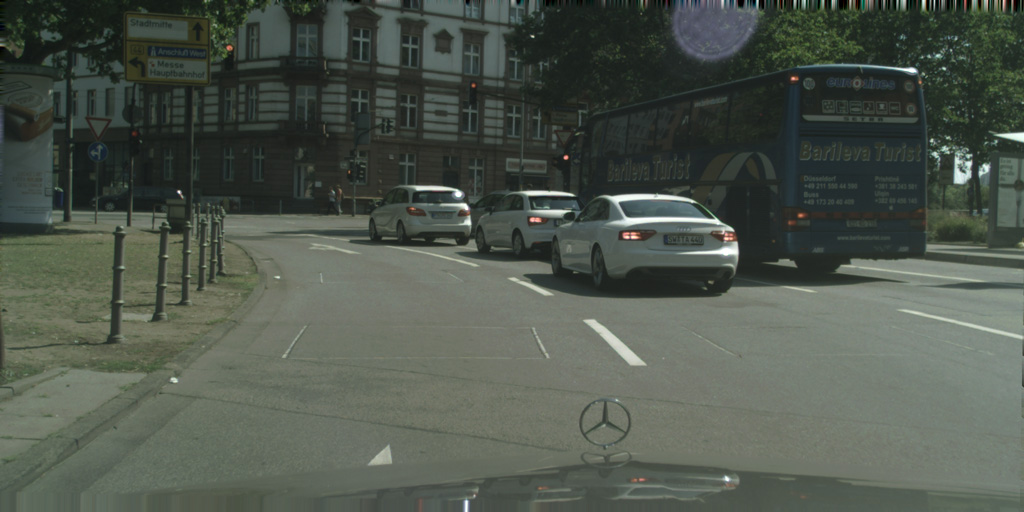}
\includegraphics[width=0.245\linewidth, trim={16cm 4cm 0 0},clip]{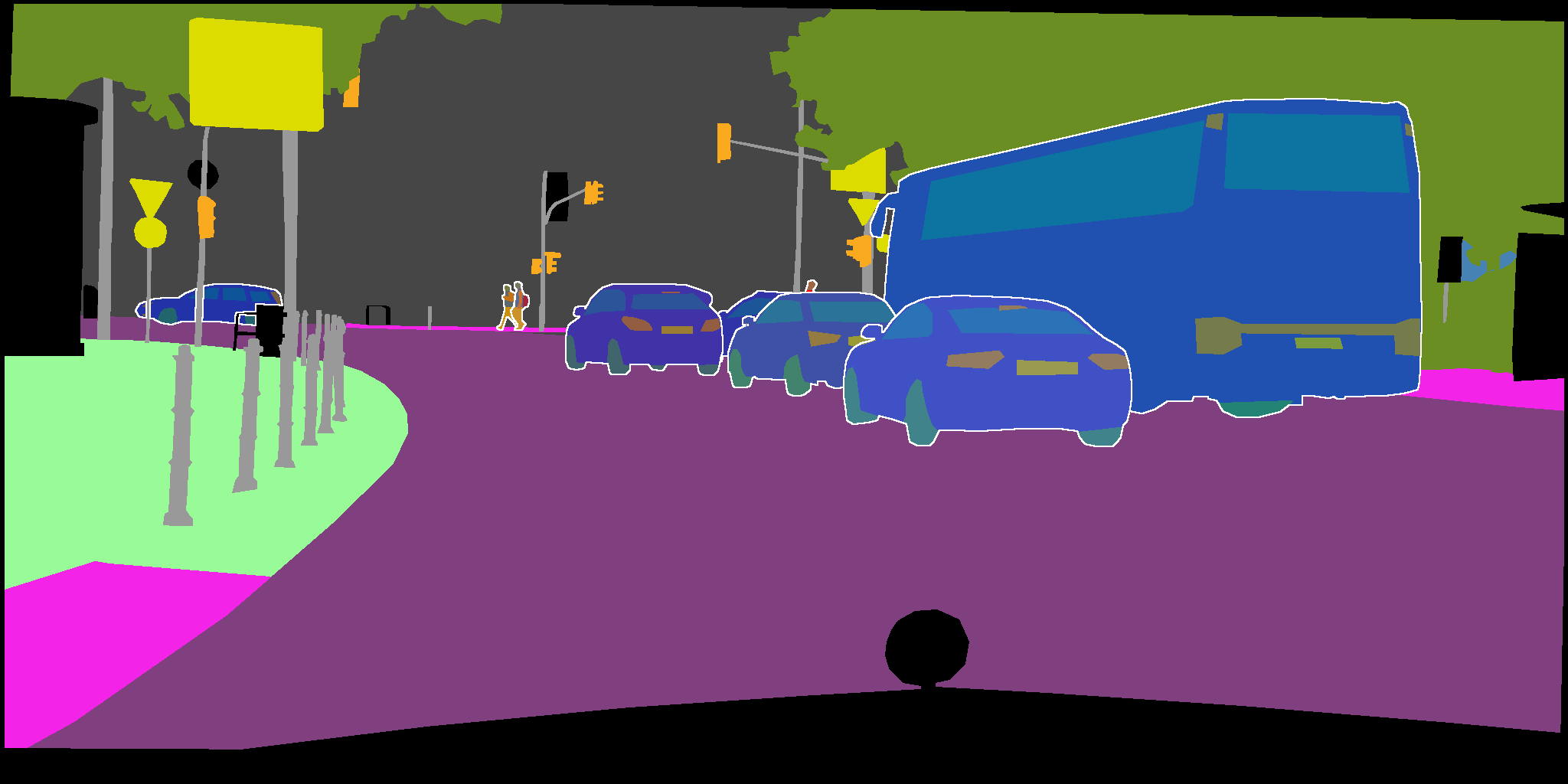}
\includegraphics[width=0.245\linewidth, trim={16cm 4cm 0 0},clip]{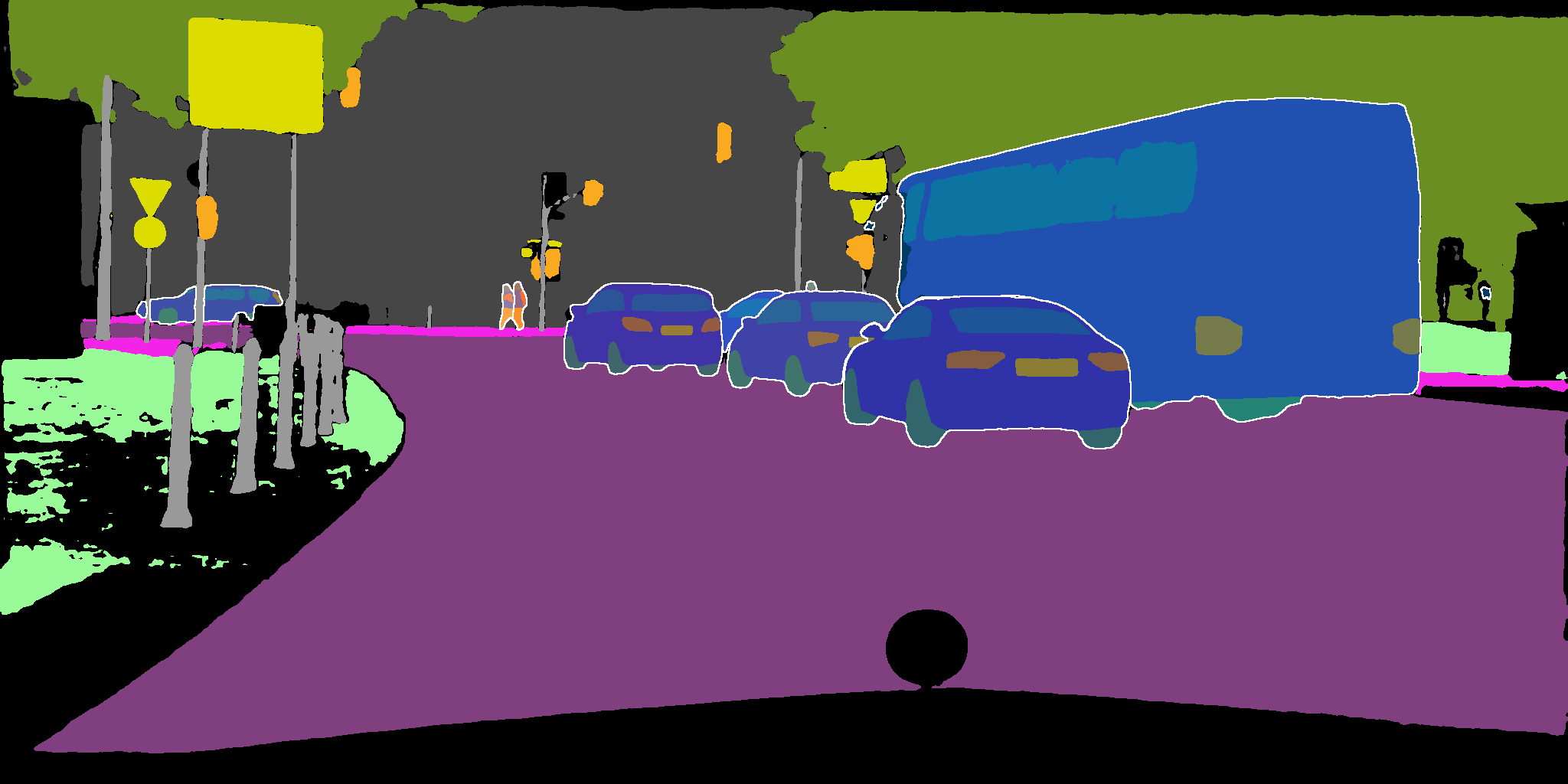}
\begin{tikzpicture}
    \node[anchor=south west,inner sep=0] (image) at (0,0) 
    {\includegraphics[width=0.245\linewidth, trim={16cm 4cm 0 0},clip]{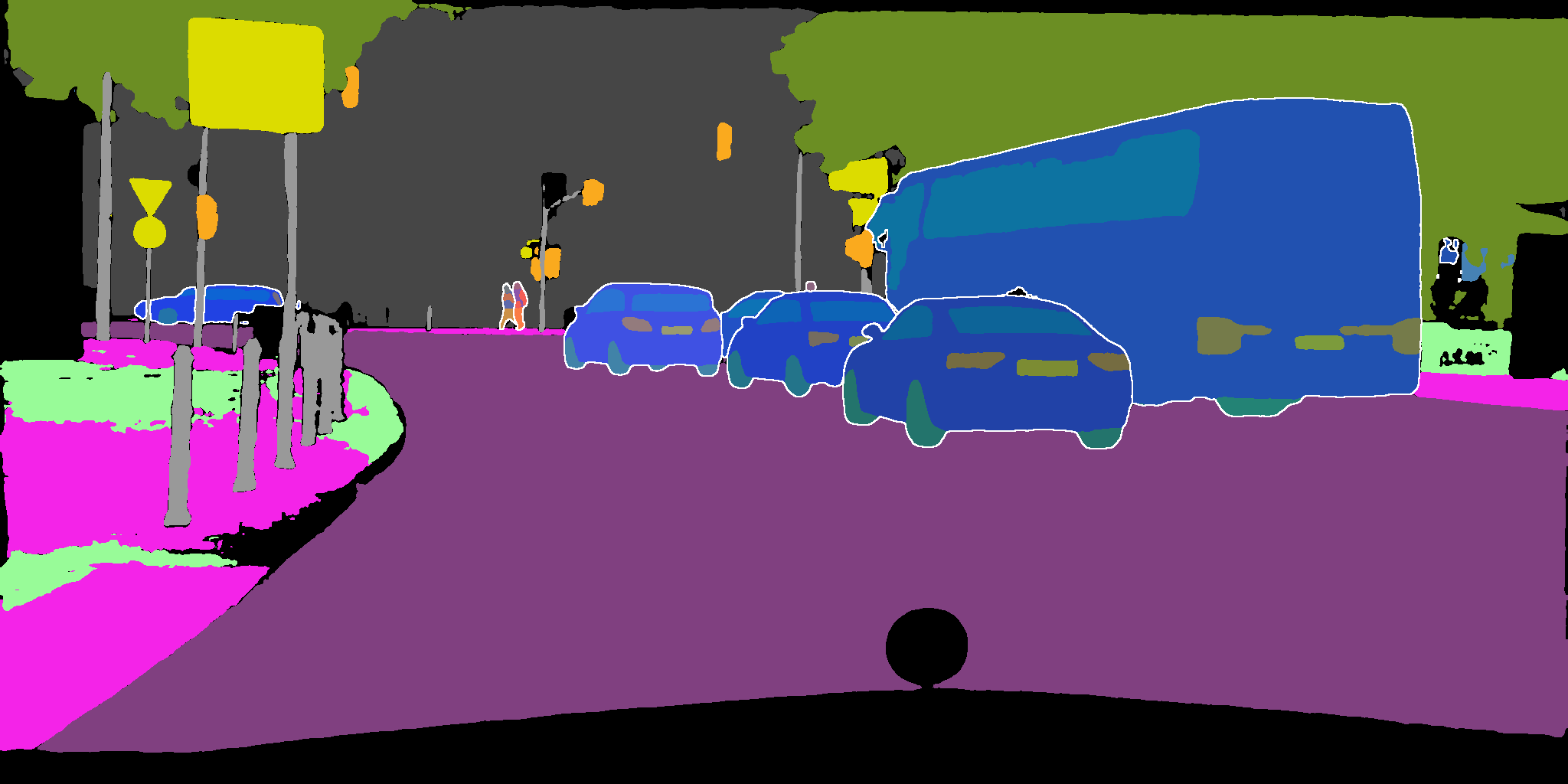}};
    \begin{scope}[x={(image.south east)},y={(image.north west)}]
    \draw[red,line width=0.5mm,rounded corners] (0.63,0.33) rectangle (0.92,0.65);
    \end{scope}
\end{tikzpicture}
\\

\includegraphics[width=0.245\linewidth, trim={0 4cm 16cm 0},clip]{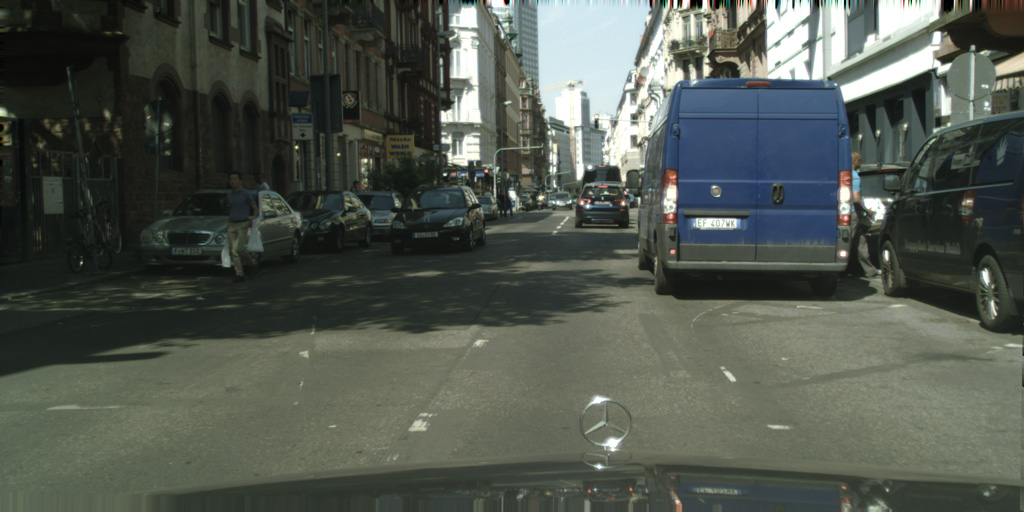}
\includegraphics[width=0.245\linewidth, trim={0 8cm 32cm 0},clip]{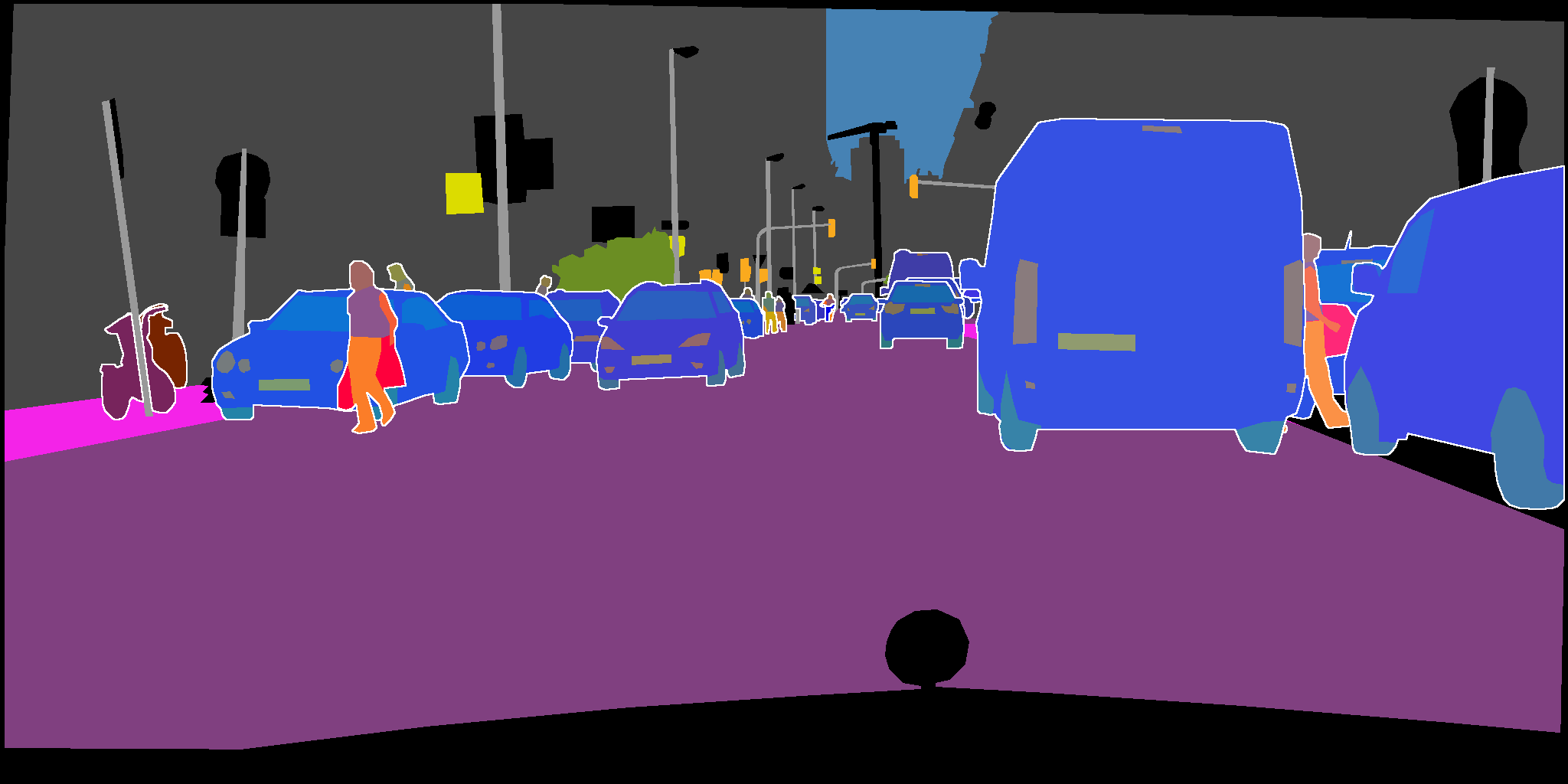}
\includegraphics[width=0.245\linewidth, trim={0 8cm 32cm 0},clip]{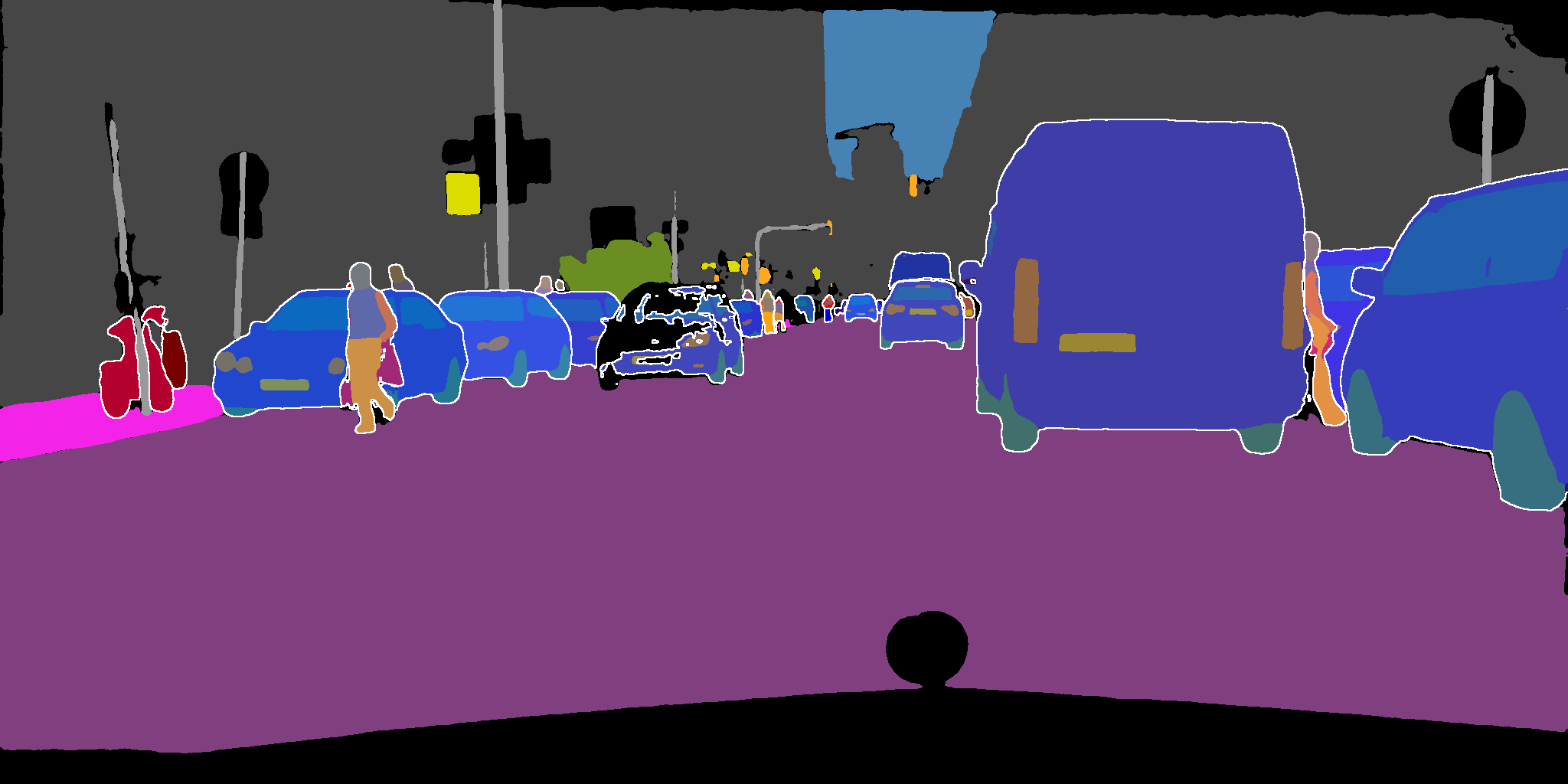}
\begin{tikzpicture}
    \node[anchor=south west,inner sep=0] (image) at (0,0) 
    {\includegraphics[width=0.245\linewidth, trim={0 8cm 32cm 0},clip]{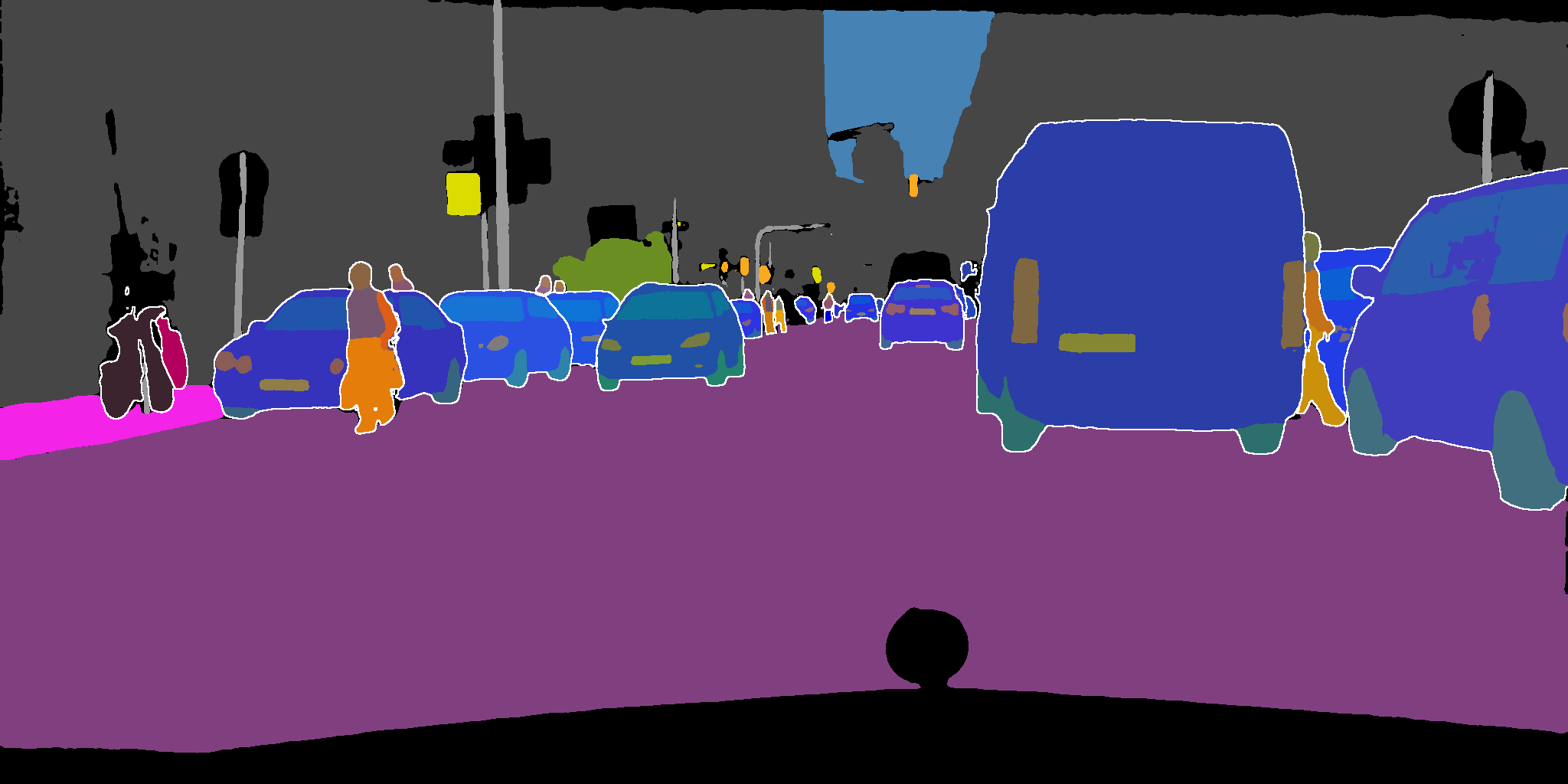}};
    \begin{scope}[x={(image.south east)},y={(image.north west)}]
    \draw[red,line width=0.5mm,rounded corners] (0.63,0.3) rectangle (0.92,0.6);
    \end{scope}
\end{tikzpicture}
\\

\includegraphics[width=0.245\linewidth, trim={12cm 2cm 0 0},clip]{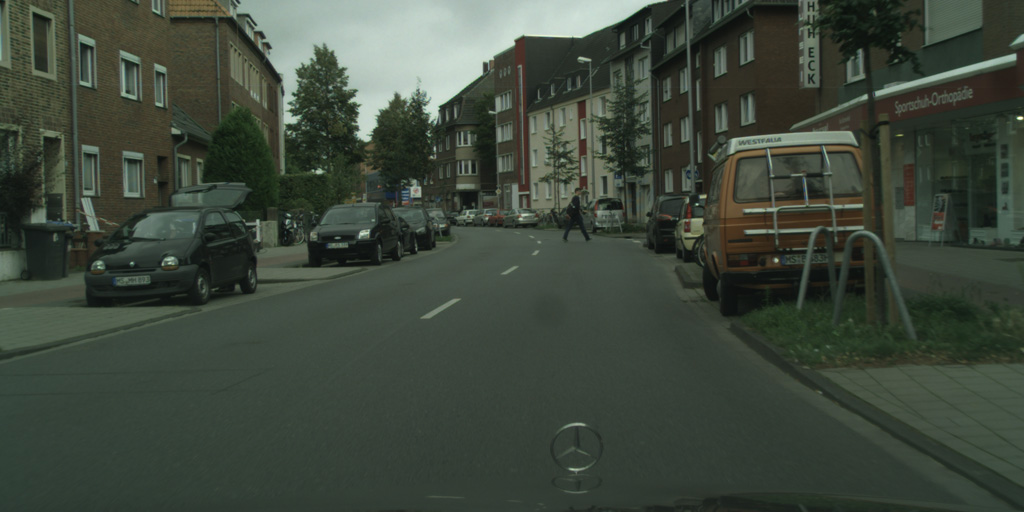}
\includegraphics[width=0.245\linewidth, trim={24cm 4cm 0 0},clip]{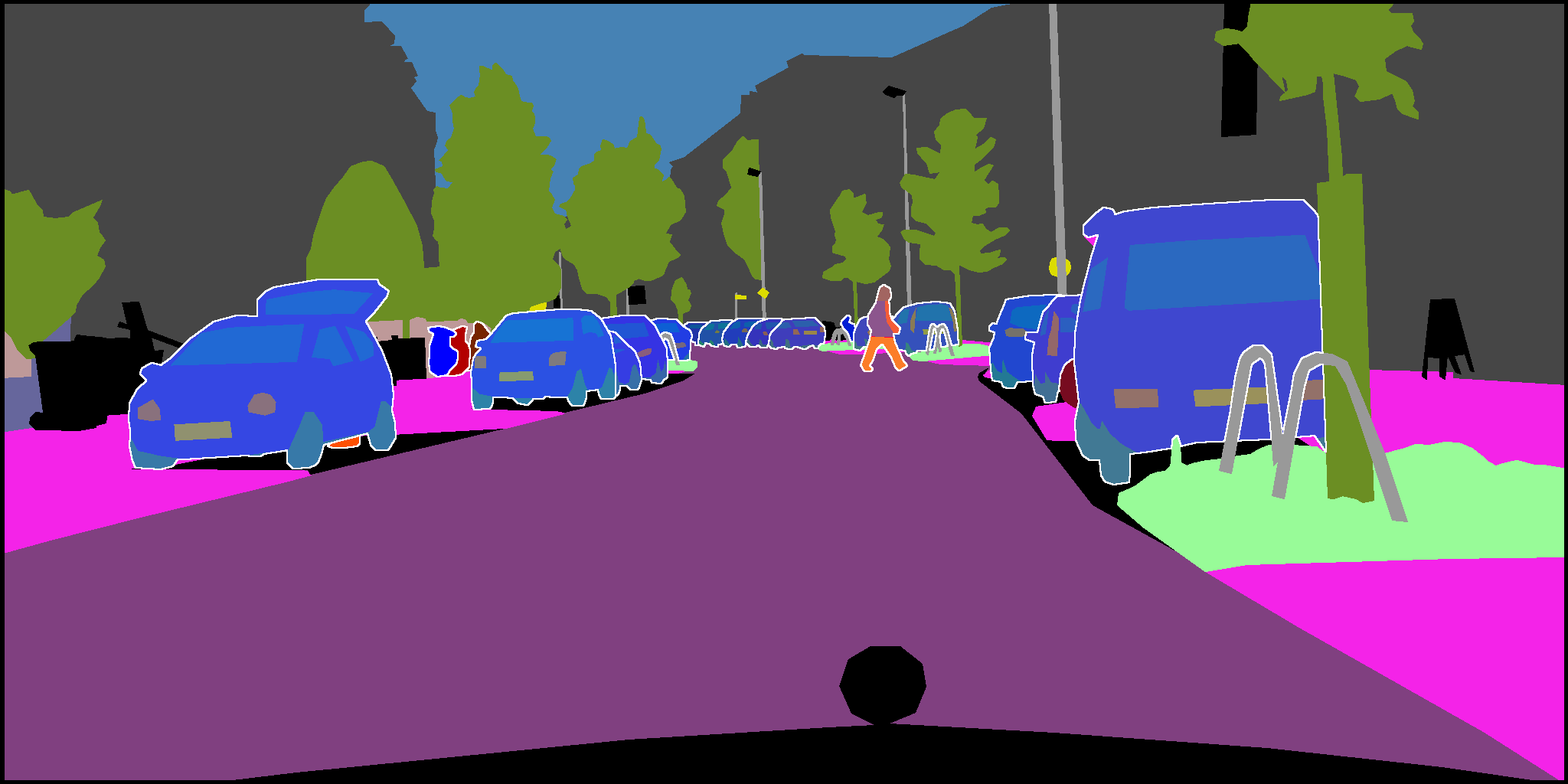}
\includegraphics[width=0.245\linewidth, trim={24cm 4cm 0 0},clip]{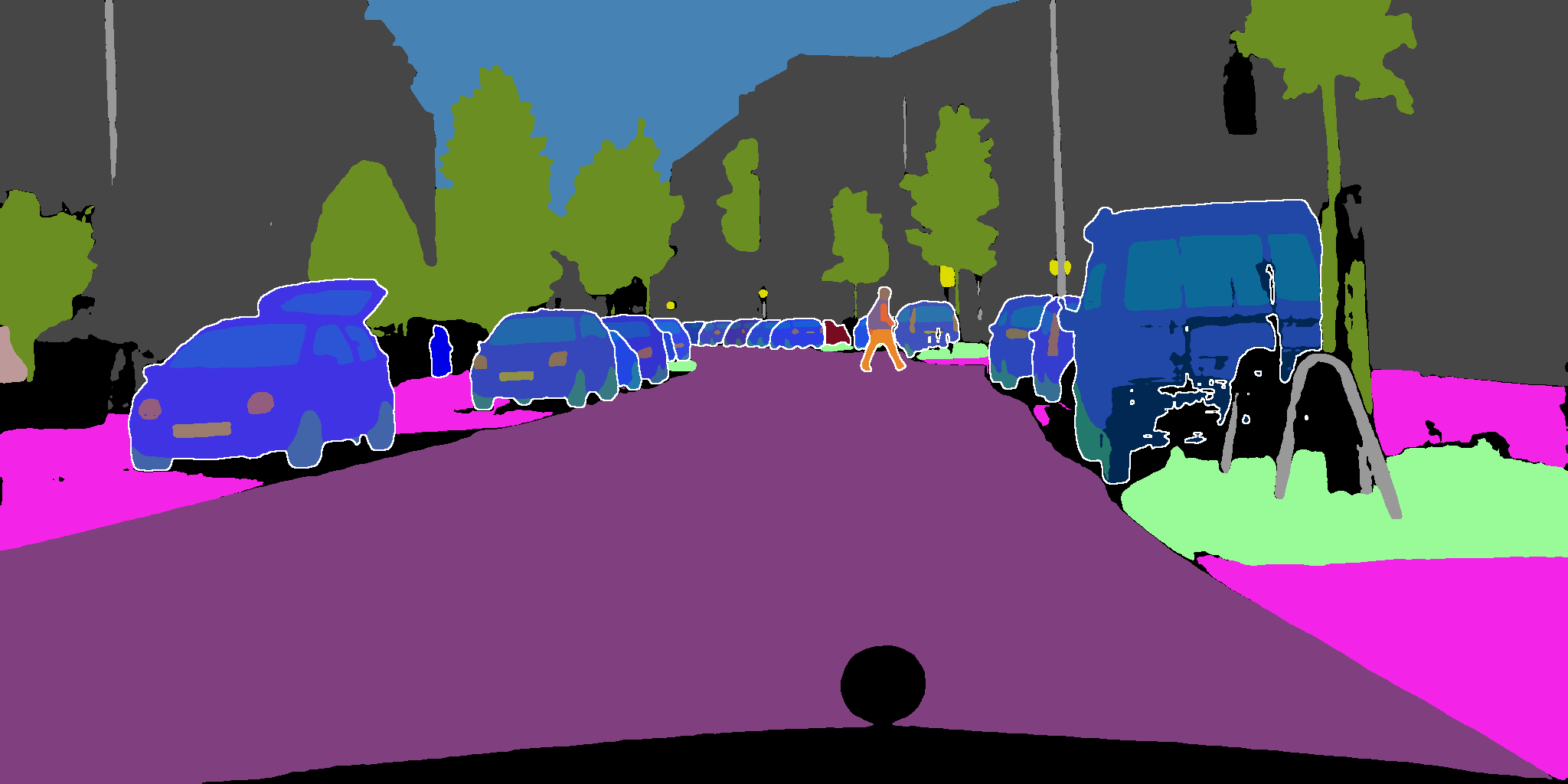}
\begin{tikzpicture}
    \node[anchor=south west,inner sep=0] (image) at (0,0) 
    {\includegraphics[width=0.245\linewidth, trim={24cm 4cm 0 0},clip]{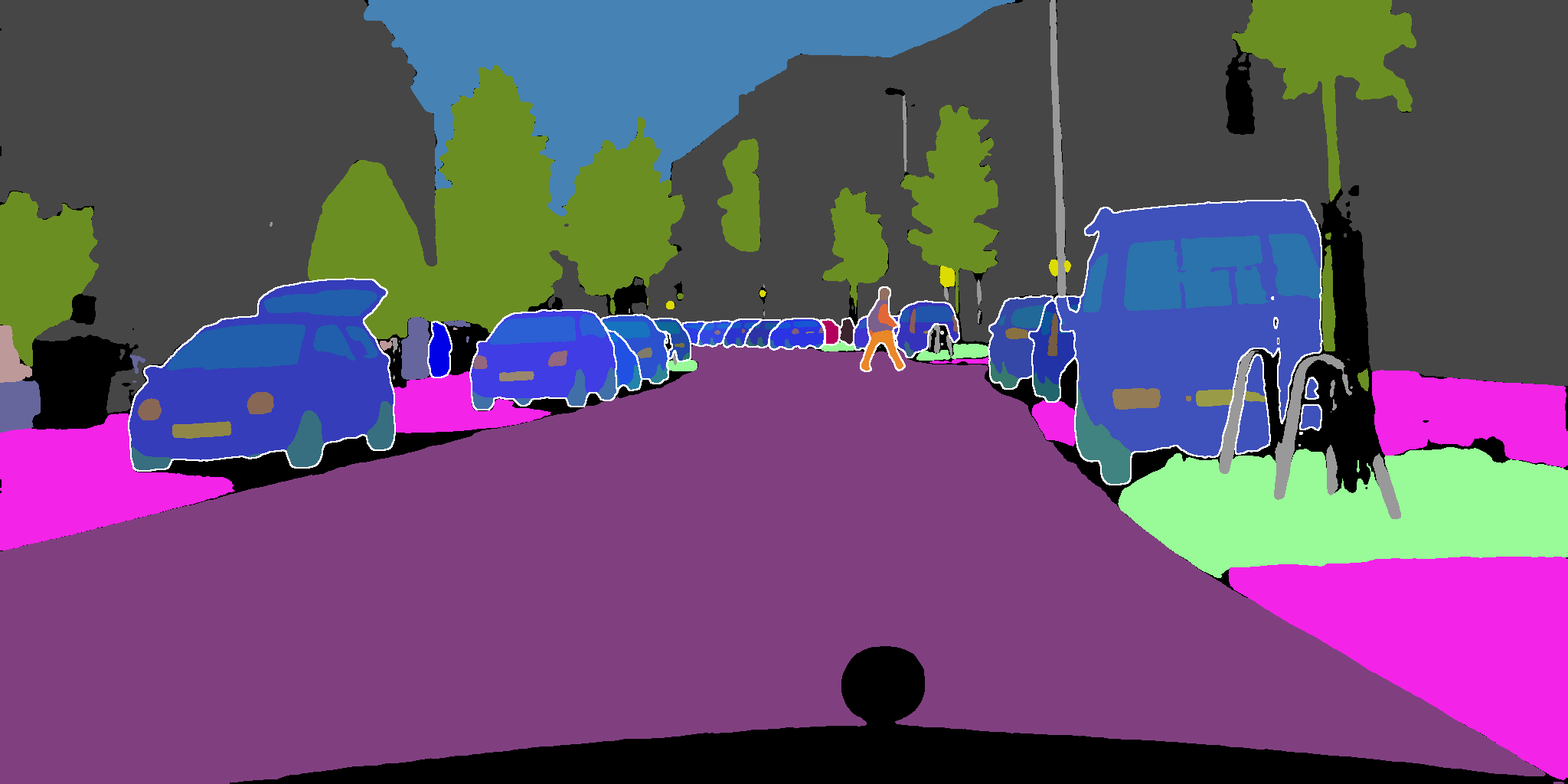}};
    \begin{scope}[x={(image.south east)},y={(image.north west)}]
    \draw[red,line width=0.5mm,rounded corners] (0.45,0.2) rectangle (0.85,0.8);
    \end{scope}
\end{tikzpicture}
\\

\includegraphics[width=0.245\linewidth, trim={16cm 4cm 0 0},clip]{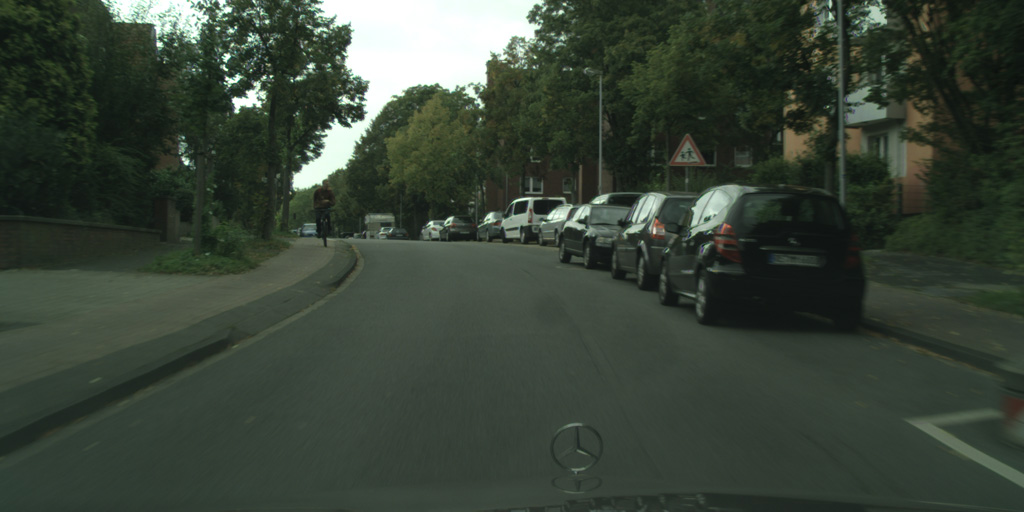}
\includegraphics[width=0.245\linewidth, trim={32cm 8cm 0 0},clip]{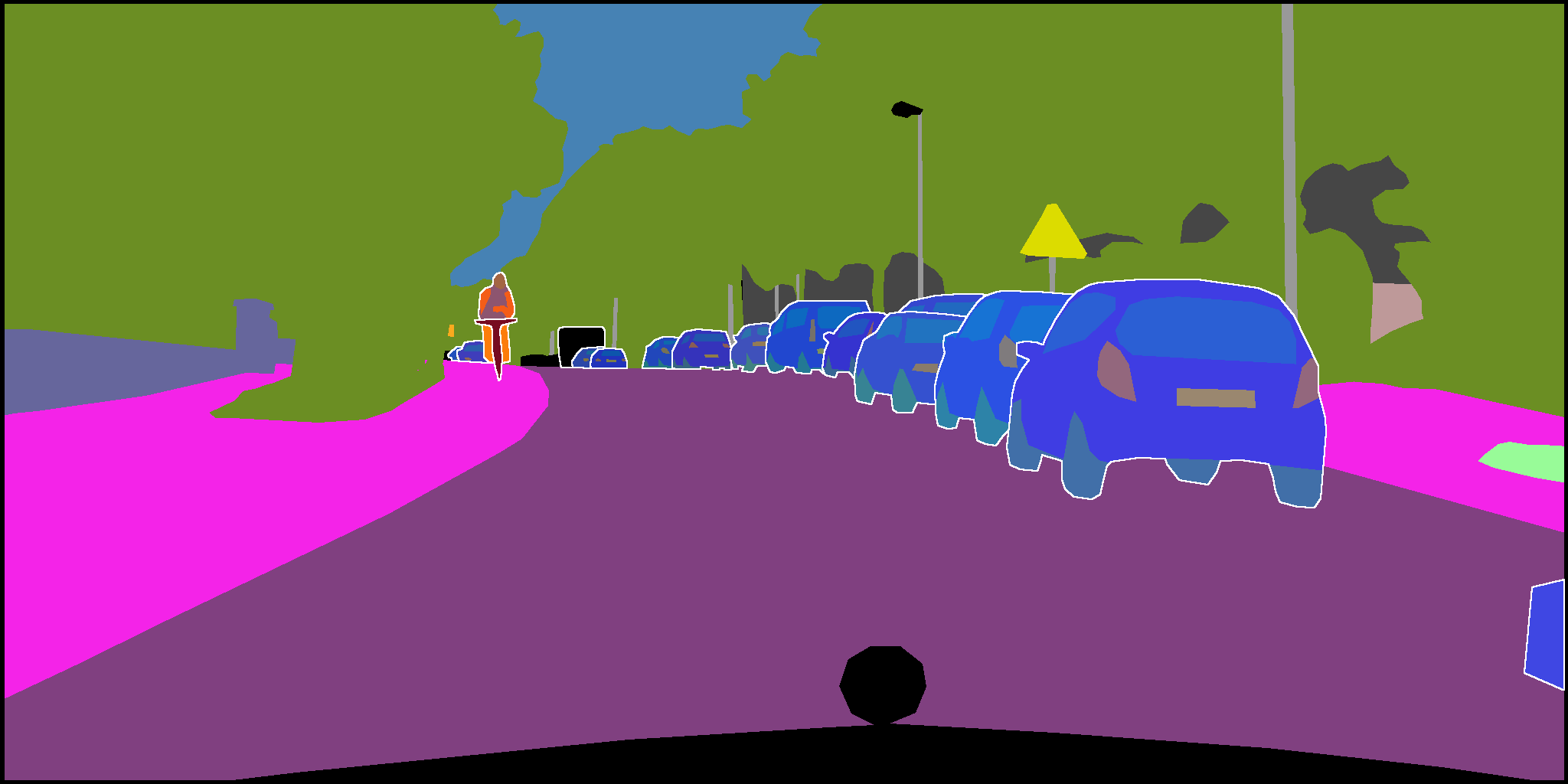}
\includegraphics[width=0.245\linewidth, trim={32cm 8cm 0 0},clip]{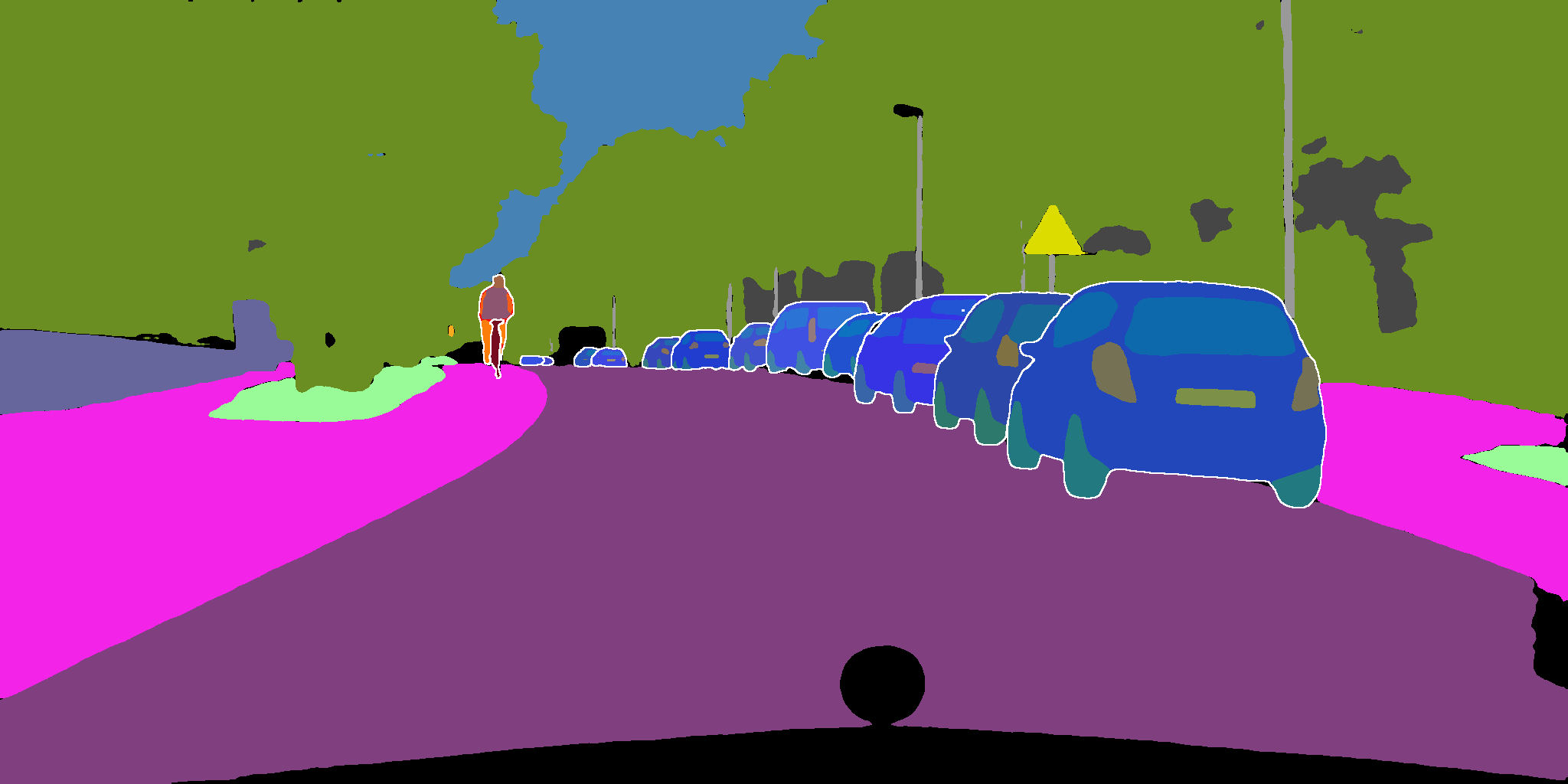}
\begin{tikzpicture}
    \node[anchor=south west,inner sep=0] (image) at (0,0) 
    {\includegraphics[width=0.245\linewidth, trim={32cm 8cm 0 0},clip]{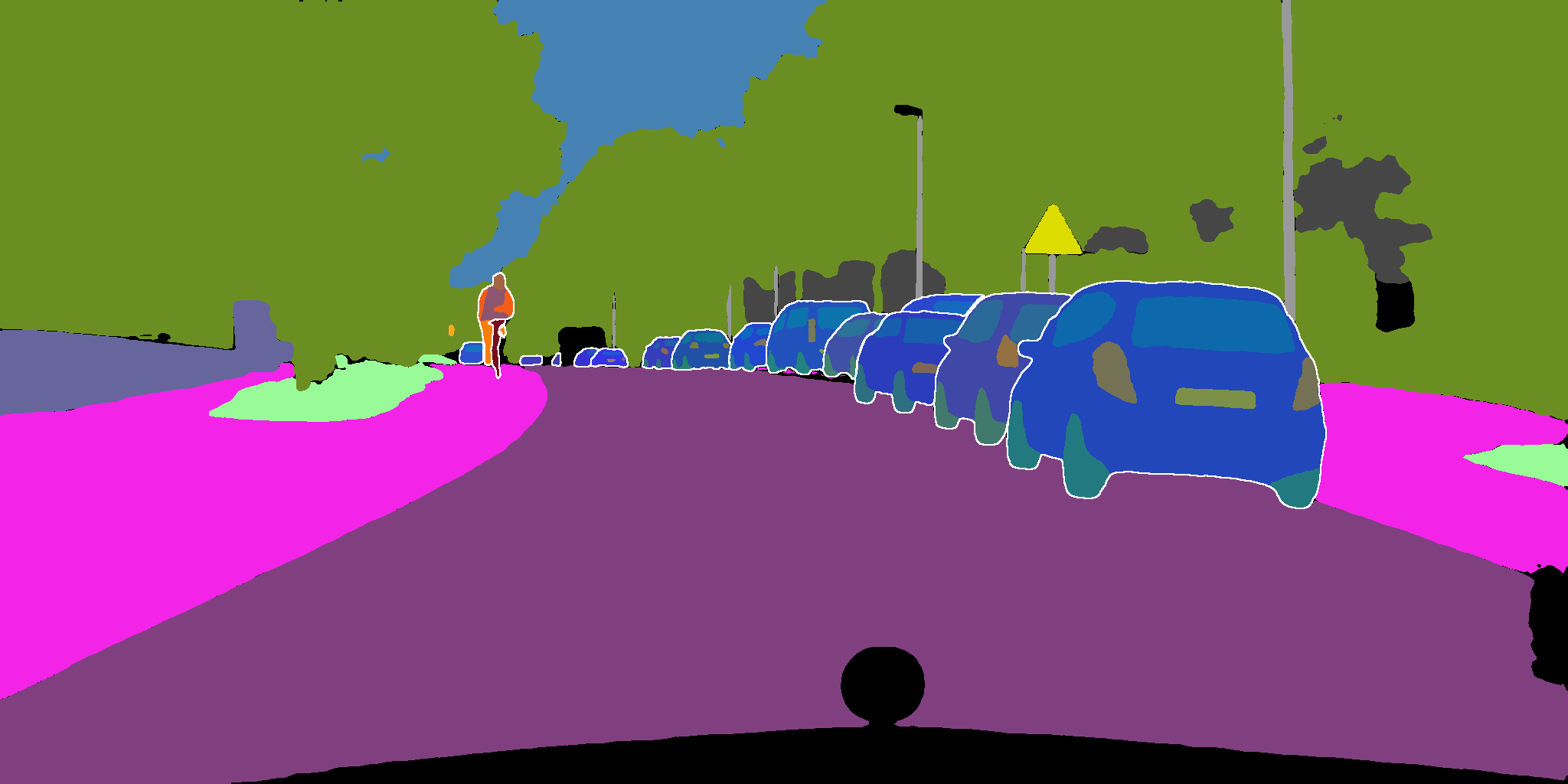}};
    \begin{scope}[x={(image.south east)},y={(image.north west)}]
    \draw[red,line width=0.5mm,rounded corners] (0.15,0.4) rectangle (0.4,0.6);
    \end{scope}
\end{tikzpicture}
\\

\begin{subfigure}[b]{0.245\textwidth}
 \centering
 \caption{Input image}
\end{subfigure}
\begin{subfigure}[b]{0.245\textwidth}
 \centering
 \caption{Ground truth}
\end{subfigure}
\begin{subfigure}[b]{0.245\textwidth}
 \centering
 \caption{Baseline}
\end{subfigure}
\begin{subfigure}[b]{0.245\textwidth}
 \centering
 \caption{TAPPS (ours)}
\end{subfigure}

\caption{\textbf{Qualitative examples of TAPPS and our strong baseline on Cityscapes-PP~\cite{cordts2016cityscapes,degeus2021pps}.} Both networks use ResNet-50~\cite{he2016resnet} with COCO pre-training~\cite{lin2014coco}. White borders separate different object-level instances; color shades indicate different categories. Note that the colors of part-level categories are not identical across instances; there are different shades of the same color. In these examples, we can see how TAPPS improves both the instance separability and part segmentation quality with respect to the strong baseline. The red boxes indicate regions in which these differences are best visible. Best viewed digitally.}
\label{supp:fig:qual_results_baseline_cpp}
\end{figure*}

In \cref{supp:fig:qual_results_baseline_ppp} and \cref{supp:fig:qual_results_baseline_cpp}, we compare TAPPS against the strong baseline that we describe in \cref{supp:sec:exp_setup:impl_details:baseline}. These examples show some of the advantages of TAPPS over the baseline. Specifically, we observe that TAPPS (1) makes more accurate part segmentation predictions within identified objects, and (2) is better able to separate different object instances.

\begin{figure*}[t]
\centering

\includegraphics[width=0.245\linewidth]{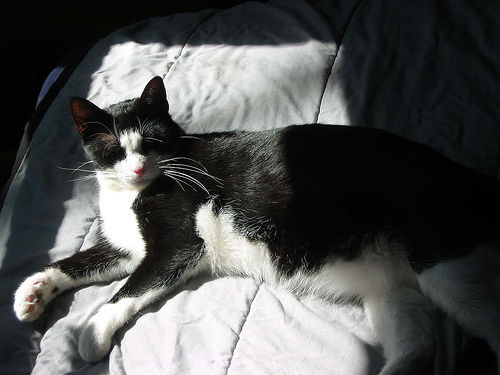}
\includegraphics[width=0.245\linewidth]{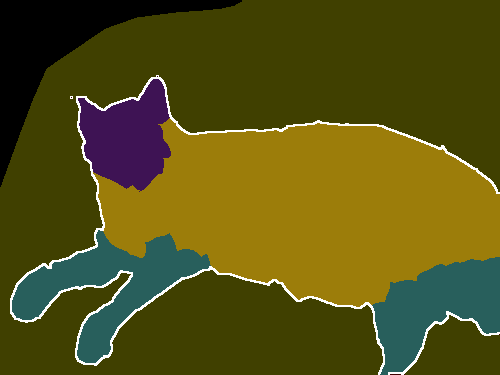}
\includegraphics[width=0.245\linewidth]{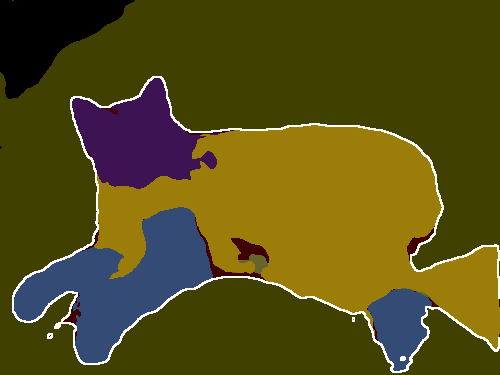}
\includegraphics[width=0.245\linewidth]{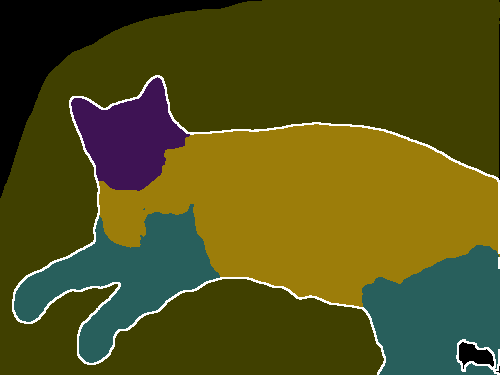}
\\

\includegraphics[width=0.245\linewidth]{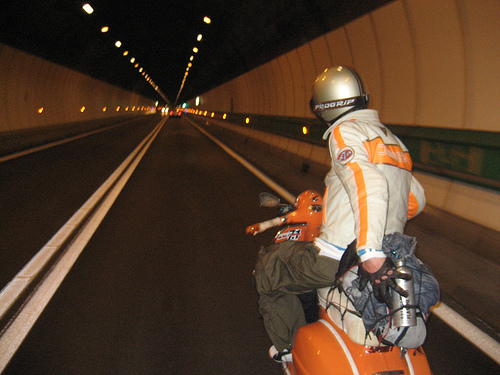}
\includegraphics[width=0.245\linewidth]{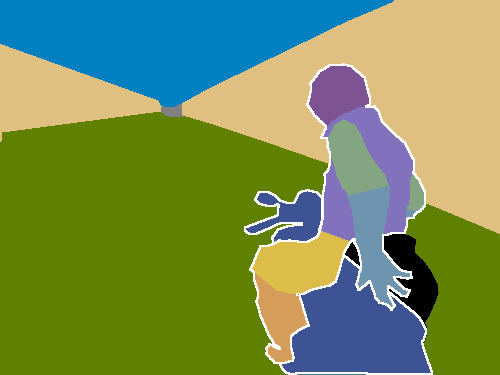}
\includegraphics[width=0.245\linewidth]{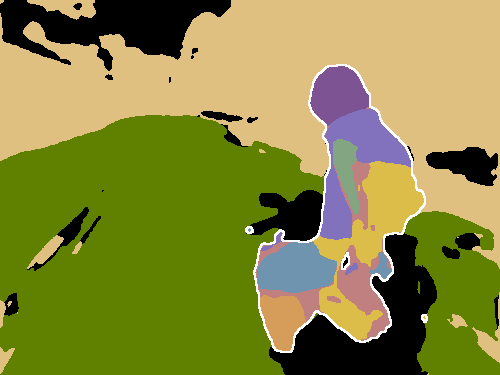}
\includegraphics[width=0.245\linewidth]{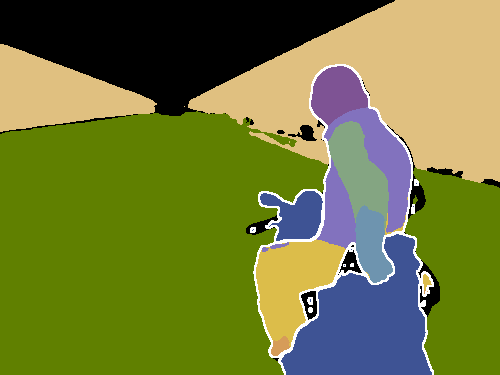}
\\

\includegraphics[width=0.245\linewidth]{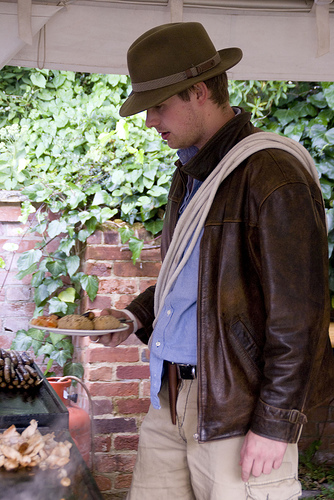}
\includegraphics[width=0.245\linewidth]{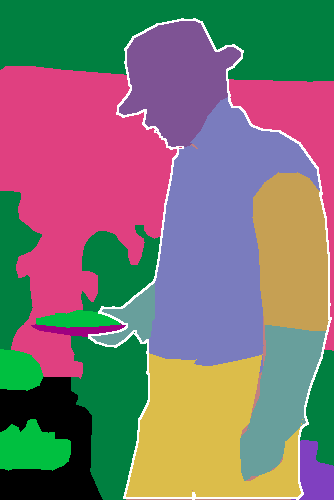}
\includegraphics[width=0.245\linewidth]{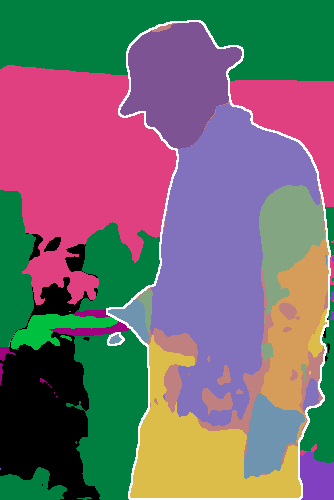}
\includegraphics[width=0.245\linewidth]{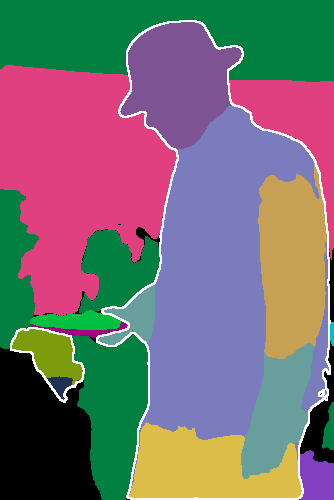}
\\

\includegraphics[width=0.245\linewidth]{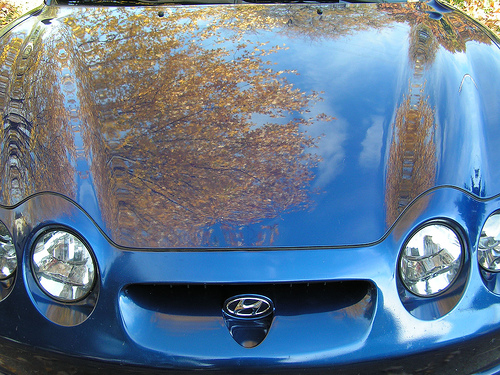}
\includegraphics[width=0.245\linewidth]{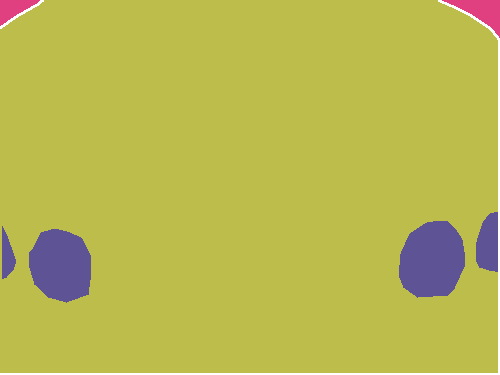}
\includegraphics[width=0.245\linewidth]{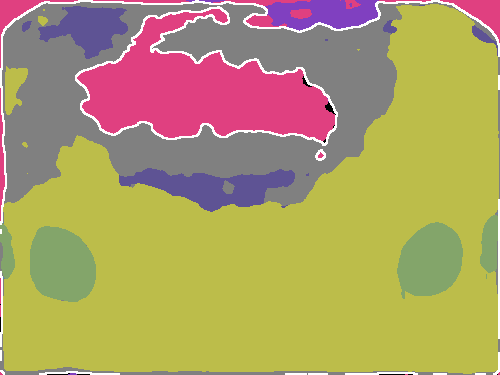}
\includegraphics[width=0.245\linewidth]{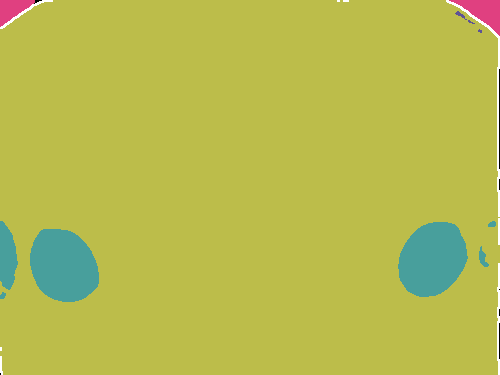}
\\

\includegraphics[width=0.245\linewidth]{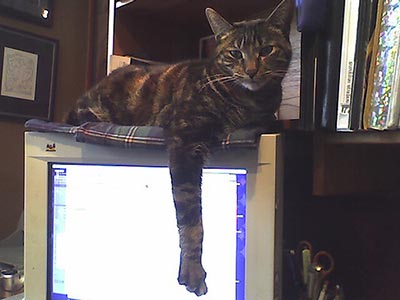}
\includegraphics[width=0.245\linewidth]{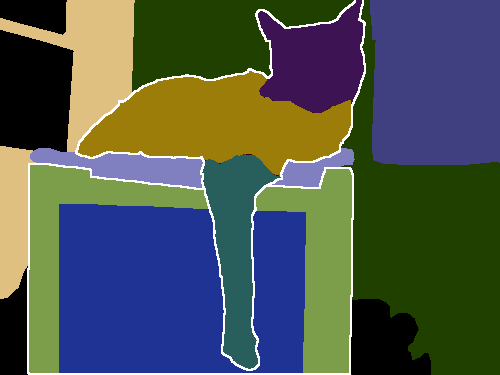}
\includegraphics[width=0.245\linewidth]{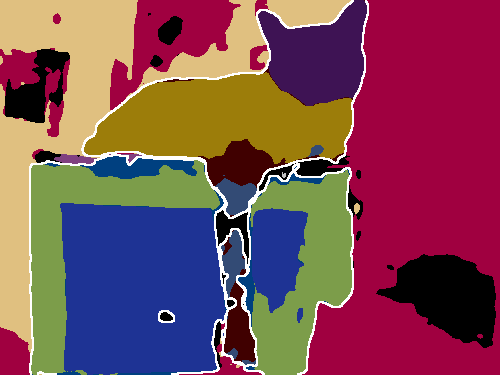}
\includegraphics[width=0.245\linewidth]{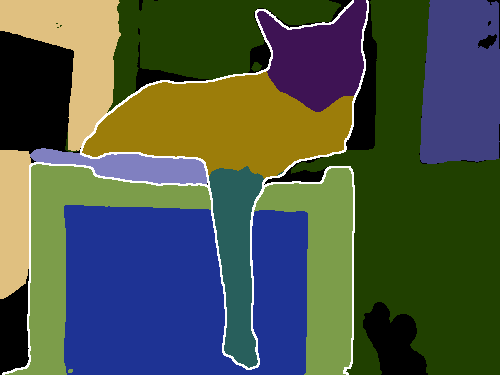}
\\

\begin{subfigure}[b]{0.245\textwidth}
 \centering
 \caption{Input image}
\end{subfigure}
\begin{subfigure}[b]{0.245\textwidth}
 \centering
 \caption{Ground truth}
\end{subfigure}
\begin{subfigure}[b]{0.245\textwidth}
 \centering
 \caption{Panoptic-PartFormer~\cite{li2022ppf}}
\end{subfigure}
\begin{subfigure}[b]{0.245\textwidth}
 \centering
 \caption{TAPPS (ours)}
\end{subfigure}

\caption{\textbf{Qualitative examples of TAPPS and Panoptic-PartFormer~\cite{li2022ppf} on Pascal-PP~\cite{chen2014pascalpart,everingham2010pascal,mottaghi14pascalcontext,degeus2021pps}.} Both networks use ResNet-50~\cite{he2016resnet} with COCO pre-training~\cite{lin2014coco}. White borders separate different object-level instances; color shades indicate different categories. Note that the colors of part-level categories are not identical across instances; there are different shades of the same color. Best viewed digitally.}
\label{supp:fig:qual_results_sota_ppp}
\end{figure*}

\begin{figure*}[t]
\centering

\includegraphics[width=0.245\linewidth, trim={4cm 5cm 16cm 5cm},clip]{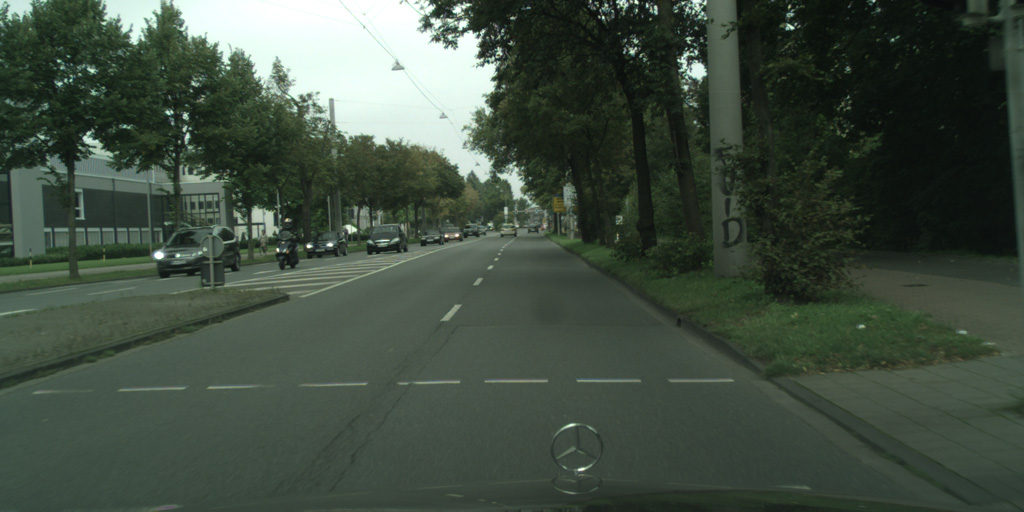}
\includegraphics[width=0.245\linewidth, trim={8cm 10cm 32cm 10cm},clip]{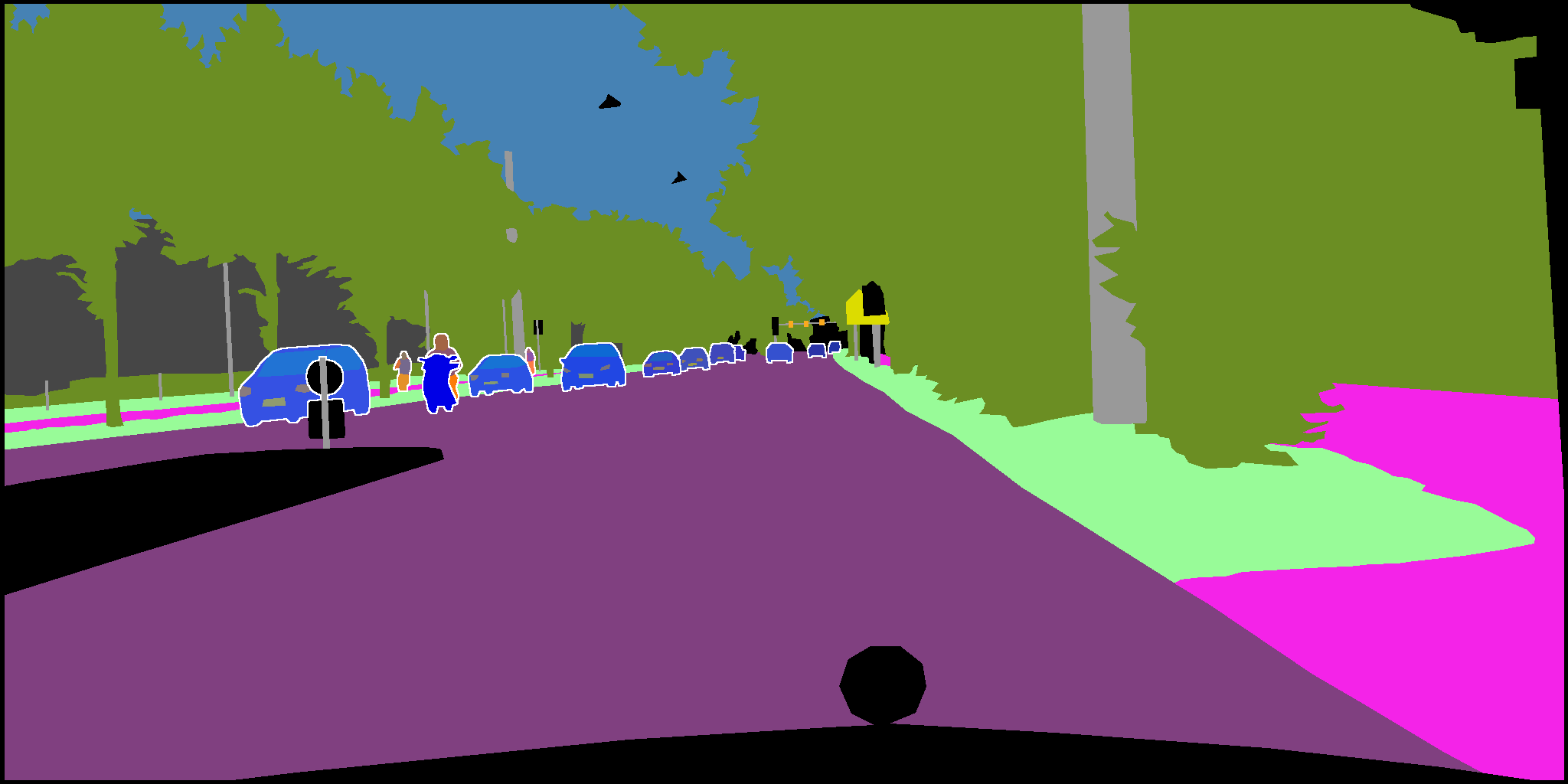}
\includegraphics[width=0.245\linewidth, trim={8cm 10cm 32cm 10cm},clip]{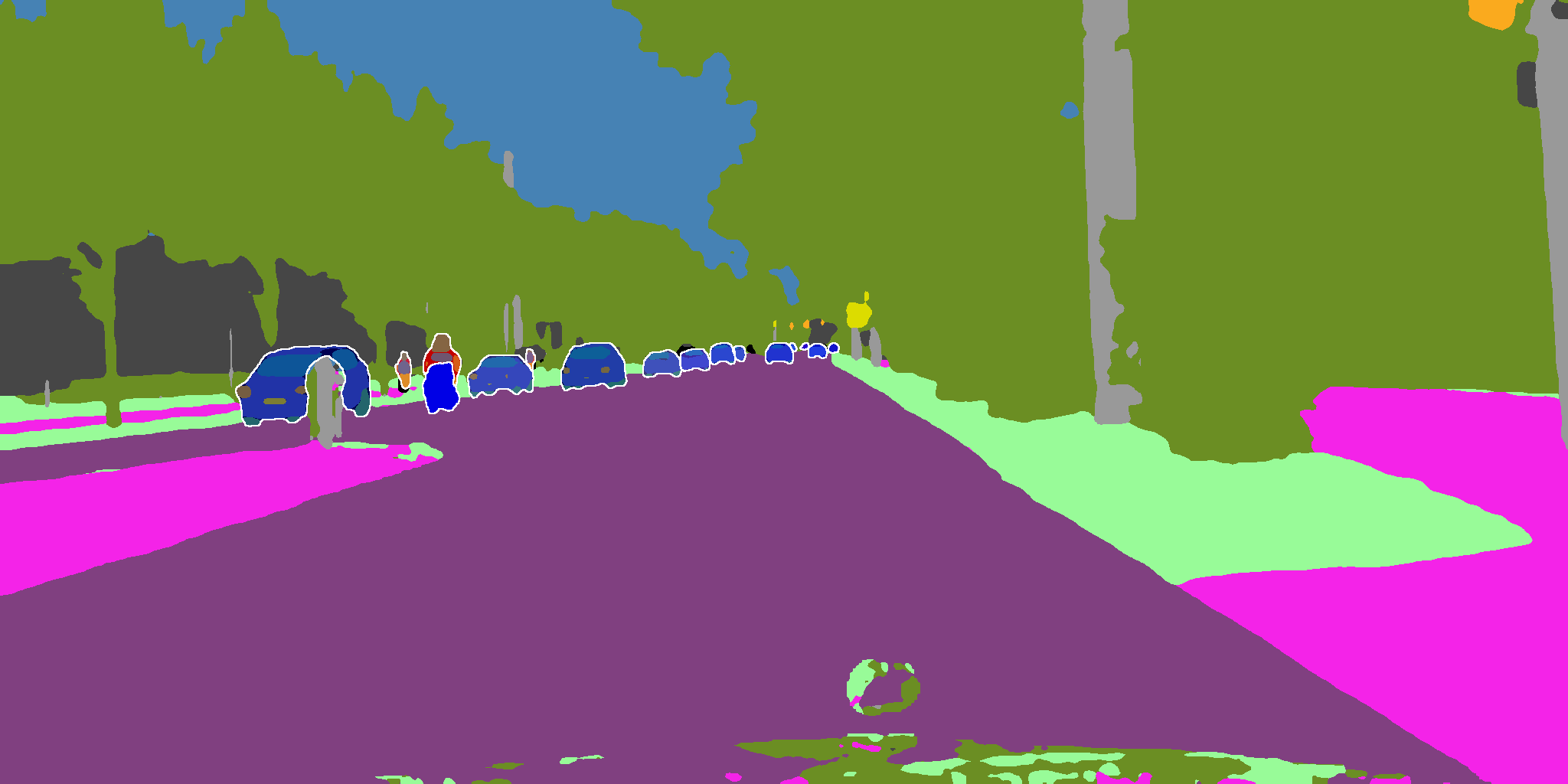}
\begin{tikzpicture}
    \node[anchor=south west,inner sep=0] (image) at (0,0) {\includegraphics[width=0.245\linewidth, trim={8cm 10cm 32cm 10cm},clip]{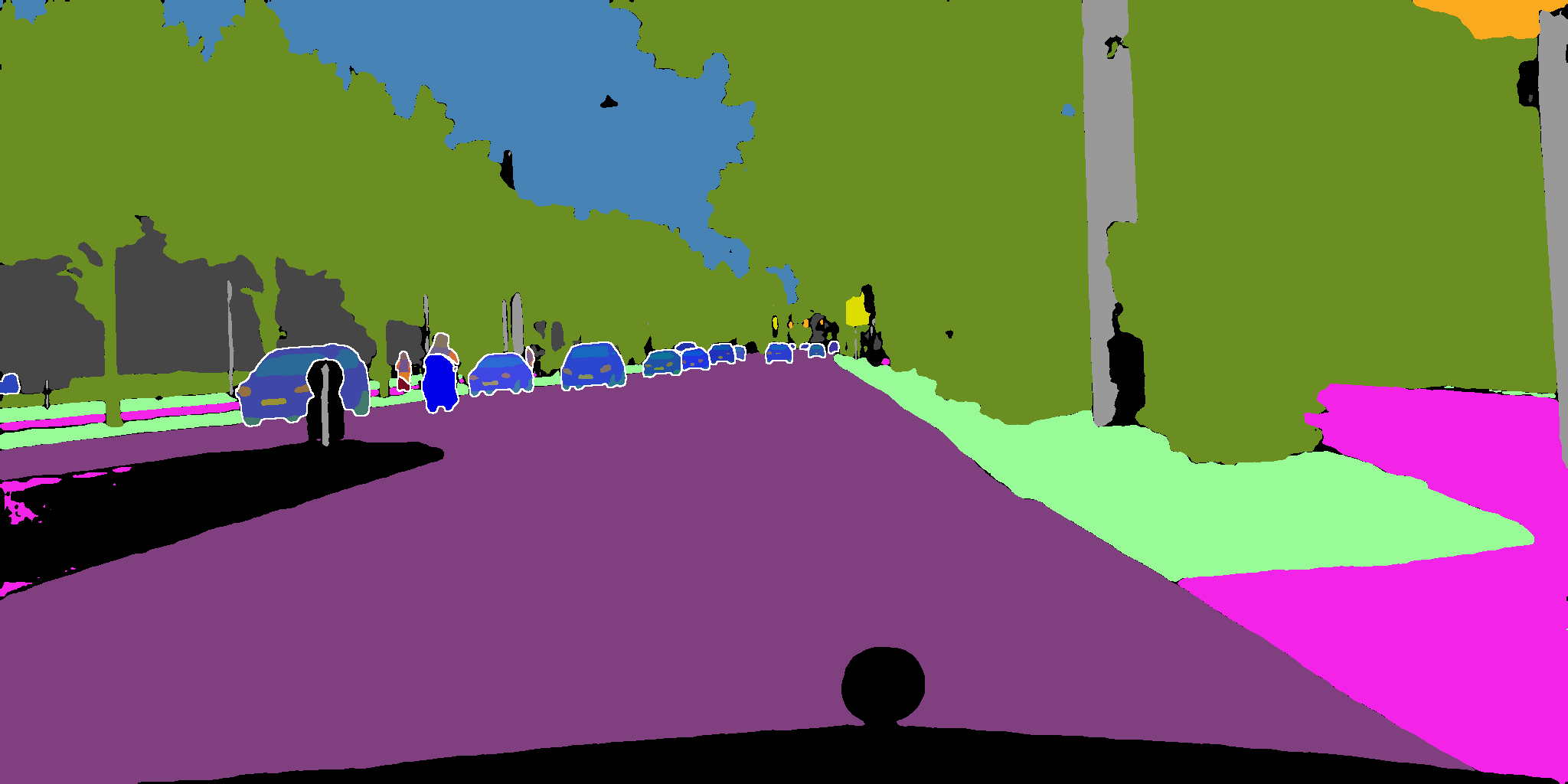}};
    \begin{scope}[x={(image.south east)},y={(image.north west)}]
        \draw[red,line width=0.5mm,rounded corners] (0.4,0.45) rectangle (0.85,0.7);
    \end{scope}
\end{tikzpicture}
\\

\includegraphics[width=0.245\linewidth]{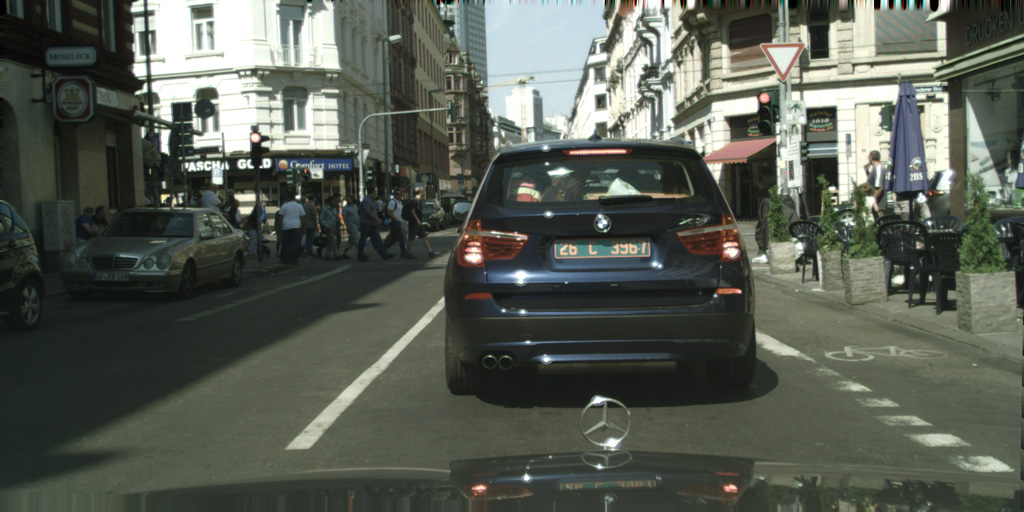}
\includegraphics[width=0.245\linewidth]{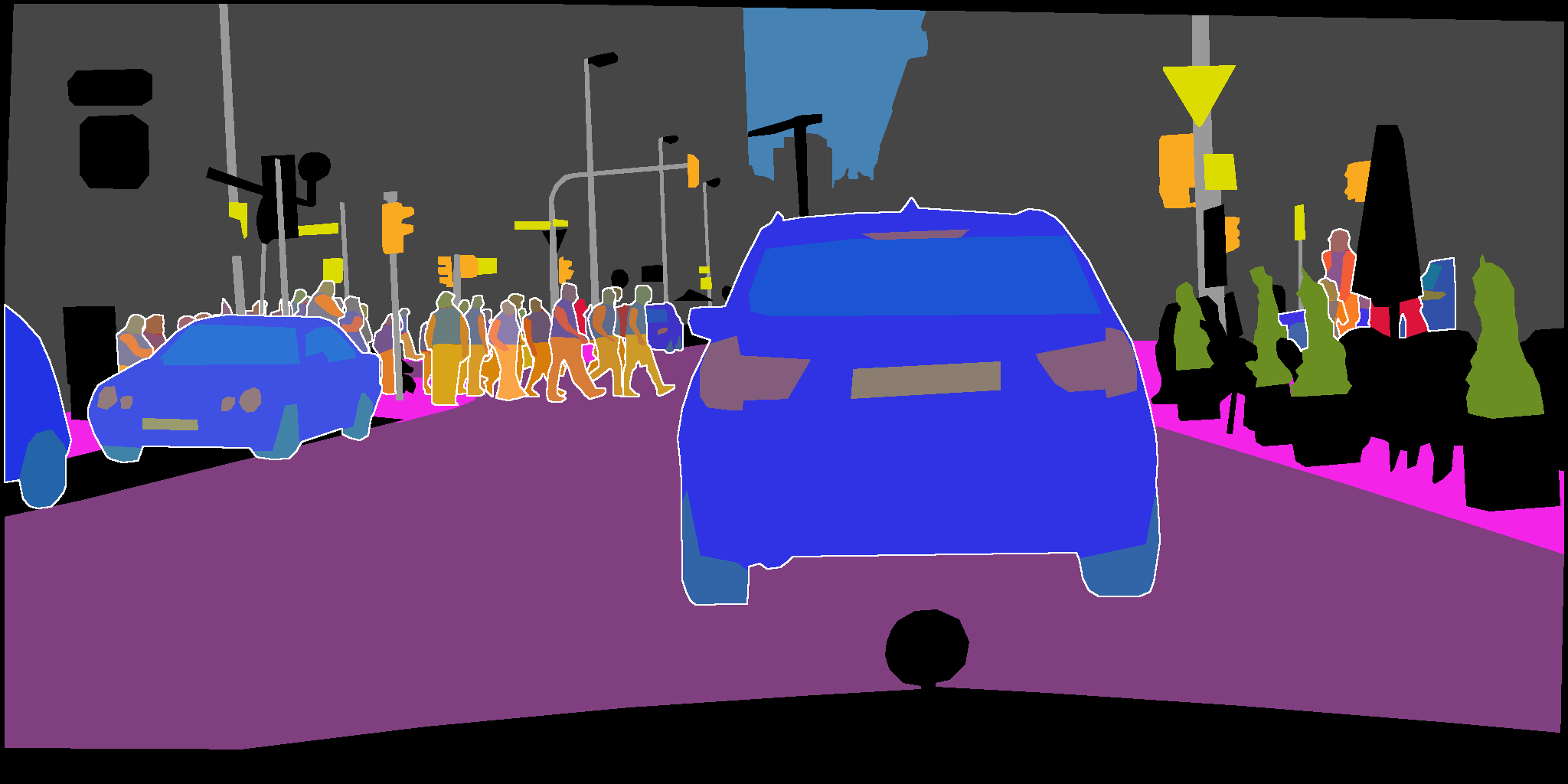}
\includegraphics[width=0.245\linewidth]{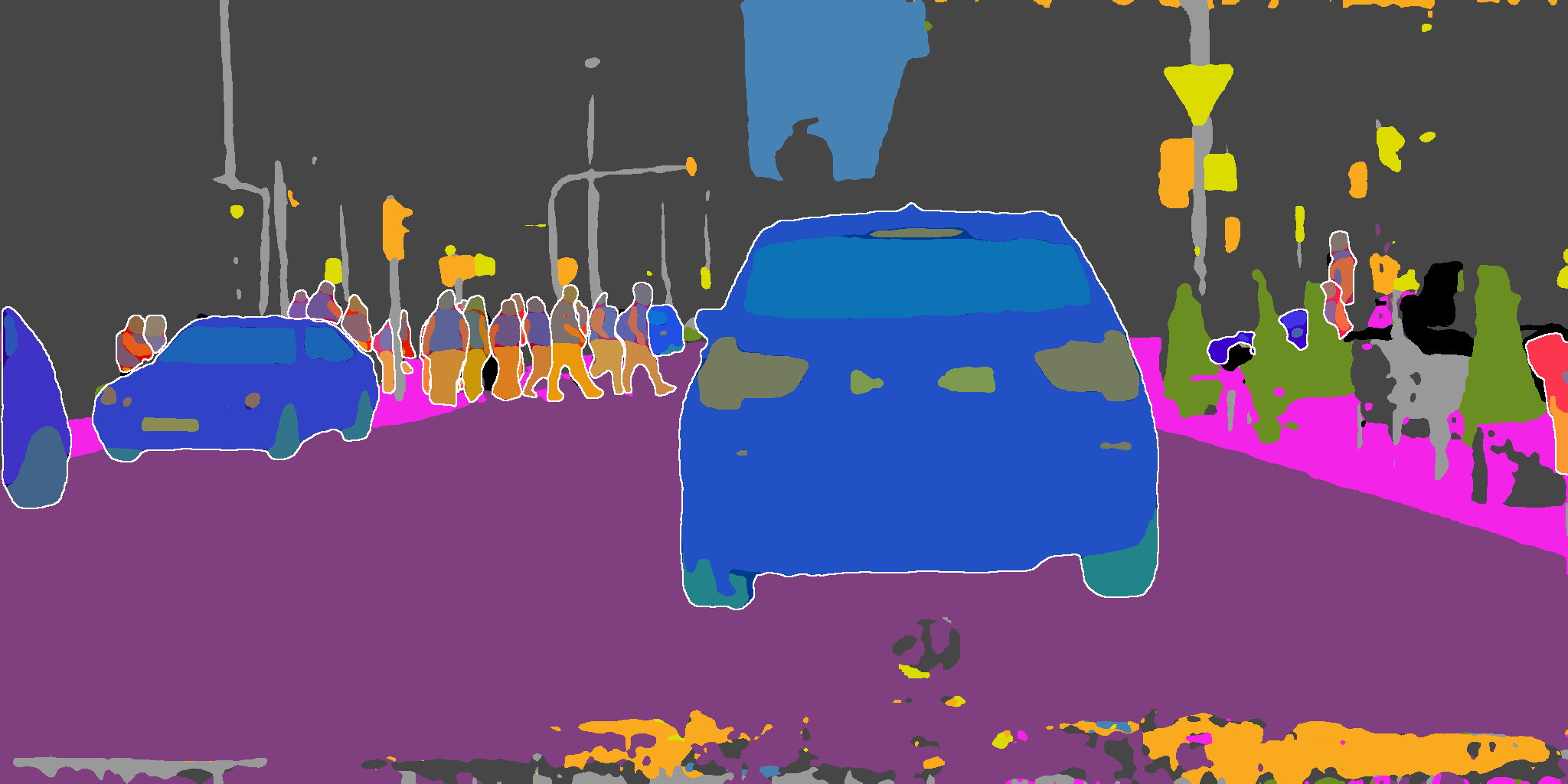}
\begin{tikzpicture}
    \node[anchor=south west,inner sep=0] (image) at (0,0) {\includegraphics[width=0.245\linewidth]{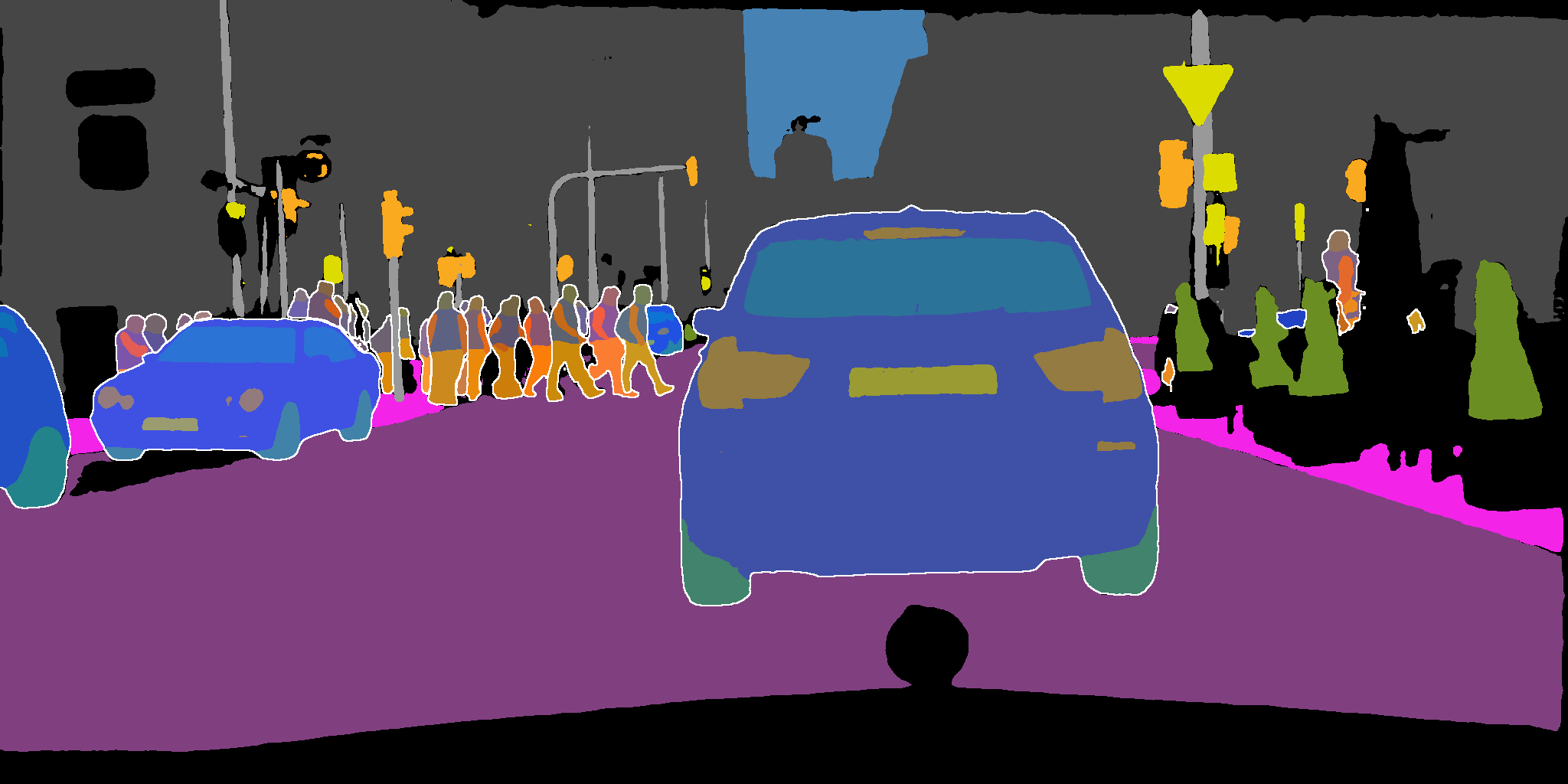}};
    \begin{scope}[x={(image.south east)},y={(image.north west)}]
        \draw[red,line width=0.5mm,rounded corners] (0.41,0.2) rectangle (0.76,0.79);
    \end{scope}
\end{tikzpicture}
\\

\includegraphics[width=0.245\linewidth, trim={14cm 5cm 0 0},clip]{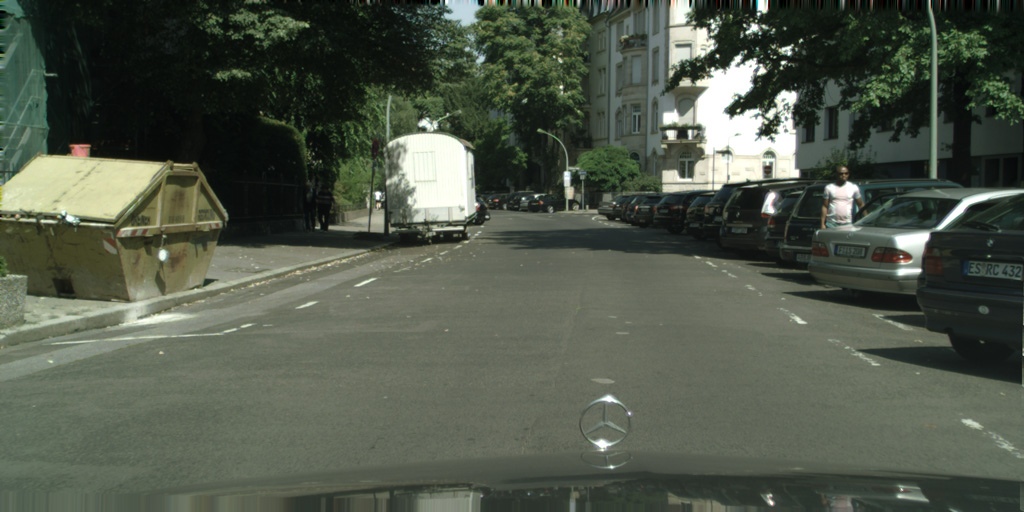}
\includegraphics[width=0.245\linewidth, trim={28cm 10cm 0 0},clip]{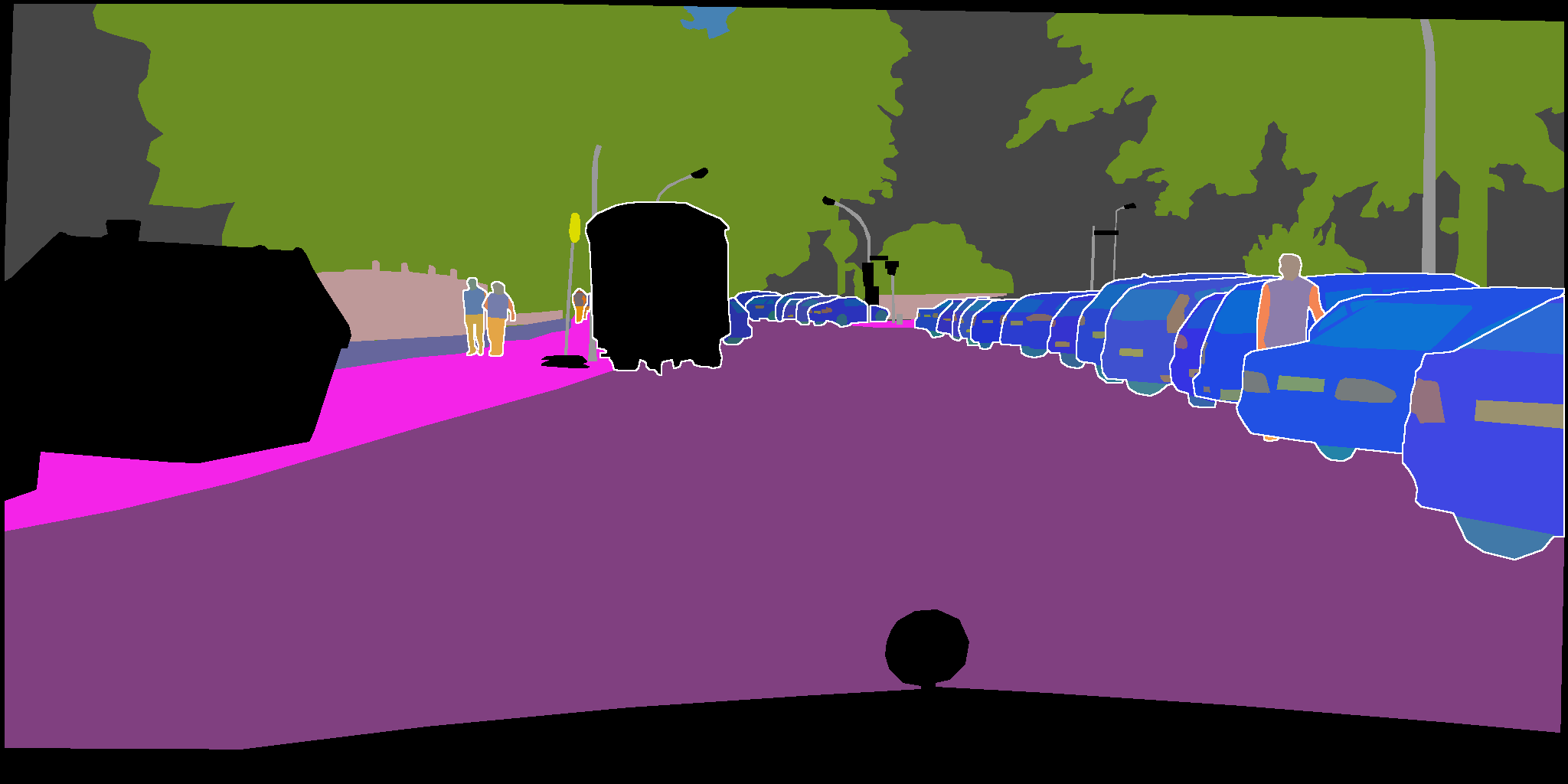}
\includegraphics[width=0.245\linewidth, trim={28cm 10cm 0 0},clip]{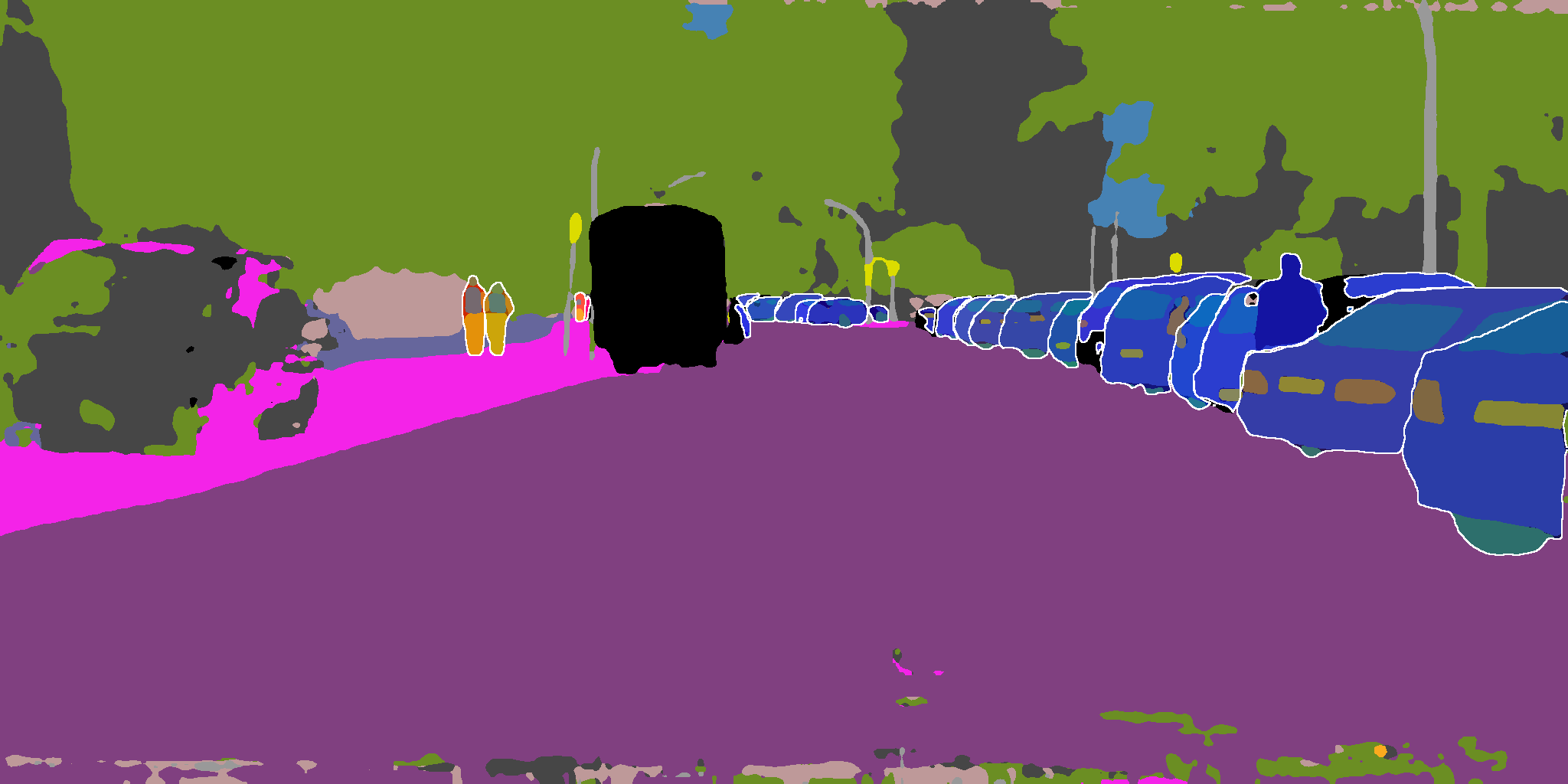}
\begin{tikzpicture}
    \node[anchor=south west,inner sep=0] (image) at (0,0) {\includegraphics[width=0.245\linewidth, trim={28cm 10cm 0 0},clip]{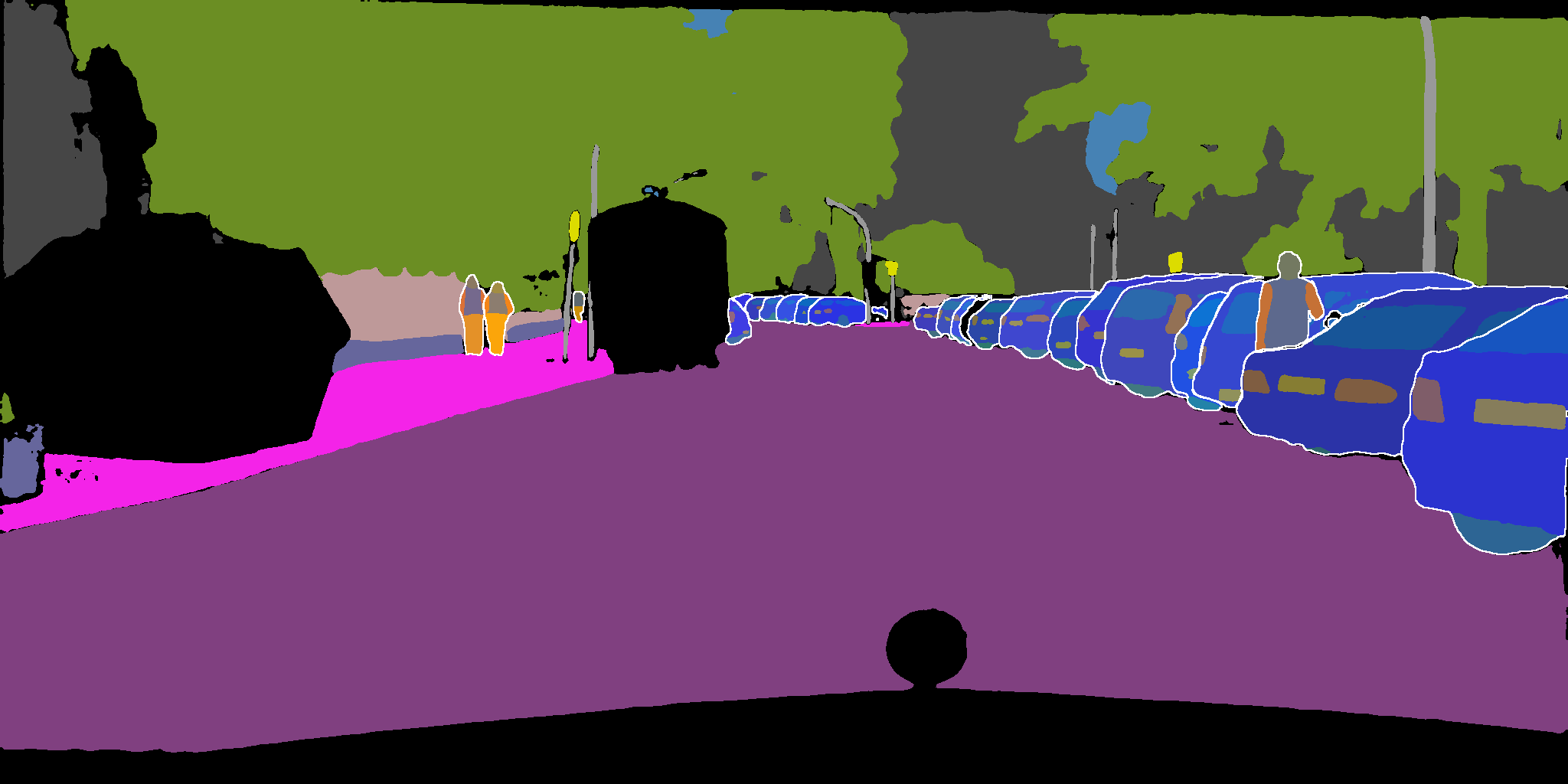}};
    \begin{scope}[x={(image.south east)},y={(image.north west)}]
        \draw[red,line width=0.5mm,rounded corners] (0.3,0.05) rectangle (0.99,0.6);
    \end{scope}
\end{tikzpicture}
\\

\includegraphics[width=0.245\linewidth, trim={14cm 5cm 0 0},clip]{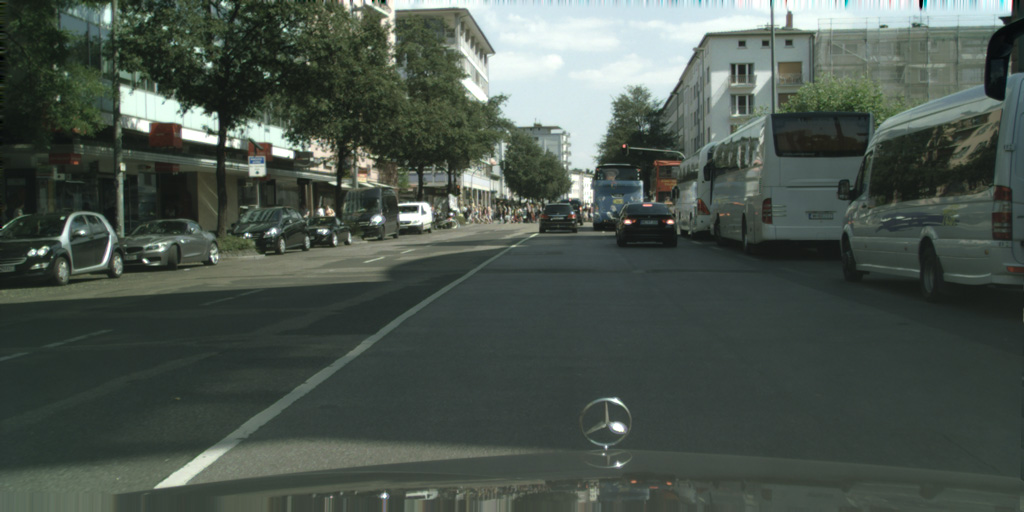}
\includegraphics[width=0.245\linewidth, trim={28cm 10cm 0 0},clip]{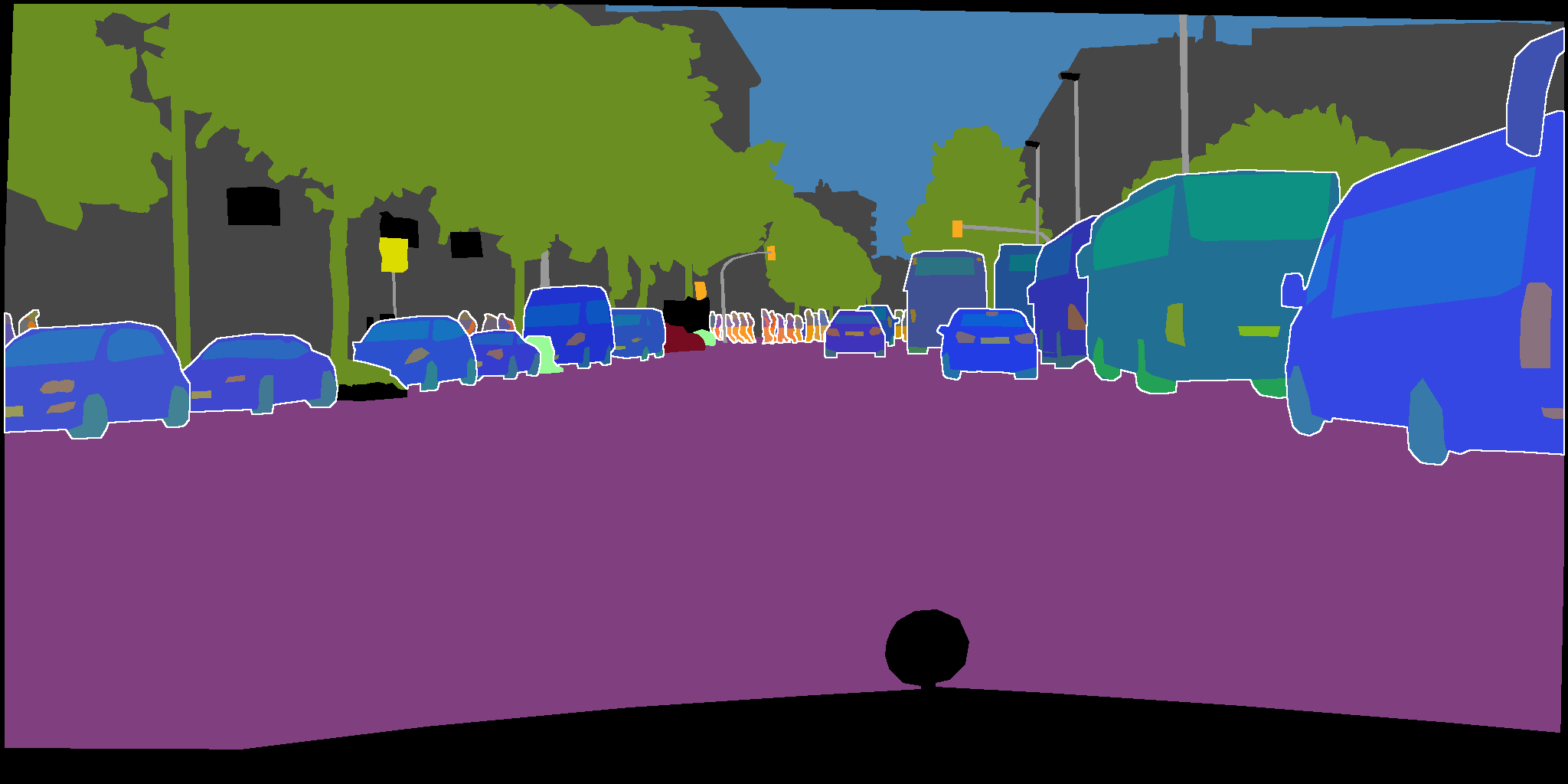}
\includegraphics[width=0.245\linewidth, trim={28cm 10cm 0 0},clip]{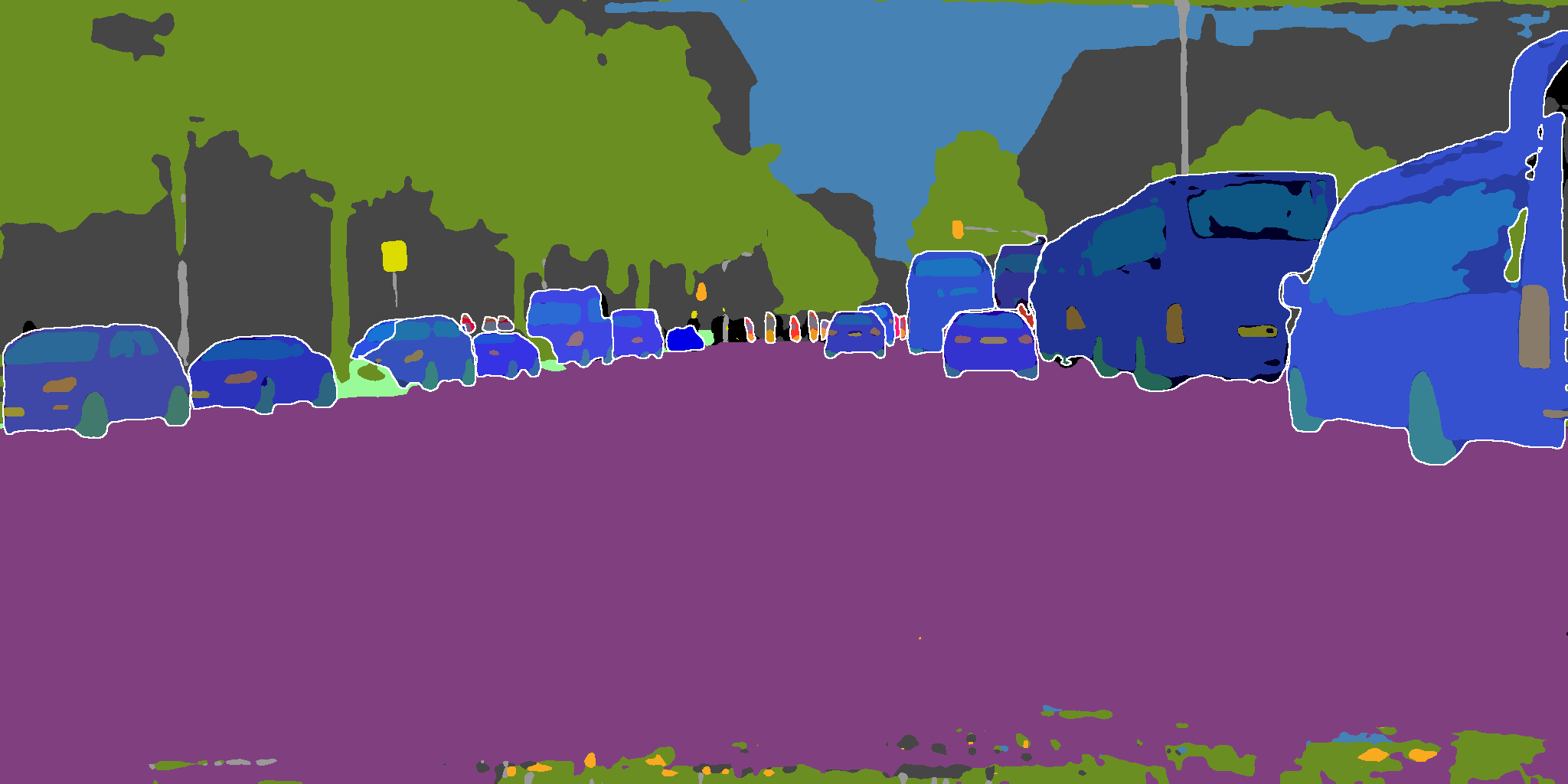}
\begin{tikzpicture}
    \node[anchor=south west,inner sep=0] (image) at (0,0) {\includegraphics[width=0.245\linewidth, trim={28cm 10cm 0 0},clip]{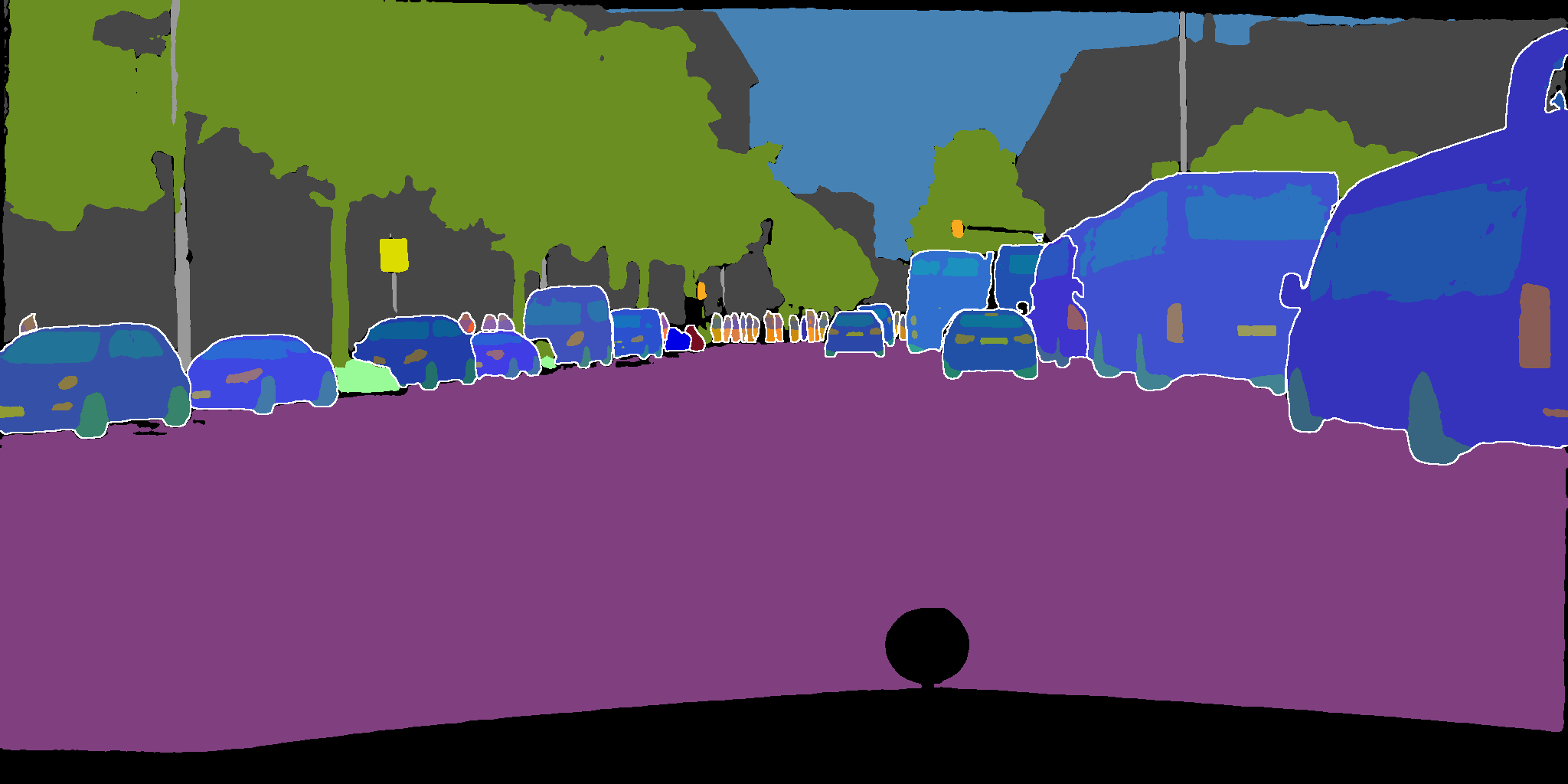}};
    \begin{scope}[x={(image.south east)},y={(image.north west)}]
        \draw[red,line width=0.5mm,rounded corners] (0.3,0.2) rectangle (0.95,0.75);
    \end{scope}
\end{tikzpicture}
\\

\includegraphics[width=0.245\linewidth]{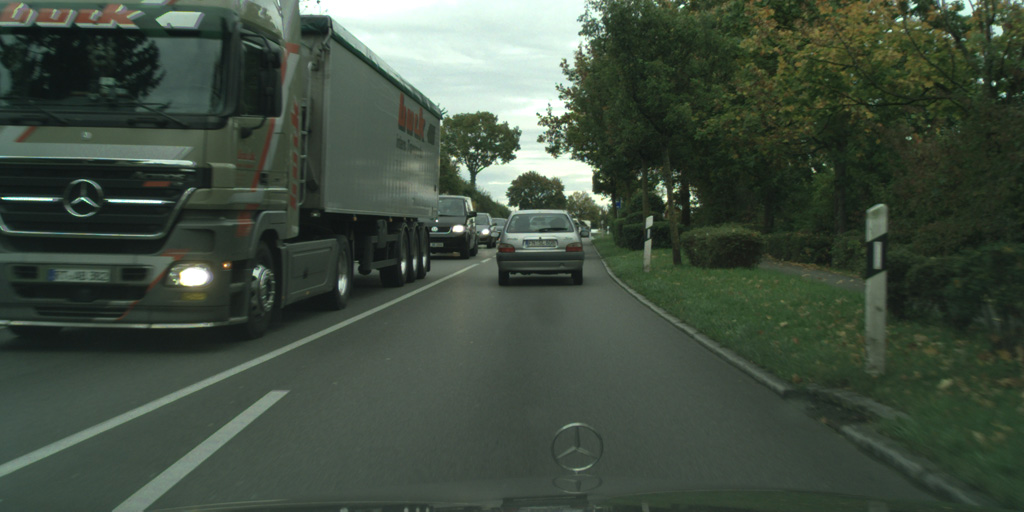}
\includegraphics[width=0.245\linewidth]{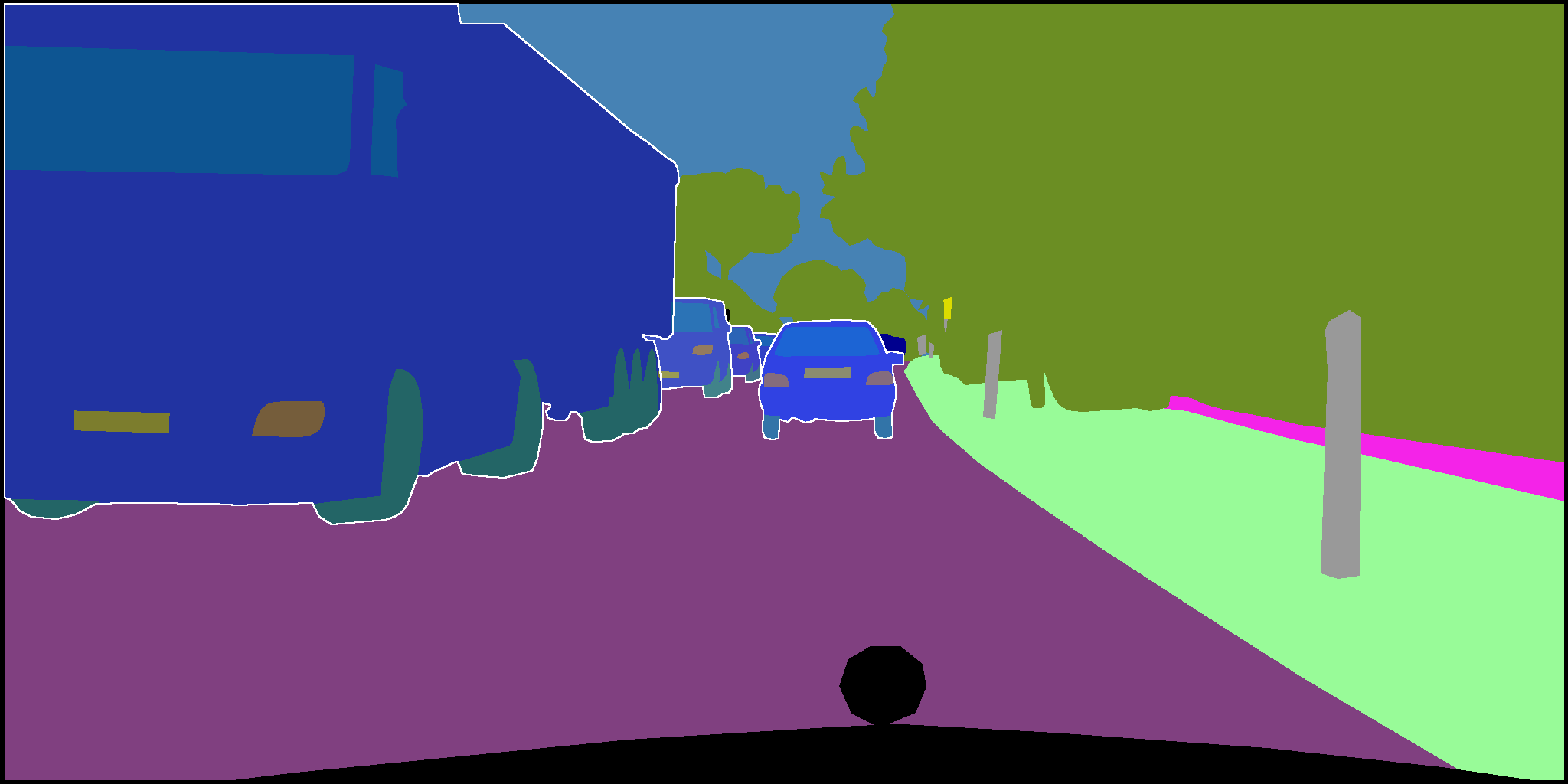}
\includegraphics[width=0.245\linewidth]{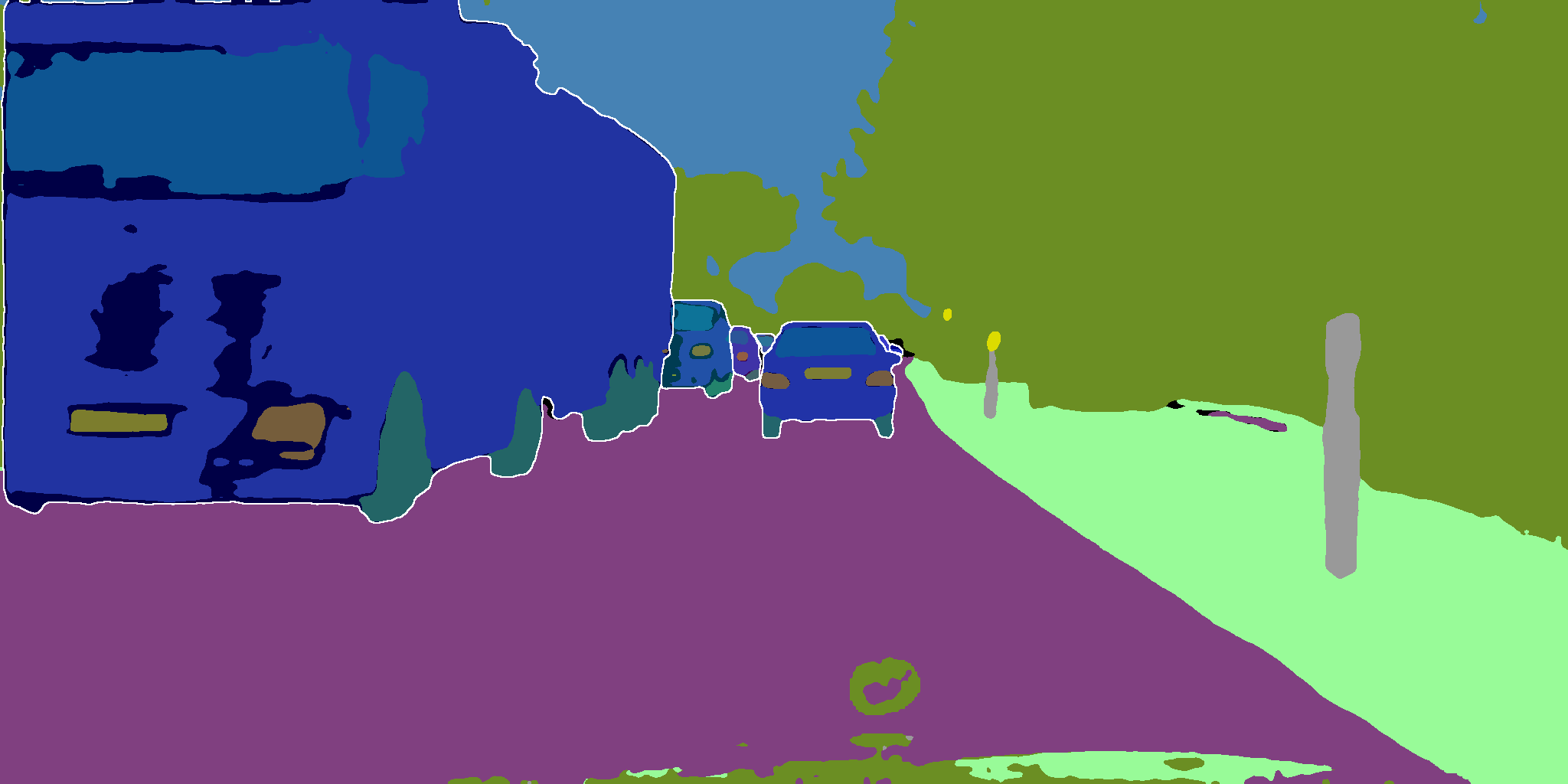}
\begin{tikzpicture}
    \node[anchor=south west,inner sep=0] (image) at (0,0) {\includegraphics[width=0.245\linewidth]{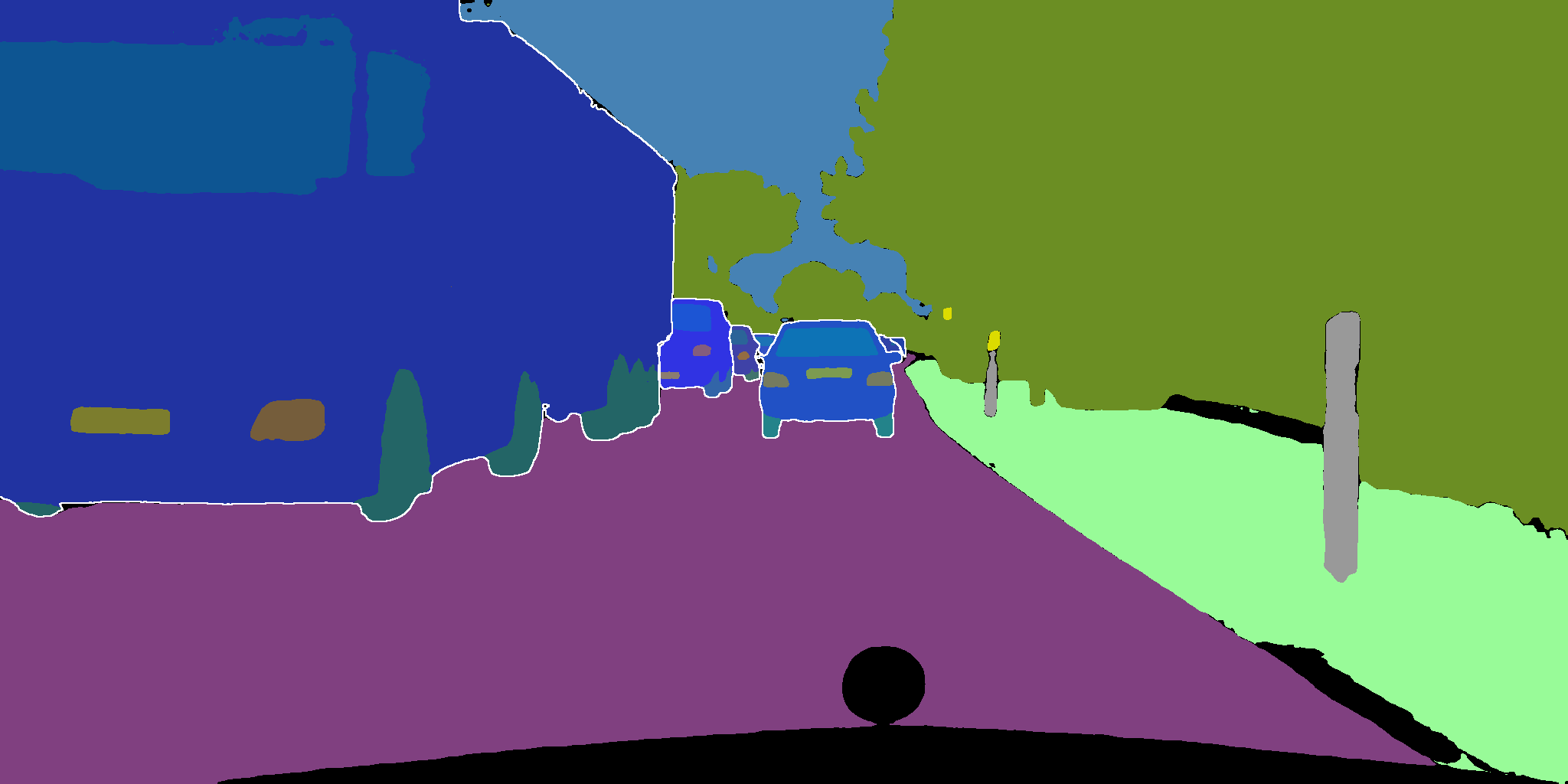}};
    \begin{scope}[x={(image.south east)},y={(image.north west)}]
        \draw[red,line width=0.5mm,rounded corners] (0.01,0.25) rectangle (0.49,0.99);
    \end{scope}
\end{tikzpicture}
\\

\includegraphics[width=0.245\linewidth, trim={10cm 5cm 0 0},clip]{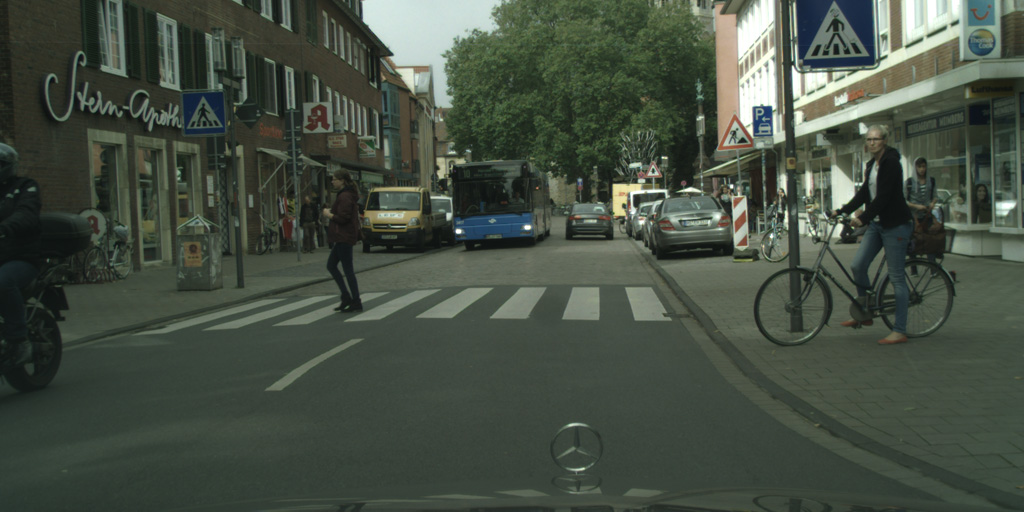}
\includegraphics[width=0.245\linewidth, trim={20cm 10cm 0 0},clip]{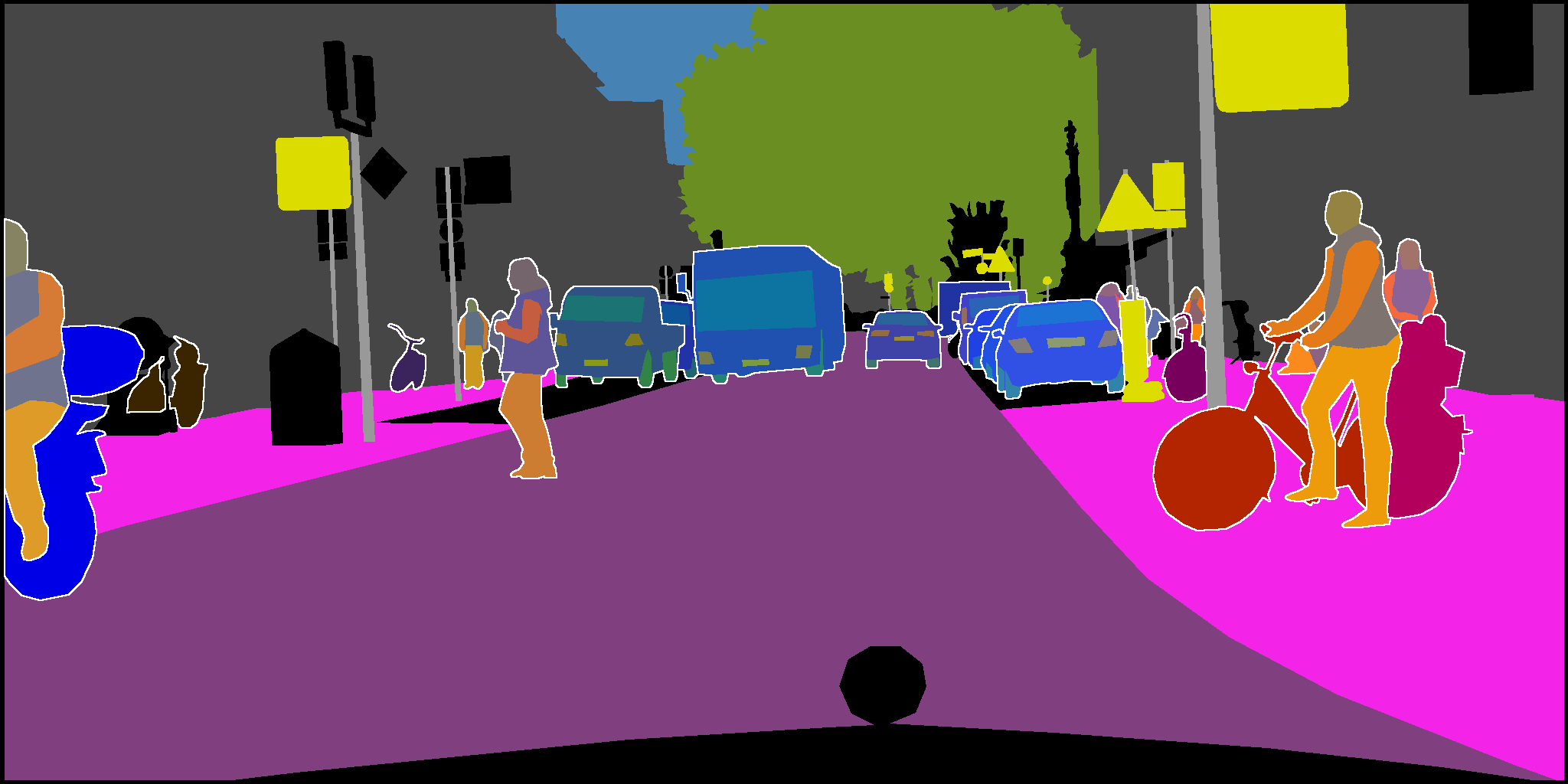}
\includegraphics[width=0.245\linewidth, trim={20cm 10cm 0 0},clip]{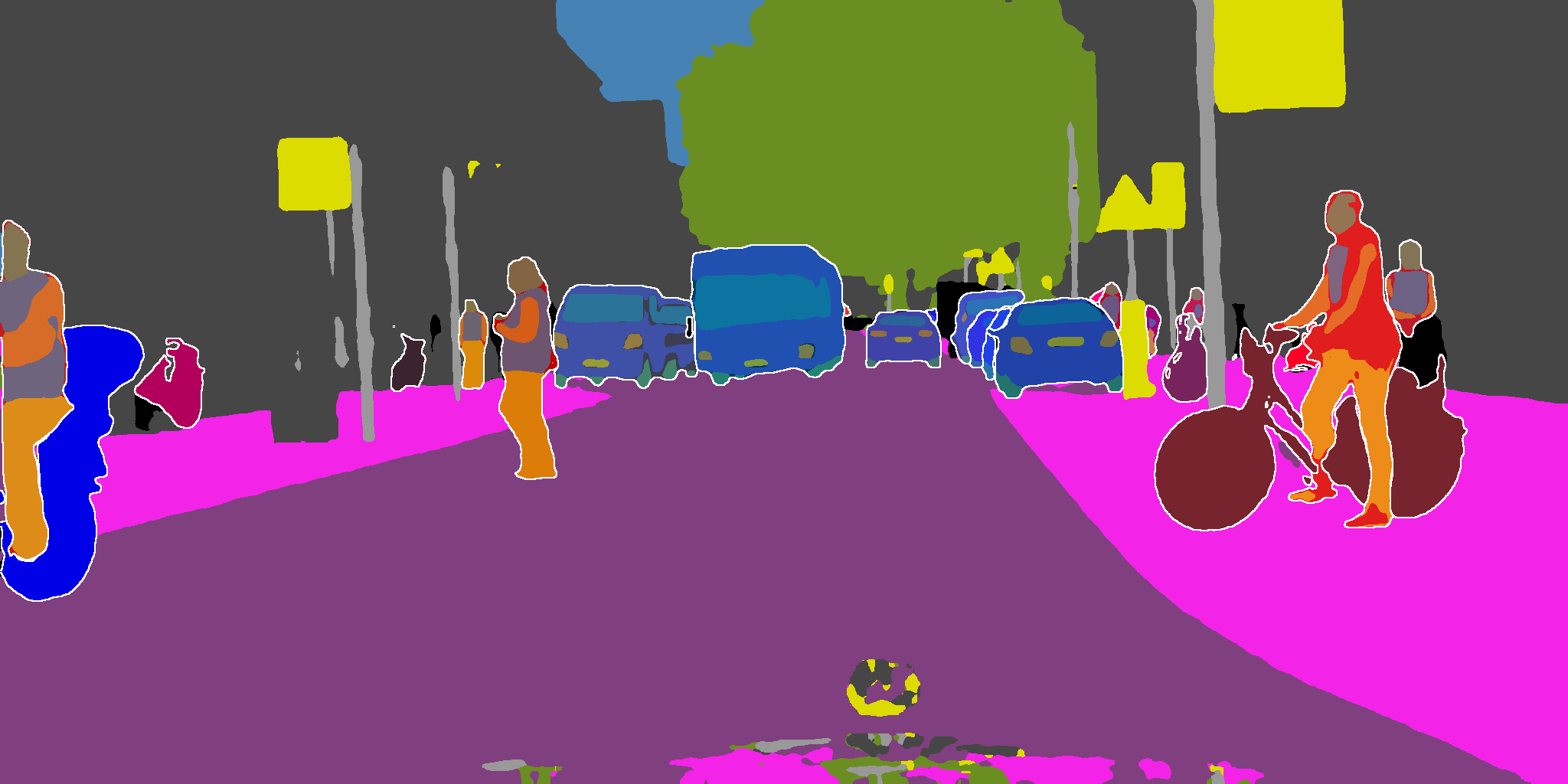}
\begin{tikzpicture}
    \node[anchor=south west,inner sep=0] (image) at (0,0) {\includegraphics[width=0.245\linewidth, trim={20cm 10cm 0 0},clip]{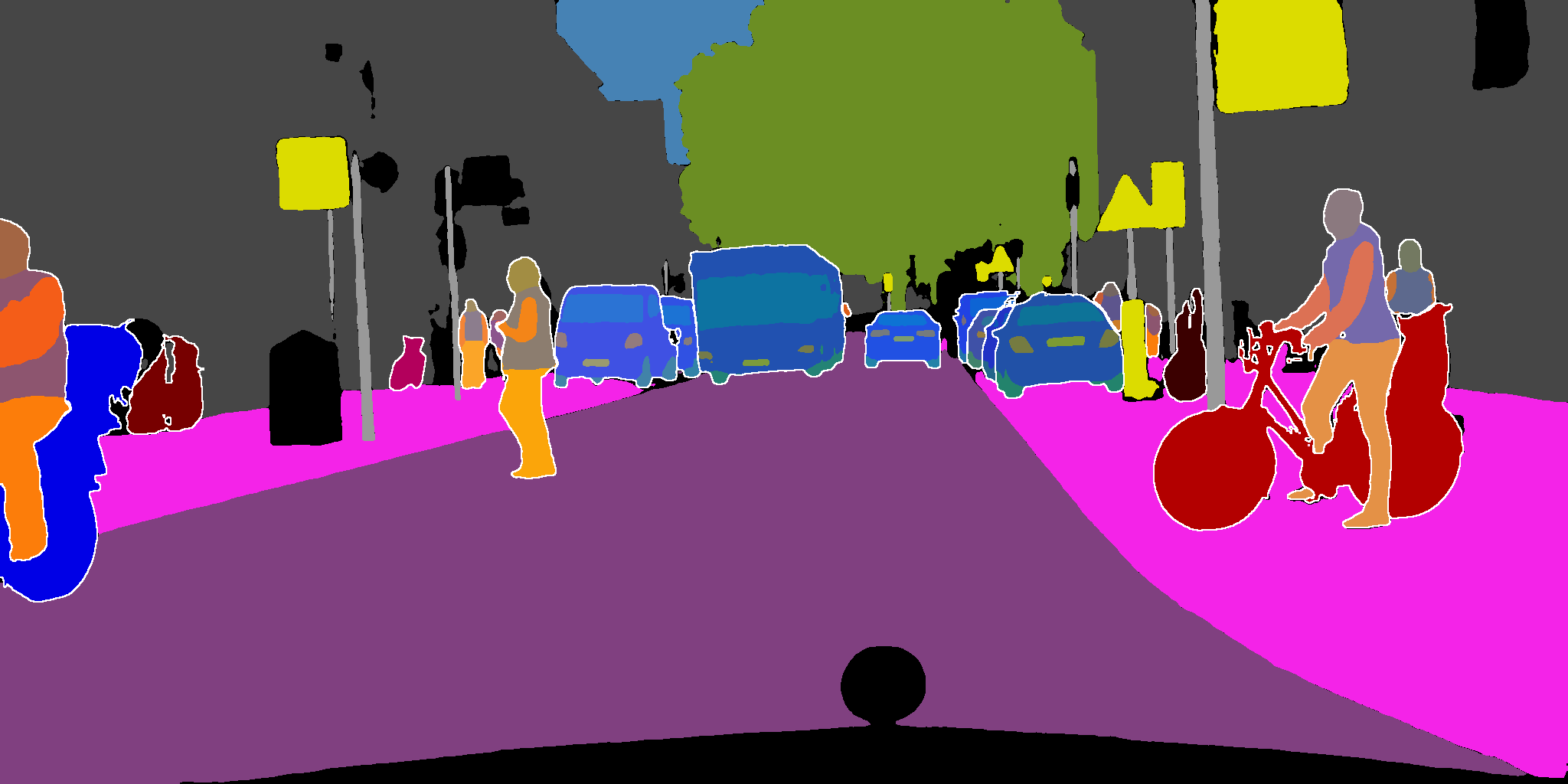}};
    \begin{scope}[x={(image.south east)},y={(image.north west)}]
        \draw[red,line width=0.5mm,rounded corners] (0.61,0.03) rectangle (0.94,0.7);
    \end{scope}
\end{tikzpicture}
\\

\includegraphics[width=0.245\linewidth]{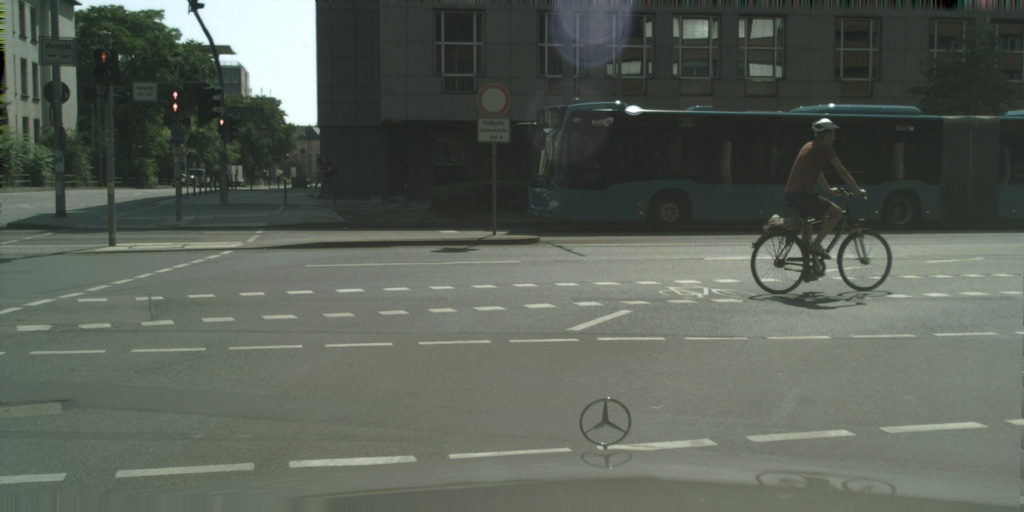}
\includegraphics[width=0.245\linewidth]{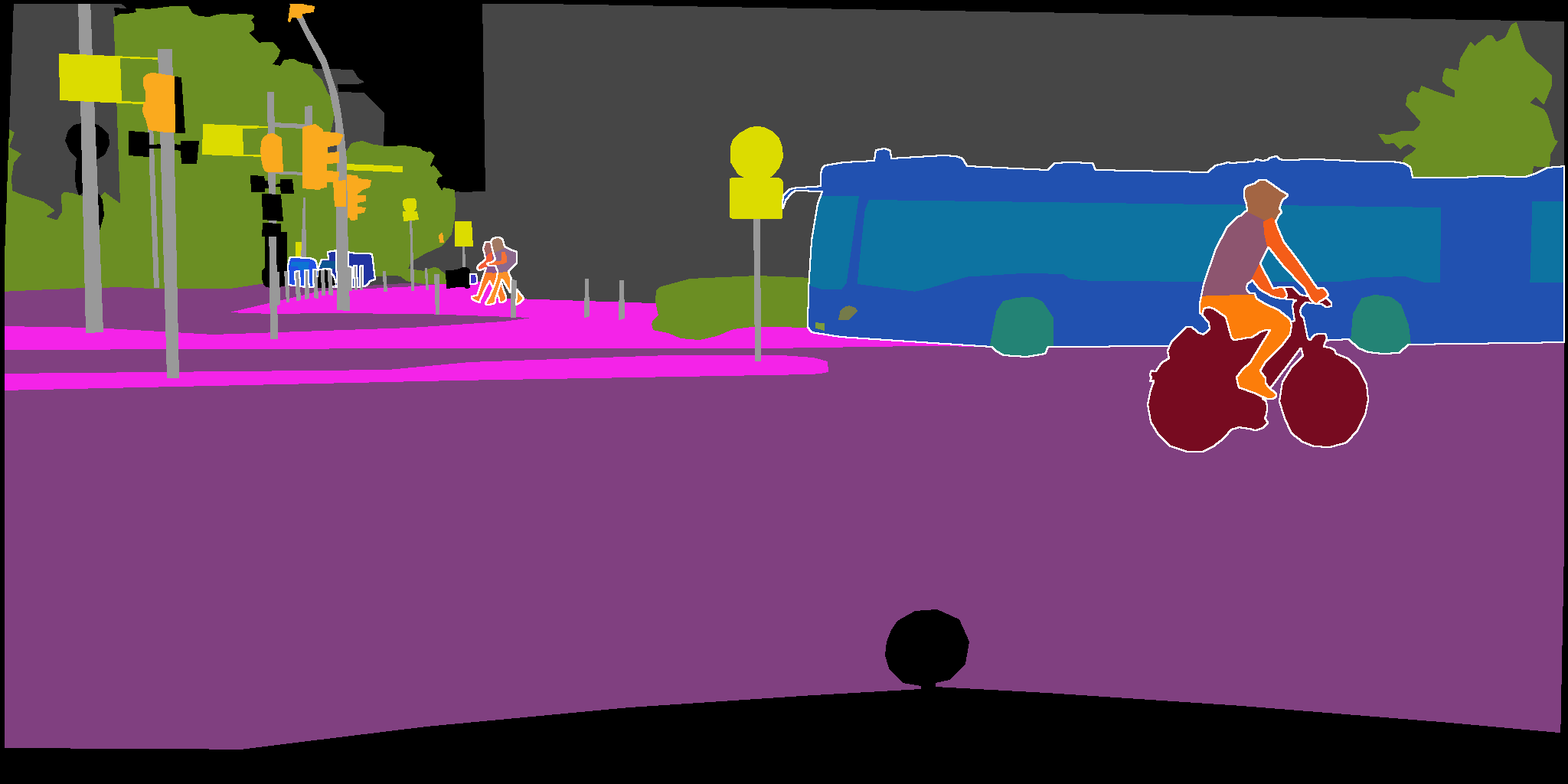}
\includegraphics[width=0.245\linewidth]{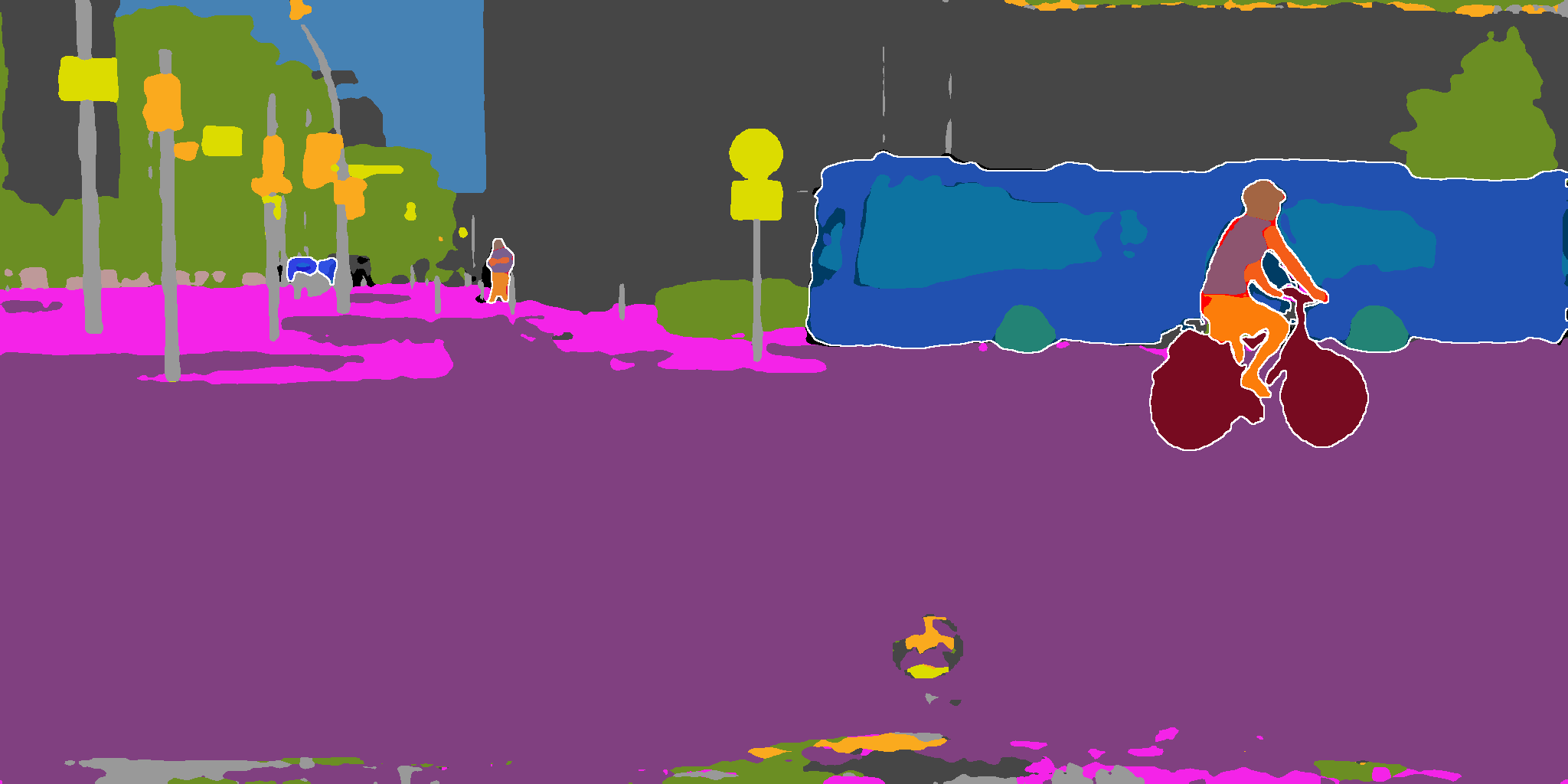}
\begin{tikzpicture}
    \node[anchor=south west,inner sep=0] (image) at (0,0) {\includegraphics[width=0.245\linewidth]{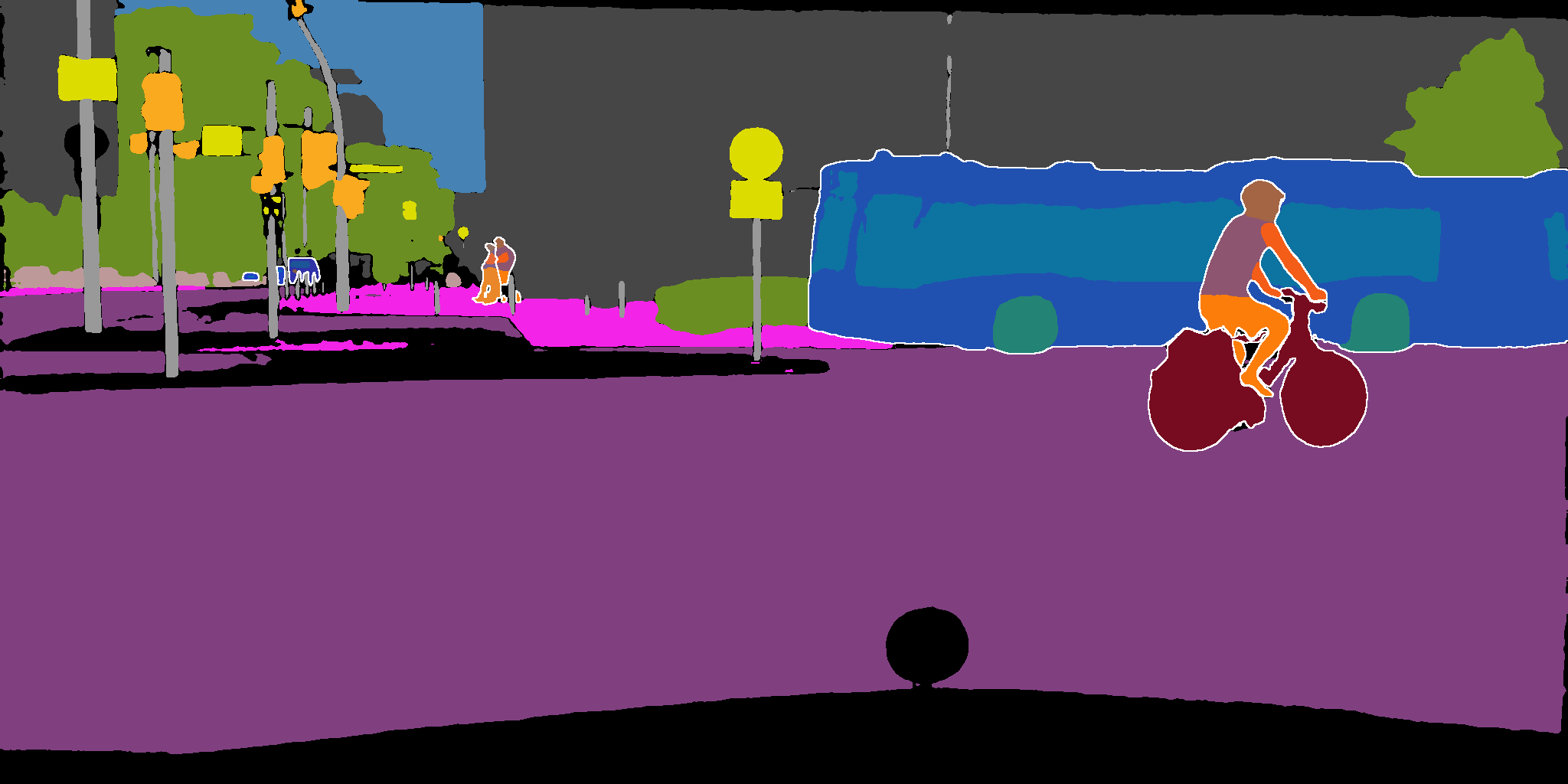}};
    \begin{scope}[x={(image.south east)},y={(image.north west)}]
        \draw[red,line width=0.5mm,rounded corners] (0.50,0.35) rectangle (0.99,0.9);
    \end{scope}
\end{tikzpicture}
\\

\begin{subfigure}[b]{0.245\textwidth}
 \centering
 \caption{Input image}
\end{subfigure}
\begin{subfigure}[b]{0.245\textwidth}
 \centering
 \caption{Ground truth}
\end{subfigure}
\begin{subfigure}[b]{0.245\textwidth}
 \centering
 \caption{Panoptic-PartFormer~\cite{li2022ppf}}
\end{subfigure}
\begin{subfigure}[b]{0.245\textwidth}
 \centering
 \caption{TAPPS (ours)}
\end{subfigure}

\caption{\textbf{Qualitative examples of TAPPS and Panoptic-PartFormer~\cite{li2022ppf} on Cityscapes-PP~\cite{cordts2016cityscapes,degeus2021pps}.} Both networks use ResNet-50~\cite{he2016resnet} with COCO pre-training~\cite{lin2014coco}. White borders separate different object-level instances; color shades indicate different categories. Note that the colors of part-level categories are not identical across instances; there are different shades of the same color. Best viewed digitally.}
\label{supp:fig:qual_results_sota_cpp}
\end{figure*}

Comparing TAPPS to state-of-the-art existing model Panoptic-PartFormer~\cite{li2022ppf} in \cref{supp:fig:qual_results_sota_ppp} and \cref{supp:fig:qual_results_sota_cpp}, we observe even more significant differences. In addition to the object-level segmentation quality, the part segmentation quality within objects is considerably better for TAPPS. This applies to large and small objects across different classes.

\begin{figure*}[t]
\centering

\includegraphics[width=0.320\linewidth]{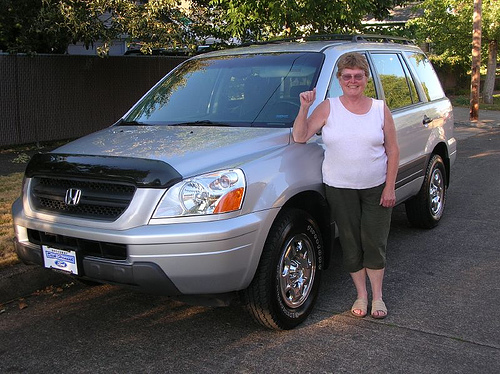}
\includegraphics[width=0.320\linewidth]{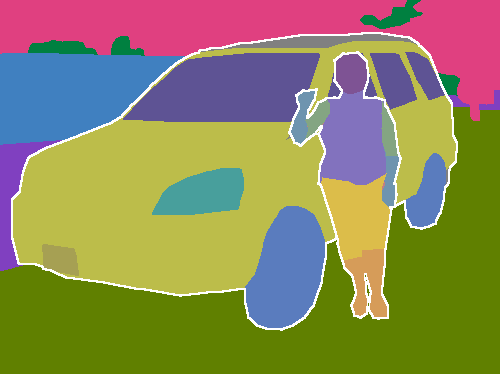}
\includegraphics[width=0.320\linewidth]{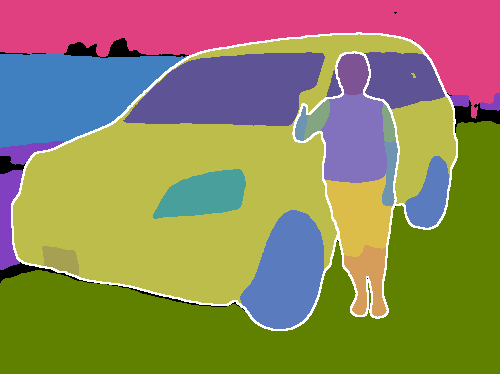}
\\

\includegraphics[width=0.320\linewidth]{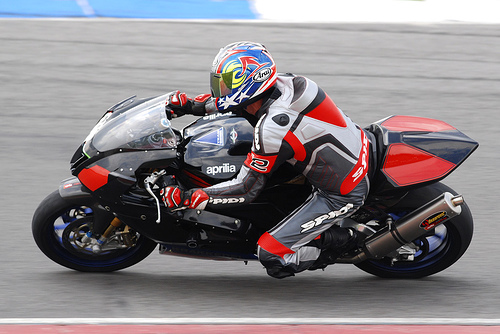}
\includegraphics[width=0.320\linewidth]{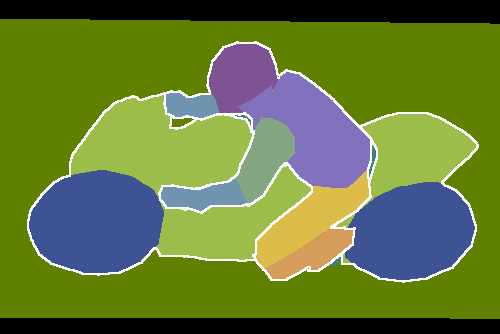}
\includegraphics[width=0.320\linewidth]{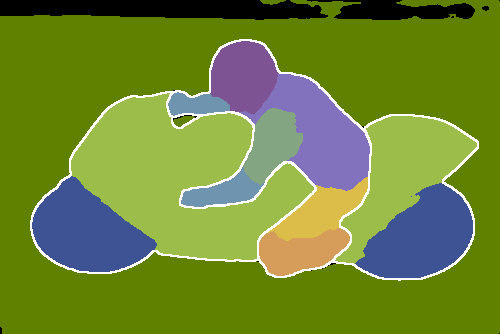}
\\

\includegraphics[width=0.320\linewidth, trim={0 0 0 3cm},clip]{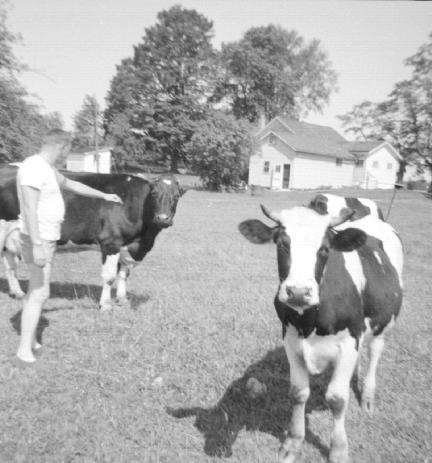}
\includegraphics[width=0.320\linewidth, trim={0 0 0 3cm},clip]{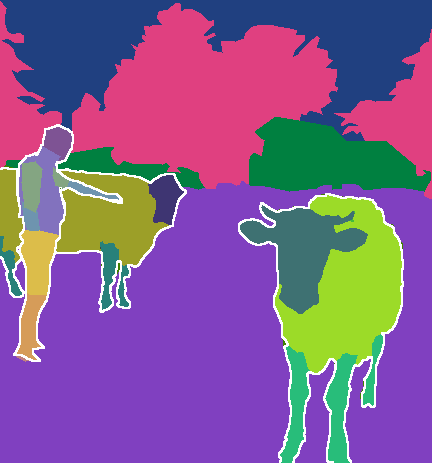}
\includegraphics[width=0.320\linewidth, trim={0 0 0 3cm},clip]{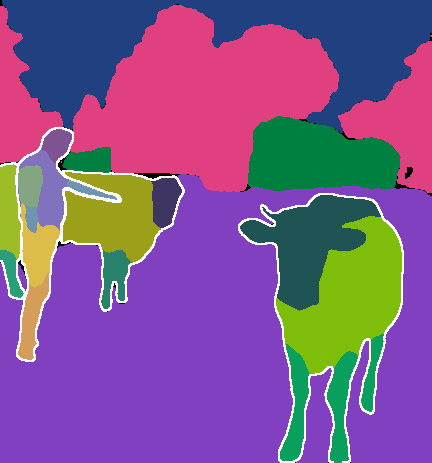}
\\

\includegraphics[width=0.320\linewidth]{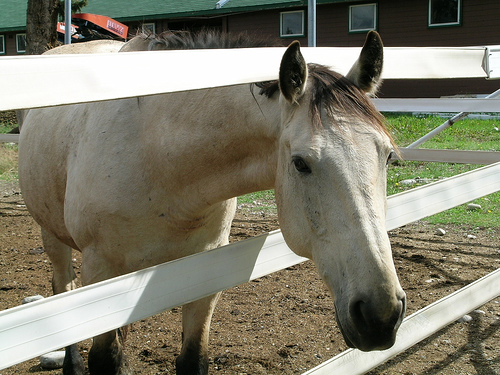}
\includegraphics[width=0.320\linewidth]{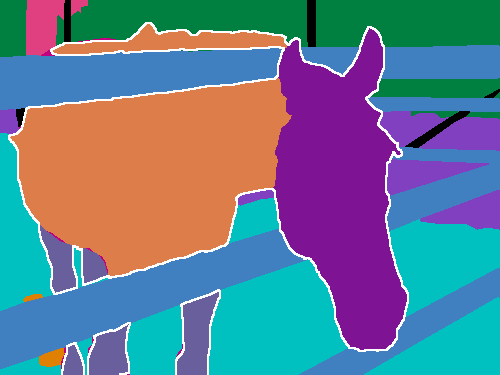}
\includegraphics[width=0.320\linewidth]{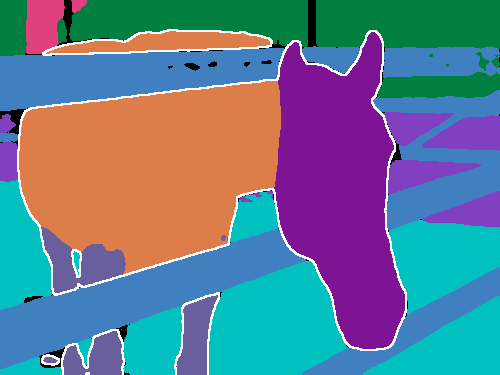}
\\

\includegraphics[width=0.320\linewidth, trim={0 0 0 .25cm},clip]{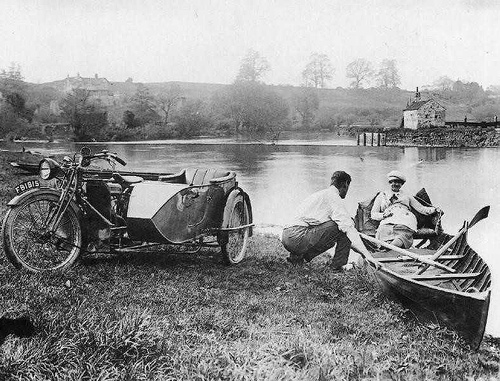}
\includegraphics[width=0.320\linewidth, trim={0 0 0 1cm},clip]{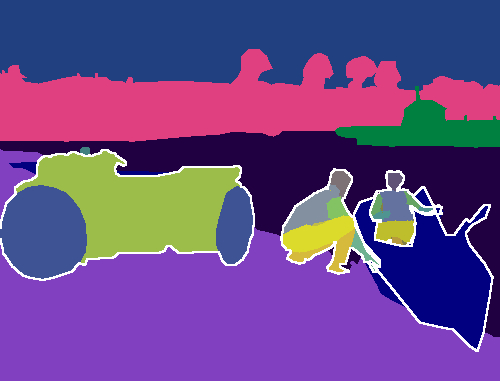}
\includegraphics[width=0.320\linewidth, trim={0 0 0 1cm},clip]{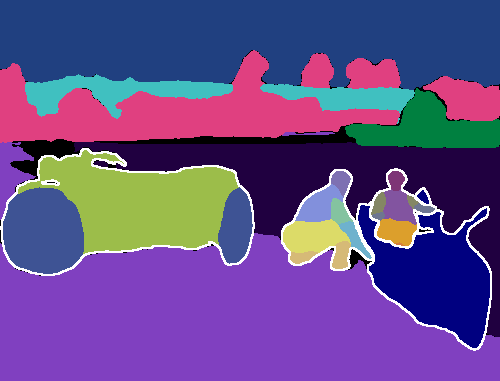}
\\

\begin{subfigure}[b]{0.320\textwidth}
 \centering
 \caption{Input image}
\end{subfigure}
\begin{subfigure}[b]{0.320\textwidth}
 \centering
 \caption{Ground truth}
\end{subfigure}
\begin{subfigure}[b]{0.320\textwidth}
 \centering
 \caption{TAPPS (ours)}
\end{subfigure}

\caption{\textbf{TAPPS with Swin-B~\cite{liu2021swin} on Pascal-PP~\cite{degeus2021pps,chen2014pascalpart,everingham2010pascal,mottaghi14pascalcontext}.} The Swin-B backbone is pre-trained on COCO panoptic~\cite{lin2014coco}. White borders separate different object-level instances; color shades indicate different categories. Note that the colors of part-level categories are not identical across instances; there are different shades of the same color. Best viewed digitally.}
\label{supp:fig:tapps_swinb_ppp}
\end{figure*}

\begin{figure*}[t]
\centering

\includegraphics[width=0.320\linewidth]{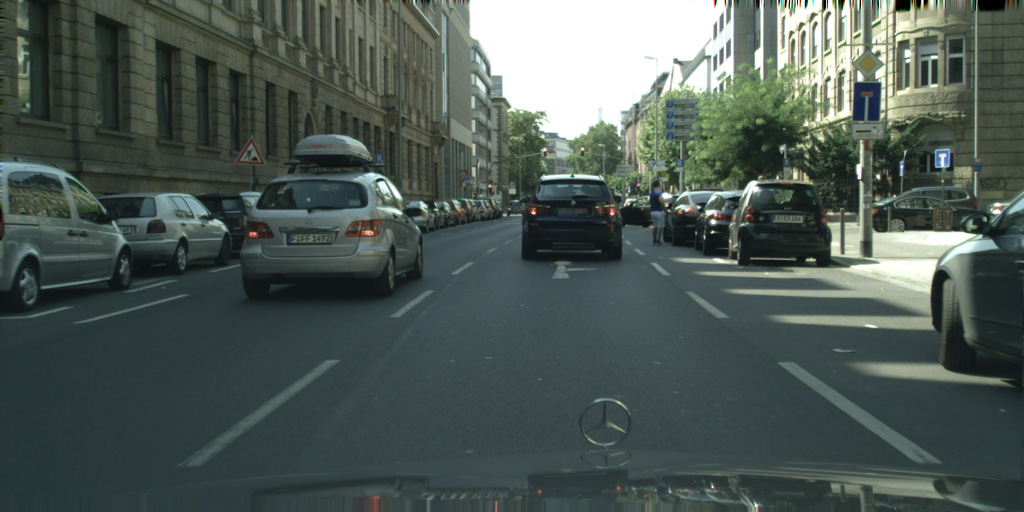}
\includegraphics[width=0.320\linewidth]{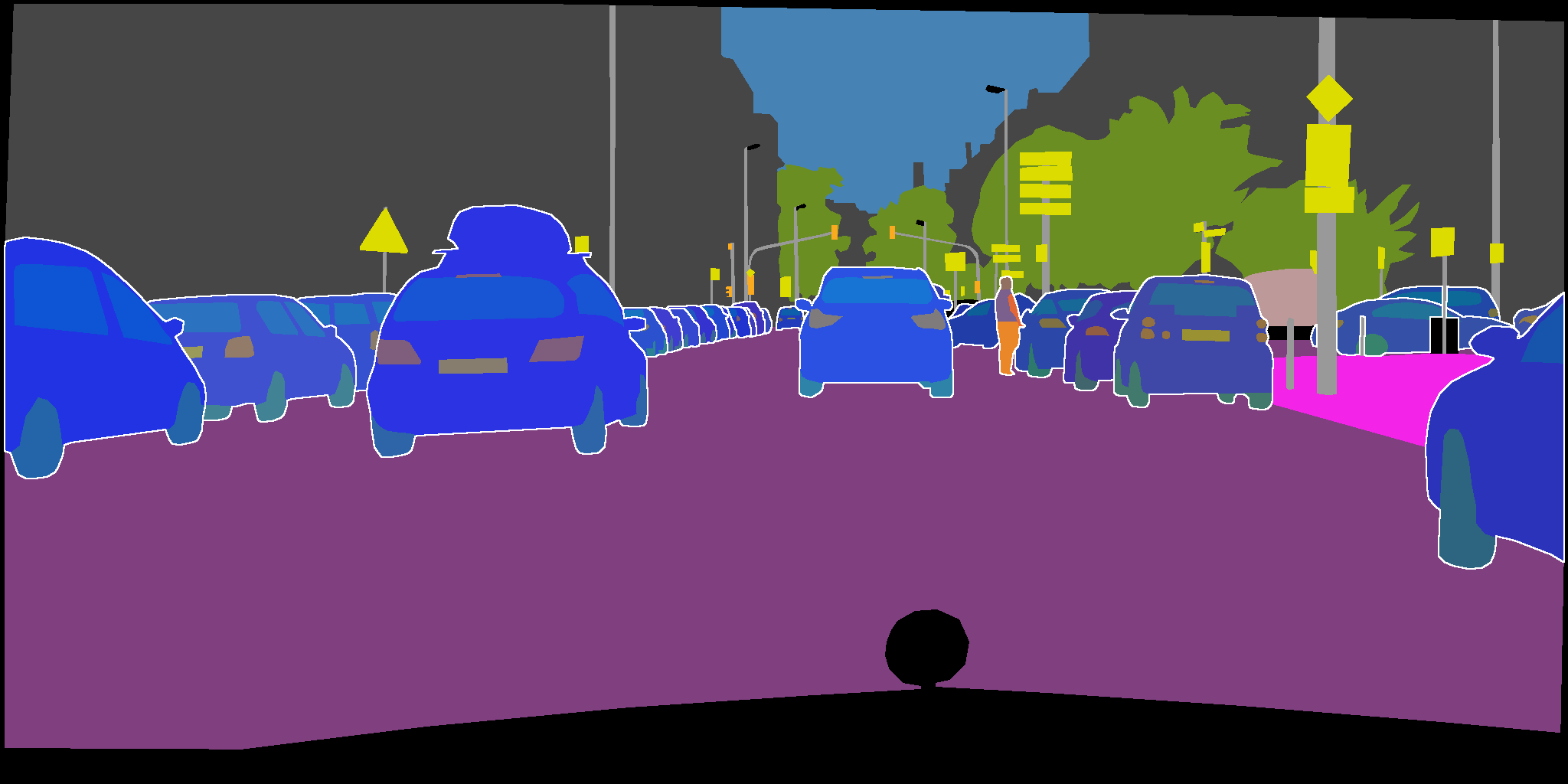}
\includegraphics[width=0.320\linewidth]{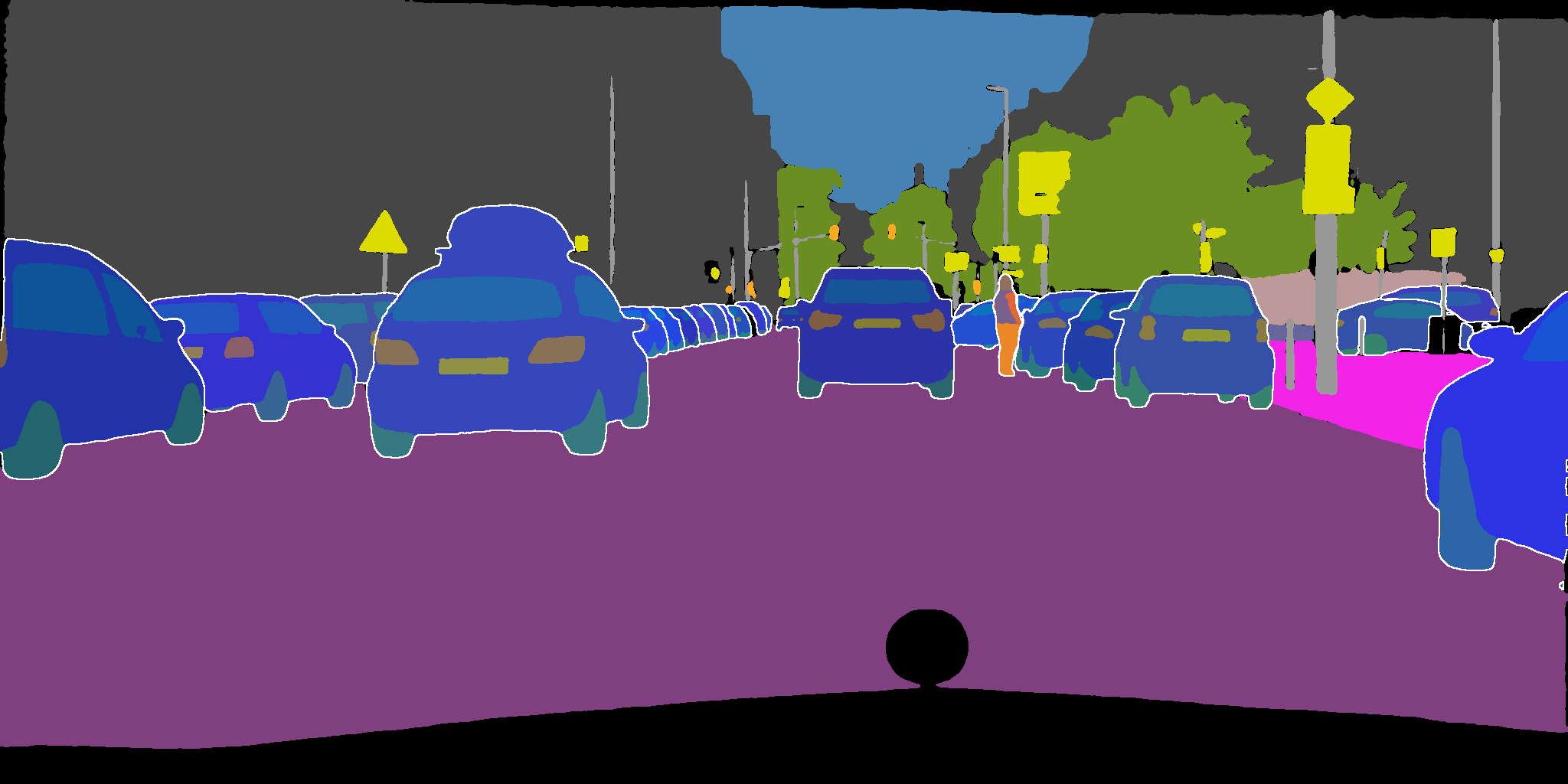}
\\

\includegraphics[width=0.320\linewidth]{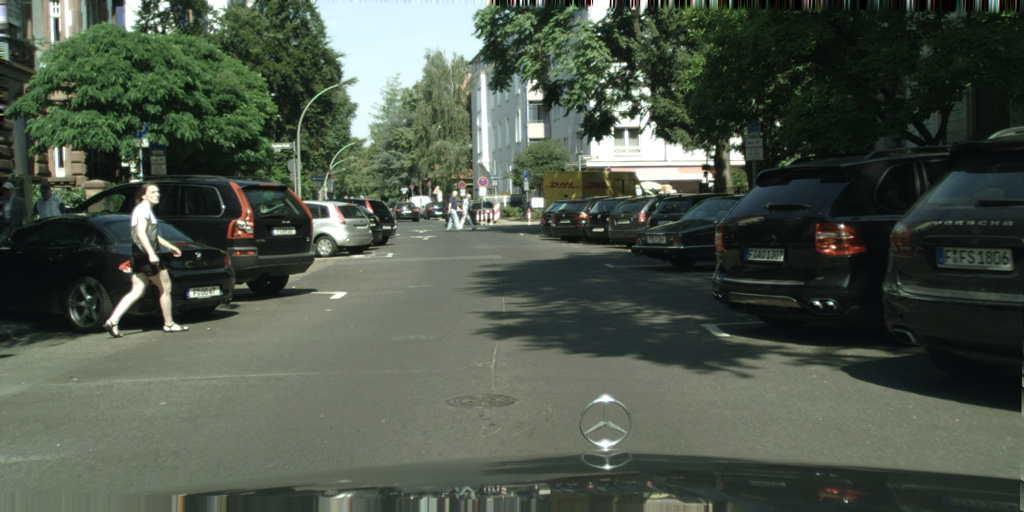}
\includegraphics[width=0.320\linewidth]{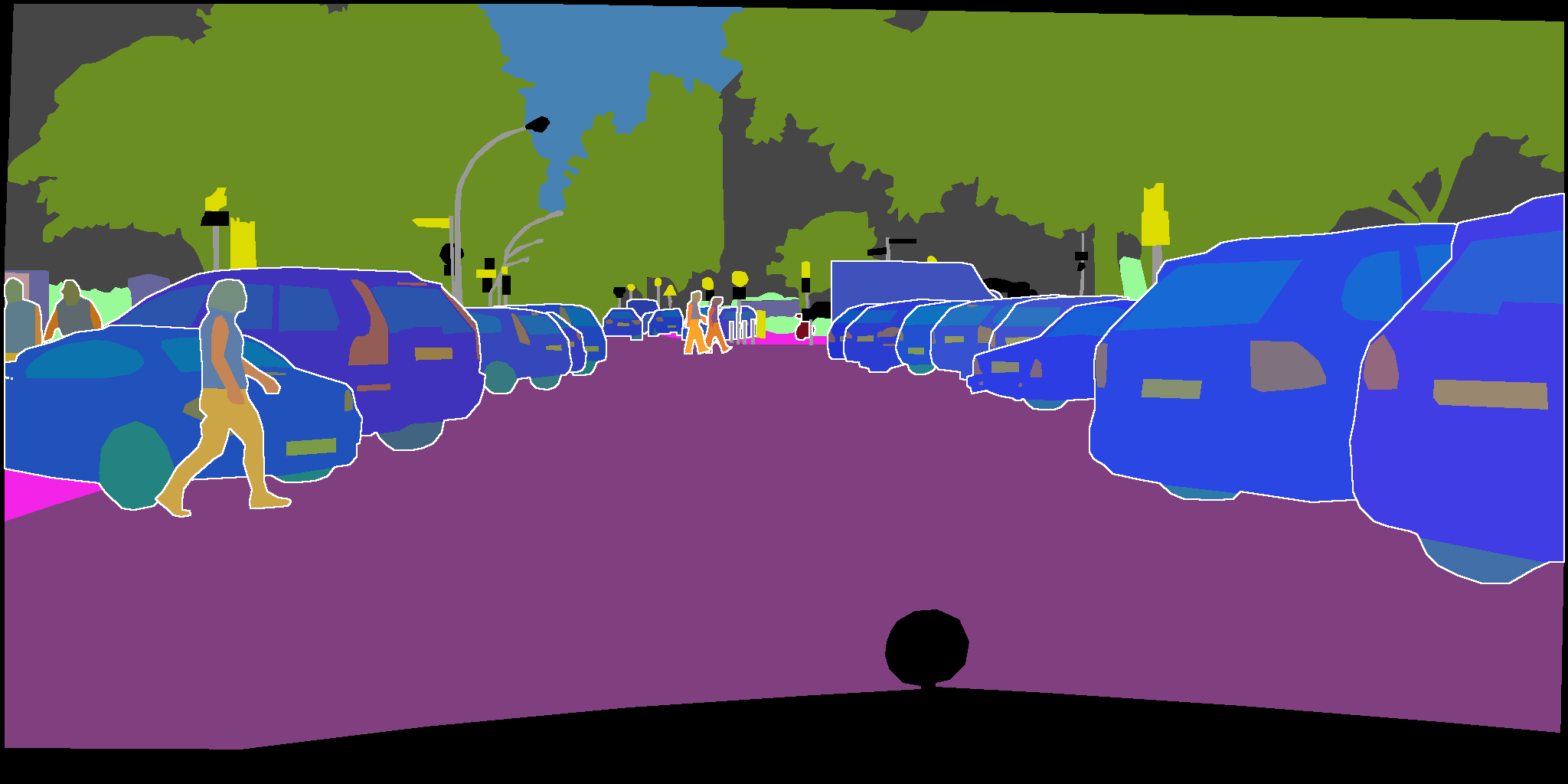}
\includegraphics[width=0.320\linewidth]{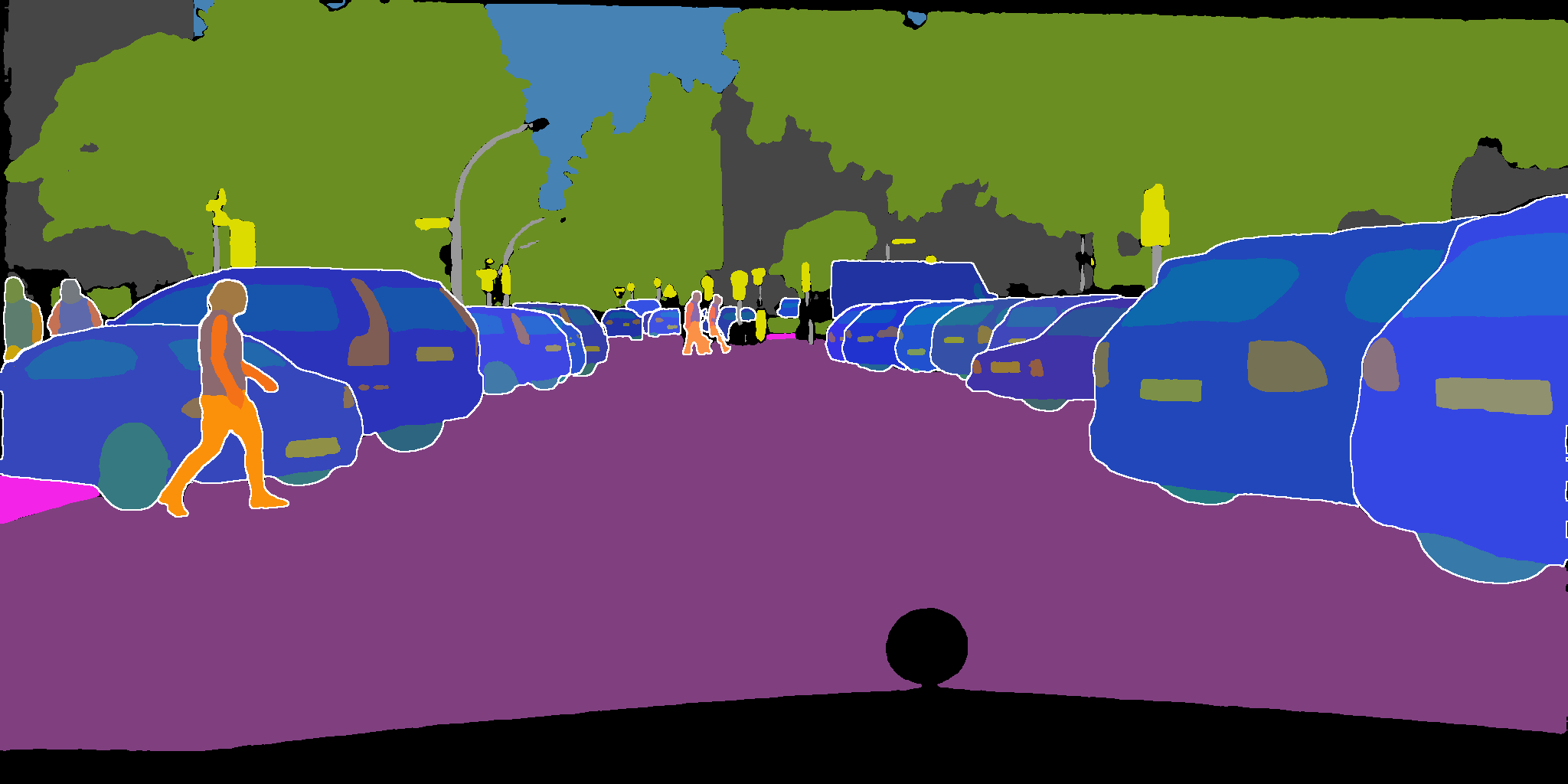}
\\

\includegraphics[width=0.320\linewidth]{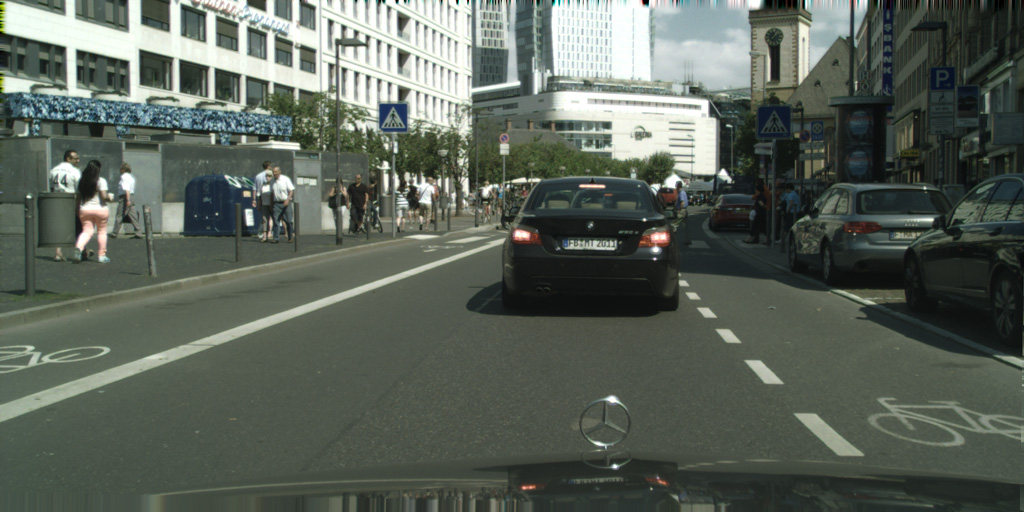}
\includegraphics[width=0.320\linewidth]{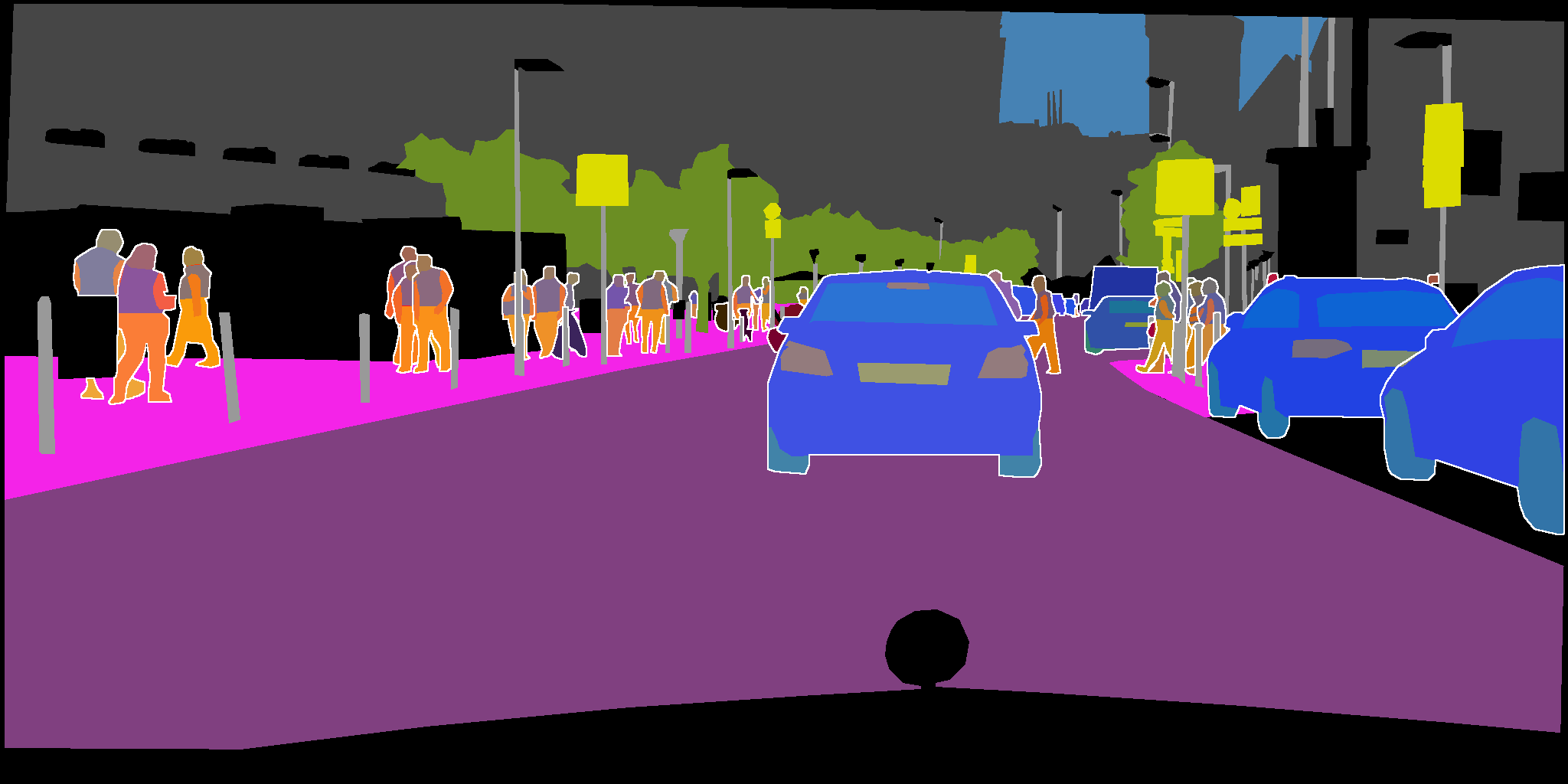}
\includegraphics[width=0.320\linewidth]{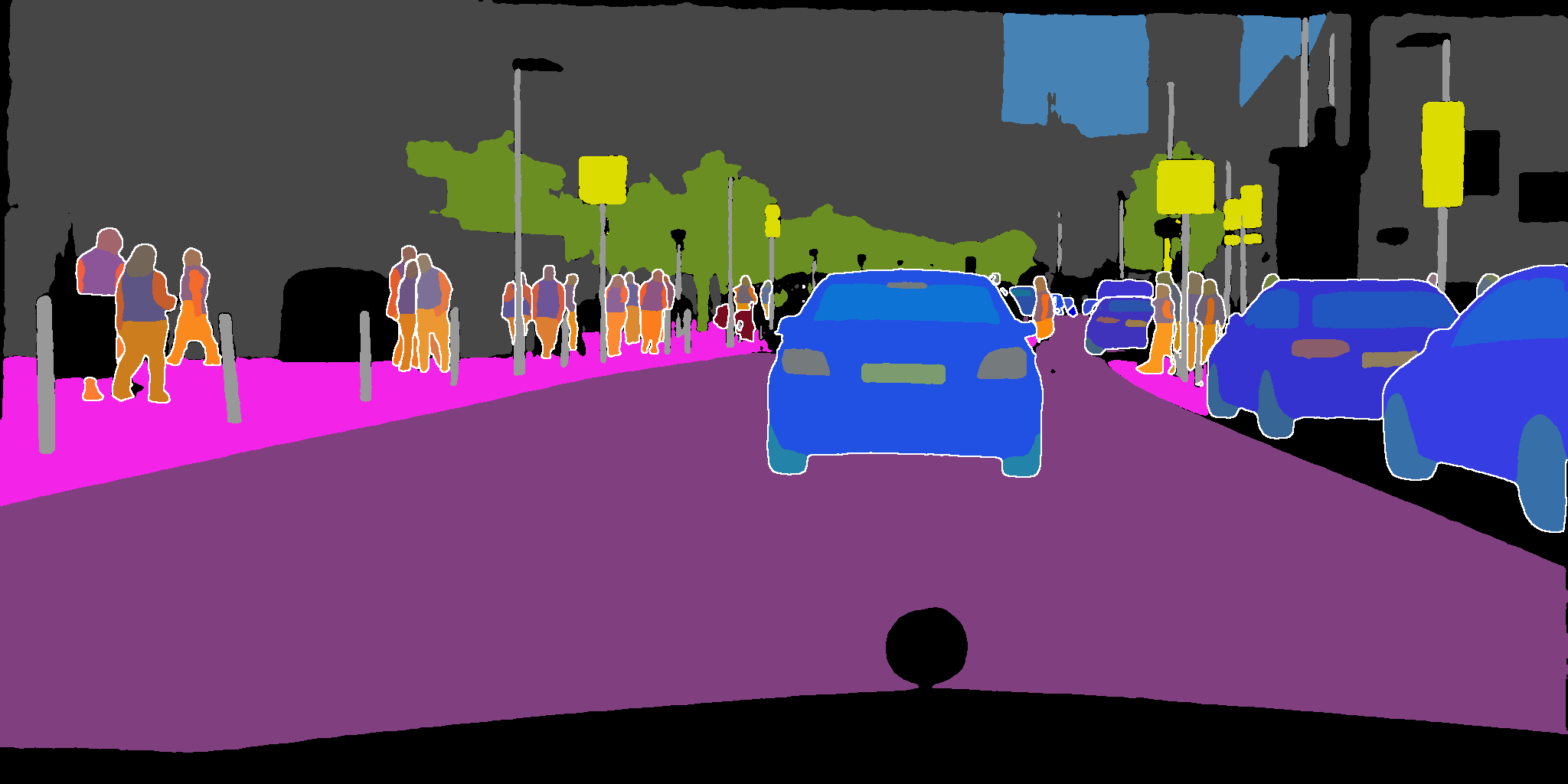}
\\

\includegraphics[width=0.320\linewidth]{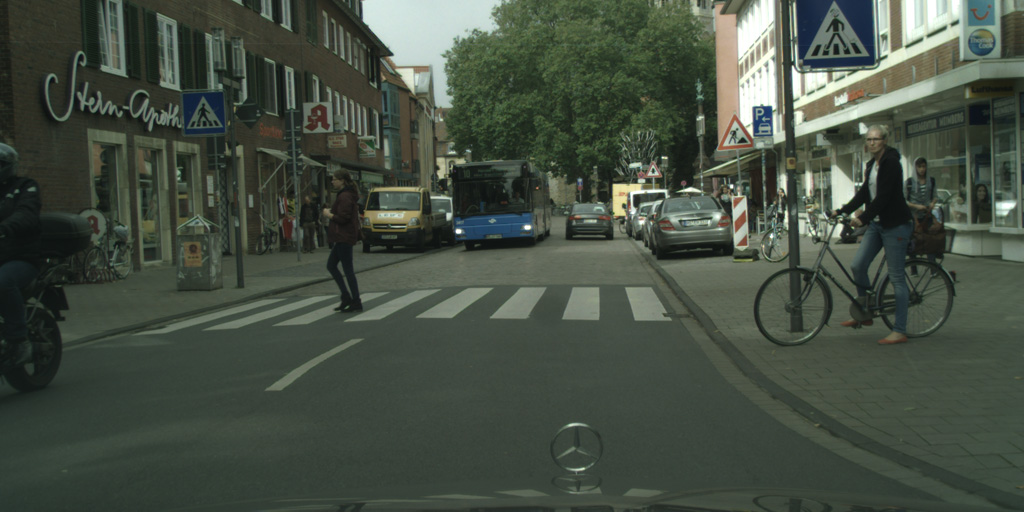}
\includegraphics[width=0.320\linewidth]{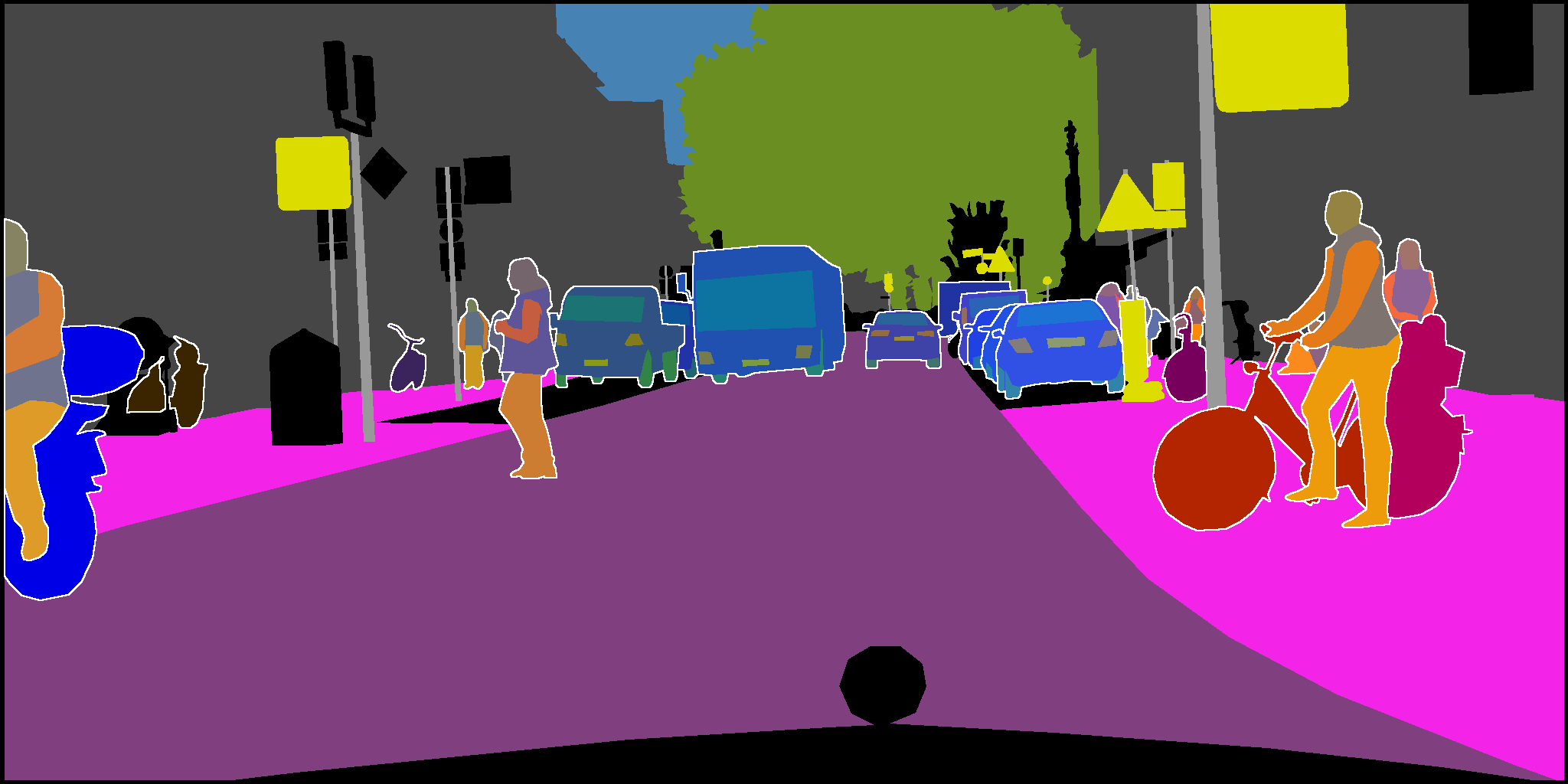}
\includegraphics[width=0.320\linewidth]{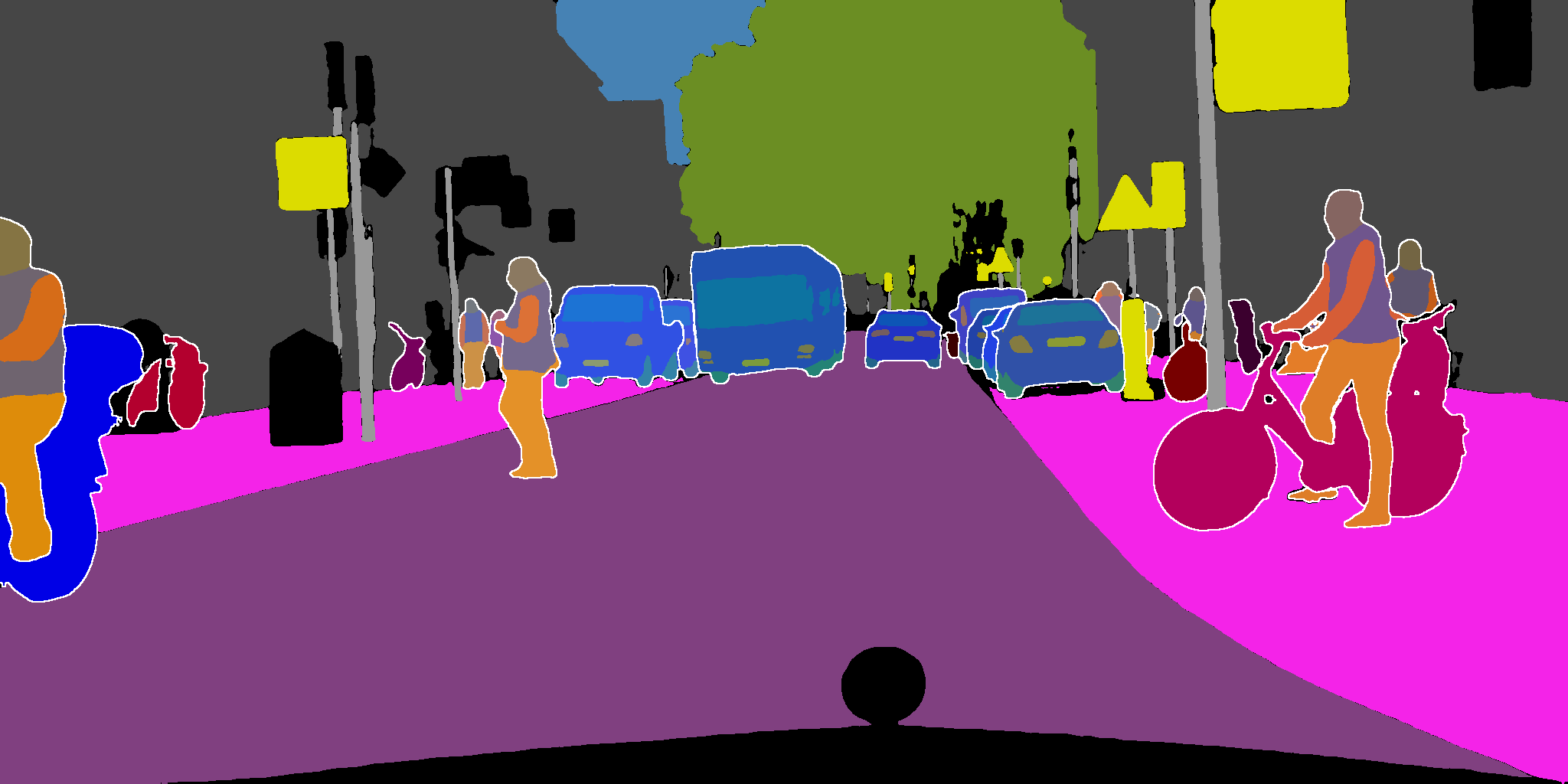}
\\

\includegraphics[width=0.320\linewidth]{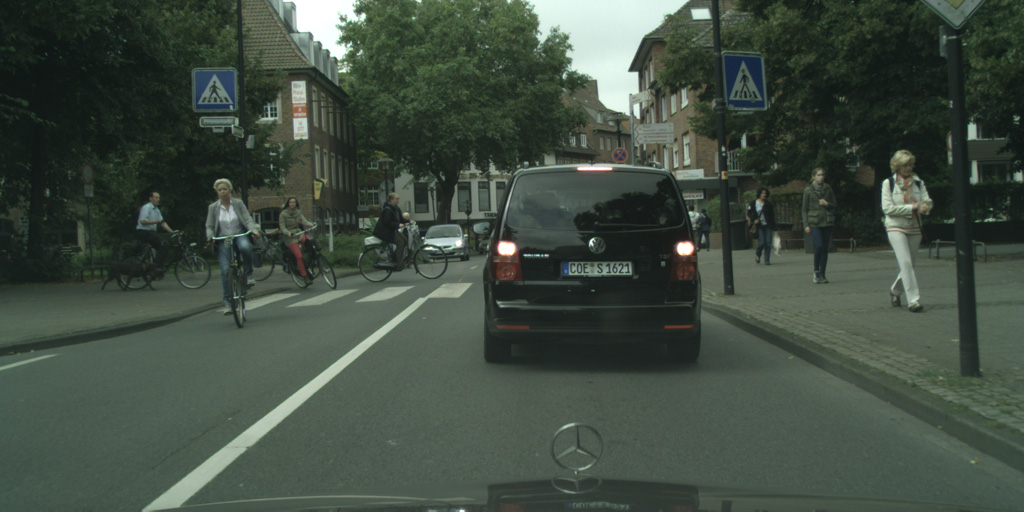}
\includegraphics[width=0.320\linewidth]{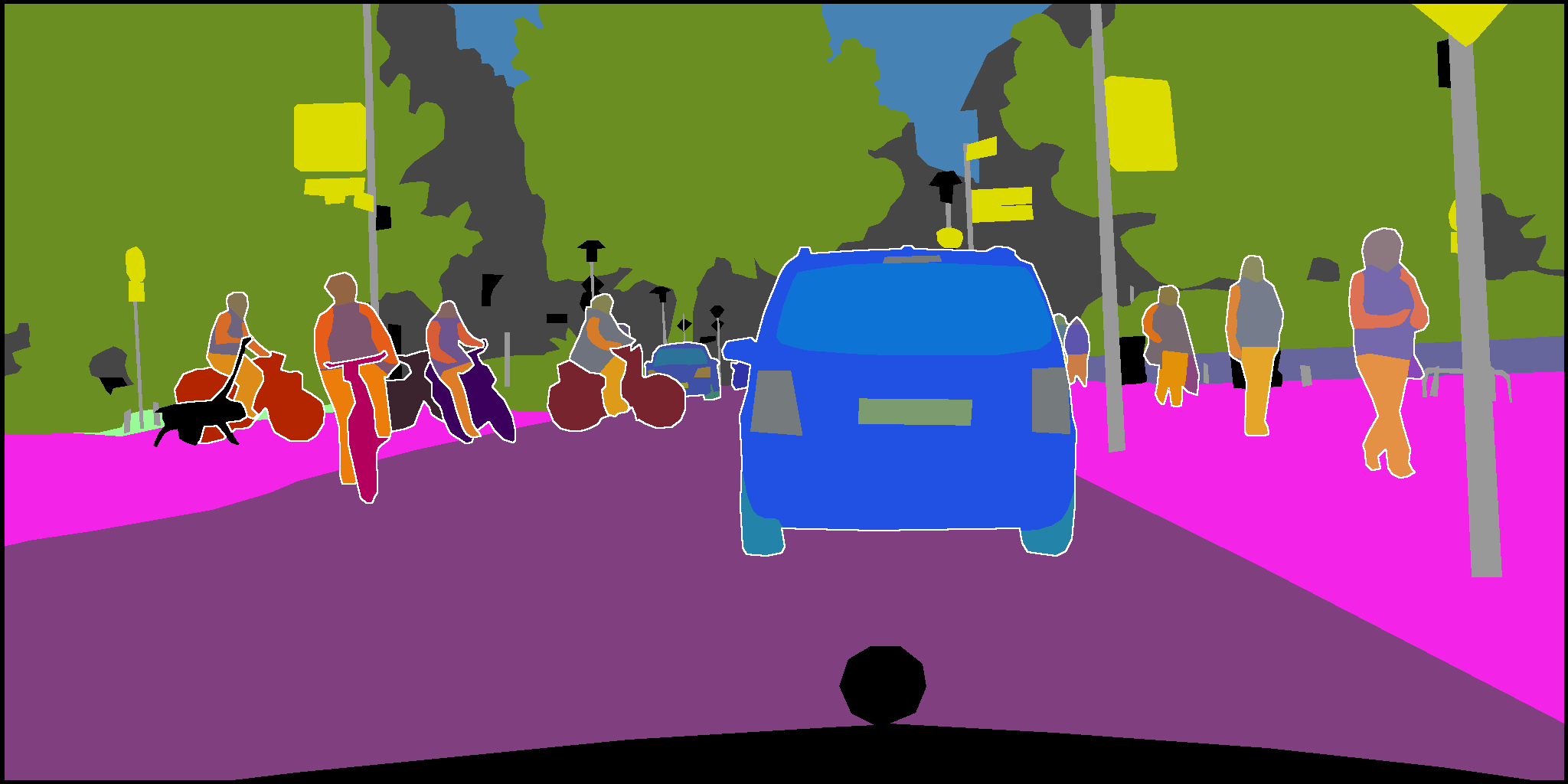}
\includegraphics[width=0.320\linewidth]{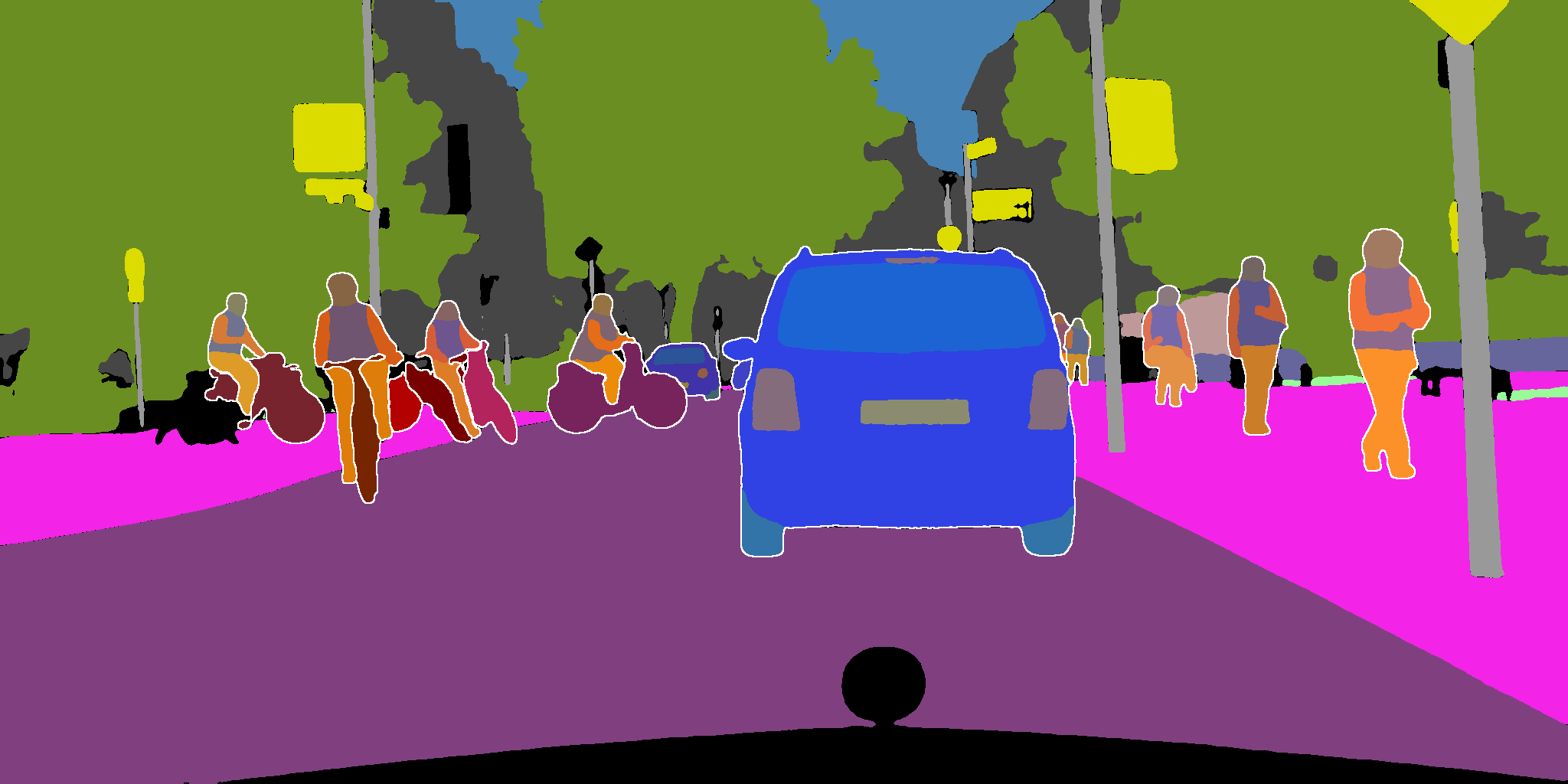}
\\

\includegraphics[width=0.320\linewidth]{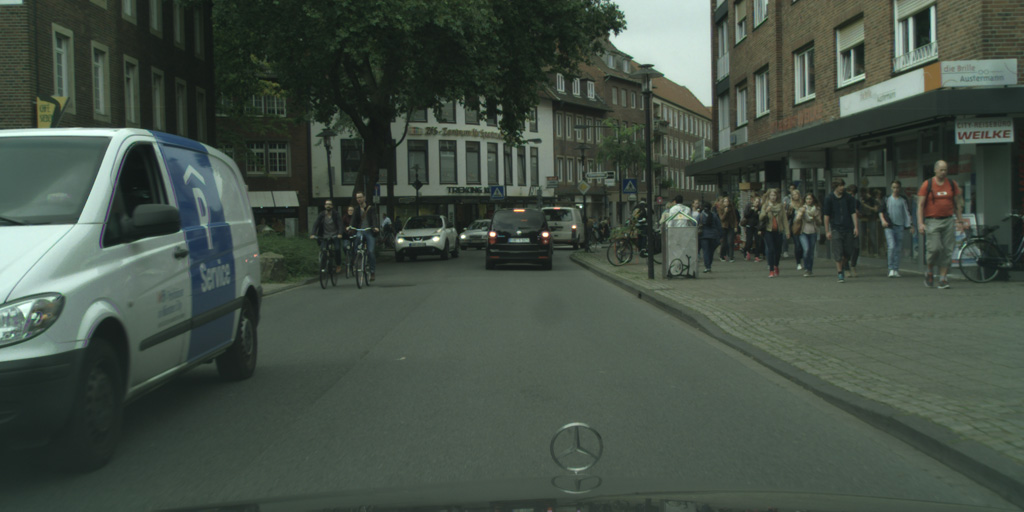}
\includegraphics[width=0.320\linewidth]{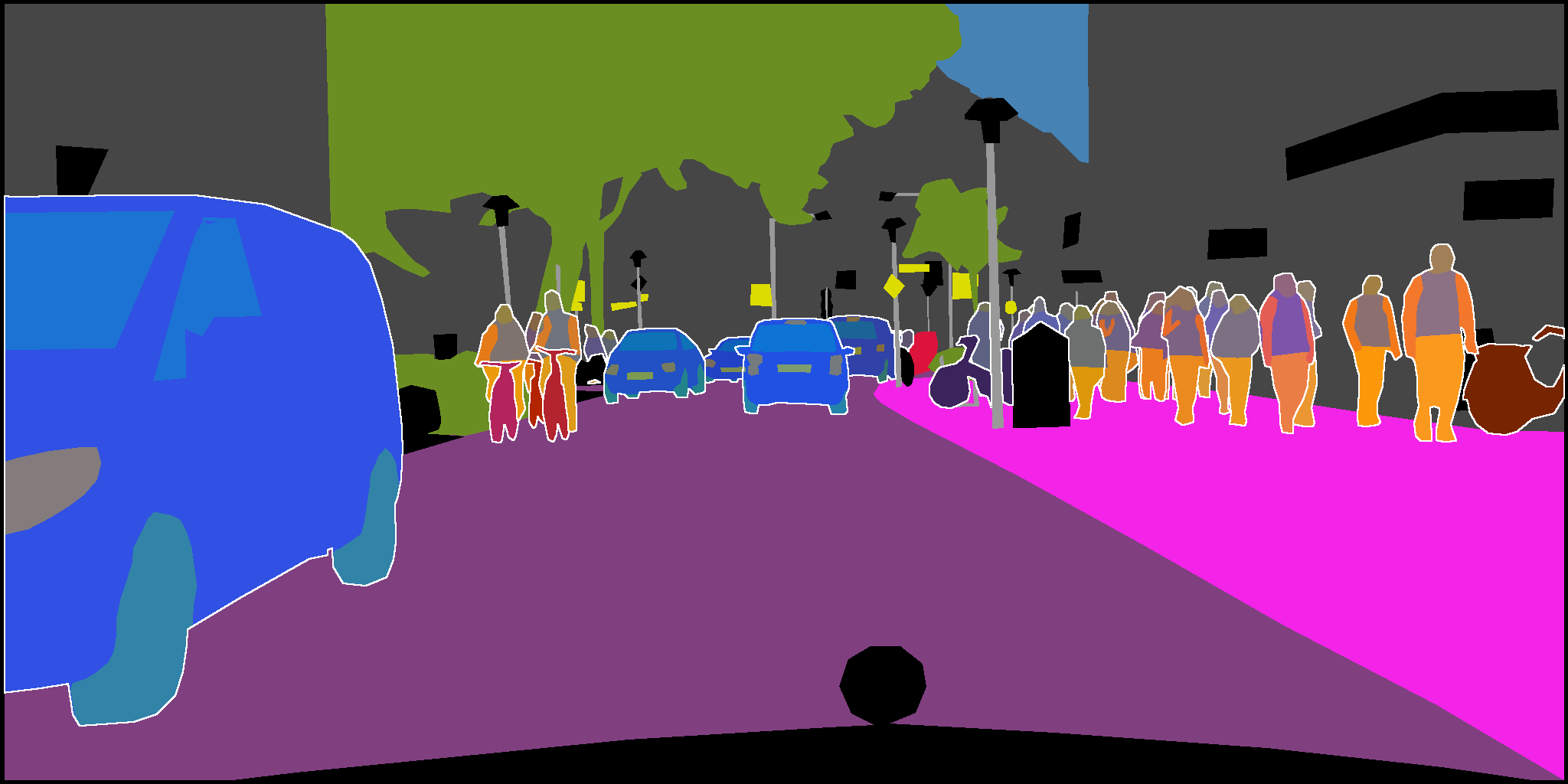}
\includegraphics[width=0.320\linewidth]{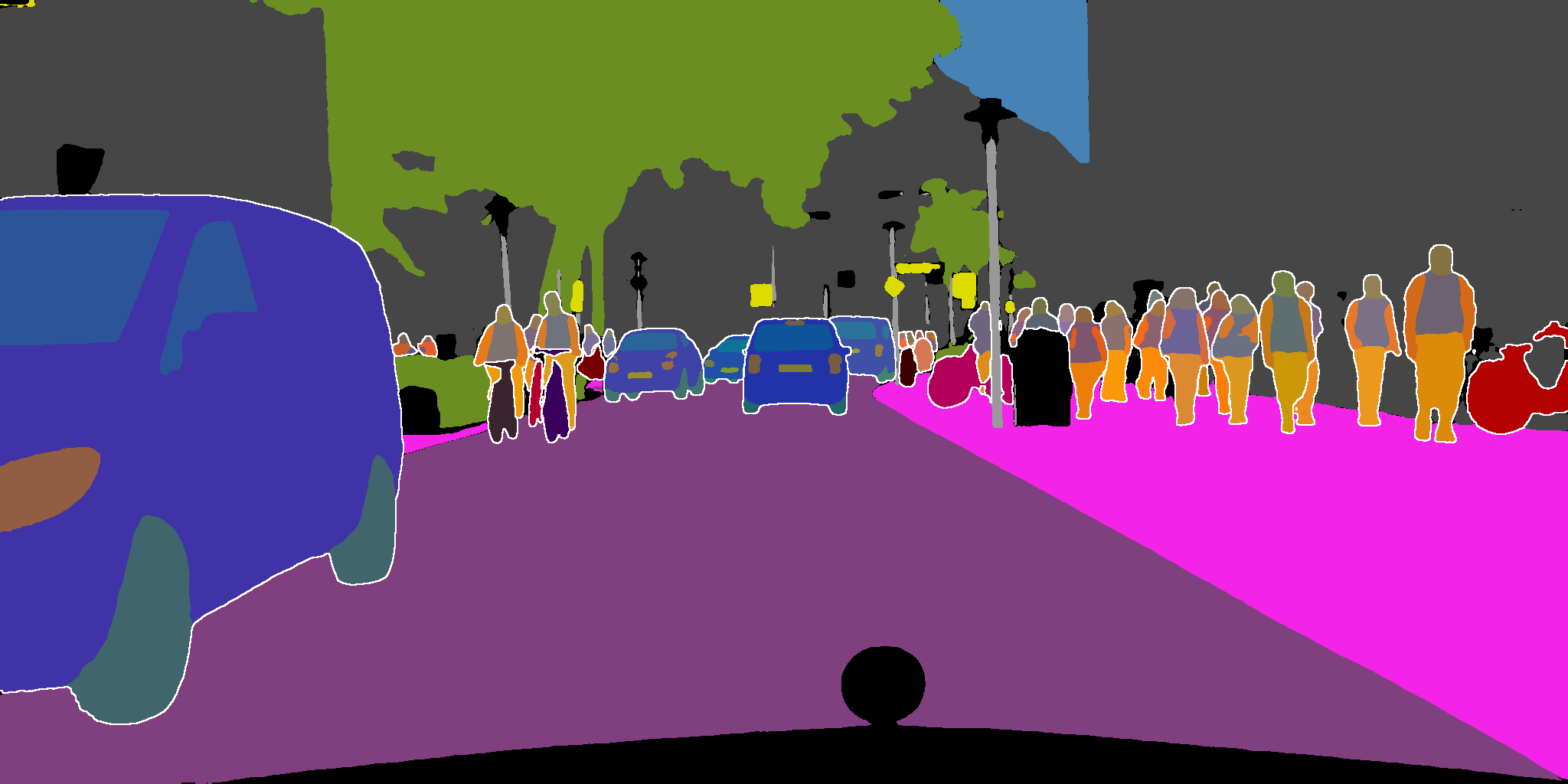}
\\

\includegraphics[width=0.320\linewidth]{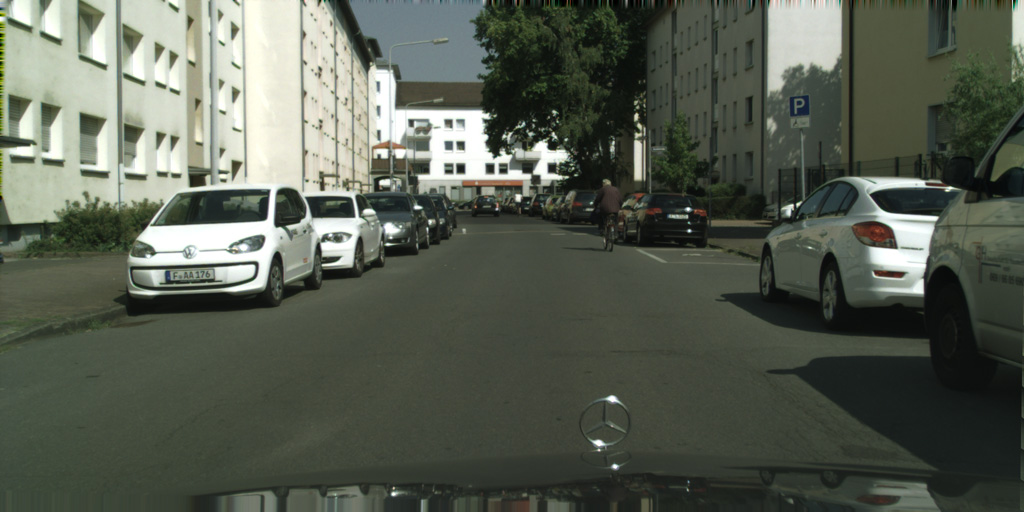}
\includegraphics[width=0.320\linewidth]{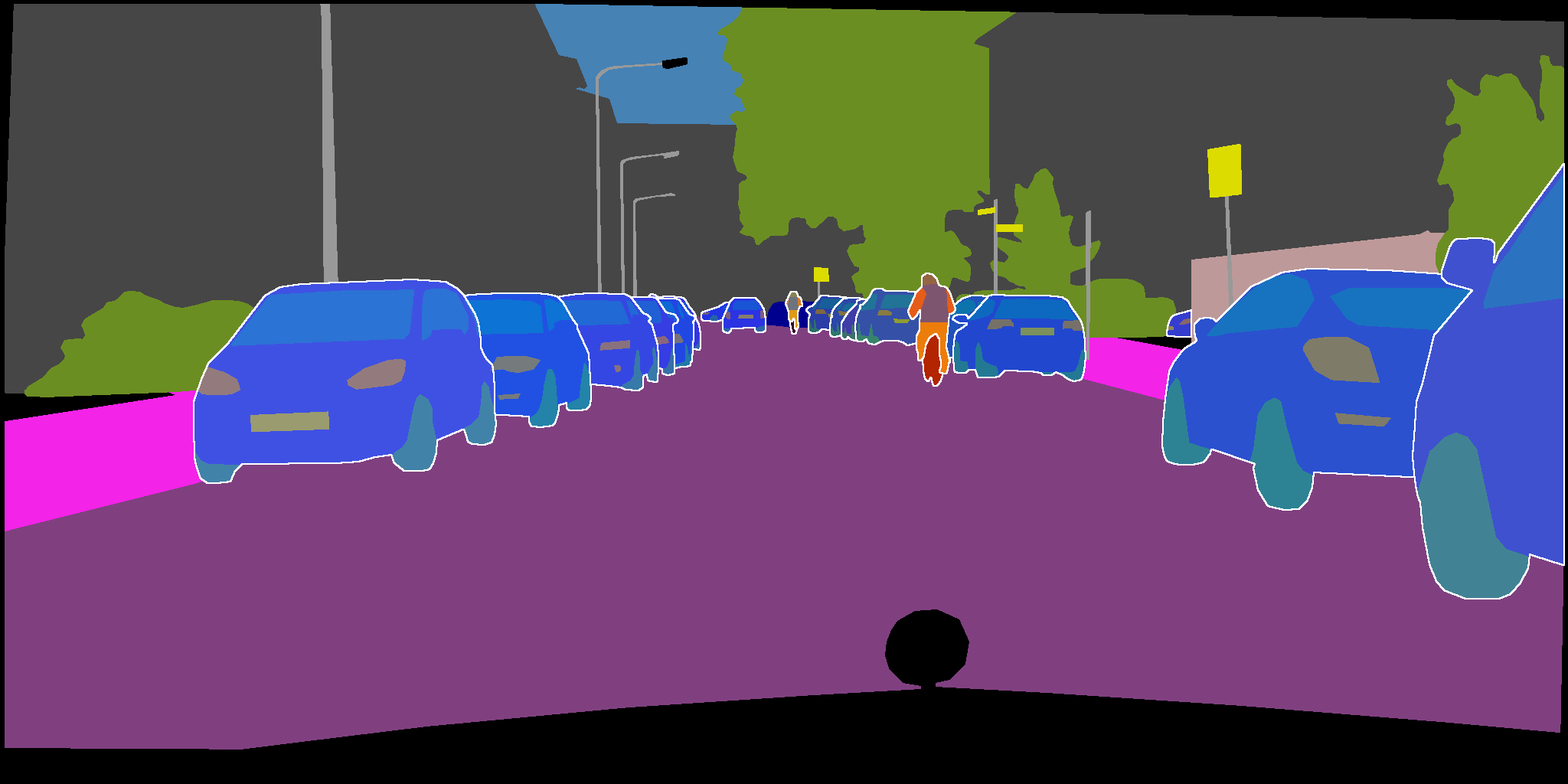}
\includegraphics[width=0.320\linewidth]{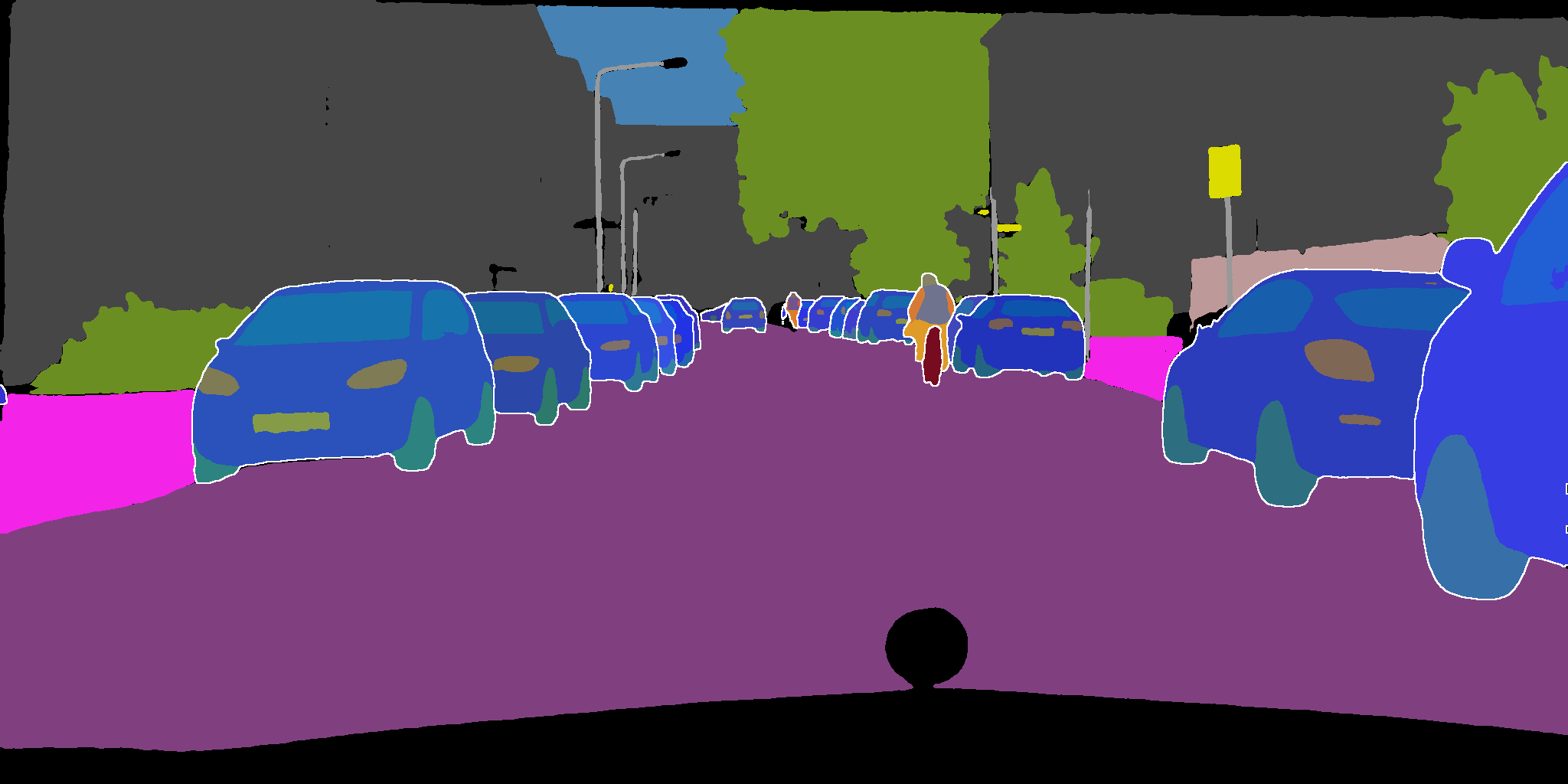}
\\

\begin{subfigure}[b]{0.320\textwidth}
 \centering
 \caption{Input image}
\end{subfigure}
\begin{subfigure}[b]{0.320\textwidth}
 \centering
 \caption{Ground truth}
\end{subfigure}
\begin{subfigure}[b]{0.320\textwidth}
 \centering
 \caption{TAPPS (ours)}
\end{subfigure}

\caption{\textbf{TAPPS with Swin-B~\cite{liu2021swin} on Cityscapes-PP~\cite{cordts2016cityscapes,degeus2021pps}.} The Swin-B backbone is pre-trained on COCO panoptic~\cite{lin2014coco}. White borders separate different object-level instances; color shades indicate different categories. Note that the colors of part-level categories are not identical across instances; there are different shades of the same color. Best viewed digitally.}
\label{supp:fig:tapps_swinb_cpp}
\end{figure*}

In \cref{supp:fig:tapps_swinb_ppp} and \cref{supp:fig:tapps_swinb_cpp}, we show examples of predictions by TAPPS with a Swin-B~\cite{liu2021swin} backbone, which achieves new state-of-the-art PPS performance. These examples show the high segmentation quality that TAPPS can achieve, across different types of objects and classes.

\begin{figure*}[t]
\centering

\includegraphics[width=0.320\linewidth]{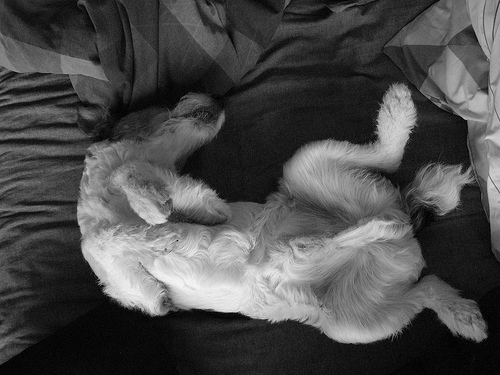}
\includegraphics[width=0.320\linewidth]{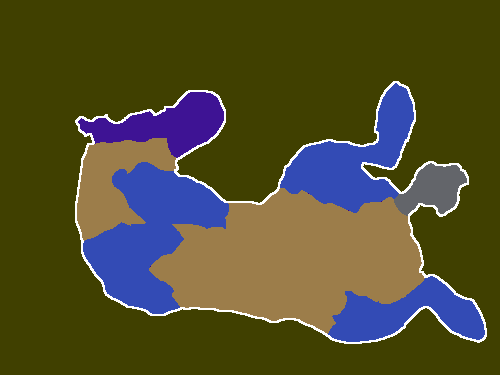}
\includegraphics[width=0.320\linewidth]{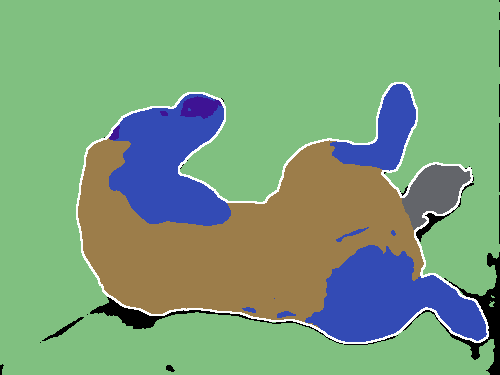}
\\

\includegraphics[width=0.320\linewidth]{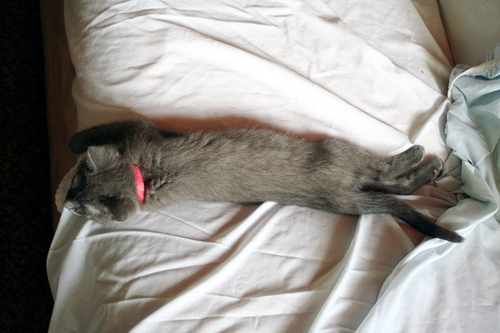}
\includegraphics[width=0.320\linewidth]{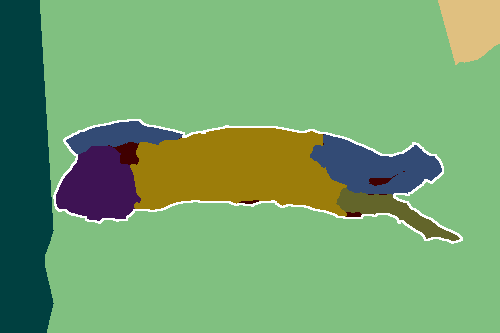}
\includegraphics[width=0.320\linewidth]{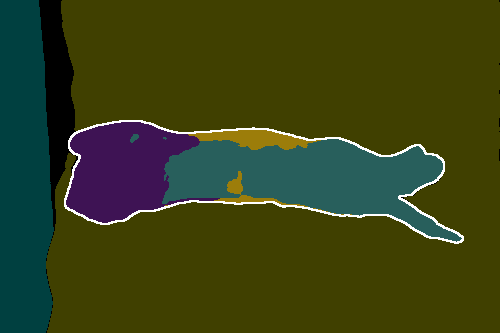}
\\

\includegraphics[width=0.320\linewidth]{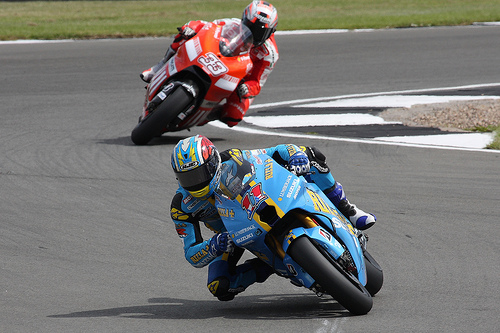}
\includegraphics[width=0.320\linewidth]{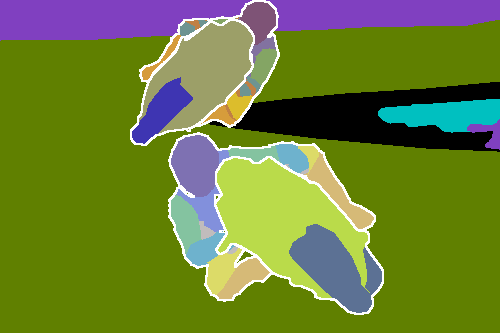}
\includegraphics[width=0.320\linewidth]{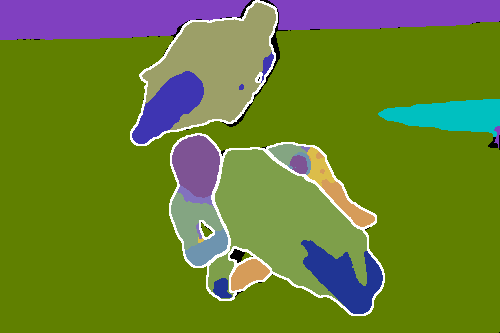}
\\

\includegraphics[width=0.320\linewidth, trim={0 4cm 8cm 0},clip]{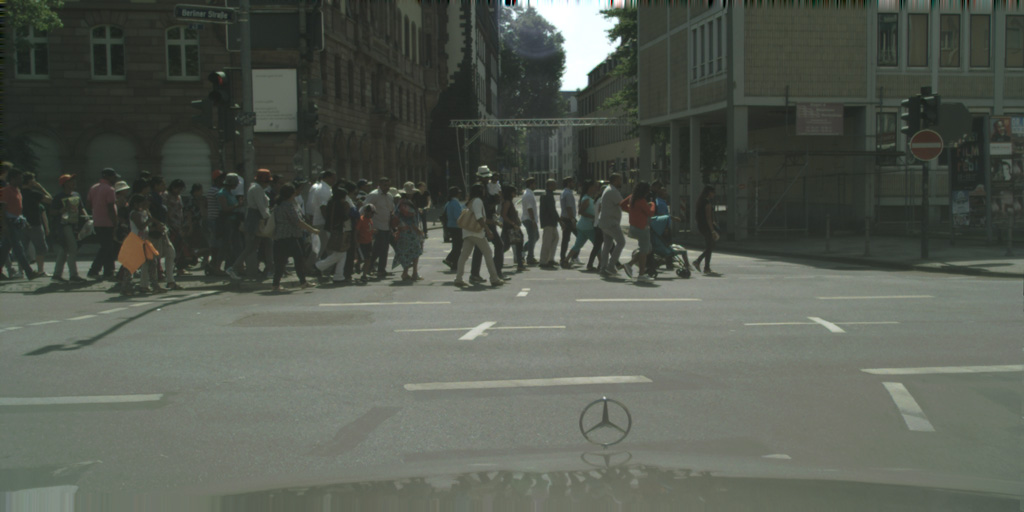}
\includegraphics[width=0.320\linewidth, trim={0 8cm 16cm 0},clip]{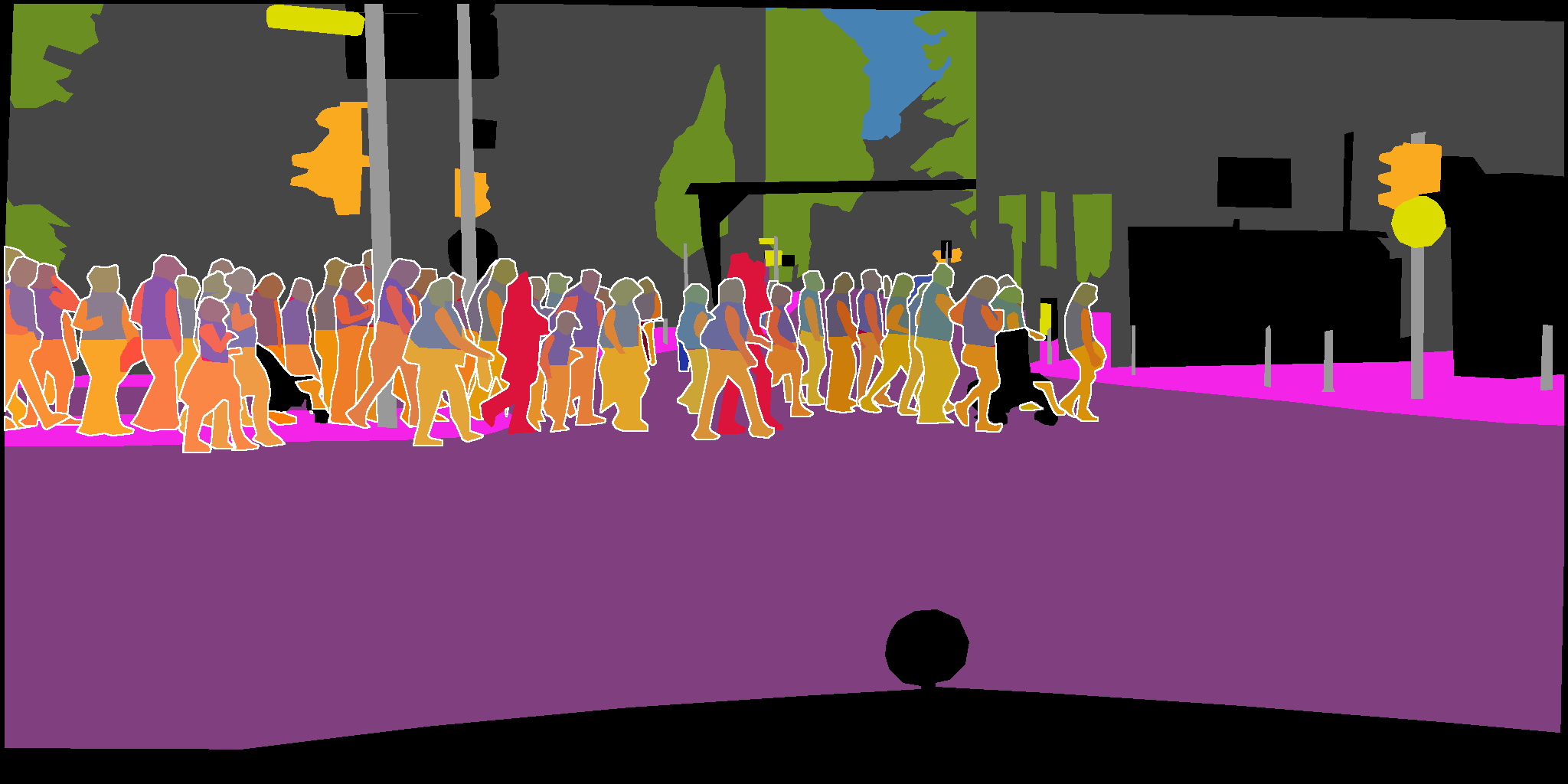}
\includegraphics[width=0.320\linewidth, trim={0 8cm 16cm 0},clip]{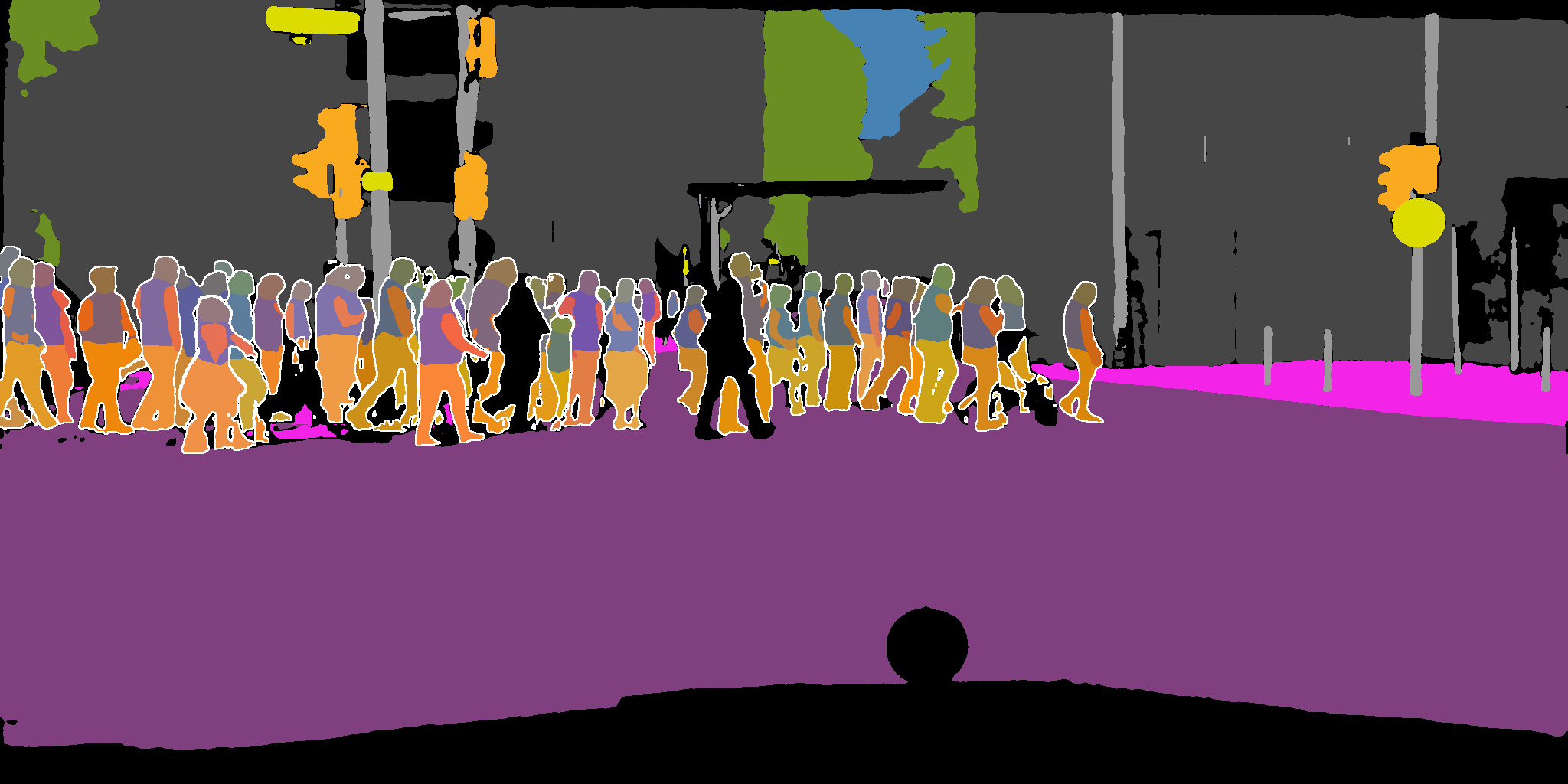}
\\

\includegraphics[width=0.320\linewidth, trim={6cm 7cm 15cm 4cm},clip]{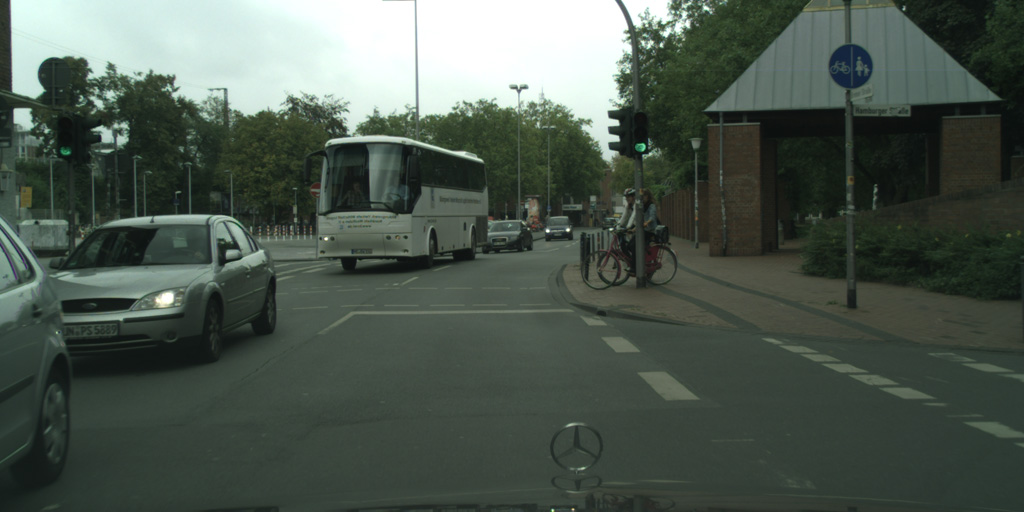}
\includegraphics[width=0.320\linewidth, trim={12cm 14cm 30cm 8cm},clip]{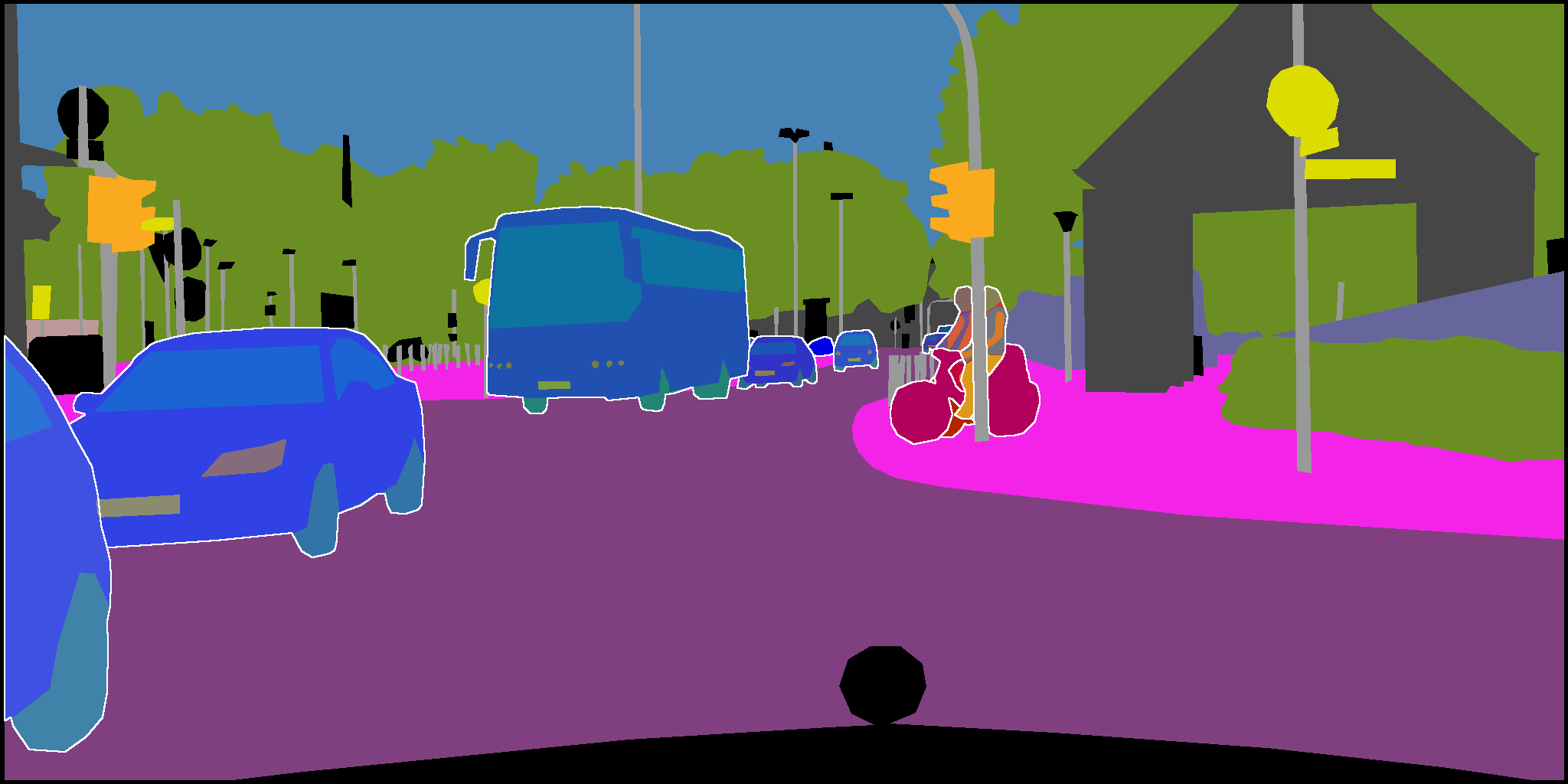}
\begin{tikzpicture}
    \node[anchor=south west,inner sep=0] (image) at (0,0) {\includegraphics[width=0.320\linewidth, trim={12cm 14cm 30cm 8cm},clip]{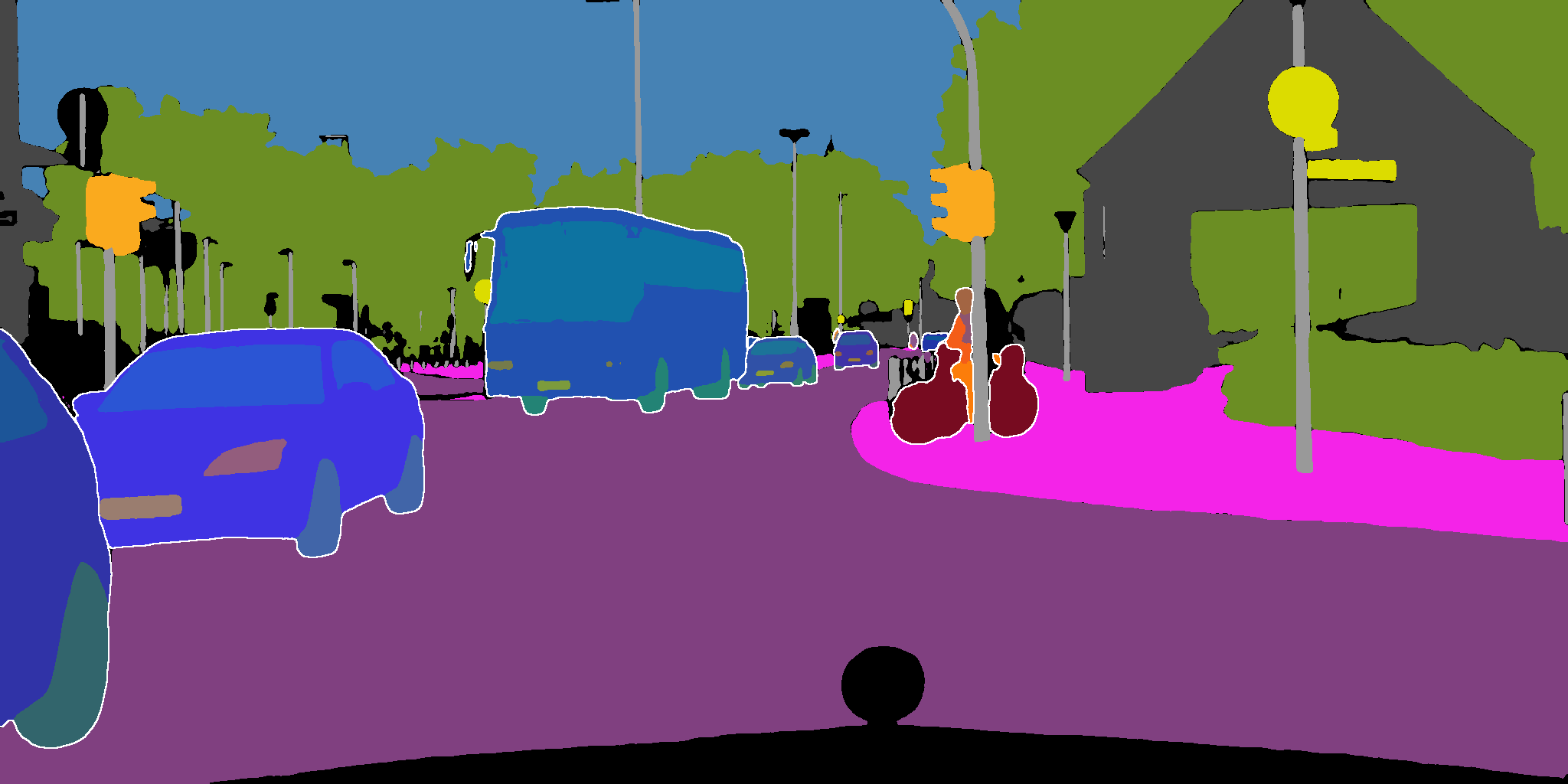}};
    \begin{scope}[x={(image.south east)},y={(image.north west)}]
        \draw[red,line width=0.5mm,rounded corners] (0.3,0.2) rectangle (0.77,0.95);
    \end{scope}
\end{tikzpicture}
\\

\includegraphics[width=0.320\linewidth, trim={15cm 3cm 0 3cm},clip]{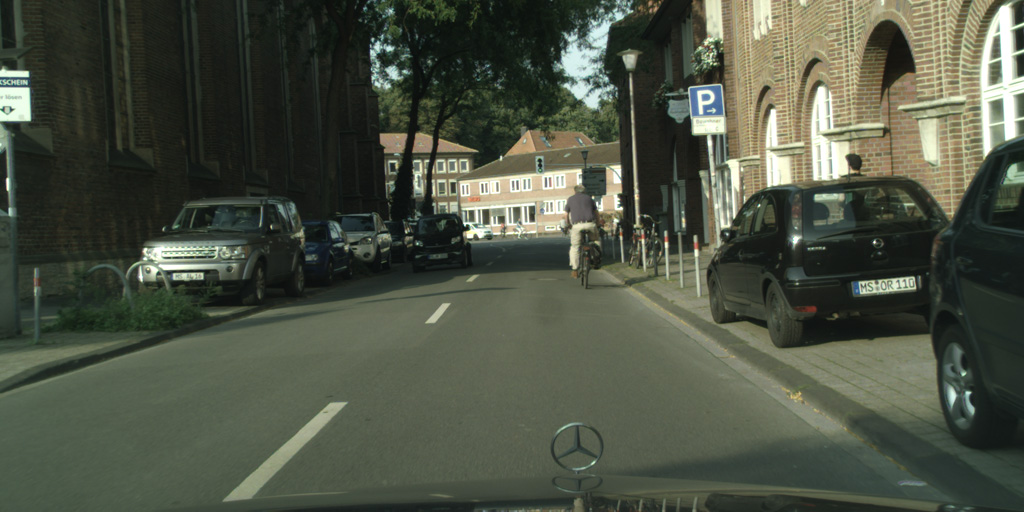}
\includegraphics[width=0.320\linewidth, trim={30cm 6cm 0 6cm},clip]{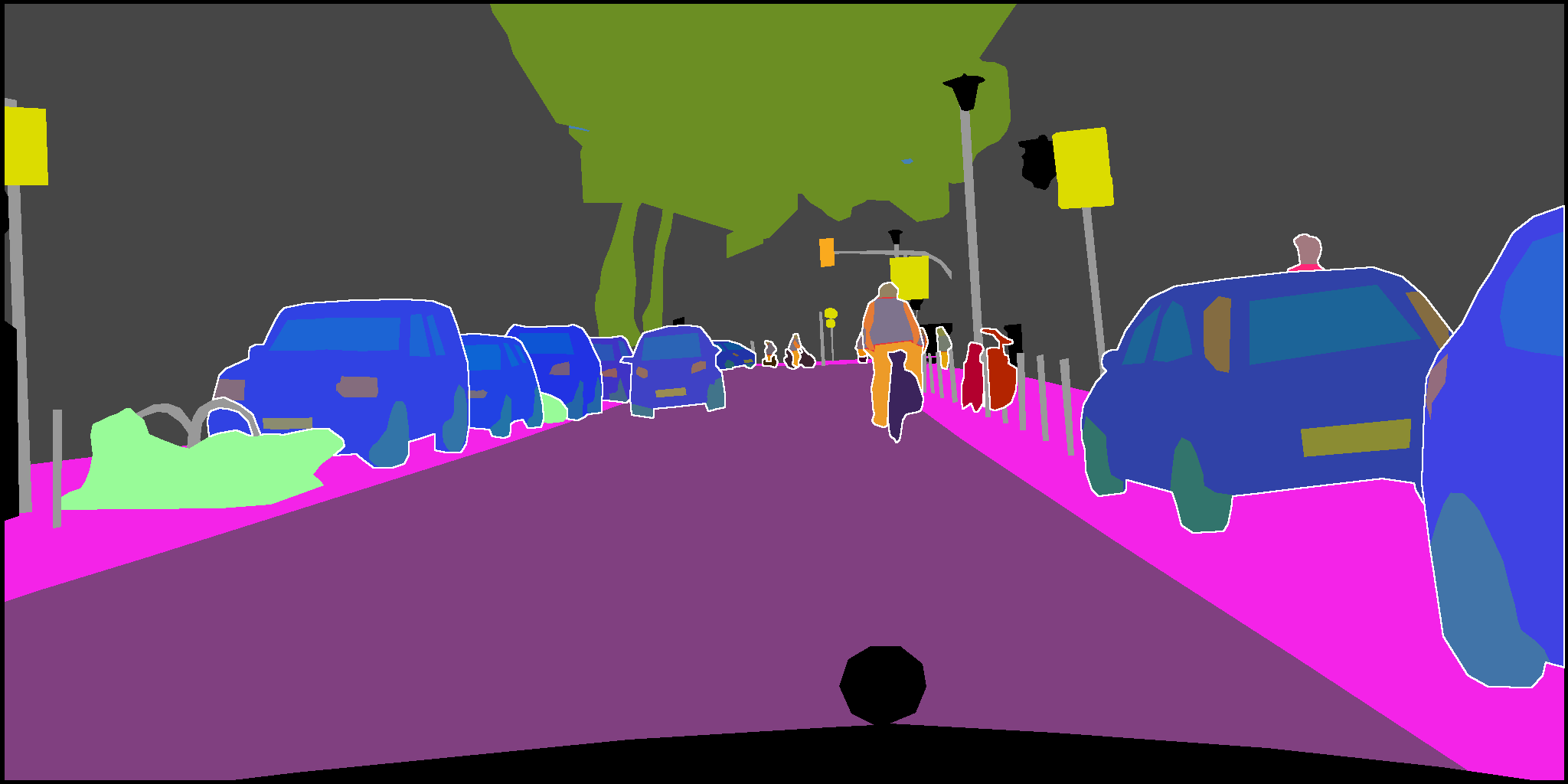}
\begin{tikzpicture}
    \node[anchor=south west,inner sep=0] (image) at (0,0) {\includegraphics[width=0.320\linewidth, trim={30cm 6cm 0 6cm},clip]{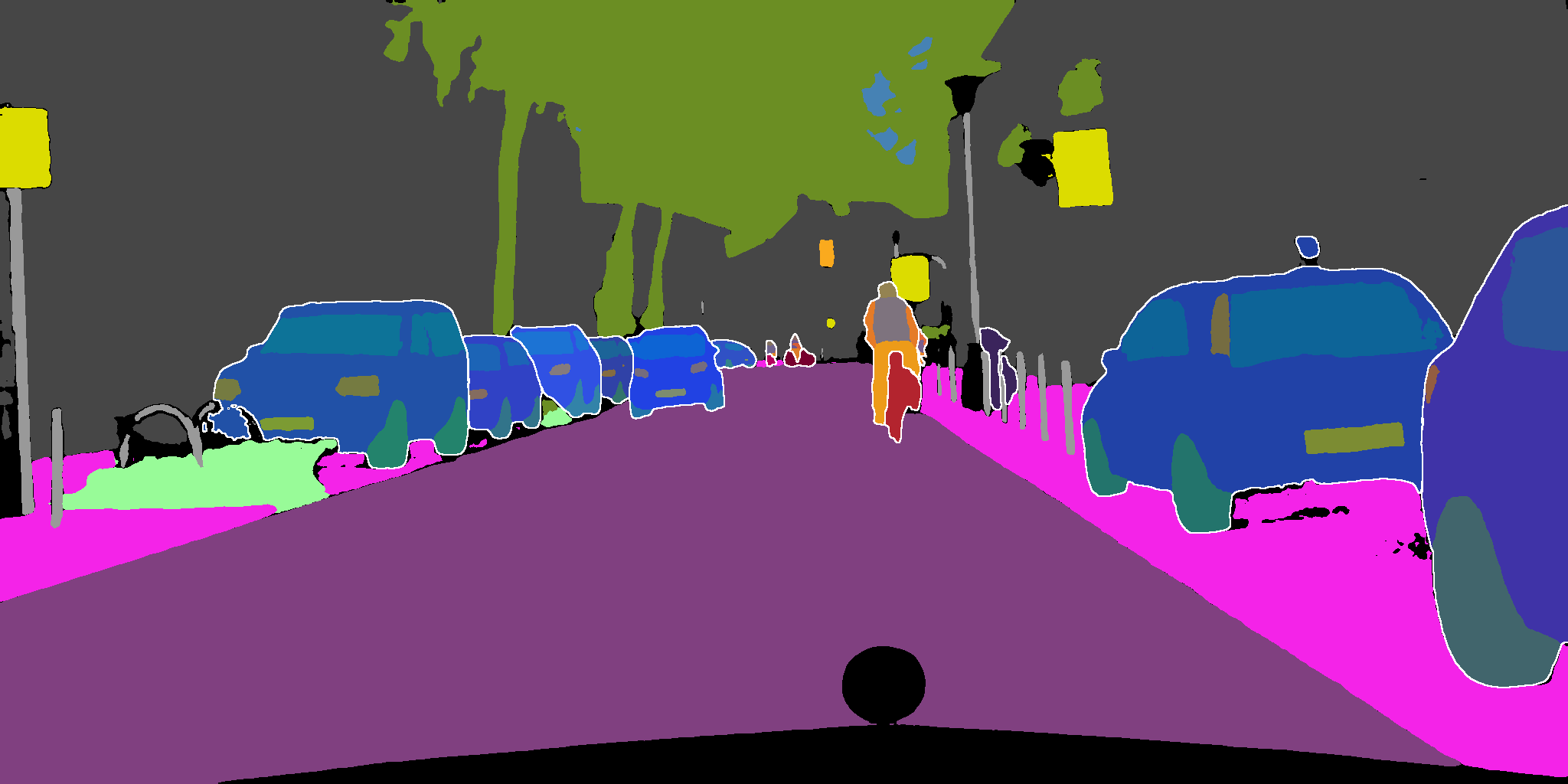}};
    \begin{scope}[x={(image.south east)},y={(image.north west)}]
        \draw[red,line width=0.5mm,rounded corners] (0.5,0.4) rectangle (0.9,0.85);
    \end{scope}
\end{tikzpicture}
\\

\begin{subfigure}[b]{0.320\textwidth}
 \centering
 \caption{Input image}
\end{subfigure}
\begin{subfigure}[b]{0.320\textwidth}
 \centering
 \caption{Ground truth}
\end{subfigure}
\begin{subfigure}[b]{0.320\textwidth}
 \centering
 \caption{TAPPS (ours)}
\end{subfigure}

\caption{\textbf{Examples of errors in TAPPS predictions.} The predictions are made by TAPPS that uses a ResNet-50~\cite{he2016resnet} backbone pre-trained on COCO panoptic~\cite{lin2014coco}. Top three images are from Pascal-PP validation~\cite{everingham2010pascal,chen2014pascalpart,mottaghi14pascalcontext,degeus2021pps}, bottom three images are from Cityscapes-PP val~\cite{cordts2016cityscapes,degeus2021pps}. White borders separate different object-level instances; color shades indicate different categories. Note that the colors of part-level categories are not identical across instances; there are different shades of the same color. Best viewed digitally.}
\label{supp:fig:tapps_errors}
\end{figure*}

Finally, \cref{supp:fig:tapps_errors} shows examples of typical errors made by TAPPS. Notably, TAPPS struggles with images in which objects are seen from uncommon perspectives, and images with many objects and complex occlusions.

{
    \small
    \bibliographystyle{ieeenat_fullname}
    \bibliography{main}

\begin{thebibliography}{53}
\providecommand{\natexlab}[1]{#1}
\providecommand{\url}[1]{\texttt{#1}}
\expandafter\ifx\csname urlstyle\endcsname\relax
  \providecommand{\doi}[1]{doi: #1}\else
  \providecommand{\doi}{doi: \begingroup \urlstyle{rm}\Url}\fi

\bibitem[Chen et~al.(2014)Chen, Mottaghi, Liu, Fidler, Urtasun, and Yuille]{chen2014pascalpart}
Xianjie Chen, Roozbeh Mottaghi, Xiaobai Liu, Sanja Fidler, Raquel Urtasun, and Alan Yuille.
\newblock {Detect What You Can: Detecting and Representing Objects using Holistic Models and Body Parts}.
\newblock In \emph{CVPR}, 2014.

\bibitem[Cheng et~al.(2021)Cheng, Schwing, and Kirillov]{cheng2021maskformer}
Bowen Cheng, Alexander~G. Schwing, and Alexander Kirillov.
\newblock {Per-Pixel Classification is Not All You Need for Semantic Segmentation}.
\newblock In \emph{NeurIPS}, 2021.

\bibitem[Cheng et~al.(2022)Cheng, Misra, Schwing, Kirillov, and Girdhar]{cheng2022mask2former}
Bowen Cheng, Ishan Misra, Alexander~G Schwing, Alexander Kirillov, and Rohit Girdhar.
\newblock {Masked-attention Mask Transformer for Universal Image Segmentation}.
\newblock In \emph{CVPR}, 2022.

\bibitem[Cordts et~al.(2016)Cordts, Omran, Ramos, Rehfeld, Enzweiler, Benenson, Franke, Roth, and Schiele]{cordts2016cityscapes}
Marius Cordts, Mohamed Omran, Sebastian Ramos, Timo Rehfeld, Markus Enzweiler, Rodrigo Benenson, Uwe Franke, Stefan Roth, and Bernt Schiele.
\newblock {The Cityscapes Dataset for Semantic Urban Scene Understanding}.
\newblock In \emph{CVPR}, 2016.

\bibitem[de~Geus et~al.(2021)de~Geus, Meletis, Lu, Wen, and Dubbelman]{degeus2021pps}
Daan de Geus, Panagiotis Meletis, Chenyang Lu, Xiaoxiao Wen, and Gijs Dubbelman.
\newblock {Part-aware Panoptic Segmentation}.
\newblock In \emph{CVPR}, 2021.

\bibitem[Everingham et~al.(2010)Everingham, Van~Gool, Williams, Winn, and Zisserman]{everingham2010pascal}
Mark Everingham, Luc Van~Gool, Christopher K.~I. Williams, John Winn, and Andrew Zisserman.
\newblock {The Pascal Visual Object Classes (VOC) Challenge}.
\newblock \emph{IJCV}, 88\penalty0 (2):\penalty0 303--338, 2010.

\bibitem[Gong et~al.(2018)Gong, Liang, Li, Chen, Yang, and Lin]{gong2018cihp}
Ke Gong, Xiaodan Liang, Yicheng Li, Yimin Chen, Ming Yang, and Liang Lin.
\newblock Instance-level human parsing via part grouping network.
\newblock In \emph{ECCV}, 2018.

\bibitem[Gong et~al.(2019)Gong, Gao, Liang, Shen, Wang, and Lin]{gong2019graphonomy}
Ke Gong, Yiming Gao, Xiaodan Liang, Xiaohui Shen, Meng Wang, and Liang Lin.
\newblock Graphonomy: Universal human parsing via graph transfer learning.
\newblock In \emph{CVPR}, 2019.

\bibitem[He et~al.(2016)He, Zhang, Ren, and Sun]{he2016resnet}
Kaiming He, Xiangyu Zhang, Shaoqing Ren, and Jian Sun.
\newblock {Deep Residual Learning for Image Recognition}.
\newblock In \emph{CVPR}, 2016.

\bibitem[Jagadeesh et~al.(2022)Jagadeesh, Schuster, and Stricker]{jagadeesh2022jppf}
Sravan~Kumar Jagadeesh, Ren{\'e} Schuster, and Didier Stricker.
\newblock {Multi-task Fusion for Efficient Panoptic-Part Segmentation}.
\newblock In \emph{ICPRAM}, 2022.

\bibitem[Jain et~al.(2023)Jain, Li, Chiu, Hassani, Orlov, and Shi]{jain2023oneformer}
Jitesh Jain, Jiachen Li, MangTik Chiu, Ali Hassani, Nikita Orlov, and Humphrey Shi.
\newblock {OneFormer: One Transformer to Rule Universal Image Segmentation}.
\newblock In \emph{CVPR}, 2023.

\bibitem[Kirillov et~al.(2019{\natexlab{a}})Kirillov, Girshick, He, and Dollar]{kirillov2019panopticfpn}
Alexander Kirillov, Ross Girshick, Kaiming He, and Piotr Dollar.
\newblock {Panoptic Feature Pyramid Networks}.
\newblock In \emph{CVPR}, 2019{\natexlab{a}}.

\bibitem[Kirillov et~al.(2019{\natexlab{b}})Kirillov, He, Girshick, Rother, and Dollar]{kirillov2019ps}
Alexander Kirillov, Kaiming He, Ross Girshick, Carsten Rother, and Piotr Dollar.
\newblock {Panoptic Segmentation}.
\newblock In \emph{CVPR}, 2019{\natexlab{b}}.

\bibitem[Li et~al.(2018)Li, Zhao, Chen, Roy, Yan, Feng, and Sim]{li2018multi}
Jianshu Li, Jian Zhao, Yunpeng Chen, Sujoy Roy, Shuicheng Yan, Jiashi Feng, and Terence Sim.
\newblock Multi-human parsing machines.
\newblock In \emph{ACM MM}, 2018.

\bibitem[Li et~al.(2022)Li, Zhou, Wang, Li, and Yang]{li2022deephierarchical}
Liulei Li, Tianfei Zhou, Wenguan Wang, Jianwu Li, and Yi Yang.
\newblock {Deep Hierarchical Semantic Segmentation}.
\newblock In \emph{CVPR}, 2022.

\bibitem[{Li} et~al.(2022){Li}, {Xu}, {Wei}, and {Yang}]{li2020self}
Peike {Li}, Yunqiu {Xu}, Yunchao {Wei}, and Yi {Yang}.
\newblock {Self-Correction for Human Parsing}.
\newblock \emph{IEEE TPAMI}, 44\penalty0 (6):\penalty0 3260--3271, 2022.

\bibitem[Li et~al.(2017)Li, Arnab, and Torr]{li2017holistic}
Qizhu Li, Anurag Arnab, and Philip~HS Torr.
\newblock Holistic, instance-level human parsing.
\newblock In \emph{BMVC}, 2017.

\bibitem[Li et~al.(2022)Li, Xu, Yang, Cheng, Tong, and Tao]{li2022ppf}
Xiangtai Li, Shilin Xu, Yibo Yang, Guangliang Cheng, Yunhai Tong, and Dacheng Tao.
\newblock {Panoptic-PartFormer: Learning a Unified Model for Panoptic Part Segmentation}.
\newblock In \emph{ECCV}, 2022.

\bibitem[Li et~al.(2023)Li, Xu, Yang, Yuan, Cheng, Tong, Lin, and Tao]{li2023ppf++}
Xiangtai Li, Shilin Xu, Yibo Yang, Haobo Yuan, Guangliang Cheng, Yunhai Tong, Zhouchen Lin, and Dacheng Tao.
\newblock {PanopticPartFormer++: A Unified and Decoupled View for Panoptic Part Segmentation}.
\newblock \emph{arXiv preprint arXiv:2301.00954}, 2023.

\bibitem[Liang et~al.(2018)Liang, Gong, Shen, and Lin]{liang2018lip}
Xiaodan Liang, Ke Gong, Xiaohui Shen, and Liang Lin.
\newblock {Look into Person: Joint Body Parsing \& Pose Estimation Network and A New Benchmark}.
\newblock \emph{IEEE TPAMI}, 41\penalty0 (4):\penalty0 871--885, 2018.

\bibitem[Lin et~al.(2014)Lin, Maire, Belongie, Hays, Perona, Ramanan, Doll{\'a}r, and Zitnick]{lin2014coco}
Tsung-Yi Lin, Michael Maire, Serge Belongie, James Hays, Pietro Perona, Deva Ramanan, Piotr Doll{\'a}r, and C~Lawrence Zitnick.
\newblock {Microsoft COCO: Common Objects in Context}.
\newblock In \emph{ECCV}, 2014.

\bibitem[Lin et~al.(2017)Lin, Dollar, Girshick, He, Hariharan, and Belongie]{lin2017fpn}
Tsung-Yi Lin, Piotr Dollar, Ross Girshick, Kaiming He, Bharath Hariharan, and Serge Belongie.
\newblock {Feature Pyramid Networks for Object Detection}.
\newblock In \emph{CVPR}, 2017.

\bibitem[Liu et~al.(2022{\natexlab{a}})Liu, Kortylewski, Zhang, Li, Guo, Liu, Yuan, Mu, Qiu, and Yuille]{liu2022udapart}
Qing Liu, Adam Kortylewski, Zhishuai Zhang, Zizhang Li, Mengqi Guo, Qihao Liu, Xiaoding Yuan, Jiteng Mu, Weichao Qiu, and Alan Yuille.
\newblock {Learning Part Segmentation Through Unsupervised Domain Adaptation From Synthetic Vehicles}.
\newblock In \emph{CVPR}, 2022{\natexlab{a}}.

\bibitem[Liu et~al.(2021)Liu, Lin, Cao, Hu, Wei, Zhang, Lin, and Guo]{liu2021swin}
Ze Liu, Yutong Lin, Yue Cao, Han Hu, Yixuan Wei, Zheng Zhang, Stephen Lin, and Baining Guo.
\newblock {Swin Transformer: Hierarchical Vision Transformer Using Shifted Windows}.
\newblock In \emph{ICCV}, 2021.

\bibitem[Liu et~al.(2022{\natexlab{b}})Liu, Mao, Wu, Feichtenhofer, Darrell, and Xie]{liu2022convnext}
Zhuang Liu, Hanzi Mao, Chao-Yuan Wu, Christoph Feichtenhofer, Trevor Darrell, and Saining Xie.
\newblock {A ConvNet for the 2020s}.
\newblock In \emph{CVPR}, 2022{\natexlab{b}}.

\bibitem[Loshchilov and Hutter(2019)]{loshchilov2019adamw}
Ilya Loshchilov and Frank Hutter.
\newblock {Decoupled Weight Decay Regularization}.
\newblock In \emph{ICLR}, 2019.

\bibitem[Michieli et~al.(2020)Michieli, Borsato, Rossi, and Zanuttigh]{michieli2020gmnet}
Umberto Michieli, Edoardo Borsato, Luca Rossi, and Pietro Zanuttigh.
\newblock {GMNet: Graph Matching Network for Large Scale Part Semantic Segmentation in the Wild}.
\newblock In \emph{ECCV}, 2020.

\bibitem[Milletari et~al.(2016)Milletari, Navab, and Ahmadi]{milletari2016dice}
Fausto Milletari, Nassir Navab, and Seyed-Ahmad Ahmadi.
\newblock {V-Net: Fully Convolutional Neural Networks for Volumetric Medical Image Segmentatio}.
\newblock In \emph{3DV}, 2016.

\bibitem[Mottaghi et~al.(2014)Mottaghi, Chen, Liu, Cho, Lee, Fidler, Urtasun, and Yuille]{mottaghi14pascalcontext}
Roozbeh Mottaghi, Xianjie Chen, Xiaobai Liu, Nam-Gyu Cho, Seong-Whan Lee, Sanja Fidler, Raquel Urtasun, and Alan Yuille.
\newblock {The Role of Context for Object Detection and Semantic Segmentation in the Wild}.
\newblock In \emph{CVPR}, 2014.

\bibitem[Pan et~al.(2023)Pan, Liu, Chao, and Price]{pan2023ops}
Tai-Yu Pan, Qing Liu, Wei-Lun Chao, and Brian Price.
\newblock {Towards Open-World Segmentation of Parts}.
\newblock In \emph{CVPR}, 2023.

\bibitem[Qi et~al.(2023)Qi, Kuen, Guo, Gu, Lin, Du, Xu, and Yang]{qi2023aims}
Lu Qi, Jason Kuen, Weidong Guo, Jiuxiang Gu, Zhe Lin, Bo Du, Yu Xu, and Ming-Hsuan Yang.
\newblock {AIMS: All-Inclusive Multi-Level Segmentation for Anything}.
\newblock In \emph{NeurIPS}, 2023.

\bibitem[Ruan et~al.(2019)Ruan, Liu, Huang, Wei, Wei, and Zhao]{ruan2019devil}
Tao Ruan, Ting Liu, Zilong Huang, Yunchao Wei, Shikui Wei, and Yao Zhao.
\newblock Devil in the details: Towards accurate single and multiple human parsing.
\newblock In \emph{AAAI}, 2019.

\bibitem[Russakovsky et~al.(2015)Russakovsky, Deng, Su, Krause, Satheesh, Ma, Huang, Karpathy, Khosla, Bernstein, Berg, and Fei-Fei]{russakovsky2015imagenet}
Olga Russakovsky, Jia Deng, Hao Su, Jonathan Krause, Sanjeev Satheesh, Sean Ma, Zhiheng Huang, Andrej Karpathy, Aditya Khosla, Michael Bernstein, Alexander~C. Berg, and Li Fei-Fei.
\newblock {ImageNet} {Large} {Scale} {Visual} {Recognition} {Challenge}.
\newblock \emph{IJCV}, 115\penalty0 (3):\penalty0 211--252, 2015.

\bibitem[Singh et~al.(2022)Singh, Gupta, Shenoy, and Sarvadevabhatla]{singh2022float}
Rishubh Singh, Pranav Gupta, Pradeep Shenoy, and Ravikiran Sarvadevabhatla.
\newblock {FLOAT: Factorized Learning of Object Attributes for Improved Multi-Object Multi-Part Scene Parsing}.
\newblock In \emph{CVPR}, 2022.

\bibitem[Sun et~al.(2023)Sun, Chen, Zhu, Xiao, Luo, Xie, and Yan]{sun2023vlpart}
Peize Sun, Shoufa Chen, Chenchen Zhu, Fanyi Xiao, Ping Luo, Saining Xie, and Zhicheng Yan.
\newblock Going denser with open-vocabulary part segmentation.
\newblock In \emph{ICCV}, 2023.

\bibitem[Tan and Le(2019)]{tan2019efficientnet}
Mingxing Tan and Quoc Le.
\newblock {EfficientNet: Rethinking Model Scaling for Convolutional Neural Networks}.
\newblock In \emph{ICML}, 2019.

\bibitem[Tan et~al.(2021)Tan, Xu, Ye, Hao, and Ma]{tan2021partparsing}
Xin Tan, Jiachen Xu, Zhou Ye, Jinkun Hao, and Lizhuang Ma.
\newblock {Confident Semantic Ranking Loss for Part Parsing}.
\newblock In \emph{ICME}, 2021.

\bibitem[Tang et~al.(2021)Tang, Chen, Li, Li, Zhang, and Hu]{tang2021bpr}
Chufeng Tang, Hang Chen, Xiao Li, Jianmin Li, Zhaoxiang Zhang, and Xiaolin Hu.
\newblock {Look Closer To Segment Better: Boundary Patch Refinement for Instance Segmentation}.
\newblock In \emph{CVPR}, 2021.

\bibitem[Tang et~al.(2023)Tang, Xie, Zhang, Hu, and Tian]{tang2022virreq}
Chufeng Tang, Lingxi Xie, Xiaopeng Zhang, Xiaolin Hu, and Qi Tian.
\newblock {Visual Recognition by Request}.
\newblock In \emph{CVPR}, 2023.

\bibitem[Tian et~al.(2020)Tian, Shen, and Chen]{tian2020condinst}
Zhi Tian, Chunhua Shen, and Hao Chen.
\newblock {Conditional Convolutions for Instance Segmentation}.
\newblock In \emph{ECCV}, 2020.

\bibitem[Wang et~al.(2021)Wang, Zhu, Adam, Yuille, and Chen]{wang2021maxdeeplab}
Huiyu Wang, Yukun Zhu, Hartwig Adam, Alan Yuille, and Liang-Chieh Chen.
\newblock {MaX-DeepLab: End-to-End Panoptic Segmentation With Mask Transformers}.
\newblock In \emph{CVPR}, 2021.

\bibitem[Wang et~al.(2020)Wang, Zhu, Dai, Pang, Shen, and Shao]{wang2020hierarchical}
Wenguan Wang, Hailong Zhu, Jifeng Dai, Yanwei Pang, Jianbing Shen, and Ling Shao.
\newblock Hierarchical human parsing with typed part-relation reasoning.
\newblock In \emph{CVPR}, 2020.

\bibitem[Wei et~al.(2023)Wei, Yue, Zhang, Kong, Liu, and Pang]{wei2023ovparts}
Meng Wei, Xiaoyu Yue, Wenwei Zhang, Shu Kong, Xihui Liu, and Jiangmiao Pang.
\newblock {OV-PARTS: Towards Open-Vocabulary Part Segmentation}.
\newblock In \emph{NeurIPS}, 2023.

\bibitem[Wu et~al.(2019)Wu, Kirillov, Massa, Lo, and Girshick]{wu2019detectron2}
Yuxin Wu, Alexander Kirillov, Francisco Massa, Wan-Yen Lo, and Ross Girshick.
\newblock {Detectron2}.
\newblock \url{https://github.com/facebookresearch/detectron2}, 2019.

\bibitem[Xiao et~al.(2018)Xiao, Liu, Zhou, Jiang, and Sun]{xiao2018upernet}
Tete Xiao, Yingcheng Liu, Bolei Zhou, Yuning Jiang, and Jian Sun.
\newblock Unified perceptual parsing for scene understanding.
\newblock In \emph{ECCV}, 2018.

\bibitem[Xie et~al.(2021)Xie, Wang, Yu, Anandkumar, Alvarez, and Luo]{xie2021segformer}
Enze Xie, Wenhai Wang, Zhiding Yu, Anima Anandkumar, Jose~M. Alvarez, and Ping Luo.
\newblock {SegFormer: Simple and Efficient Design for Semantic Segmentation with Transformers}.
\newblock In \emph{NeurIPS}, 2021.

\bibitem[Yang et~al.(2019)Yang, Song, Wang, and Jiang]{yang2019parsingrcnn}
Lu Yang, Qing Song, Zhihui Wang, and Ming Jiang.
\newblock {Parsing R-CNN for Instance-Level Human Analysis}.
\newblock In \emph{CVPR}, 2019.

\bibitem[Yang et~al.(2020)Yang, Song, Wang, Hu, Liu, Xin, Jia, and Xu]{yang2020rprcnn}
Lu Yang, Qing Song, Zhihui Wang, Mengjie Hu, Chun Liu, Xueshi Xin, Wenhe Jia, and Songcen Xu.
\newblock {Renovating Parsing R-CNN for Accurate Multiple Human Parsing}.
\newblock In \emph{ECCV}, 2020.

\bibitem[Yang et~al.(2021)Yang, Song, Wang, Hu, and Liu]{yang2021hierrcnn}
Lu Yang, Qing Song, Zhihui Wang, Mengjie Hu, and Chun Liu.
\newblock {Hier R-CNN: Instance-Level Human Parts Detection and A New Benchmark}.
\newblock \emph{IEEE TIP}, 30:\penalty0 39--54, 2021.

\bibitem[Zhang et~al.(2022)Zhang, Cao, Qi, Song, and Zhou]{zhang2022aiparsing}
Sanyi Zhang, Xiaochun Cao, Guo-Jun Qi, Zhanjie Song, and Jie Zhou.
\newblock {AIParsing: Anchor-Free Instance-Level Human Parsing}.
\newblock \emph{IEEE TIP}, 31:\penalty0 5599--5612, 2022.

\bibitem[Zhang et~al.(2021)Zhang, Pang, Chen, and Loy]{zhang2021knet}
Wenwei Zhang, Jiangmiao Pang, Kai Chen, and Chen~Change Loy.
\newblock {K-Net: Towards Unified Image Segmentation}.
\newblock In \emph{NeurIPS}, 2021.

\bibitem[Zhao et~al.(2018)Zhao, Li, Cheng, Sim, Yan, and Feng]{zhao2018mhp}
Jian Zhao, Jianshu Li, Yu Cheng, Terence Sim, Shuicheng Yan, and Jiashi Feng.
\newblock Understanding humans in crowded scenes: Deep nested adversarial learning and a new benchmark for multi-human parsing.
\newblock In \emph{ACM MM}, 2018.

\bibitem[Zhao et~al.(2019)Zhao, Li, Zhang, and Tian]{zhao2019bsanet}
Yifan Zhao, Jia Li, Yu Zhang, and Yonghong Tian.
\newblock {Multi-Class Part Parsing With Joint Boundary-Semantic Awareness}.
\newblock In \emph{ICCV}, 2019.

\end{thebibliography}
}

\end{document}